%% file: acl_latex.tex
\pdfoutput=1

\documentclass[11pt]{article}
\usepackage[]{acl}

\usepackage{times}
\usepackage{latexsym}

\usepackage[T1]{fontenc}
\usepackage[utf8]{inputenc}

\usepackage{microtype}

\usepackage{inconsolata}

\usepackage{graphicx}
\usepackage{booktabs}
\usepackage{multirow}
\usepackage{color}
\usepackage{hyperref}
\usepackage{amsmath}
\usepackage{amsfonts}
\usepackage{amsthm}
\usepackage{scalefnt}
\usepackage{xspace}
\usepackage{CJKutf8}
\usepackage{authblk}

\theoremstyle{plain}

\def\secref#1{\S\ref{sec:#1}}
\def\seclabel#1{\label{sec:#1}}

\newcommand{\falsenegatives}{\textbf{\emph{SCLFP}}\xspace}
\newcommand{\truenegatives}{\textbf{\emph{UWLFP}}\xspace}

\newcounter{notecounter}

\newcommand{\enoteson}{\long\gdef\enote##1##2{{
\stepcounter{notecounter}
{\large\bf
\hspace{0cm}\arabic{notecounter} $<<<$ ##1: ##2
$>>>$\hspace{1cm}}}}}

\enoteson
\title{Tracing Multilingual Factual Knowledge Acquisition in Pretraining}

\author[1,2]{\bf Yihong Liu}
\author[1,2,3]{\bf Mingyang Wang}
\author[1,2]{\bf Amir Hossein Kargaran}
\author[1,2]{\bf Felicia Körner}
\author[1,2]{\\ \bf Ercong Nie}
\author[1,2]{\bf Barbara Plank}
\author[4]{\bf François Yvon}
\author[1,2]{\bf Hinrich Sch\"utze}

\affil[1]{Center for Information and Language Processing, LMU Munich} 
\affil[2]{Munich Center for Machine Learning (MCML)} \affil[3]{Bosch Center for Artificial Intelligence}
\affil[4]{Sorbonne Université, CNRS, ISIR, France
 \protect\\ \texttt{\{yihong, mingyang, amir, fkoerner, nie\}@cis.lmu.de}}

\begin{document}
\maketitle
\begin{abstract}

Large Language Models (LLMs) are capable of recalling multilingual factual knowledge present in their pretraining data.
However, most studies evaluate only the final model, leaving
the development of \emph{factual recall}
and \emph{crosslingual consistency} throughout pretraining
%process
%either "the pretraining process" or "pretraining" i prefer
%the latter: "throughout" makes clear it's a process
largely unexplored. 
In this work, we trace how factual recall and crosslingual consistency evolve during pretraining, focusing on OLMo-7B as a case study.
We find that both accuracy and consistency improve over time for most languages.
We show that this improvement is primarily driven by the \emph{fact frequency} in the pretraining corpus: more frequent facts are more likely to be recalled correctly, regardless of language. 
Yet, some low-frequency facts in non-English languages can still be correctly recalled.
Our analysis reveals that these instances largely benefit from crosslingual transfer of their English counterparts -- an effect that emerges predominantly in the early stages of pretraining.
We pinpoint two distinct pathways through which multilingual factual knowledge acquisition occurs:
(1) \textbf{\emph{frequency-driven learning}}, which is dominant and language-agnostic, and (2) \textbf{\emph{crosslingual transfer}}, which is limited in scale and typically constrained to relation types involving named entities. 
We release our code and data to facilitate further research at {\url{https://github.com/cisnlp/multilingual-fact-tracing}}.

\end{abstract}

\section{Introduction}

Despite being predominantly trained on English-centric data, LLMs exhibit surprisingly strong multilingual capabilities across a wide range of tasks \citep{jiang2023mistral7b,touvron2023llama2openfoundation,zhang-etal-2024-getting,zhao2025surveylargelanguagemodels}. 
Notably, they can recall factual knowledge in multiple languages \citep{petroni-etal-2019-language,jiang-etal-2020-x,kassner-etal-2021-multilingual}.
However, these models frequently exhibit \emph{crosslingual inconsistencies} -- answering a factual query correctly in one language but failing to do so in another \citep{qi-etal-2023-cross,chua2025crosslingual,wang2025multilinguality}. 
% This behavior contrasts with human bilinguals, who can typically recall a fact learned in one language across all their capable languages. 
Although bilinguals typically recall information more effectively when the language of encoding matches the language of retrieval, they can usually recall factual knowledge learned in one language using their other proficient language \citep{marian2000language,CHUNG2019149} -- highlighting a flexibility that contrasts with the inefficiencies seen in LLMs.
Understanding this discrepancy requires deeper insight into how multilingual factual knowledge is acquired.
% in the first place 

\begin{figure}
    \centering
    \includegraphics[width=0.48\textwidth]{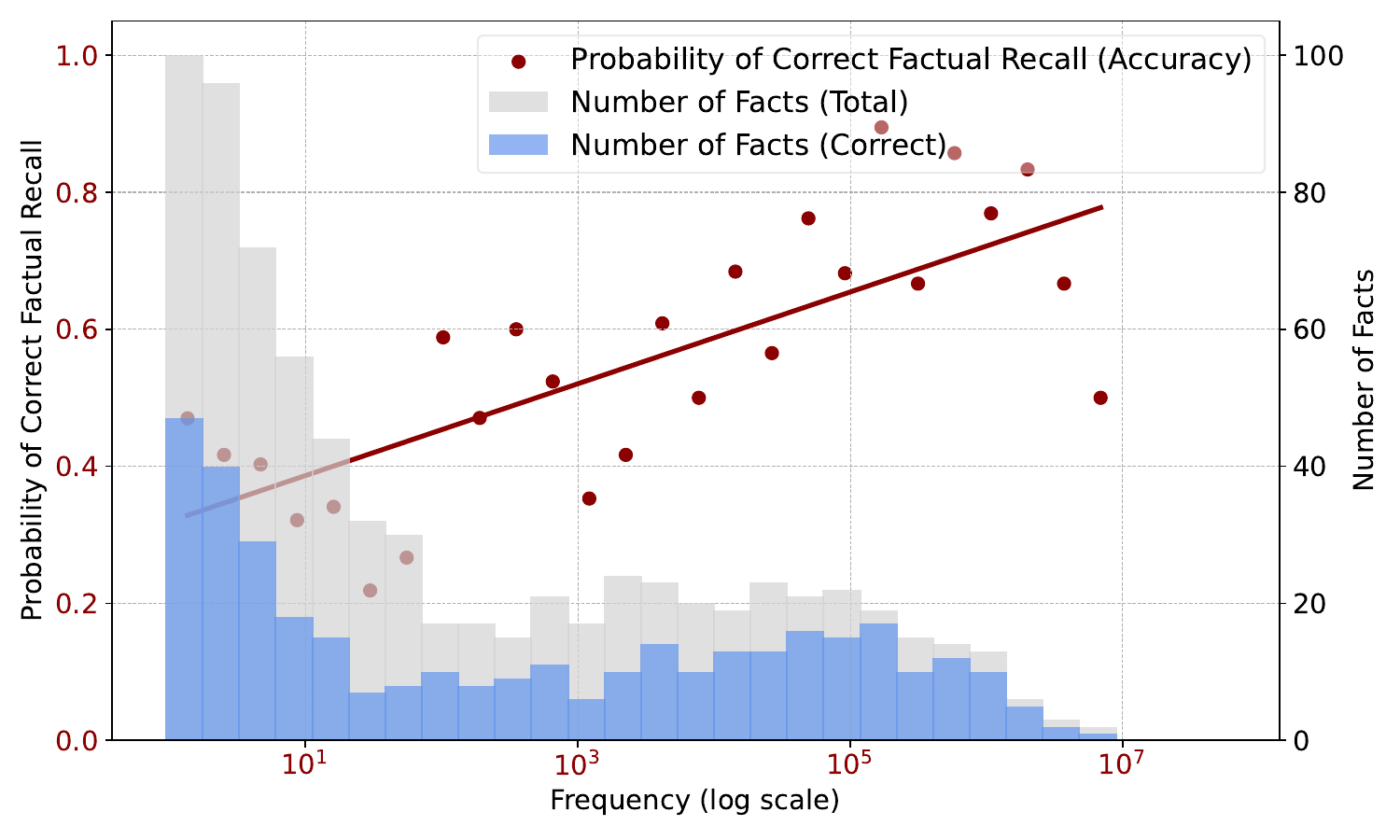}
    \caption{
        Relationship between fact frequency and factual recall in \textbf{Catalan}. 
        High-frequency facts are more likely to be correctly recalled, indicating the effect of \emph{frequency-based learning}. 
        Meanwhile, the correct recall of some low-frequency facts suggests the influence of \emph{crosslingual transfer} from other languages.
    }
    \label{fig:catalan_first_page_figure}
\end{figure}

While prior work has investigated mechanisms of (multilingual) factual recall \citep{geva-etal-2023-dissecting,zhao-etal-2024-tracing,Chang2024Factual,fierro2024multilingualmodelsrememberinvestigating,liu2025relation} and analyzed sources of crosslingual inconsistency \citep{qi-etal-2023-cross,wang2025multilinguality}, 
these studies have largely focused on \emph{final models}, 
% drawing conclusions from static evaluations at the end of pretraining. 
drawing conclusions solely from the end of pretraining.
As a result, the developmental process by which LLMs acquire factual knowledge across languages remains poorly understood. 

To address this gap, we trace the dynamics of multilingual factual recall and crosslingual consistency throughout pretraining. 
Rather than treating factual recall as a static outcome, we analyze its emergence across checkpoints using OLMo-7B \citep{groeneveld2024olmo}, an English-centric decoder-only LLM pretrained on Dolma \citep{soldaini-etal-2024-dolma}. 
Our analysis evaluates both accuracy within individual languages and consistency across languages for facts that are parallel in all languages.

In addition, we investigate the key factors that contribute to correct multilingual factual recall. 
Prior work has shown that the frequency of an instance can significantly influence performance relating to it, including factual prediction \citep{razeghi-etal-2022-impact, elazar2023measuring,McCoy2024embers,merullo2025linear}. 
Motivated by these findings, we hypothesize that \emph{fact frequency} in the pretraining corpus plays a central role in multilingual factual recall.
To test this, we compute the frequency of each fact and systematically link it to factual recall across languages and pretraining stages.

We summarize the key findings of this paper:
\begin{itemize}
    \item[(i)] \textbf{The capacity for multilingual factual recall develops progressively during pretraining} (\secref{dynamics}).
    English and languages distant from English converge in early stages, while languages more similar to English (e.g., those sharing the Latin script) continue to improve with extended pretraining.
    \item [(ii)] \textbf{The correctness of factual recall is largely explained by a single factor: fact frequency in the pretraining corpus} (\secref{frequency}).
    High-frequency facts are consistently recalled more accurately across languages (e.g., Catalan in Figure~\ref{fig:catalan_first_page_figure}).
    In addition, this frequency-correctness relationship emerges early and strengthens throughout pretraining.
    \item [(iii)] \textbf{Some low-frequency facts in non-English languages are recalled correctly mainly via crosslingual transfer} (\secref{transferred_facts}).
    High-frequency counterparts in English mainly enable these cases. However, the scale of transfer is limited and constrained to certain relation types.
\end{itemize}

\section{Related Work}

% \paragraph{Factual Recall}
\paragraph{Multilingual Factual Recall and Consistency} 
Several studies have investigated the factual knowledge stored in models through knowledge probing. 
\citet{jiang-etal-2020-x} and \citet{kassner-etal-2021-multilingual} assess factual recall by translating English prompts into multiple languages, revealing notable performance disparities across languages. 
\citet{yin-etal-2022-geomlama} extend this analysis to region-specific commonsense knowledge, finding that the best-performing language for querying facts about a country (e.g., China) is often English rather than its native language (e.g., Chinese), indicating the English-centric bias of models. 
% \paragraph{Crosslingual Consistency}
Building on multilingual probing studies, \citet{qi-etal-2023-cross} and \citet{aggarwal2025language} investigate crosslingual consistency and find that LLMs often return different answers for equivalent queries in different languages.
\citet{wang2025multilinguality} further explore the underlying causes of these inconsistencies through mechanistic interpretability, revealing how internal representations contribute to divergent outputs across languages. 
Following this line of research, our work traces the development of factual recall and crosslingual consistency throughout pretraining, shedding light on how these capabilities emerge and evolve.

\paragraph{Pretraining Trajectory Investigation}

Several studies have investigated how Transformer-based models \citep{transformer2017vaswani} acquire linguistic or task-specific knowledge during different phases of pretraining, in both monolingual \citep{choshen-etal-2022-grammar,xia-etal-2023-training,muller-eberstein-etal-2023-subspace,chen2024sudden,Chang2024Factual} and multilingual settings \citep{blevins-etal-2022-analyzing,wang-etal-2024-probing-emergence}.
A concurrent study by \citet{merullo2025linear} most closely resembles our work; they demonstrate that fact frequency is a strong predictor of both factual recall and the emergence of linear factual representations (e.g., subject-to-object mappings via linear transformation) \citep{Hernandez2024Linearity}.
However, their analysis is conducted in a purely monolingual context. In contrast, our work examines multilingual factual knowledge acquisition and shows that while fact frequency remains a key driver of factual recall, crosslingual knowledge transfer provides additional -- albeit limited -- benefits in enhancing multilingual factual recall.

\section{Experiment Setups}

\subsection{Languages and Model Checkpoints}\seclabel{checkpoints}

\paragraph{Languages} We consider 12 languages that span 6 language families and use 7 different scripts: Arabic (\textbf{ara\_Arab}), Catalan (\textbf{cat\_Latn}), Chinese (\textbf{zho\_Hans}), English (\textbf{eng\_Latn}), French (\textbf{fra\_Latn}), Greek (\textbf{ell\_Grek}), Japanese (\textbf{jpn\_Jpan}), Korean (\textbf{kor\_Kore}), Russian (\textbf{rus\_Cyrl}), Spanish (\textbf{spa\_Latn}), Turkish (\textbf{tur\_Latn}), Ukrainian (\textbf{ukr\_Cyrl}).\footnote{Some languages, e.g., Ukrainian, are much less resourced than others, according to our exploration of the multilingual coverage of Dolma \citep{soldaini-etal-2024-dolma} (cf.\ \secref{dolma}).}

\paragraph{Model Checkpoints} 
We use the open-source OLMo-1.7 7B model \citep{groeneveld2024olmo} (referred to as OLMo) in our study. 
OLMo is a decoder-only LLM pretrained on Dolma \citep{soldaini-etal-2024-dolma}, an English-centric corpus with some multilingual coverage.
% \footnote{We explore the multilingual coverage of Dolma in \secref{dolma}.} 
To capture the dynamics of factual knowledge acquisition throughout pretraining, we select model checkpoints at two granularities. 
Based on preliminary experiments showing that changes are more pronounced in the early pretraining stages, we include checkpoints every 1,000 steps from step 0 to step 50,000. 
Beyond 50,000 steps, we consider every 5,000 steps up to step 400,000. 
This setup enables us to trace the model’s development from initialization to a mature stage with good multilingual capability (trained on approximately 1.7T tokens).

\subsection{Multilingual Factual Dataset}\seclabel{dataset}

We use KLAR \citep{wang2025multilinguality}, a multilingual factual knowledge probing dataset, for our investigation. 
We use \textbf{1,197} facts grouped into \textbf{12}~relation categories (cf.\ Table~\ref{tab:relation_fact_counts} in \secref{klar}).
Each fact is represented as a triple of subject, relation, and object. 
KLAR also provides a prompt template for each relation in each language, structured as ``\texttt{<Question> The answer is:}''. For example, for triple (\emph{France, capital, Paris}), the template will then be expanded as ``\emph{Where is France's capital located? The answer is:}'', with expected answer ``\emph{Paris}'' in English.
All facts and prompt templates are available in all 12 languages.
We therefore transform each fact into a query $q_i^l$ with expected answer $o_i^l$ in language $l$; 
for each fact $i$, $q_i^l$ and $q_i^{l'}$ are translations of the same query in languages $l$ and $l'$.
We denote the resulting set of queries as $Q$.\footnote{In this paper, we only consider a single prompt template for each relation in each language, since the influence of prompt variation is expected to be limited due to the simplicity of factual recall. We present an additional study on how different prompt templates affect the factual recall in \secref{prompt_variation}.}

\subsection{Evaluation}\seclabel{evaluation}

To evaluate \textbf{consistency}, we compute the overlapping ratio of correct predictions, following \citet{jiang-etal-2020-x} and \citet{wang2025multilinguality}.
Since OLMo is an English-centric model due to the predominance of English in Dolma's documents (cf.\ \secref{dolma}), we treat English as a \emph{reference language} and compute how consistent the predictions from other languages are compared to predictions made in English:\footnote{We present a complementary investigation of holistic crosslingual consistency across all language pairs in \secref{holistic}.}
% Specifically, we use English as a ``pivot'' language and compute how consistent the predictions from other languages are compared to predictions from English:
$$
\text{CO}(l) = \frac{\sum_{i}^{|Q|}\mathbf{1}(\mathcal{M}(q_i^l) = o_i^l \land \mathcal{M}(q_i^\text{eng}) = o_i^\text{eng})}{\sum_{i}^{|Q|}\mathbf{1}(\mathcal{M}(q_i^l) = o_i^l \lor \mathcal{M}(q_i^\text{eng}) = o_i^\text{eng})}
$$
where $q_i^\text{eng}$ and $o_i^\text{eng}$ are the query and expected answer for the $i$th query in English, $\mathbf{1}(\cdot)$ is the indicator function, and $\mathcal{M}(\cdot)$ is the LLM's prediction function. When assessing correctness ($\mathcal{M}(q_i^l) = o_i^l$), we rely on the model's \emph{complete generation}, checking whether it contains $o_i^l$. 
We depart here from previous
work \citep{geva-etal-2023-dissecting,qi-etal-2023-cross,Hernandez2024Linearity}
that just checks the first predicted token, which
can be misleading due to ambiguity and tokenization issues.\footnote{Even though the first token is correct, the final prediction can be wrong because the object is split into multiple tokens. For example, ``Antwerp'' and ``Antananarivo'' share the same first token ``Ant''. It is therefore ambiguous which city the model is trying to generate based on just the token ``Ant''.}
We also compute the per language \textbf{accuracy}:
$
\text{ACC}(l) = \frac{\sum_{i}^{|Q|}\mathbf{1}(\mathcal{M}(q_i^l) = o_i^l)}{|Q|}
$
which allows us to trace how well factual recall is performed.
% in each language.

\begin{figure*}
    \centering
    \includegraphics[width=0.24\textwidth]{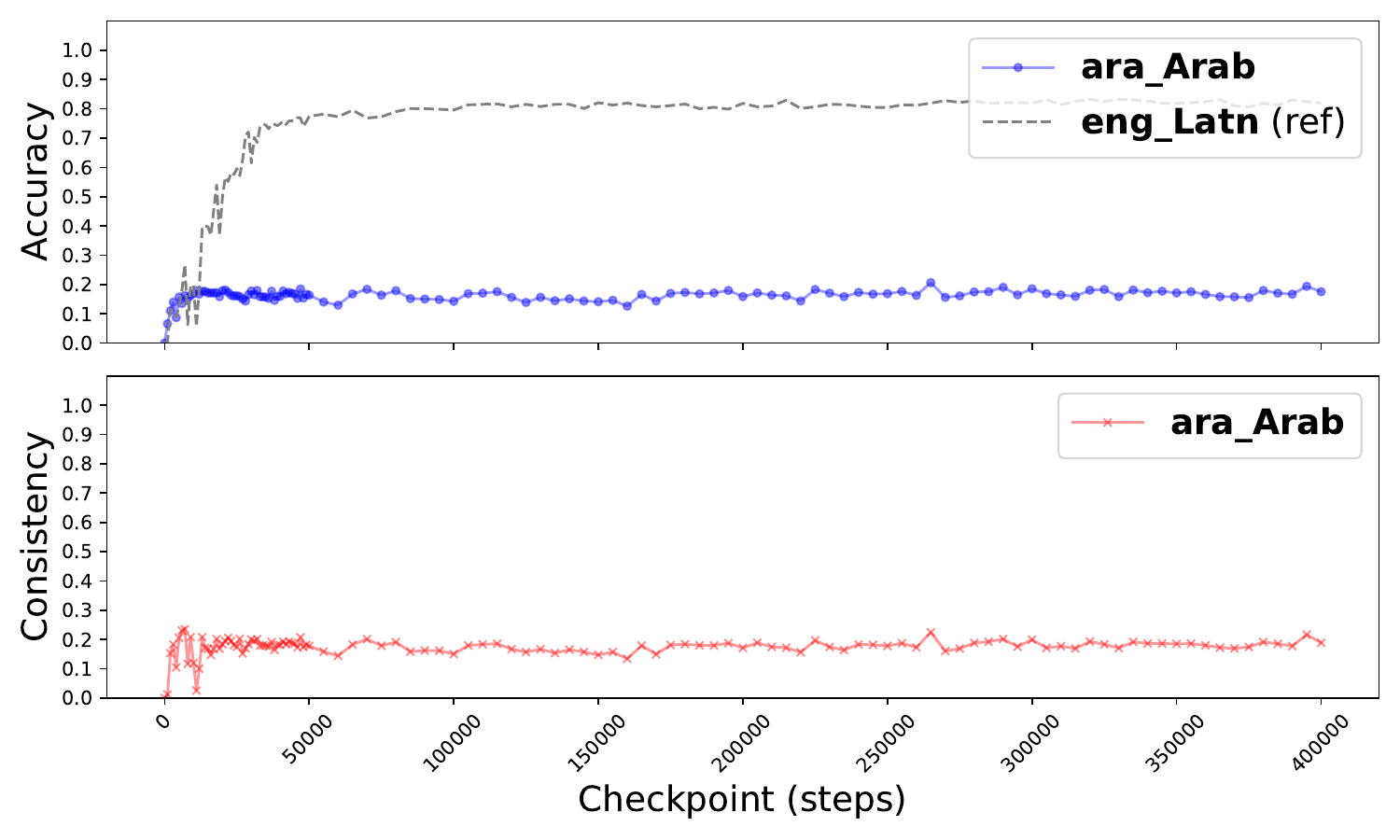}
    \includegraphics[width=0.24\textwidth]{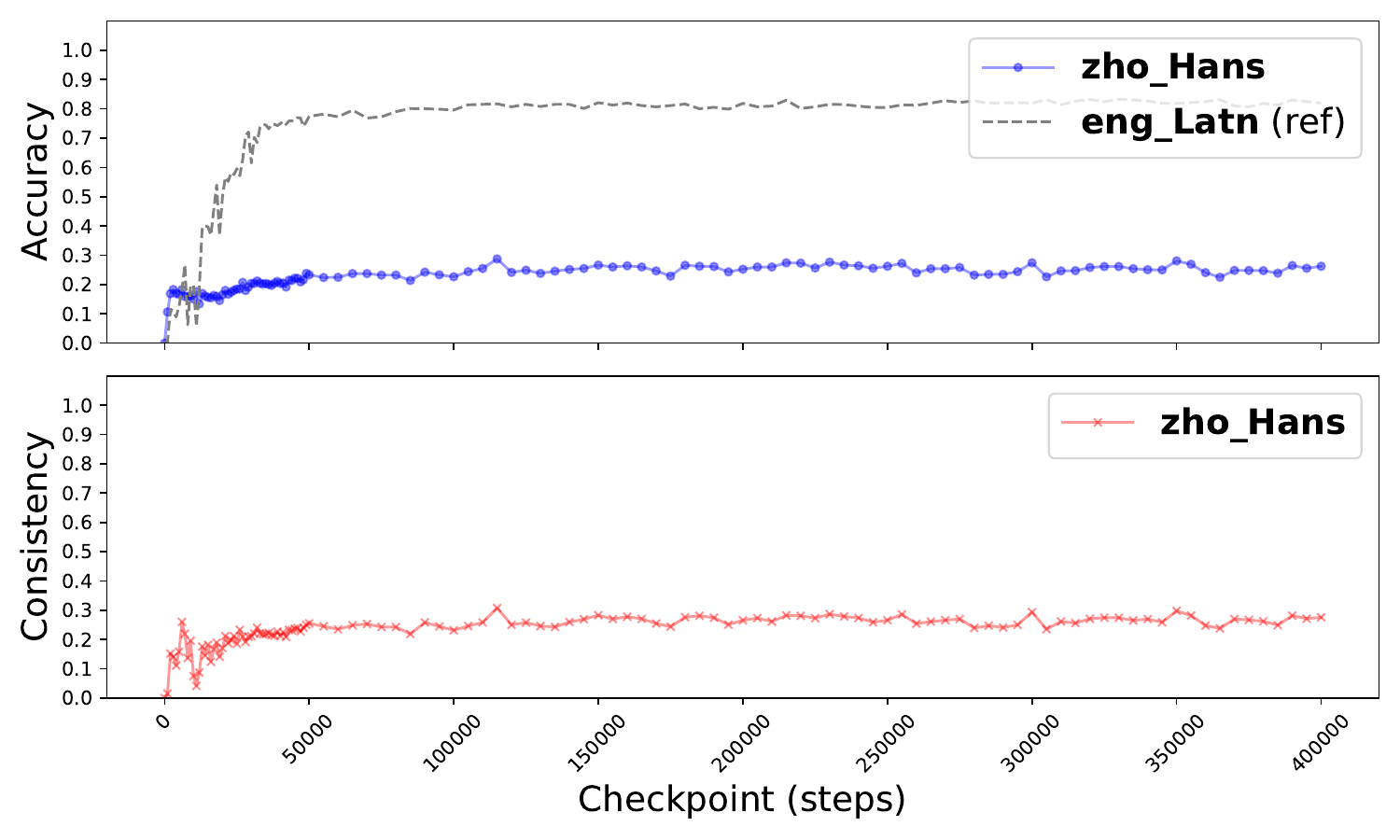}
    \includegraphics[width=0.24\textwidth]{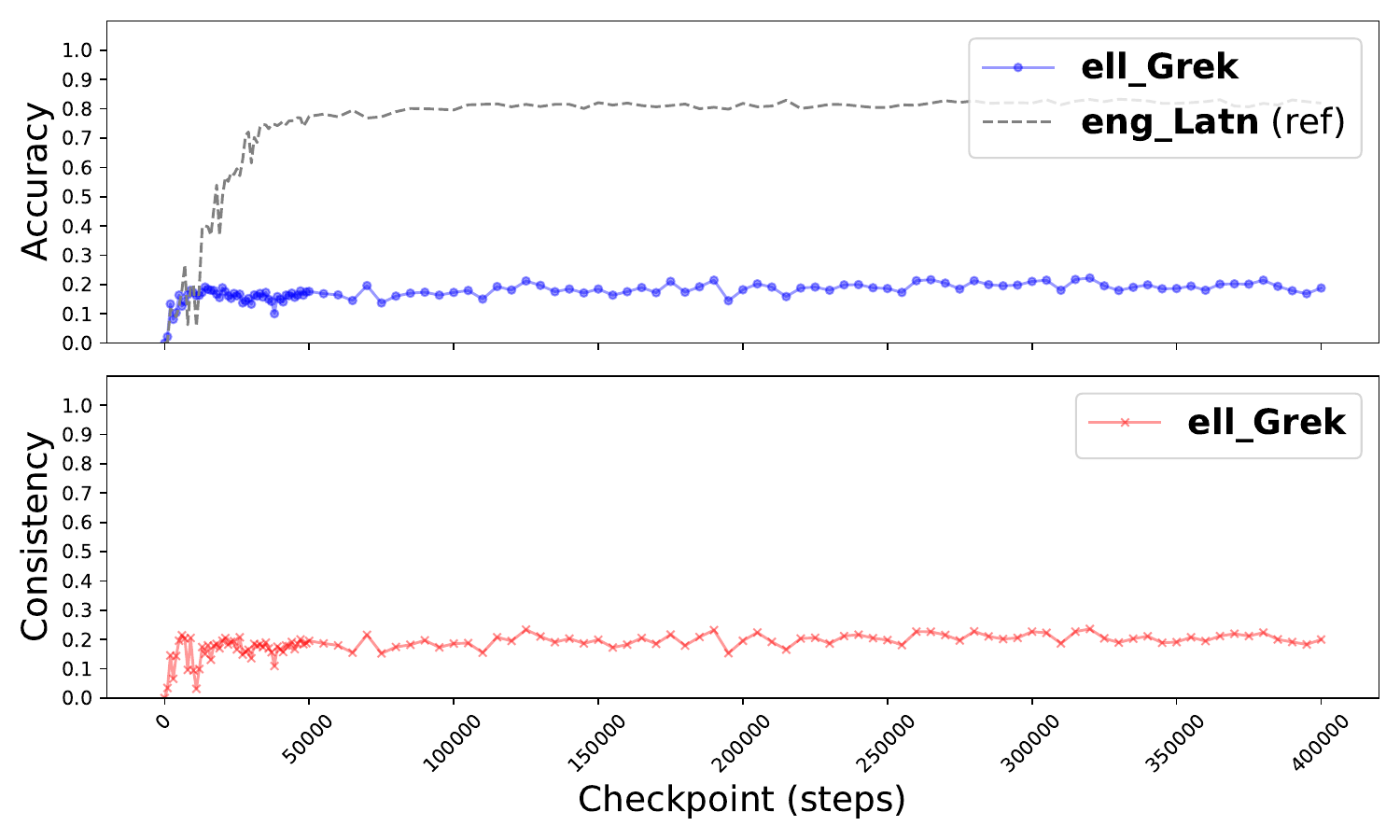}
    \includegraphics[width=0.24\textwidth]{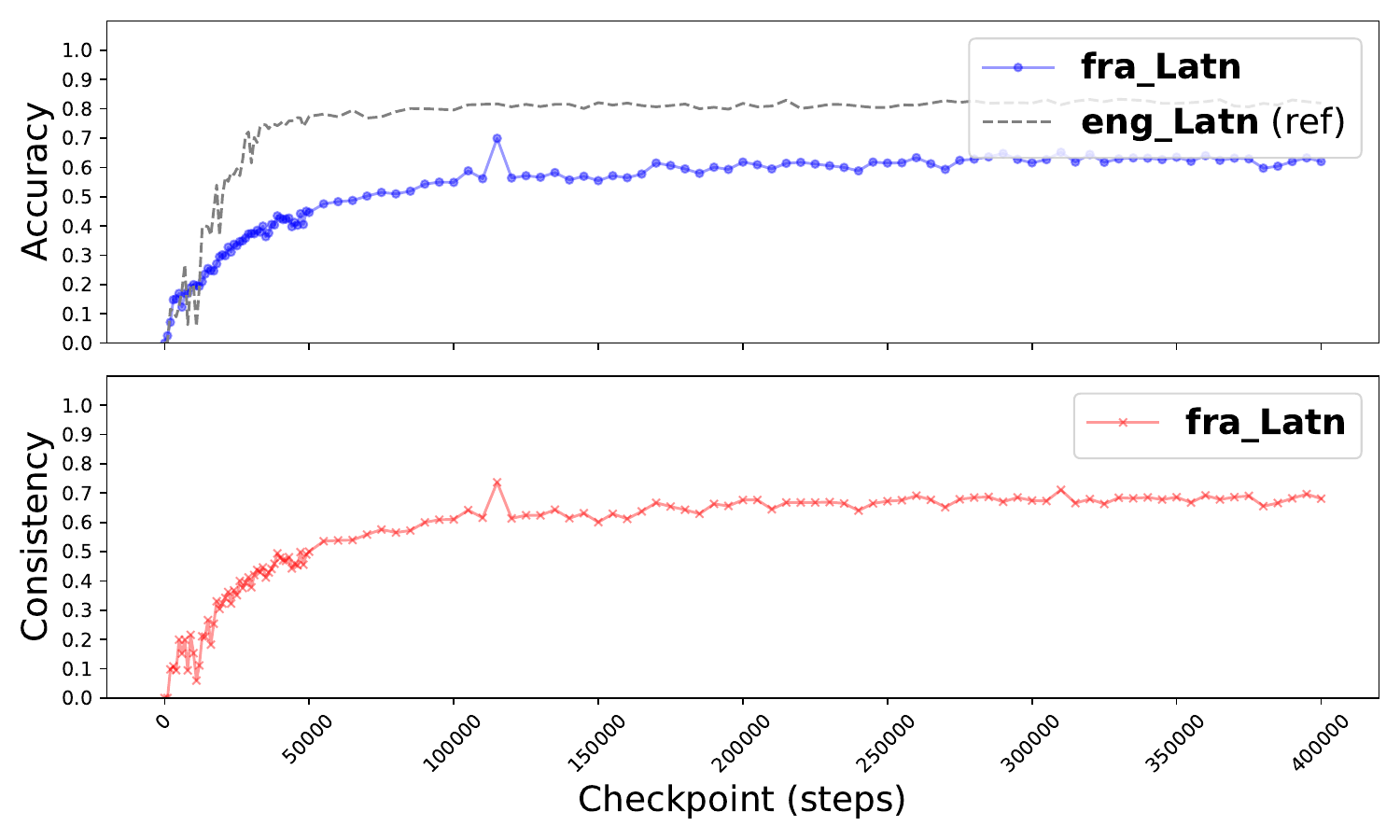}
    \includegraphics[width=0.24\textwidth]{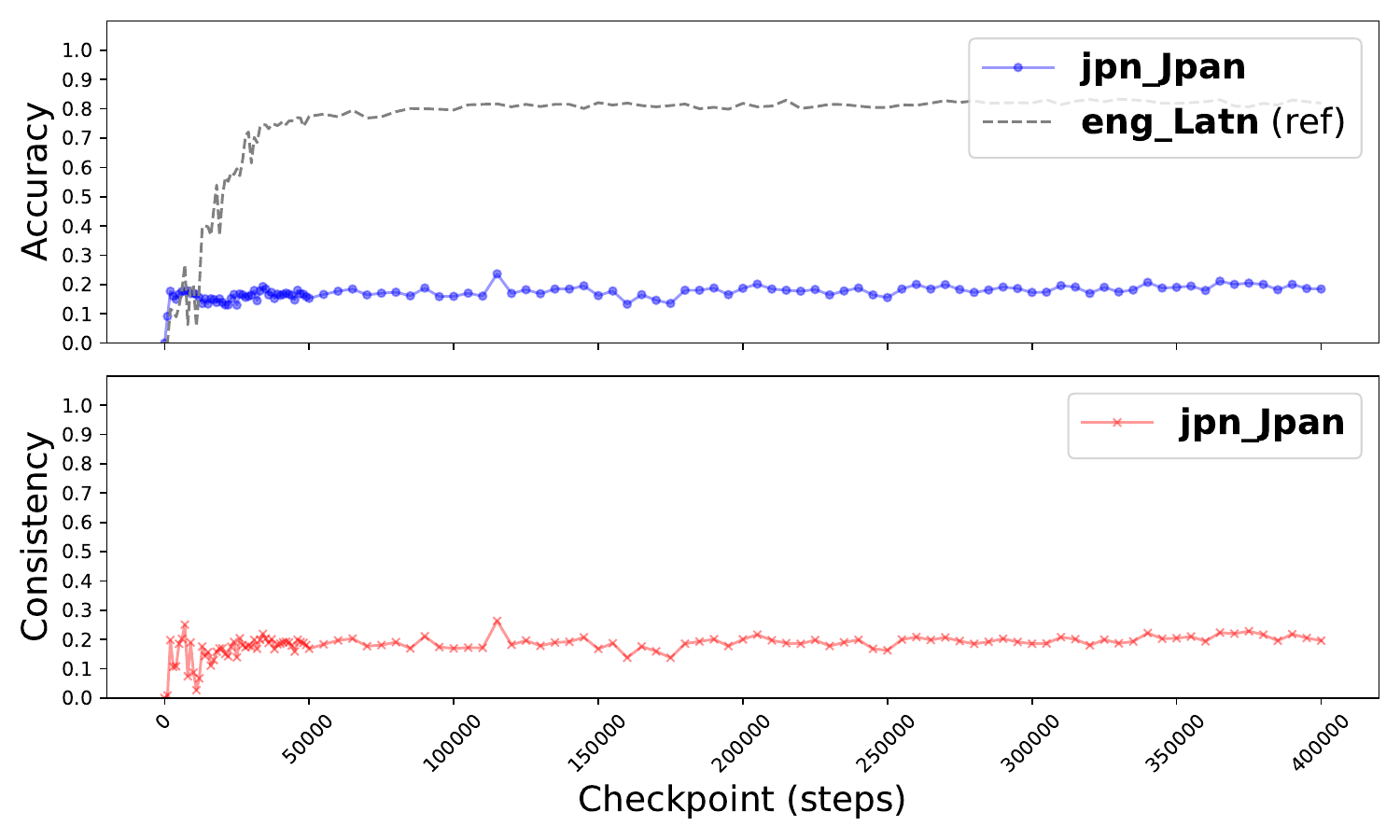}
    \includegraphics[width=0.24\textwidth]{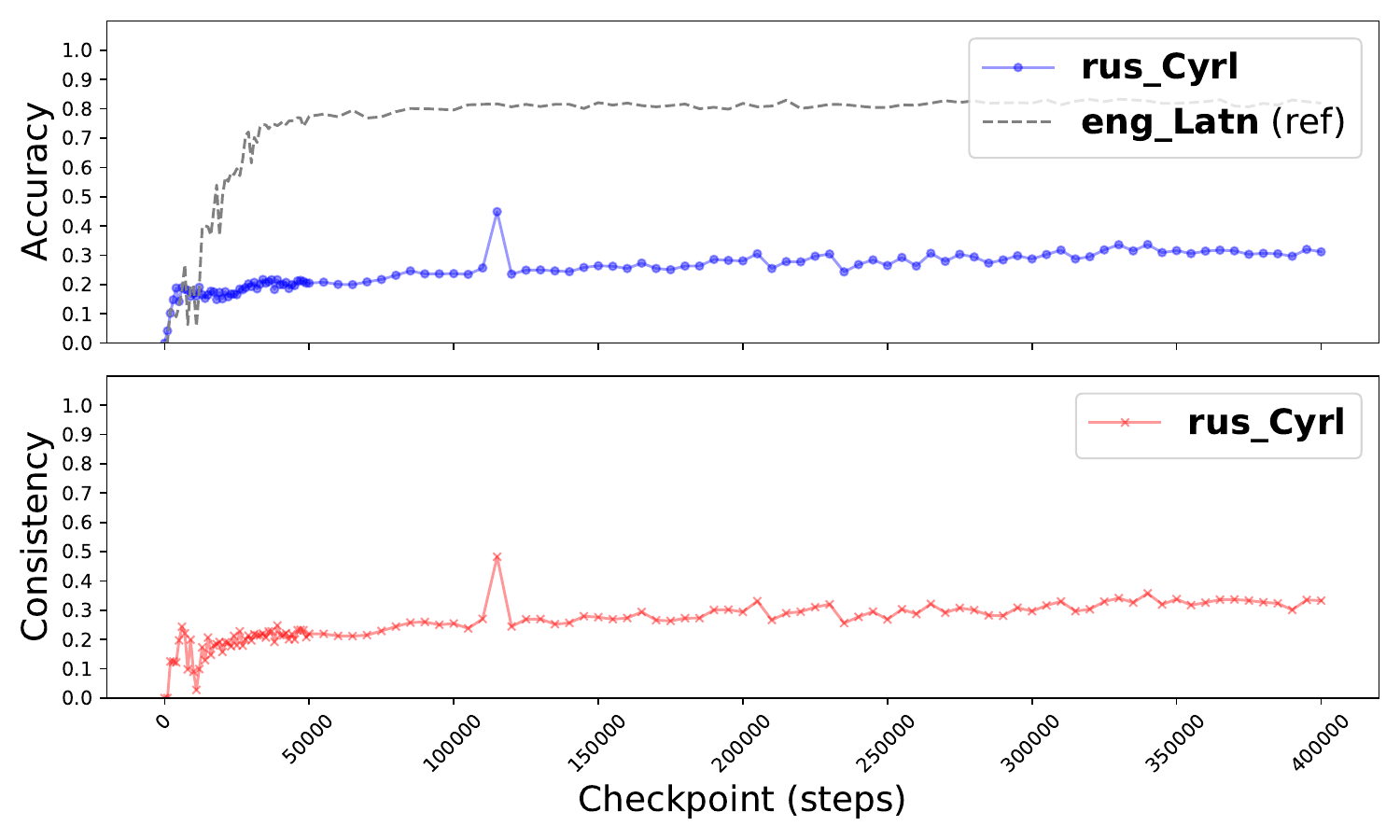}
    \includegraphics[width=0.24\textwidth]{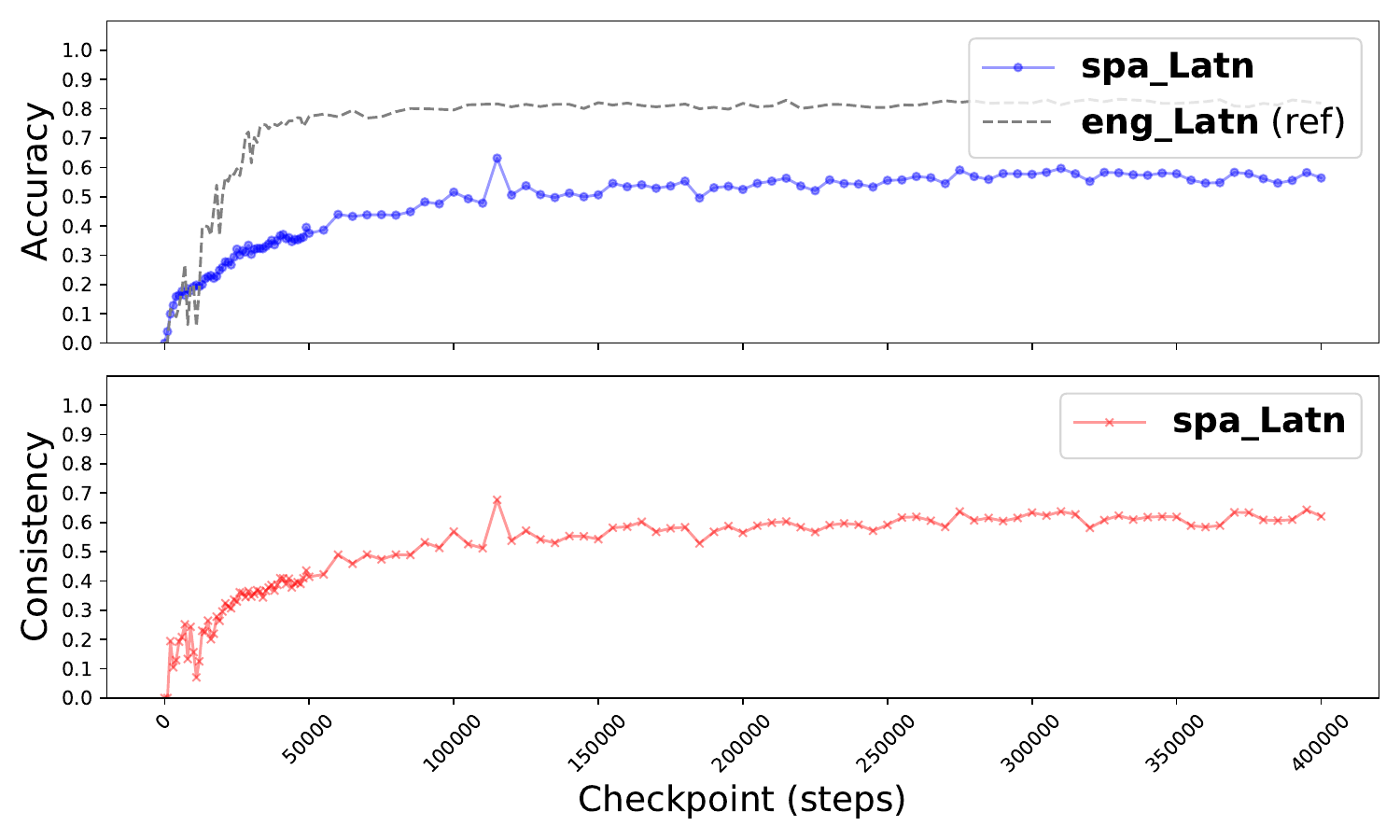}
    \includegraphics[width=0.24\textwidth]{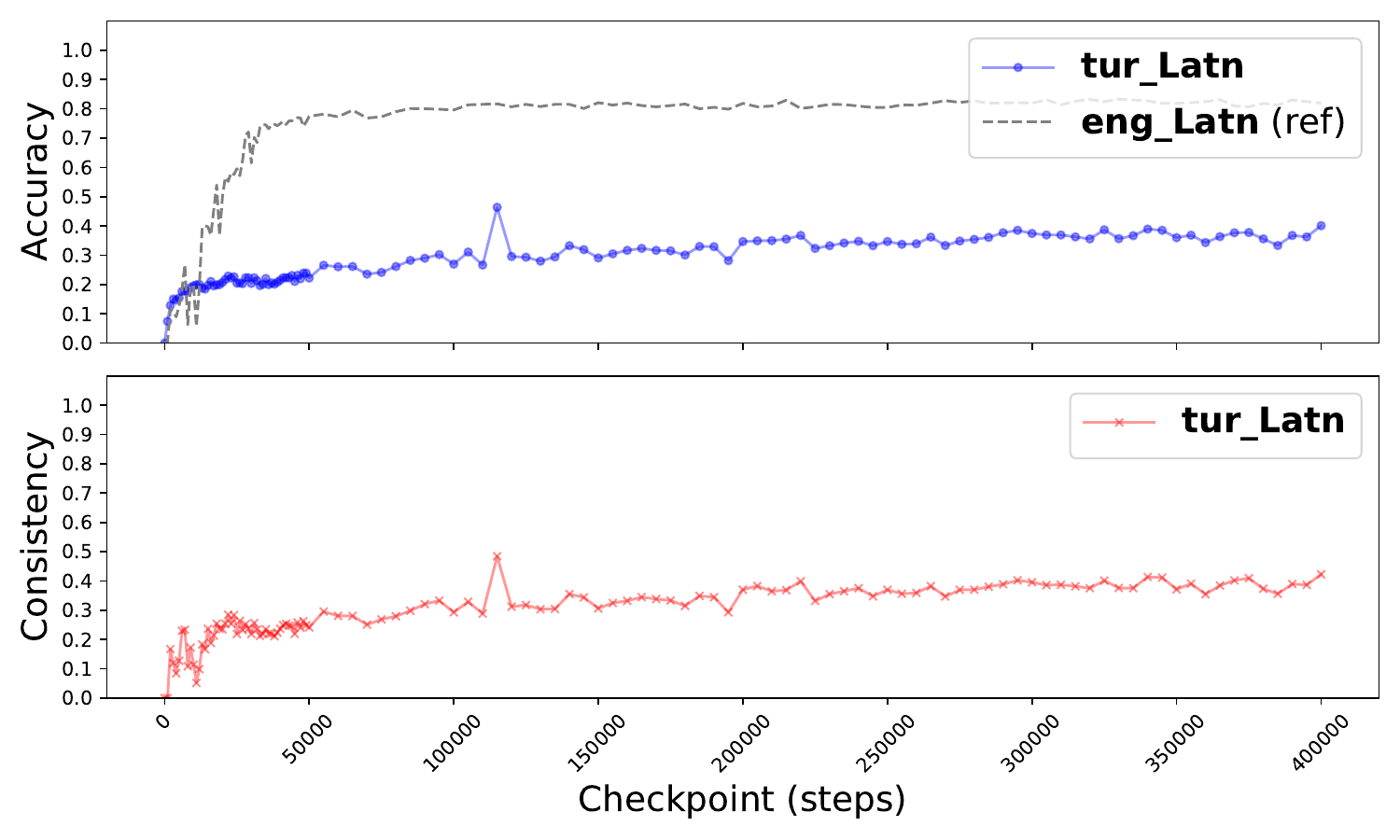}
    \caption{Factual accuracy (ACC) and crosslingual consistency (CO). While factual knowledge is rapidly acquired during the early stages of pretraining and is reasonably high in many languages, a substantial performance gap remains between English and most other languages, highlighting the limitations of crosslingual knowledge transfer.}
    \label{fig:performance_over_checkpoints}
\end{figure*}

\subsection{Fact Frequencies}\seclabel{frequencies}

We approximate a fact's frequency by counting the number of documents where \textbf{its subject and object co-occur} in the pretraining corpus.
This co-occurrence-based approximation has been widely used
%in prior work
and shown to be reliable \citep{elsahar-etal-2018-rex,elazar2023measuring,merullo2025linear,liu2025relation}.
For some languages, this approximation is fairly accurate due to the uniqueness of their scripts -- for example, the subject-object pair (\begin{CJK}{UTF8}{gbsn}法国\end{CJK}, \begin{CJK}{UTF8}{gbsn}巴黎\end{CJK}) in Chinese is unlikely to appear in texts from other languages. 
However, ambiguity arises in languages that share scripts, such as English and French. 
The same pair (\emph{France, Paris}), for instance, may appear in either language, resulting in an \emph{aggregated frequency} count shared across both.
We analyze the impact of this identical-fact effect and show that it does not compromise the robustness of our findings (cf.\ \secref{exclude}).
% For example, the frequency of the fact triple (\emph{France, capital, Paris}) is estimated by the number of times ``France'' and ``Paris'' co-occur in the same document.
To efficiently obtain these co-occurrence counts, we use the ElasticSearch API provided by \textbf{WIMBD} \citep{elazar2024wimbd}, a tool designed for scalable search and frequency analysis over large corpora.\footnote{A public demo of WIMBD is available at: \url{https://wimbd.apps.allenai.org/}. 
An alternative with similar functionality is Infini-gram \citep{liu2025infinigram}: \url{https://infini-gram.readthedocs.io/en/latest/api.html}.
}
All fact frequencies in our analysis are computed over the Dolma v1.7 corpus \citep{soldaini-etal-2024-dolma} used to pretrain OLMo, by measuring the number of subject-object co-occurrences for each fact in KLAR.

\section{Multilingual Factual Recall Dynamics}\seclabel{dynamics}

We begin our analysis by tracing how factual recall performance evolves throughout pretraining across different languages. 
Specifically, we examine both \emph{accuracy} and \emph{crosslingual consistency} at each checkpoint of OLMo (cf.~\secref{checkpoints}) using the KLAR dataset. 
Figure~\ref{fig:performance_over_checkpoints} summarizes these results for eight languages (see \secref{complete_dynamics} for full results).

\paragraph{Crosslingual consistency is tightly coupled with non-English performance.}
We observe that the trajectory of crosslingual consistency in each language $l \neq \text{eng\_Latn}$ closely mirrors its own factual accuracy throughout pretraining. 
This suggests that consistency is primarily driven by whether the fact is correctly recalled in $l$, which almost always implies that it is also recalled in English.
%  as English almost always recalls the fact if it is recalled elsewhere. as facts recalled in a language different from English
The implication is twofold. (1)
For non-English languages,
the consistency of a language (CO) is
effectively gated by its performance (ACC).
(2) The limited capability of the model to
transfer knowledge from English to other languages, referred
to as the \emph{crosslingual knowledge barrier} \citep{chua2025crosslingual},
% , is not merely a final-state issue, but 
is a persistent problem throughout pretraining.

\paragraph{Factual knowledge is acquired rapidly in early pretraining phases.}
We observe that factual recall performance (ACC) improves very quickly in the early stages of pretraining for many languages. 
For example, English reaches approximately 80\% accuracy after only 50K steps (roughly 209B tokens), with minimal gains beyond that point. 
This indicates that factual knowledge is acquired rapidly early 
% -- within the first 10\% of total pretraining --
and does not substantially benefit from further pretraining steps. 
While longer pretraining is known to improve other capabilities of LLMs \citep{kaplan2020scalinglaw,le-scao-etal-2022-language,xiong2024temporal}, factual recall appears to rely on simpler mechanisms gained in early-stage training, likely tied to memorization of frequent co-occurrences, for which we give empirical evidence in \secref{frequency}.

\begin{figure*}
    \centering
    \setlength{\abovecaptionskip}{-0.01cm}
    \includegraphics[width=0.24\textwidth]{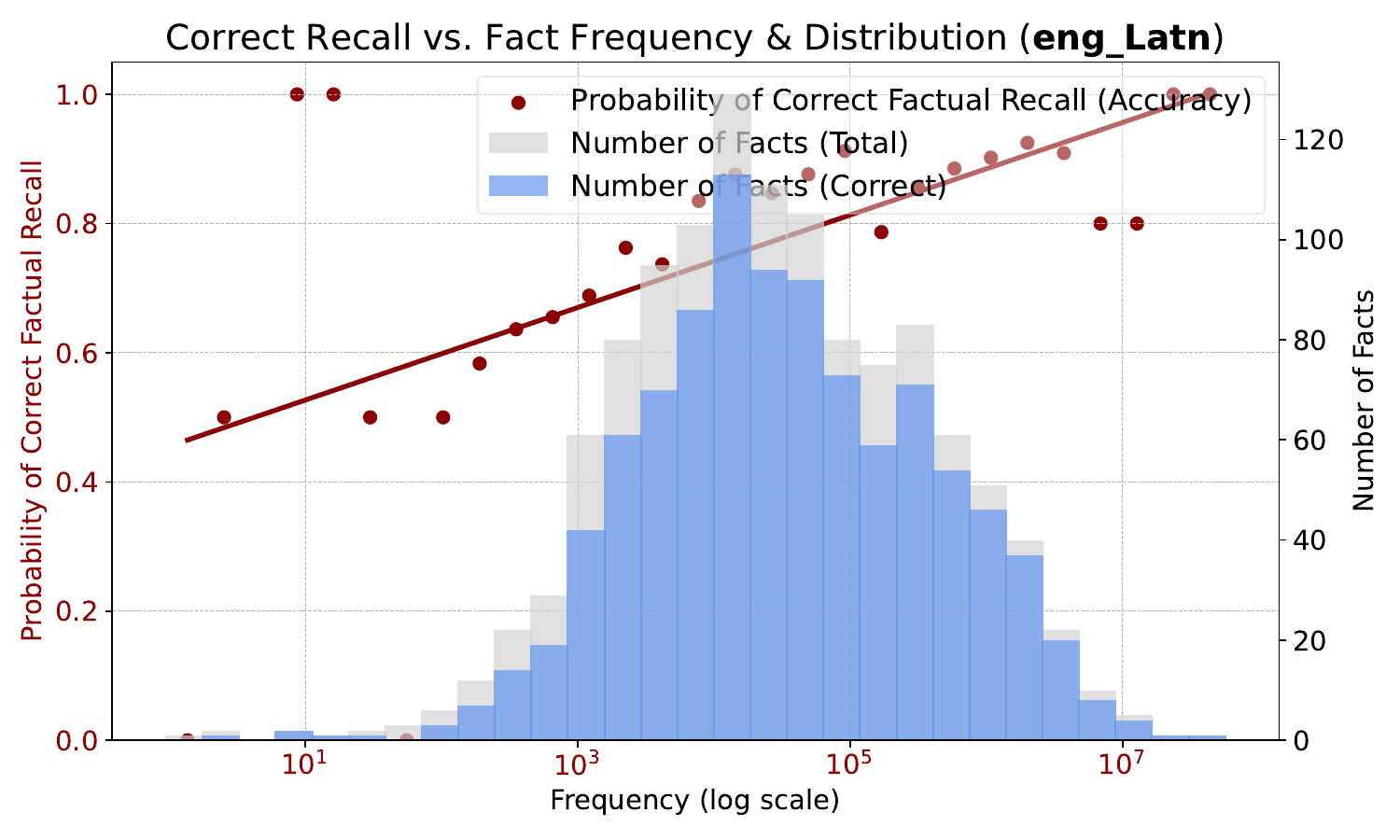}
    \includegraphics[width=0.24\textwidth]{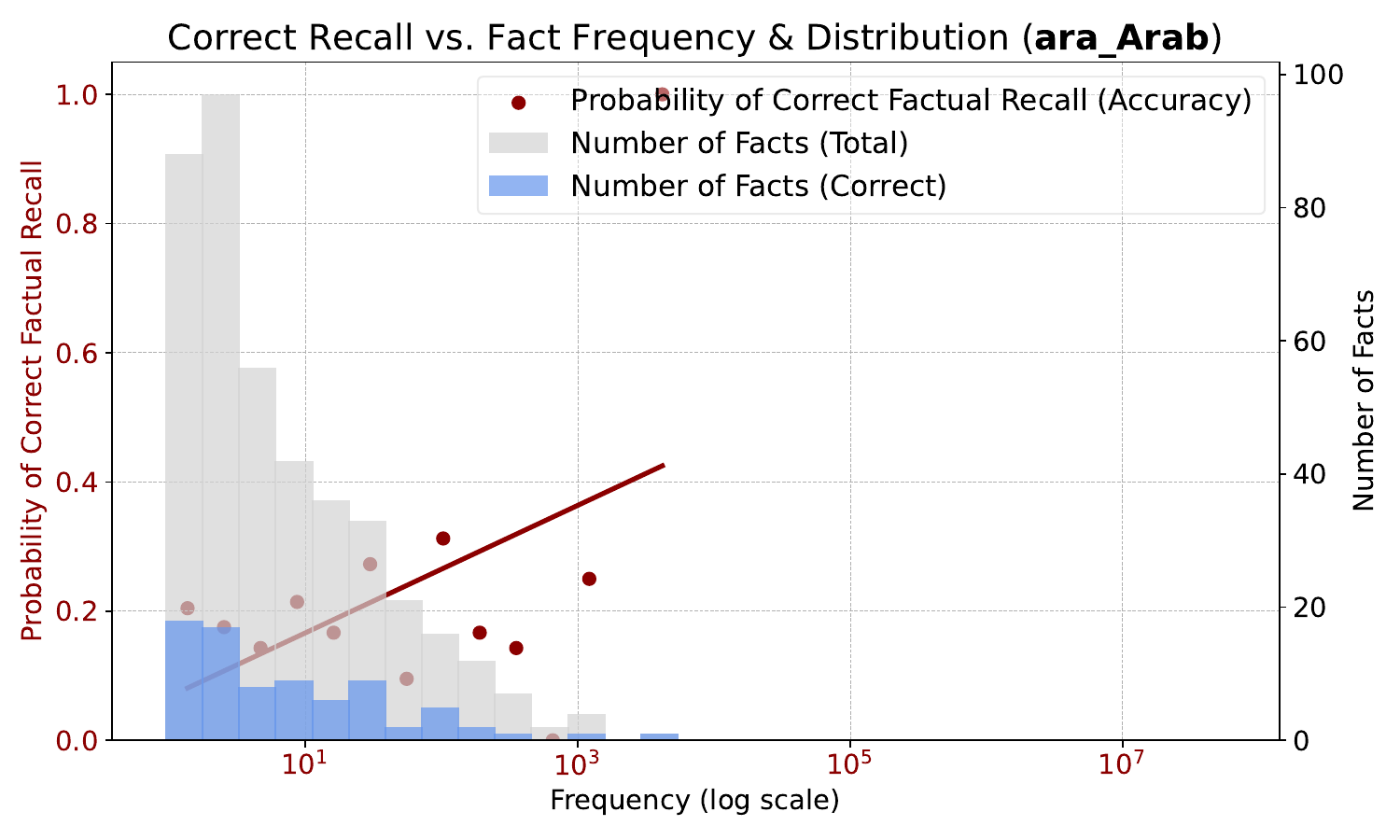}
    \includegraphics[width=0.24\textwidth]{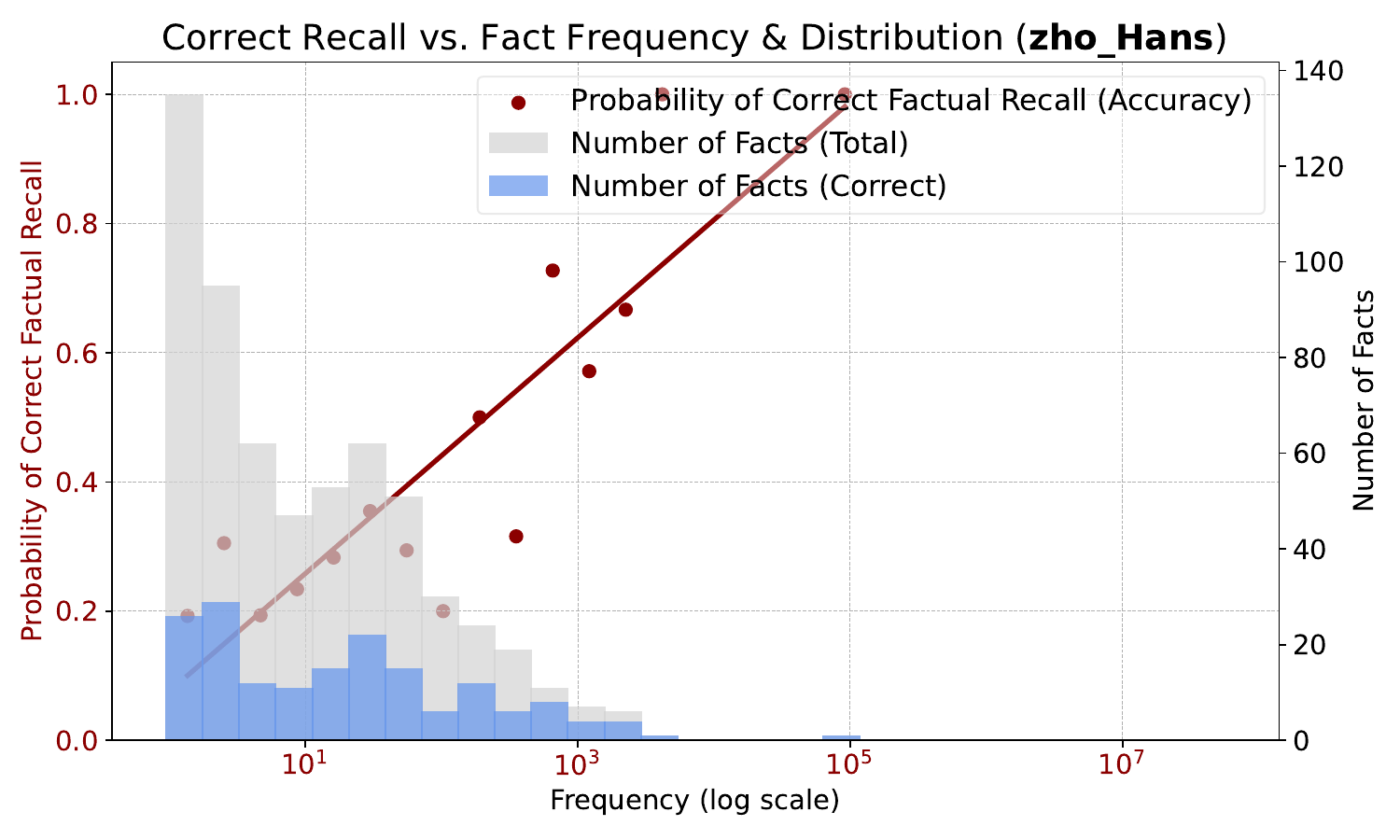}
    \includegraphics[width=0.24\textwidth]{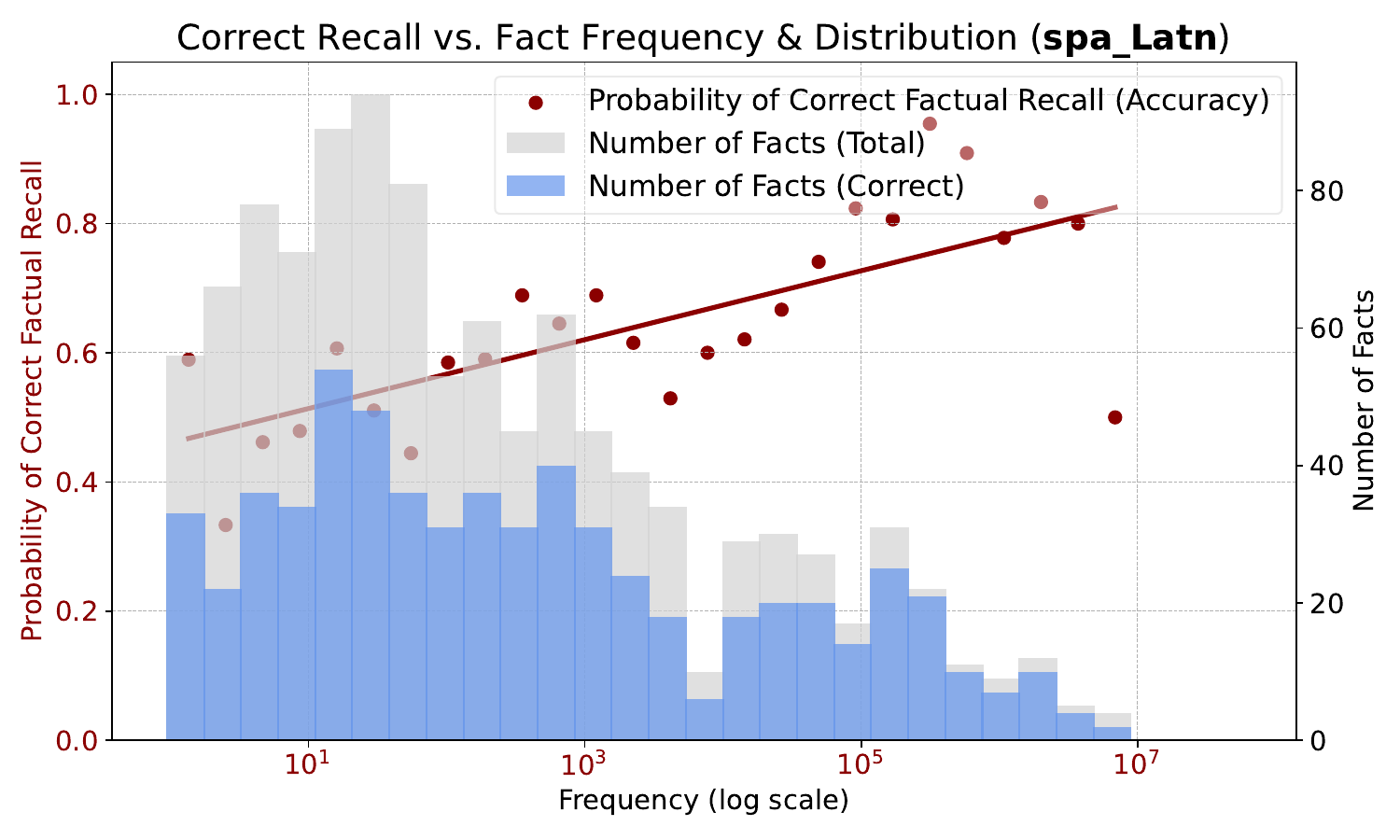}
    \includegraphics[width=0.24\textwidth]{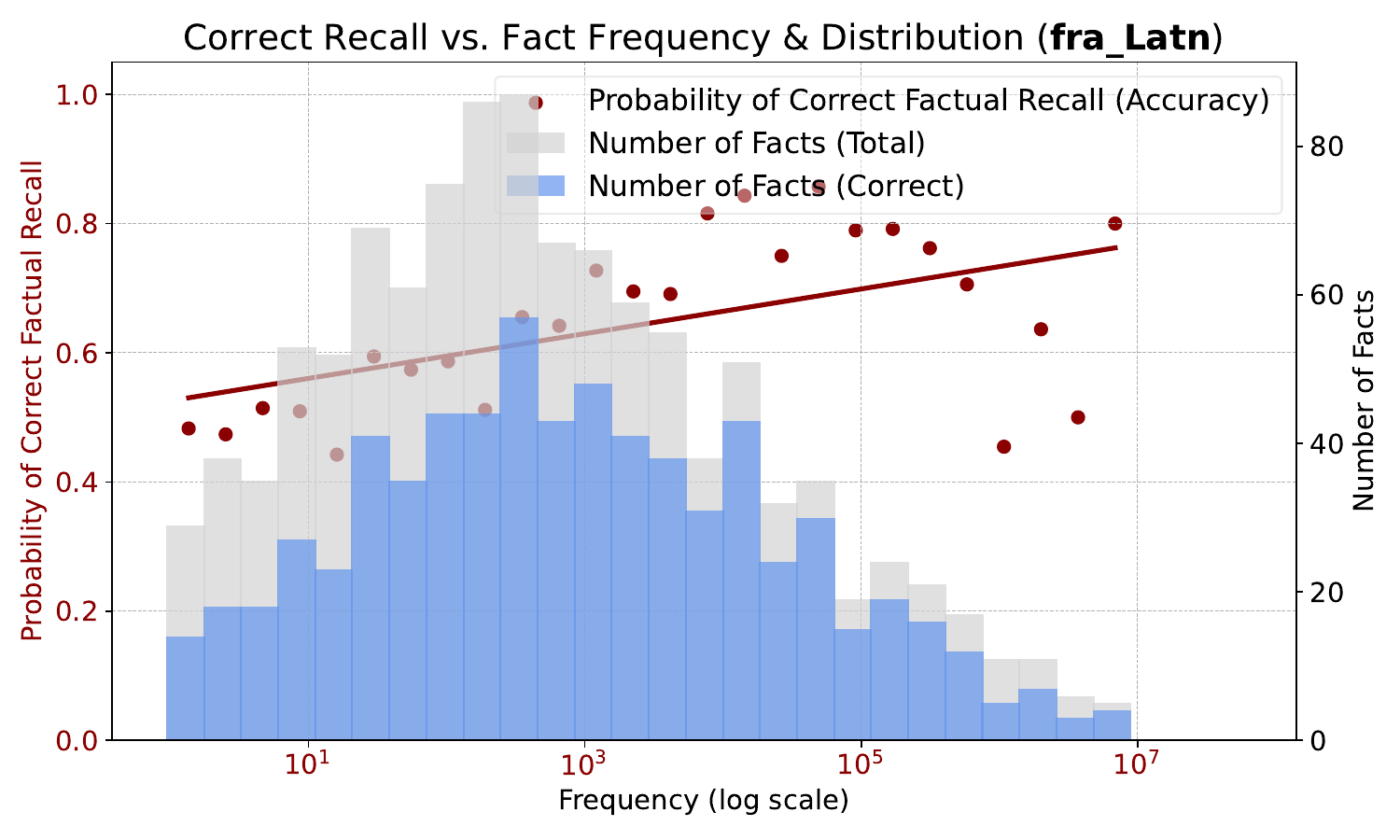}
    \includegraphics[width=0.24\textwidth]{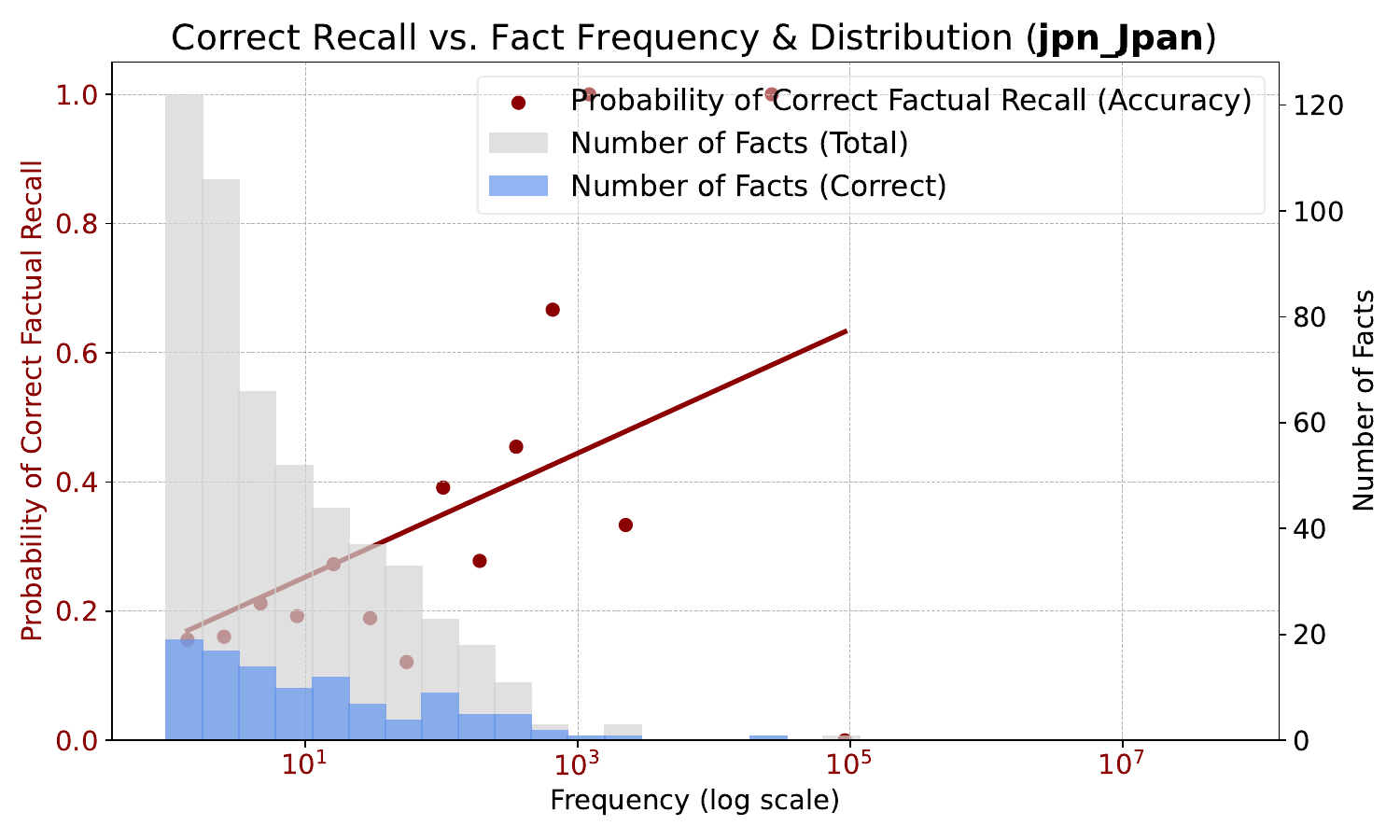}
    \includegraphics[width=0.24\textwidth]{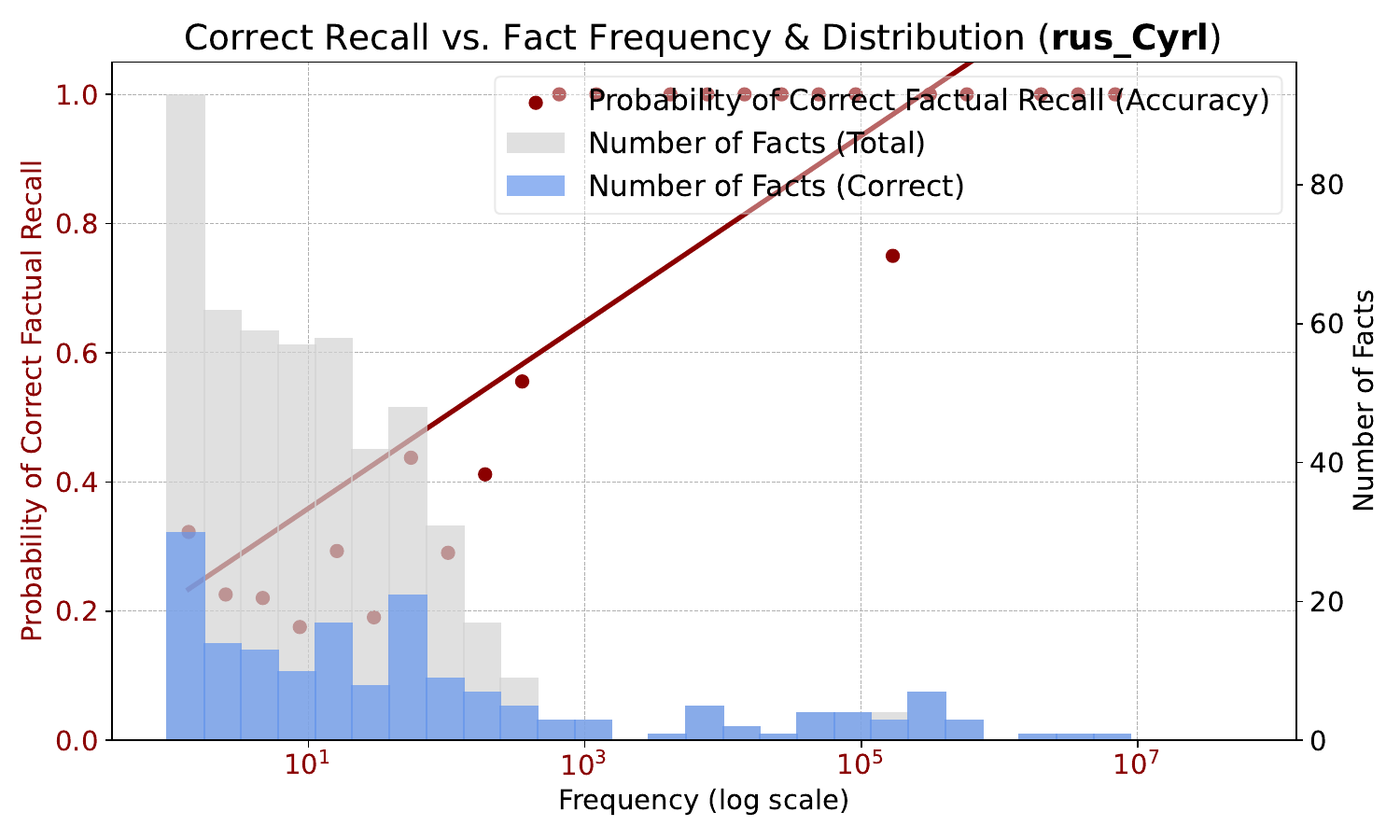}
    \includegraphics[width=0.24\textwidth]{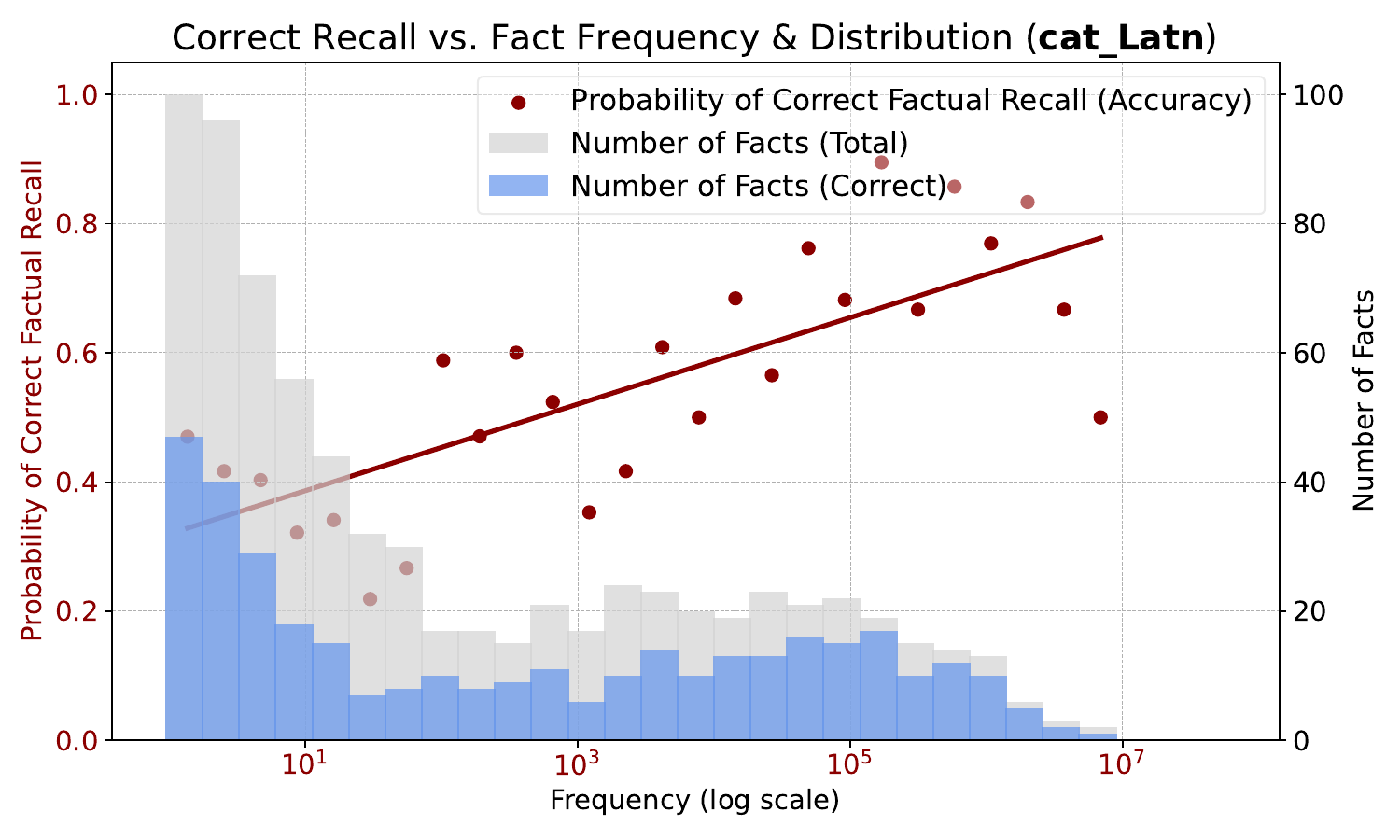}
\caption{Relationship between fact frequency and the probability of correct factual recall. A consistent upward trend across individual languages indicates that higher-frequency facts are more likely to be recalled by the model.}
    \label{fig:correctness_frequency_local}
\end{figure*}

\paragraph{Script plays a more important role than language family in sustained improvements.}
Languages such as ara\_Arab, jpn\_Jpan, and kor\_Kore, which neither use the Latin script nor belong to the Indo-European family, reach early saturation in performance -- typically even before 2K steps.
In contrast, Latin-script languages such as cat\_Latn, fra\_Latn, and spa\_Latn, continue to improve with more training steps.
Interestingly, ell\_Grek, despite being an Indo-European language, saturates early as well, whereas tur\_Latn, from the Turkic family, benefits from extended pretraining.
This pattern suggests that surface features like script similarity are more influential for possible crosslingual knowledge transfer than deeper typological relationships, as we further investigate in \secref{transferred_facts}.

\section{Fact Frequency As Predictor}\seclabel{frequency}

A notable observation in \secref{dynamics} is that factual
recall performance
(ACC)
rapidly converges for many languages, including English. 
This suggests that the model acquires much of its factual knowledge in the early stages of pretraining and is able to recall it reliably when appropriately prompted (cf.\ \secref{dataset}).
We hypothesize that this behavior reflects a form of memorization, where frequent exposure to specific facts in the pretraining corpus enables the model to retrieve them accurately.
To investigate this, we approximate the frequency of all facts in the KLAR dataset (cf.\ \secref{frequencies}) and analyze the relationship between frequency and factual recall performance both ``\textbf{globally}'' -- across all languages -- and ``\textbf{locally}'' -- within individual languages.

\subsection{Global Results Across All Languages}\seclabel{global_results}

\begin{figure}
    \centering
\includegraphics[width=0.45\textwidth]{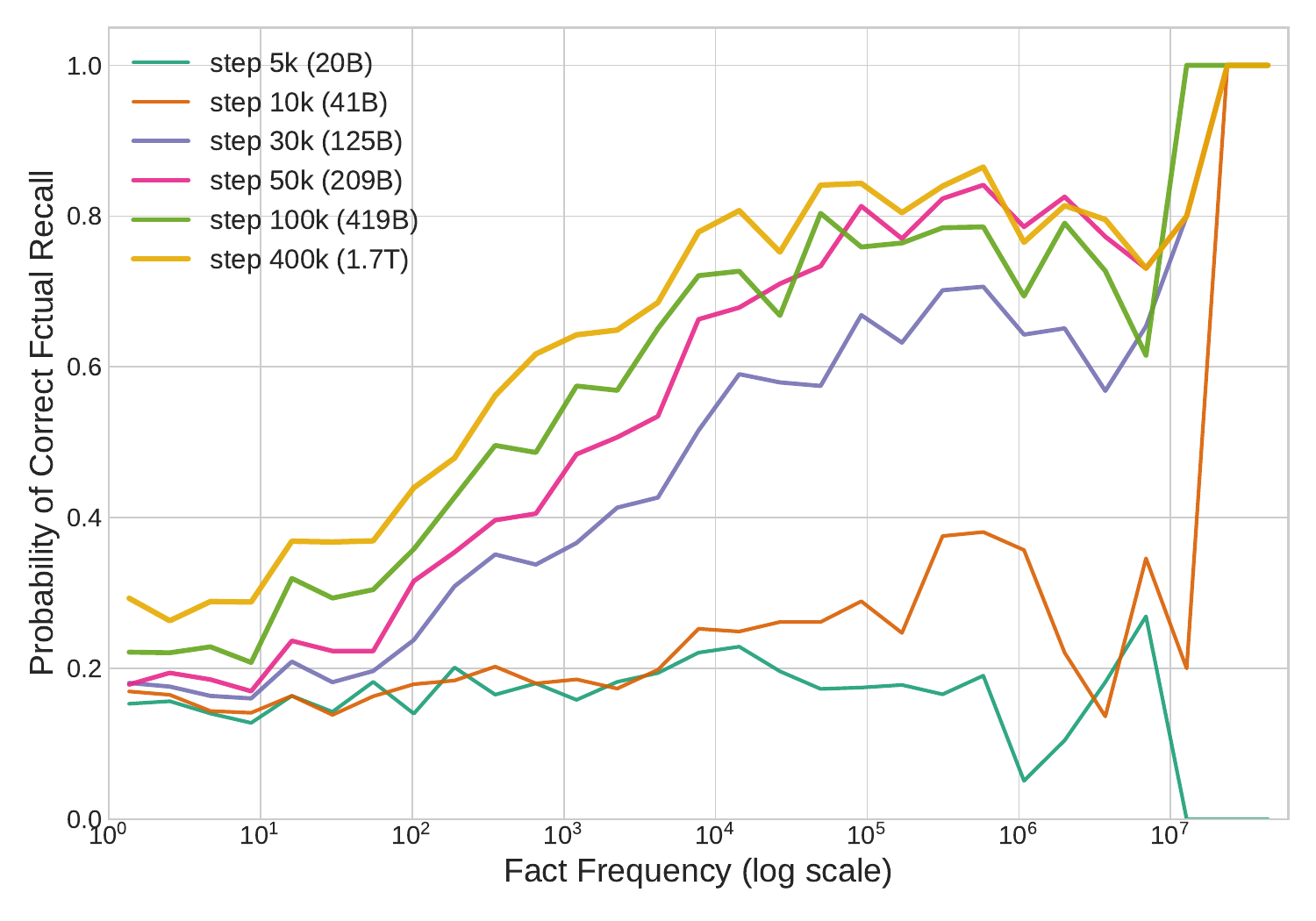}
    \caption{Relationship between fact frequency and factual
 recall for all languages and six
pretraining checkpoints. 
    High-frequency facts are more likely to be correctly recalled than rare ones. 
    This frequency-correctness correlation emerges early in pretraining and becomes more pronounced over time.
    }
    \label{fig:correctness_frequency_globle}
\end{figure}

We analyze the relationship between fact frequency (in log
scale) and probability of correct factual recall across six
OLMo checkpoints: 5K, 10K, 30K, 50K, 100K and 400K. % pretraining steps, respectively. 
The results are displayed in Figure~\ref{fig:correctness_frequency_globle}. 
Results for more checkpoints are reported in \secref{appendix:overall_result}.

\paragraph{Fact frequency strongly predicts factual recall performance.}
At the 400K-step checkpoint (corresponding to approximately 1.7T tokens), we observe a strong positive correlation between the fact log frequency and the probability of correct factual recall, with a Pearson correlation coefficient of $r = 0.93$ ($p<0.001$). 
This indicates a robust linear relationship between the two variables and supports our hypothesis that fact frequency in the pretraining corpus is a key determinant of factual recall performance across languages.

\paragraph{This correlation emerges early in pretraining.}
% Even at early stages of pretraining, a clear positive correlation is already observable. 
While the 5K-step and 10k-step checkpoints (around 20B and 41B tokens, respectively) show
weak correlation, the 30K-step checkpoint
(around 125B tokens) has Pearson coefficients $r = 0.95$, 
indicating strong correlation. 
Together with the high factual recall accuracy observed in early checkpoints (cf.\ Figure~\ref{fig:performance_over_checkpoints}), these results suggest that the model is exposed to and memorizes many high-frequency facts early in pretraining, enabling accurate recall even before large-scale exposure, aligned with findings from \citet{merullo2025linear}.

\subsection{Analysis per Language}\seclabel{local_results}

We further investigate whether the relationship between fact
frequency and factual recall accuracy holds consistently
across individual languages. We focus on the 400k-step checkpoint.

\paragraph{High-frequency facts are more likely to be correctly recalled within individual languages.}
Figure~\ref{fig:correctness_frequency_local} shows the distribution of fact frequencies and corresponding factual recall probabilities for 8 representative languages (results for additional languages are in \secref{appendix:per_language_result}).
Across all cases, we observe a clear trend: facts that occur more frequently in the pretraining corpus are more likely to be correctly recalled. 
This pattern is not limited to English; languages such as
rus\_Cyrl exhibit particularly strong effects -- for
instance, when fact frequency exceeds $10^3$, the model
recalls the fact with near-perfect accuracy.
Similar trends are observed in other languages as well, suggesting that fact frequency plays a consistently central role in determining factual recall performance across languages.

\subsection{Recall Prediction with Frequencies}\seclabel{prediction}

\begin{table}[t]
\centering
\small
\setlength{\tabcolsep}{10pt}
\begin{tabular}{lrrrrr}
\toprule
\textbf{Lang} & \textbf{Threshold} & \textbf{Accuracy} & \textbf{FN} \\
\midrule
ara\_Arab & 3485  & 0.83 & 209 \\
cat\_Latn & 2506  & 0.63 & 384 \\
ell\_Grek & 483   & 0.84 & 190 \\
eng\_Latn & 108   & 0.82 & 7   \\
fra\_Latn & 19    & 0.64 & 134 \\
jpn\_Jpan & 352   & 0.82 & 212 \\
kor\_Kore & 402   & 0.80 & 238 \\
rus\_Cyrl & 370   & 0.72 & 330 \\
spa\_Latn & 12    & 0.60 & 169 \\
tur\_Latn & 3068  & 0.64 & 373 \\
ukr\_Cyrl & 385   & 0.79 & 248 \\
zho\_Hans & 502   & 0.75 & 296 \\
\bottomrule
\end{tabular}
\caption{Best threshold, accuracy, and false negatives
when using fact frequency as a predictor of factual recall.
We interpret FN
as \textbf{s}urprising \textbf{c}orrect \textbf{l}ow-\textbf{f}requency \textbf{p}redictions
(\falsenegatives) -- predictions that are correct even
though the underlying fact frequency is low. 
Good accuracy
on assessing fact frequency as a
predictor for correct fact recall 
is achieved for most languages with this classifier as shown
in column ``Accuracy''.}
\label{tab:frequency_threshold}
\end{table}

% \begin{table}[t]
% % \setlength{\abovecaptionskip}{-0.01cm}
% \setlength{\belowcaptionskip}{-0.5cm}
% \centering
% \small
% \setlength{\tabcolsep}{5pt}
% \begin{tabular}{lrrrrr}
% \toprule
% \textbf{Lang} & \textbf{Threshold} & \textbf{Accuracy} & \textbf{FP} & \textbf{FN} \\
% \midrule
% ara\_Arab & 3485  & 0.83 & 0   & 209 \\
% cat\_Latn & 2506  & 0.63 & 64  & 384 \\
% ell\_Grek & 483   & 0.84 & 2   & 190 \\
% eng\_Latn & 108   & 0.82 & 205 & 7   \\
% fra\_Latn & 19    & 0.64 & 302 & 134 \\
% jpn\_Jpan & 352   & 0.82 & 6   & 212 \\
% kor\_Kore & 402   & 0.80 & 1   & 238 \\
% rus\_Cyrl & 370   & 0.72 & 2   & 330 \\
% spa\_Latn & 12    & 0.60 & 304 & 169 \\
% tur\_Latn & 3068  & 0.64 & 60  & 373 \\
% ukr\_Cyrl & 385   & 0.79 & 0   & 248 \\
% zho\_Hans & 502   & 0.75 & 7   & 296 \\
% \bottomrule
% \end{tabular}
% \caption{Best threshold, accuracy, and error breakdown (false positives and false negatives) when using fact frequency as a predictor of factual recall correctness.
% We interpret FN
% as \textbf{s}urprising \textbf{c}orrect \textbf{l}ow-\textbf{f}requency \textbf{p}redictions
% (\falsenegatives) -- predictions that are correct even
% though the underlying fact frequency is low. 
% Good accuracy
% on assessing fact frequency as a
% predictor for correct fact recall 
% is achieved for most languages with this classifier as shown
% in column ``Accuracy''.}
% \label{tab:frequency_threshold}
% \end{table}

We observed in \secref{local_results} that the relationship between fact frequency and factual recall holds consistently across individual languages. 
This naturally leads to a further question: \textbf{Can the recallability of a fact be reliably predicted solely based on its frequency within a given language?}
To answer this, we construct a simple frequency-based classifier for each language and evaluate its effectiveness.
Again, we focus on the 400k-step checkpoint.

Formally, for each language $l$, we define a dataset $\mathcal{D}_l = \{ (f_i^l, y_i^l) \}_{i=1}^{N}$, where $f_i^l \in \mathbb{Z}_{\geq 0}$ is the frequency of fact $i$, and $y_i^l \in \{0,1\}$ indicates whether the model correctly recalled the fact ($1$ if correct, $0$ otherwise).
$\mathbb{Z}_{\geq 0}$ is the set of positive integers
including 0.
We then define a threshold-based classifier  $h_t^l(f)$ for
each language as:
$
h_t^l(f) = 
\begin{cases}
1, & \text{if } f \geq t \\
0, & \text{otherwise}
\end{cases}
$.
The optimal threshold $t^*_l$ in each language is selected to maximize classification accuracy:
$$
t^*_l = \arg\max_{t \in \mathbb{Z}_{\geq 0}} \frac{1}{N} \sum_{i=1}^{N} \mathbf{1} \left( h_t^l(f_i^l) = y_i^l \right)
$$
where $\mathbf{1}(\cdot)$ is the indicator function. 
To better understand the classification behavior, we also compute the number of false negatives (FN) under the optimal threshold, as these facts are also correctly predicted but with low frequencies.\footnote{
Other error types are not the primary focus of our further analysis presented in the main content. 
For example, the cause of false positives may be due to (1) insufficient exposure to the fact despite its high frequency, or (2) sensitivity to the specific prompt used for evaluation. We present an analysis of the classifier in \secref{sensitivity} and a complete error breakdown in \secref{exclude}.}
% To better understand the classification behavior, we also compute the number of false positives (FP) and false negatives (FN) under the optimal threshold.
% $$
% \text{FP}_l = \sum_{i=1}^{N} \mathbf{1}(h_{t^*_\ell}(f_i^l) = 1 \land y_i^l = 0)
% $$
% $$
% \text{FN}_l = \sum_{i=1}^{N} \mathbf{1}(h_{t^*_\ell}(f_i^l) = 0 \land y_i^l = 1)
% $$
Table~\ref{tab:frequency_threshold} presents the classification performance.

\begin{figure*}
    \centering
    \includegraphics[width=0.24\textwidth]{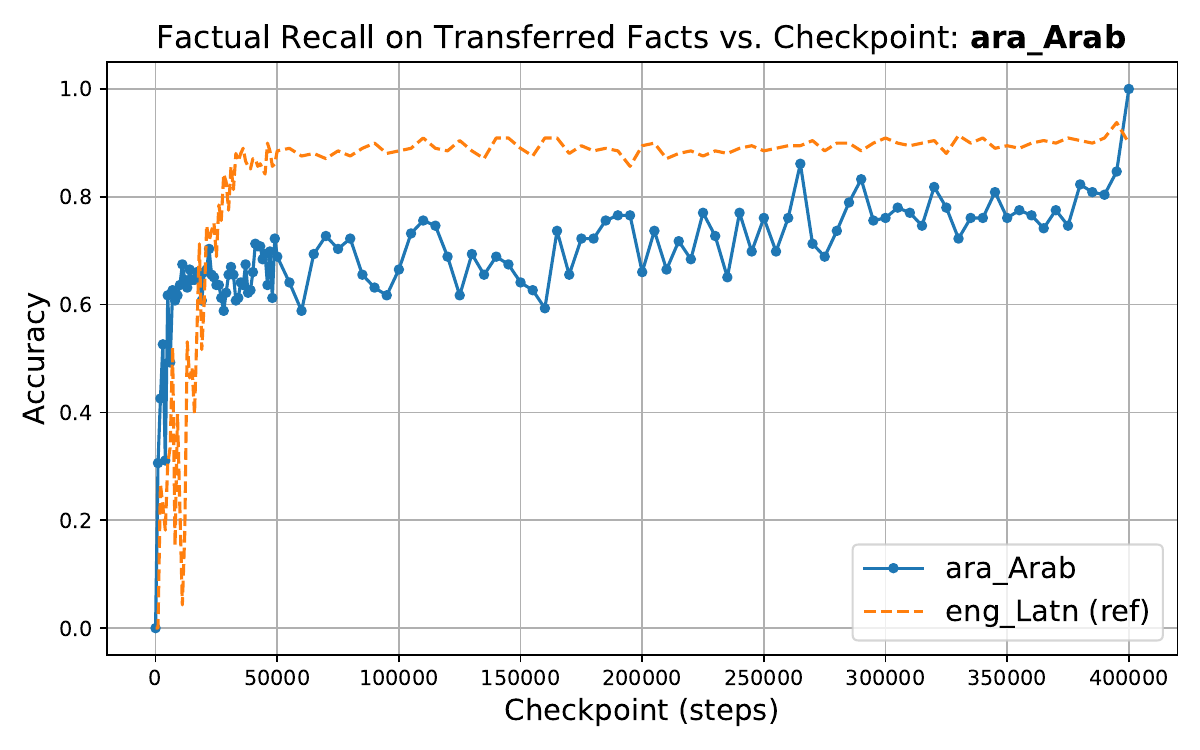}
    \includegraphics[width=0.24\textwidth]{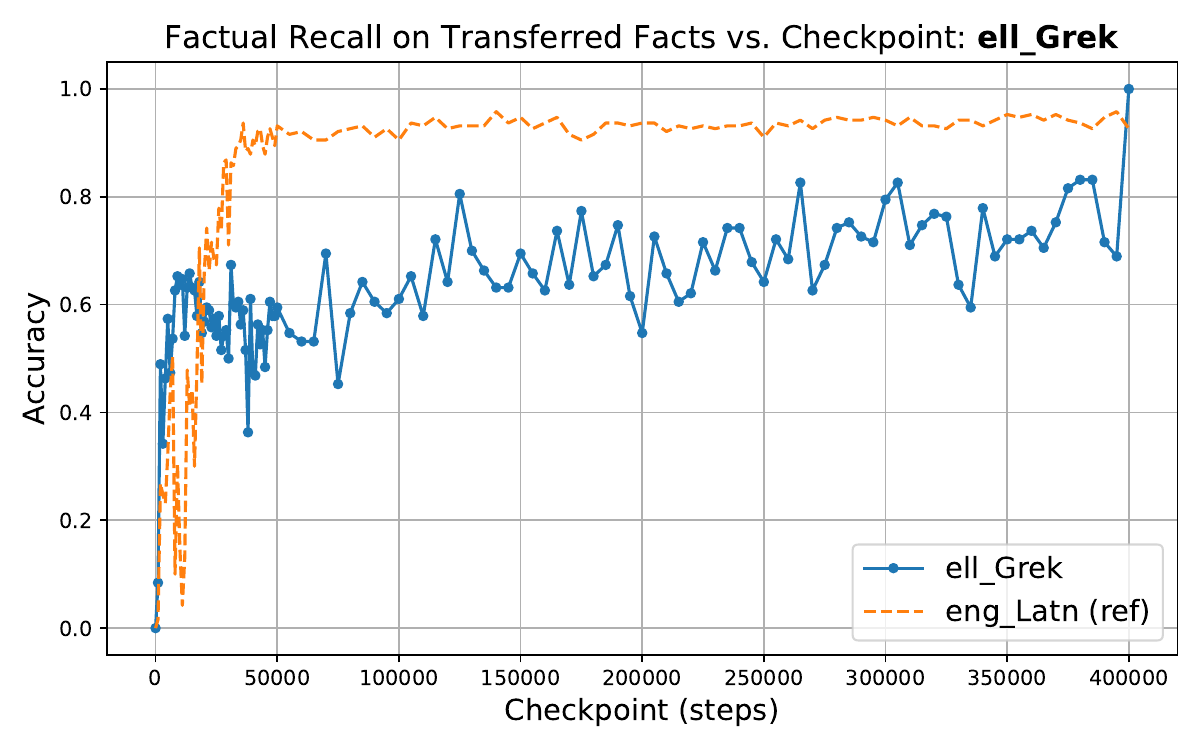}
    \includegraphics[width=0.24\textwidth]{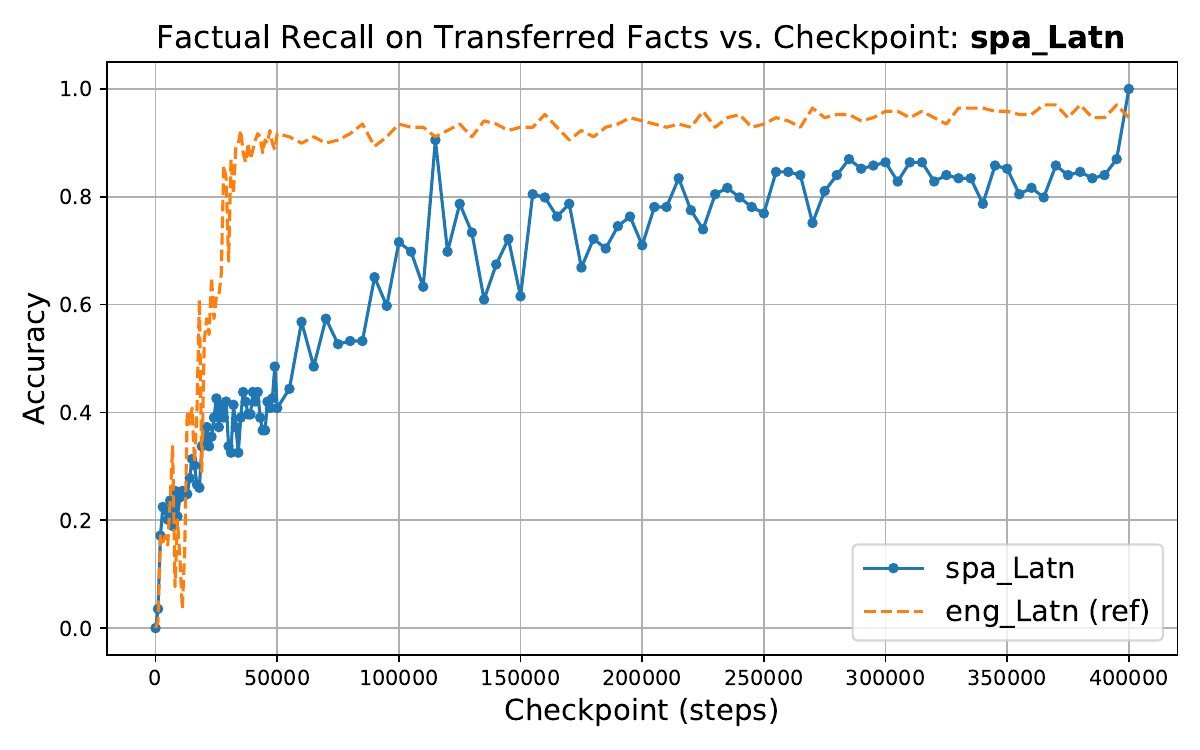}
    \includegraphics[width=0.24\textwidth]{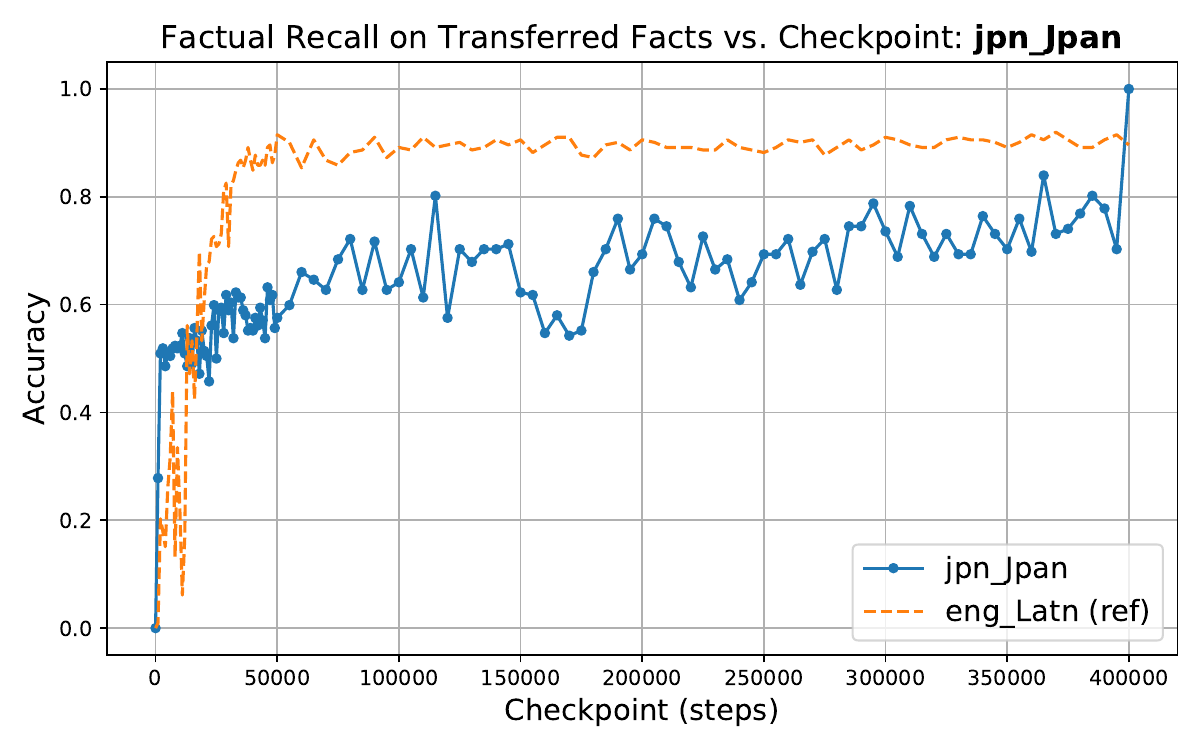}
    \includegraphics[width=0.24\textwidth]{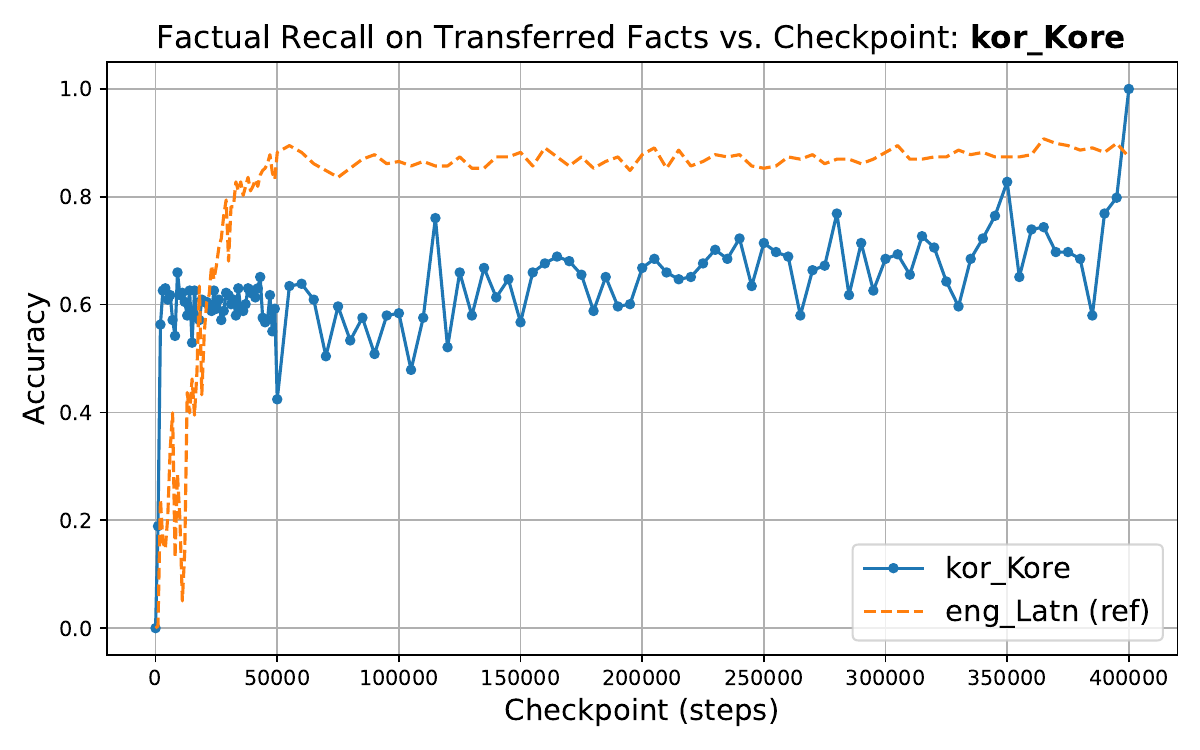}
    \includegraphics[width=0.24\textwidth]{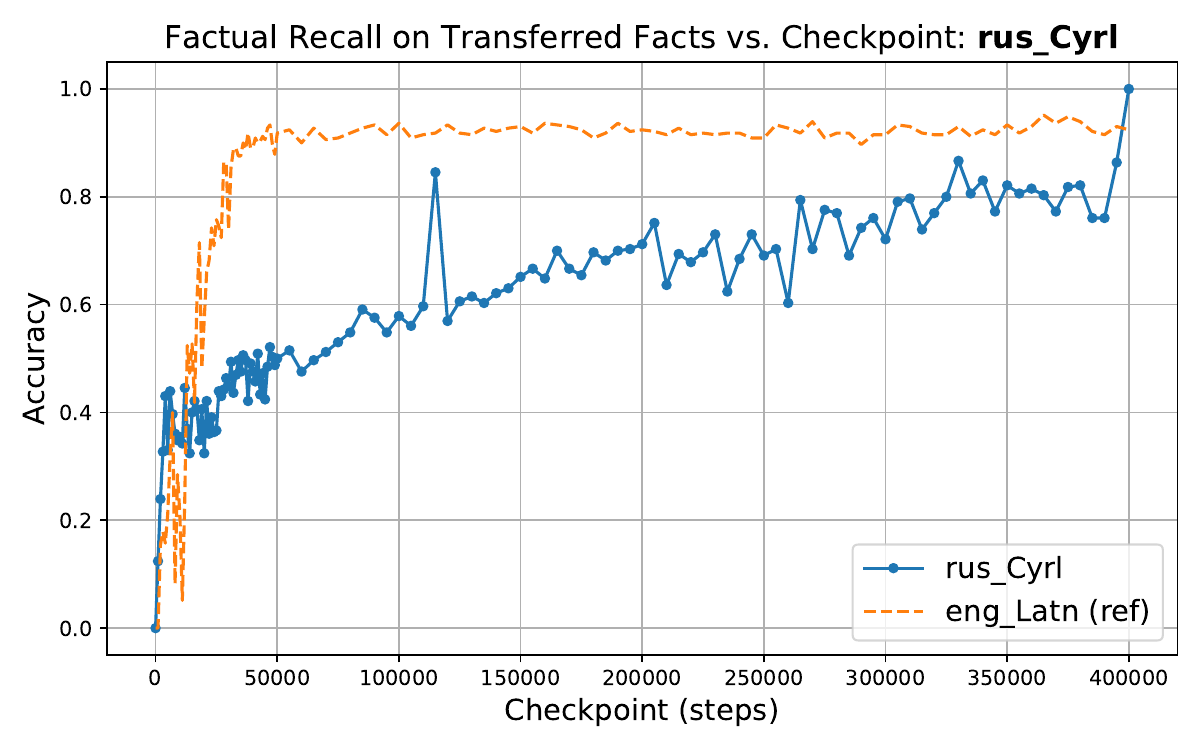}
    \includegraphics[width=0.24\textwidth]{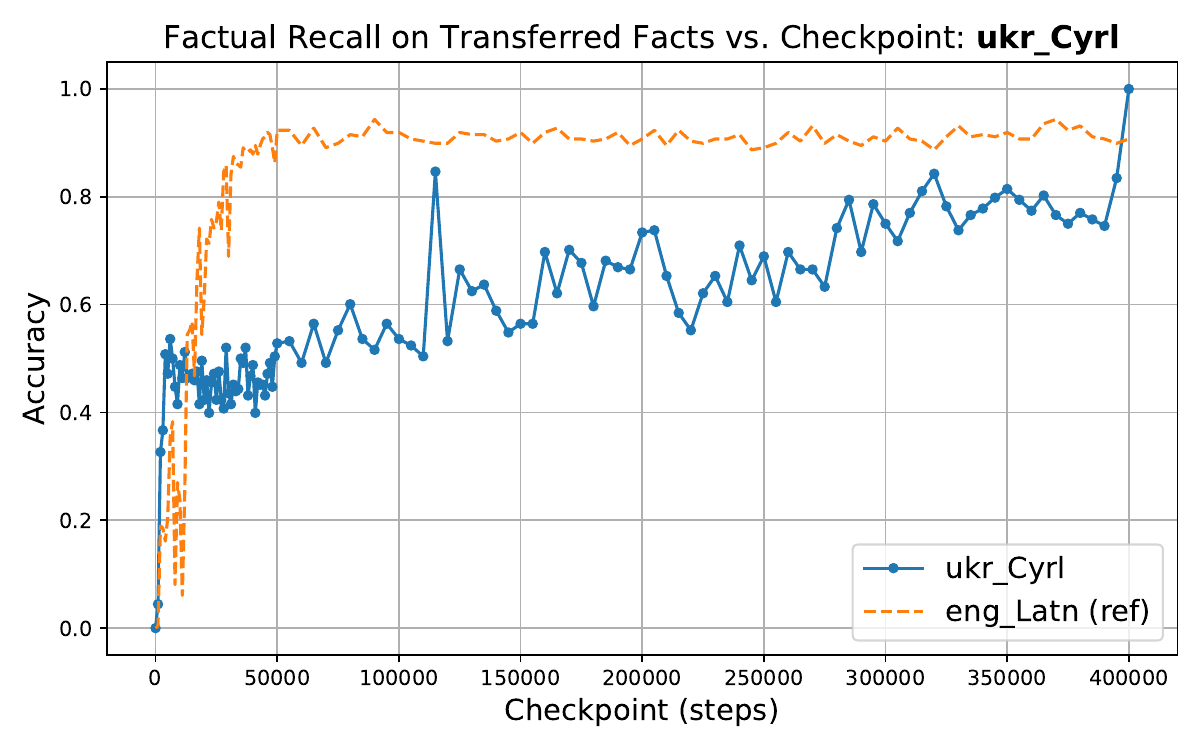}
    \includegraphics[width=0.24\textwidth]{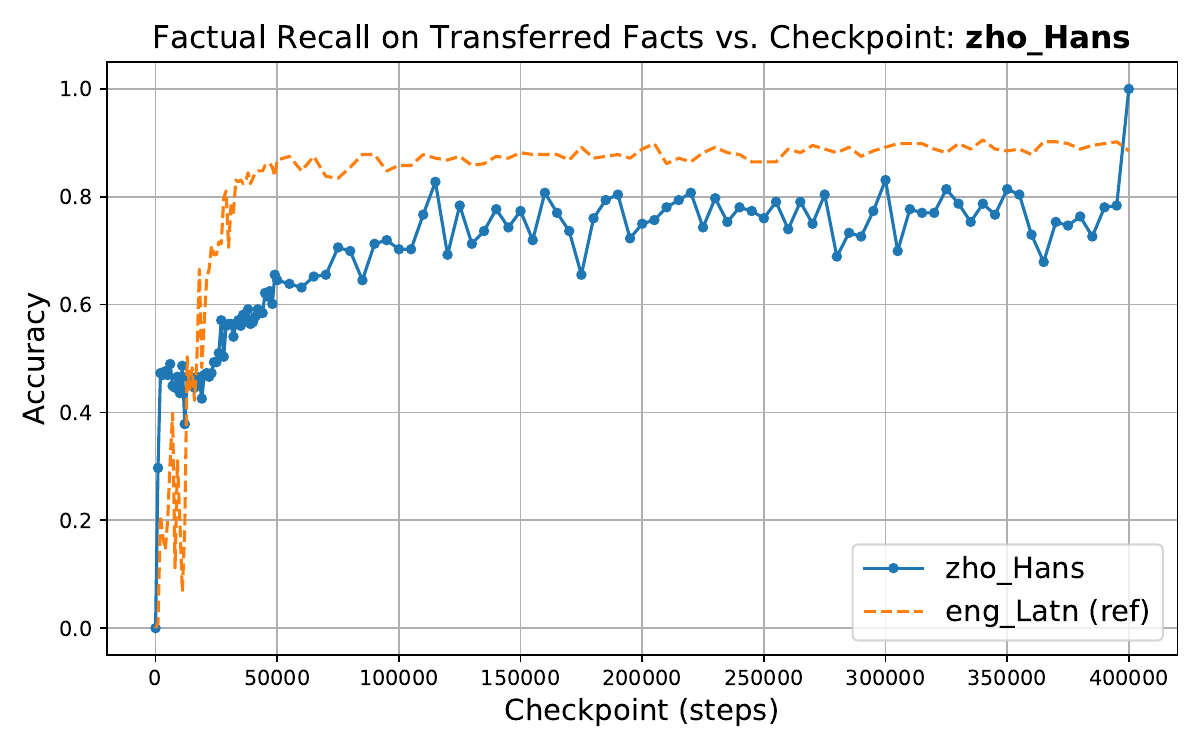}
    \caption{Dynamics of learning for \falsenegatives{}s
(surprisingly correct low frequency predictions, i.e., FNs in Table \protect\ref{tab:frequency_threshold})
across 8 languages.
Crosslingual transfer emerges early in pretraining and continues to strengthen over time.
}
    \label{fig:transferred_facts_over_checkpoints}
\end{figure*}

\paragraph{Fact frequency serves as a strong predictor of factual recall for many languages.}

Across all languages, the threshold-based classifier achieves accuracy above 0.6, indicating performance much better than random guessing.
A closer inspection reveals that all languages with relatively lower accuracy, i.e., fra\_Latn, spa\_Latn, tur\_Latn, and cat\_Latn, use the Latin script, with no exceptions.
In contrast, languages using non-Latin scripts consistently achieve higher accuracy.\footnote{We conduct a sensitivity analysis on the classifier in \secref{sensitivity} and show it is more robust in non-Latin-script languages.} 
We hypothesize that this pattern stems from extensive crosslingual transfer from English to other Latin-script languages.
As a result, many low- or mid-frequency facts in these languages may still be correctly recalled, likely due to shared vocabulary and lexical overlap, as also shown by \citet{qi-etal-2023-cross}. 
This transfer effect tends to shift the optimal classification threshold downward, enabling the threshold-based classifier to correctly predict low-frequency facts more often than expected.

\paragraph{All languages but English exhibit large false negative rates.}
This is particularly clear in languages using non-Latin scripts, such as ara\_Arab and ukr\_Cyrl, where the classifier fails to capture many low-frequency facts that are in fact recalled correctly by the model.
Even in Latin-script languages -- where the accuracy is relatively lower than in other languages due to the reasons noted above -- we still observe a substantial number of false negatives.
English stands out as the only language with 
% a large number of false positives but relatively 
few false negatives, because of the generally high fact frequencies.
This consistent trend across languages suggests that many low-frequency facts are correctly recalled, motivating a closer examination of such cases. 
We further investigate them in \secref{transferred_facts}.

\section{Investigation of Transfer Effect}\seclabel{transferred_facts}

We observed a substantial number of false negatives when using frequency as a predictor in \secref{prediction}, particularly for languages that 
use non-Latin scripts.
This is counterintuitive given the strong role frequency typically plays in factual recall.
We hypothesize that these 
% false negative 
cases are due to the \textbf{crosslingual transfer} effect -- factual knowledge is primarily learned in English and is successfully transferred to other languages.
In the following sections, we present a detailed analysis of
these false negatives identified in \secref{prediction} --
which we will refer to
as \textbf{s}urprisingly \textbf{c}orrect \textbf{l}ow-\textbf{f}requency \textbf{p}redictions
(\falsenegatives{}s).

\begin{figure}
    \centering
    \includegraphics[width=0.45\textwidth]{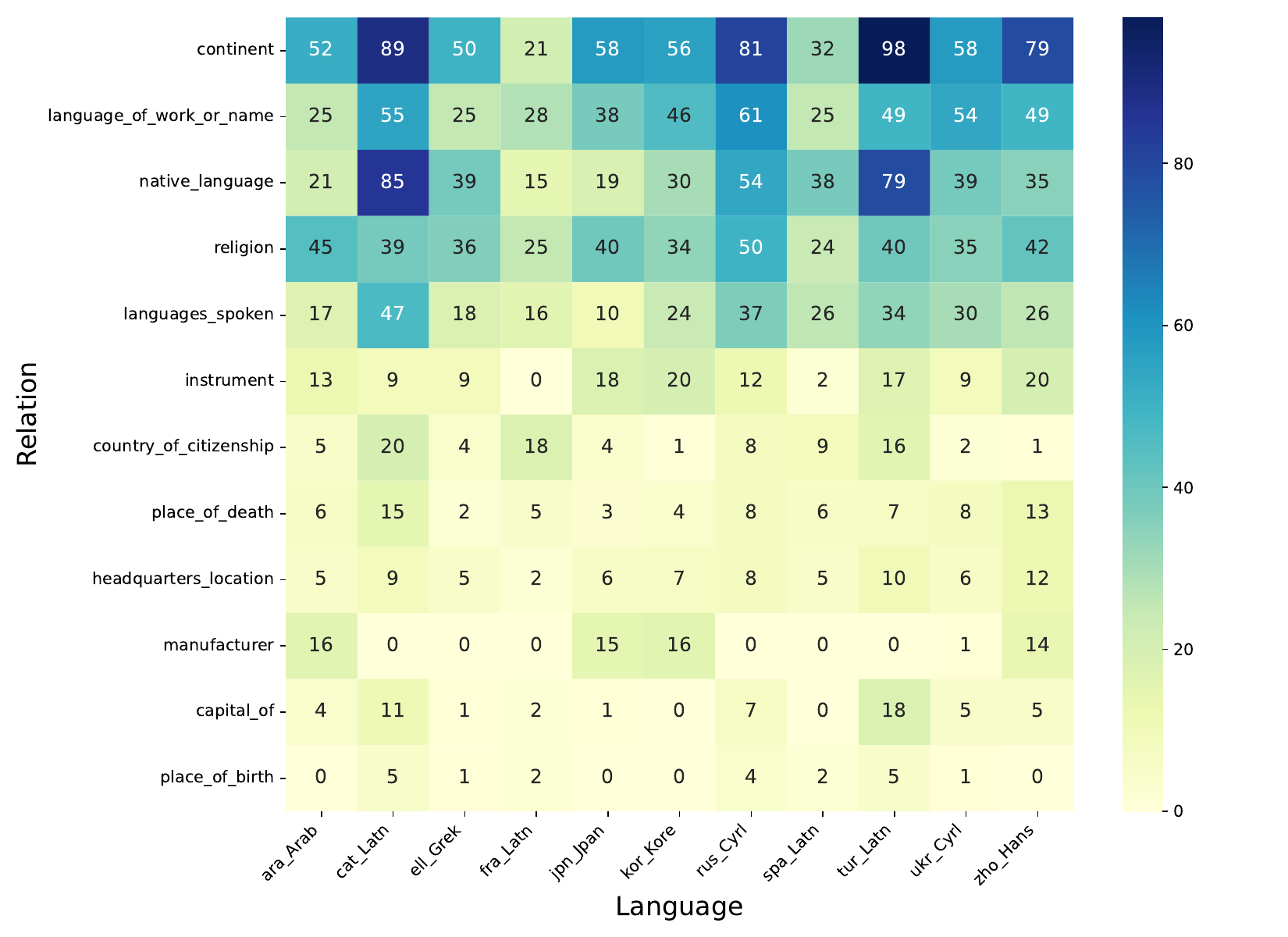}
    \caption{Distribution of  \falsenegatives{}s
(surprisingly correct low frequency predictions)
across relation types for each language.
   High \falsenegatives values are concentrated on relation
    types that involve only a small set of candidates -- which are generally named entities.
    }
    \label{fig:transferred_fact_relation}
\end{figure}

\subsection{Relation Type Distribution}\seclabel{relation_distribution}

We hypothesize that facts
that
involve named entities or shared vocabulary
are easier to transfer across languages -- e.g., the
subject-object pair France-Paris is easy to transfer from
English to French since the two named entities are identical
in English and French.
This intuition is grounded in how humans often rely on lexical similarity and recognizable entities when transferring knowledge.
To investigate this, we group \falsenegatives{}s in each language by their relation type, as shown in Figure~\ref{fig:transferred_fact_relation}.

\paragraph{\falsenegatives{}s are concentrated in relation types involving named entities.}
This trend is especially pronounced in relations with a limited set of possible candidates, such as \texttt{continent} and \texttt{religion}.
Languages that use a non-Latin script also benefit from named entity transfer, e.g., in \texttt{instrument} and \texttt{manufacturer} relations.
This observation aligns with prior work showing that named entities are more easily transferred across script boundaries, particularly in encoder-only models \citep{imanigooghari-etal-2023-glot500,liu-etal-2024-ofa}.

\paragraph{Latin-script languages benefit more broadly from crosslingual transfer.}
Compared to languages using other scripts, languages written in Latin script receive transfer benefits across a wider range of relations, such as \texttt{country\_of\_citizenship}.
This is expected, as many Latin-script languages have substantial vocabulary overlap, leading to greater token-level similarity.
Such overlap enables the transfer of identical or lexically similar entities -- e.g., ``Bulgària'' in cat\_Latn and ``Bulgaristan'' in tur\_Latn.
Moreover, higher token-level similarity in the context during pretraining can also facilitate the alignment, enhancing entity transfer (cf.\ \secref{similarity}).

\subsection{Learning Progression}

As shown in \secref{dynamics}, the model acquires a substantial amount of factual knowledge during the early stages of pretraining.
This raises a natural question: \textbf{Is crosslingual knowledge transfer similarly concentrated in the early stages, or does it continue throughout pretraining?}
To explore this, we examine the learning trajectories of \falsenegatives{}s across languages.
Figure~\ref{fig:transferred_facts_over_checkpoints} illustrates how recall factual accuracy for \falsenegatives{}s evolves over pretraining checkpoints for 8 languages (see full results in \secref{complete_fns_dynamics}).

\paragraph{Extensive crosslingual transfer occurs during early pretraining.}
Across all languages, factual recall accuracy for \falsenegatives{}s rapidly improves during the initial stages of pretraining.
This trend is especially pronounced in languages that use non-Latin scripts.
For example, ara\_Arab, ell\_Grek, and Kor\_Kore reach over 60\% accuracy within the first 20K steps, after which their performance plateaus or grows slowly, similar to the trend observed for in \secref{dynamics}.
% These findings suggest that crosslingual transfer already happen in the earlier stages.
These findings suggest that crosslingual transfer is not merely an emergent property of the final model, but rather a phenomenon that develops early in pretraining.

\paragraph{Many languages continue to benefit from transfer throughout pretraining.}
This is especially the case for languages using the Latin script, such as spa\_Latn, which display a more gradual and continuous improvement.
As discussed in \secref{relation_distribution}, these languages benefit from crosslingual transfer across a broader range of relations, facilitated by extensive lexical overlap with other Latin-script languages.
This broader scope of transferable content contributes to the prolonged learning curve.
We also observe that rus\_Cyrl and zho\_Hans benefit from continued improvements over time, which could be attributed to the comparatively larger representation of Russian and Chinese texts in the pretraining corpus (cf.\ \secref{dolma}).
Notably, ukr\_Cyrl exhibits a learning curve that rapidly and closely aligns with rus\_Cyrl, suggesting that transfer also occurs between other script-sharing languages (we show their consistency continues to improve in \secref{holistic}).

\subsection{Similarity Dynamics}\seclabel{similarity}

\begin{figure}
    \centering
    \includegraphics[width=0.23\textwidth]{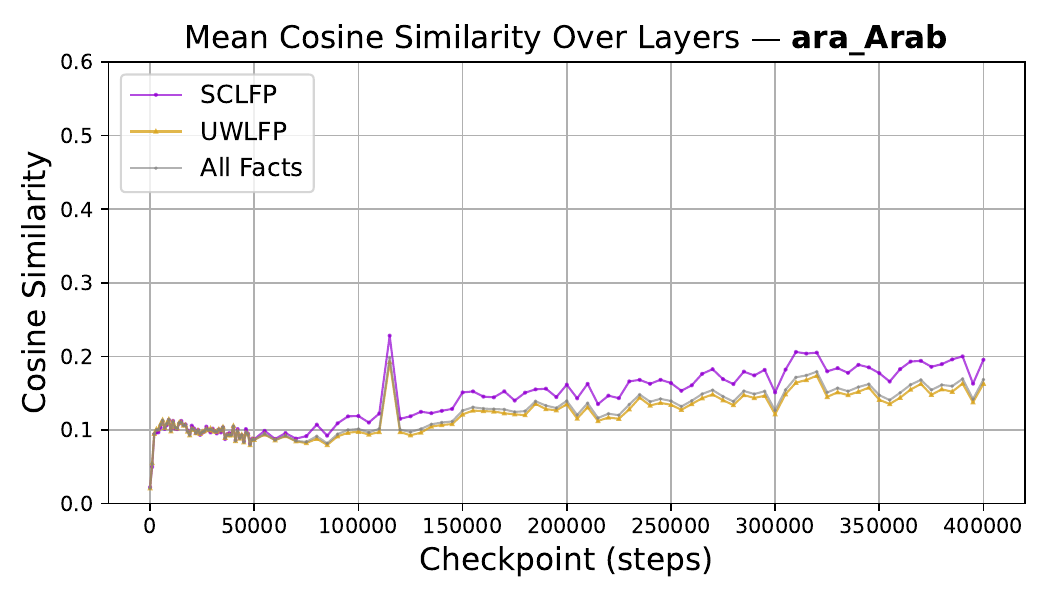}
    \includegraphics[width=0.23\textwidth]{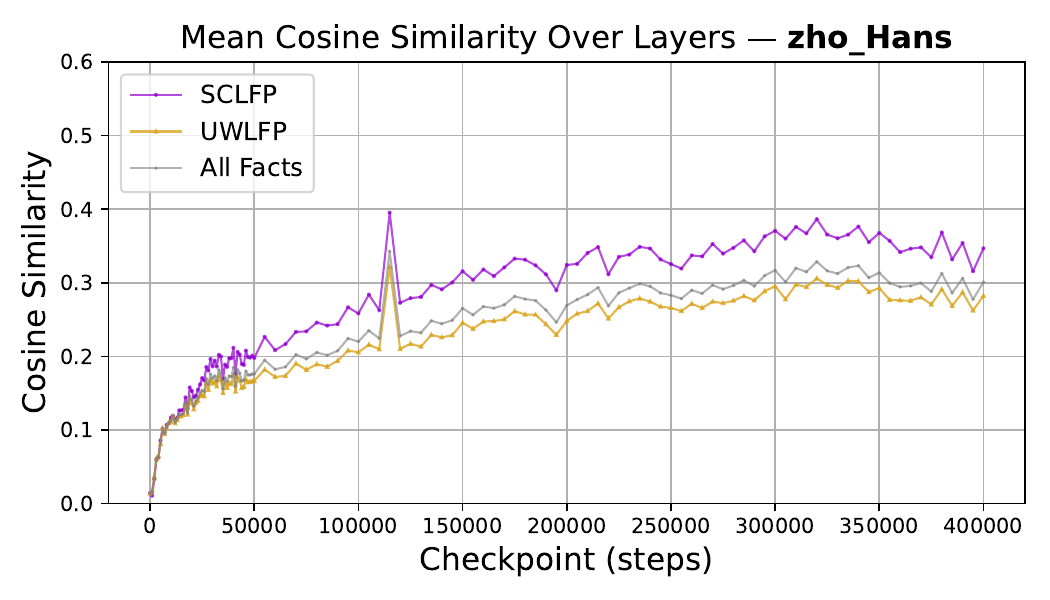}
    \includegraphics[width=0.23\textwidth]{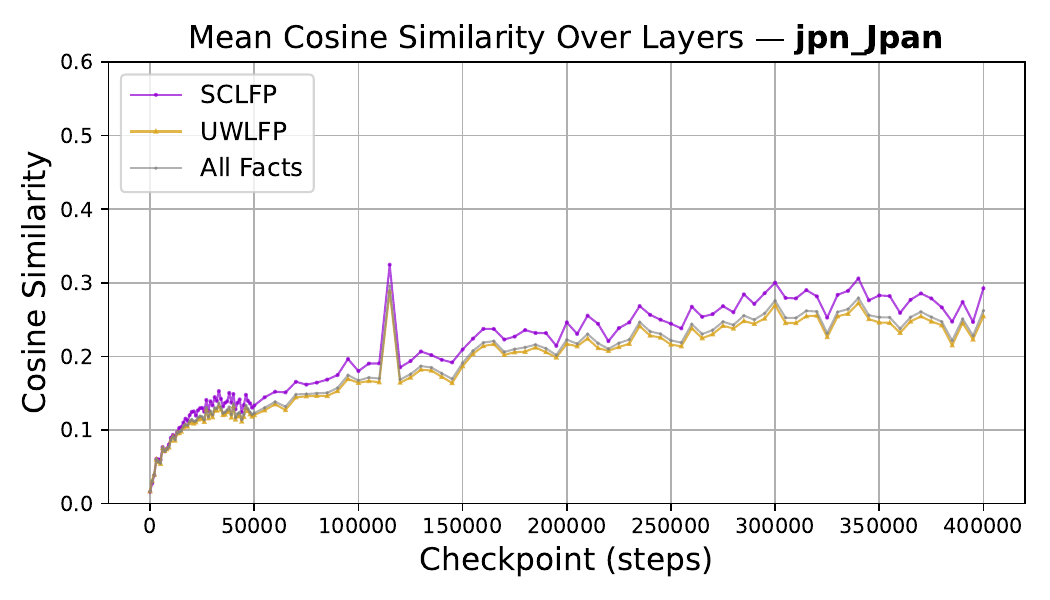}
    \includegraphics[width=0.23\textwidth]{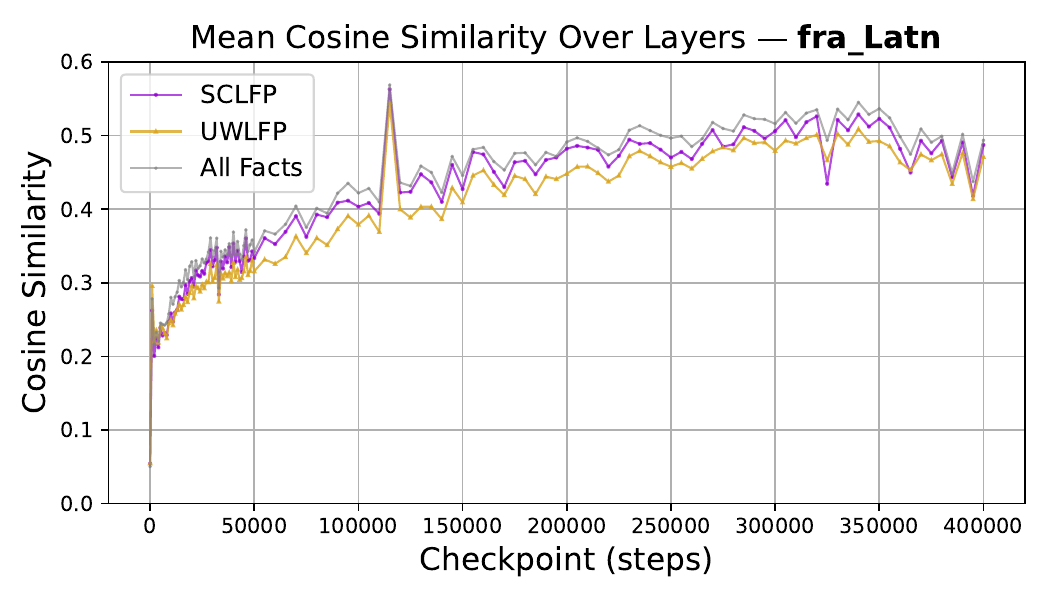}
    \includegraphics[width=0.23\textwidth]{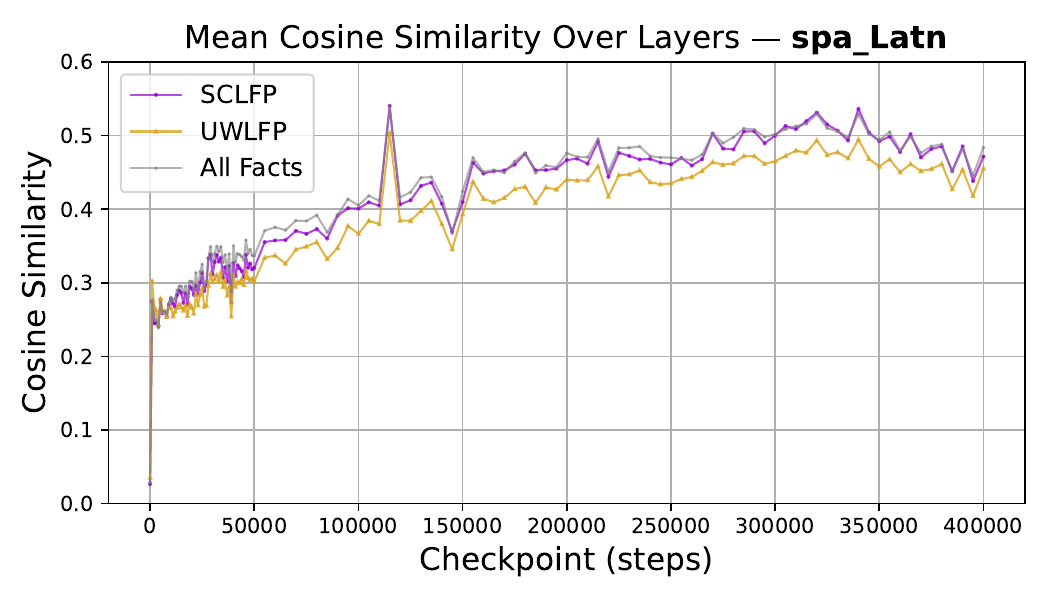}
    \includegraphics[width=0.23\textwidth]{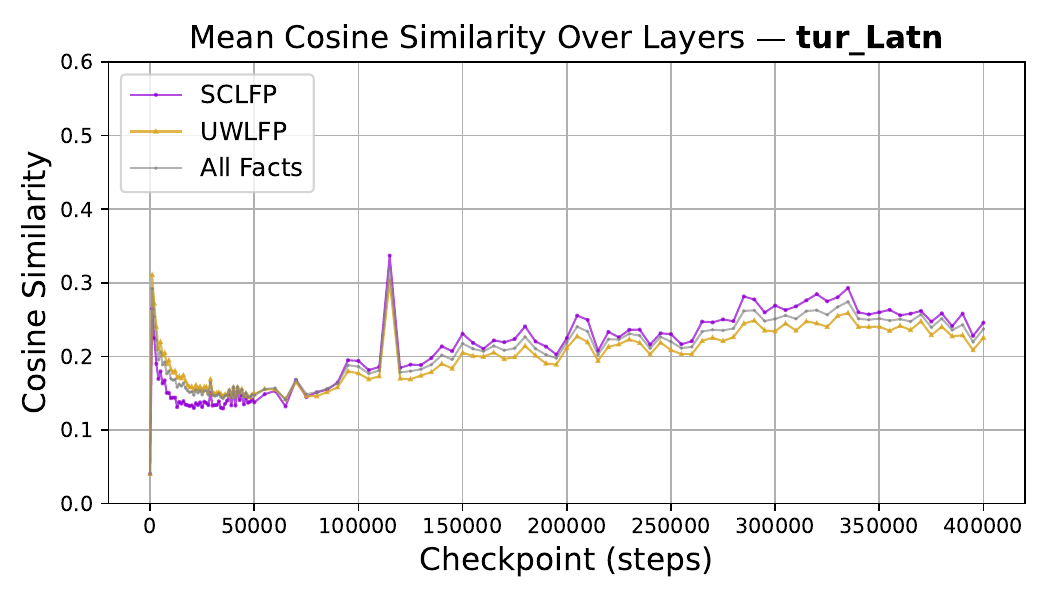}
    \caption{Mean cosine similarity between sentence-level representations of \falsenegatives, \truenegatives, and all facts for each language paired with English during pretraining. 
    All 6 languages exhibit consistently higher similarity for \falsenegatives than for \truenegatives, highlighting the emergence of crosslingual transfer through representation alignment.}
    \label{fig:similarity_over_checkpoints}
\end{figure}

To better understand why certain languages, particularly those that do not use the Latin script, benefit from knowledge acquired in English, we analyze the evolution of cosine similarity between sentence-level representations of prompts (cf.~\secref{dataset}) or fact pairs corresponding to \falsenegatives{}s during pretraining.
Specifically, we create fact pairs of \falsenegatives{}s for each language, where every pair contains one prompt in that language and its counterpart in English.
We then track the cosine similarity between these paired representations across checkpoints.\footnote{We use the contextualized embedding of the final token as the sentence-level representation. Representations are extracted at each layer, and we report the mean cosine similarity computed by averaging similarities across all layers.}
As a baseline, we also compute cosine similarities for \truenegatives{}s -- \textbf{u}nsurprisingly \textbf{w}rong \textbf{l}ow-\textbf{f}requency \textbf{p}redictions identified in our frequency-based classification (cf.~\secref{local_results}) -- as well as for \textbf{all fact} pairs in each language.
Figure~\ref{fig:similarity_over_checkpoints} illustrates the progression of similarity scores over time for 6 languages (full results are available in \secref{appendix:similarity}).\footnote{
To avoid inflated similarity, for each language, we filter out fact pairs where the object strings in that language and English are identical.
Table~\ref{tab:same_object_percentages} in \secref{appendix:transferred_facts} provides statistics of fact pairs containing identical objects across languages.}

\paragraph{Similarity remains higher for \falsenegatives{}s than for \truenegatives{}s.}
Across all languages, we observe a consistent trend: the cosine similarity for \falsenegatives{}s quickly surpasses that of \truenegatives{}s.
While both begin at comparable levels, a clear and sustained separation emerges after approximately 50K pretraining steps.
This divergence suggests that the model aligns the representations of \falsenegatives{}s with their English counterparts better than for \truenegatives{}s -- facts that are similarly low-frequency but incorrectly predicted.
These findings offer direct evidence of crosslingual knowledge transfer on \falsenegatives{}s, benefiting from better alignment with English, spanning both language and script boundaries.

\paragraph{Better alignment enables crosslingual transfer but does not guarantee correct recall.}
The consistently high similarity in Latin-script languages aligns with prior work showing that Transformer models tend to cluster representations based on shared script \citep{wen-yi-mimno-2023-hyperpolyglot, liu-etal-2024-translico}.
However, improved alignment alone is not sufficient:
for \truenegatives{}s, 
the model continues to better align them in pretraining, yet this does not lead to gains
in recall accuracy (i.e., \truenegatives{}s are not learned).
This suggests that beyond alignment, other factors -- such as language-specific understanding/generation and instruction following abilities -- also play a critical role in factual recall.

\section{Conclusion}

We investigate how multilingual factual recall and crosslingual consistency emerge during pretraining, using OLMo-7B as a case study. 
Our analysis shows that factual recall improves early and is primarily driven by fact frequency, regardless of language.
However, some low-frequency facts in non-English languages can still be recalled, mainly due to crosslingual transfer from English -- especially for relations that involve named entities. 
We therefore conclude that multilingual factual knowledge is gained through both frequency-driven learning and crosslingual transfer starting from early stages. 

\section*{Limitations}

While this work contributes to emerging efforts in exploring multilingual knowledge acquisition during the pretraining process and contributes to understanding the mechanisms of acquisition, several limitations should be acknowledged.

First, our study focuses on the checkpoints of a single English-centric model as a case study.
This choice is primarily due to the scarcity of open-source models that provide both intermediate checkpoints and detailed documentation of their pretraining corpora. 
We therefore echo~\citet{soldaini-etal-2024-dolma} and encourage greater transparency in the community, including the release of intermediate checkpoints and associated data. 
This would facilitate further research into knowledge acquisition dynamics and help deepen our understanding of LLM pretraining processes.

Second, our approximation of fact frequency in certain script-sharing languages may lack full accuracy.
As discussed in \secref{frequencies} and \secref{exclude}, this is due to the difficulty in disambiguating language identity in shared-script corpora. 
While our findings suggest this issue does not significantly affect the overall results, future work could improve precision by applying language identification techniques, especially where computational resources permit.

Finally, although we analyze the dynamics of multilingual knowledge acquisition and identify two primary mechanisms -- frequency-based learning and crosslingual transfer -- we do not investigate the conditions under which each mechanism is most effective.
Studying these underlying factors requires controlled manipulation of the pretraining corpus to observe causal effects, which falls beyond the scope of this work. 
Nonetheless, we regard this as a promising direction for future research.

\section*{Acknowledgments}

This work was funded by Deutsche Forschungsgemeinschaft (project SCHU 2246/14-1).
François Yvon has been partly funded by the French National Funding
Agency (ANR) under the France 2030 program (ref. ANR-23-IACL-0007). Barbara Plank and Felicia Körner are supported by the European Research Council (ERC) Consolidator Grant DIALECT 101043235.

\bibliography{custom}

\appendix

\section{KLAR Statistics}\seclabel{klar}

\begin{table}
% \setlength{\abovecaptionskip}{0cm}
% \small
\setlength{\belowcaptionskip}{-0.4cm}
\setlength{\tabcolsep}{0.8mm}{}
\centering
\begin{tabular}{l r}
\hline
\textbf{Relation} & \textbf{Number of Facts} \\
\hline
\texttt{capital\_of} & 212 \\
\texttt{continent} & 212 \\
\texttt{country\_of\_citizenship} & 60 \\
\texttt{headquarters\_location} & 51 \\
\texttt{instrument} & 46 \\
\texttt{language\_of\_work\_or\_name} & 108 \\
\texttt{languages\_spoken} & 104 \\
\texttt{manufacturer} & 35 \\
\texttt{native\_language} & 130 \\
\texttt{place\_of\_birth} & 35 \\
\texttt{place\_of\_death} & 79 \\
\texttt{religion} & 125 \\
\hline
\textbf{total} & \textbf{1,197} \\
\hline
\end{tabular}
\caption{Number of facts grouped by relation types.}
\label{tab:relation_fact_counts}
\end{table}

We present the statistics of the KLAR dataset \citep{wang2025multilinguality} in Table~\ref{tab:relation_fact_counts}.
KLAR is based on BMLAMA17 \citep{qi-etal-2023-cross} with some minor modifications to improve the applicability to autoregressive models. 
We use \textbf{1,197} facts grouped into \textbf{12}~relation categories.

\input{holistic_consistency}
\input{frequency_correctness}
\input{sensitivity}
\input{similarity}
\input{transferred_facts}

\input{removing_identical_facts}
\input{dolma_documents}

% \newpage

\input{per_lang_per_relation}

\input{prompt_variation}

\section{Experimental Environment and Hyperparameters}\seclabel{hyperparameters}

All experiments are conducted on NVIDIA RTX A6000 GPUs. 
For each fact in each language, we use the prompt template provided in KLAR \citep{wang2025multilinguality}. 
Each final query is accompanied by three randomly selected demonstrations to enhance pattern-matching capabilities, thereby facilitating object extraction from the model's response. 
We use vLLM to generate responses for each query, with generation parameters set to greedy decoding and a maximum output length of 10 tokens.\footnote{\url{https://docs.vllm.ai/en/latest/}}

\end{document}

%% file: holistic_consistency.tex
\section{Complete Factual Recall Dynamics}\seclabel{complete_dynamics}

\begin{figure*}
    \centering
    \includegraphics[width=0.24\textwidth]{figures/consistency/ar.pdf}
    \includegraphics[width=0.24\textwidth]{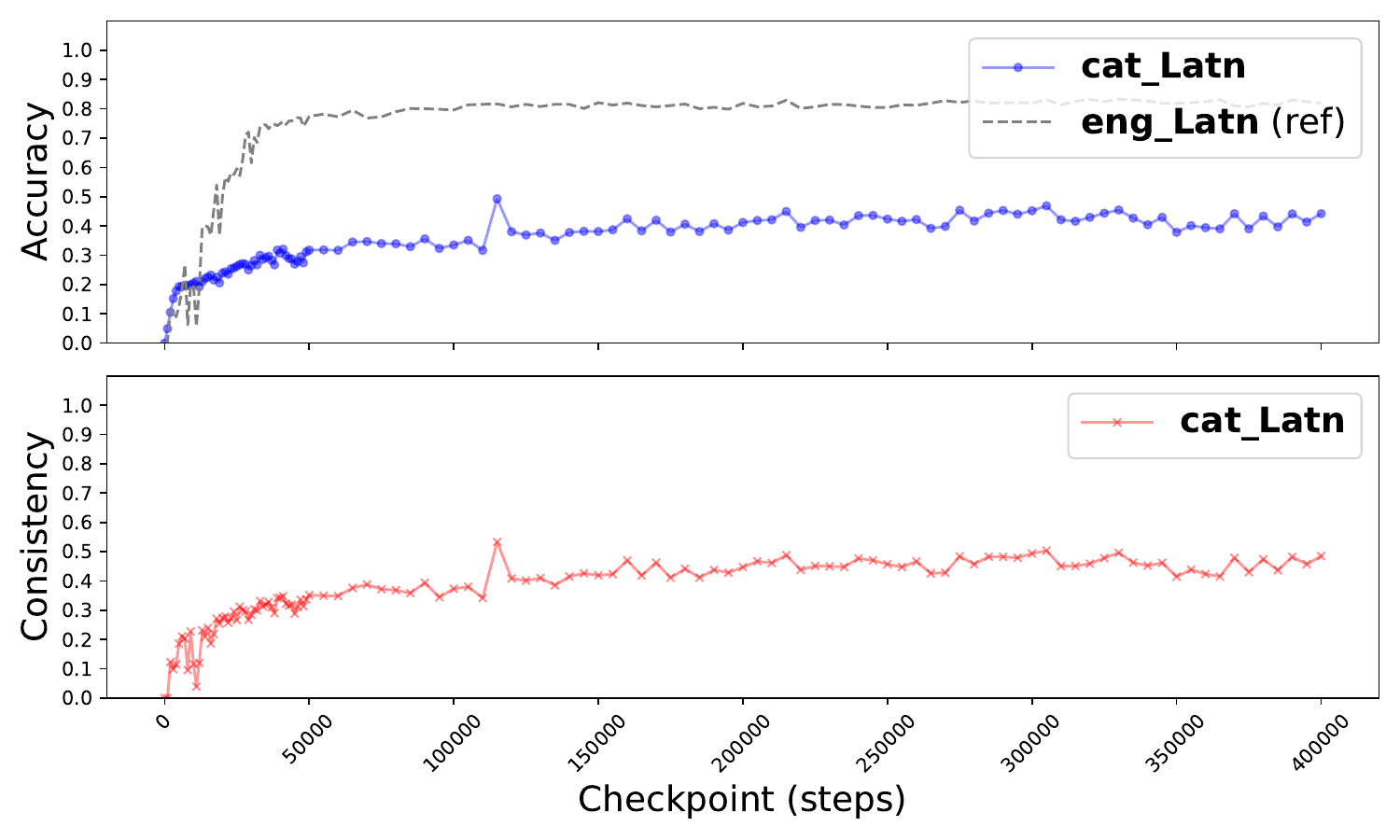}
    \includegraphics[width=0.24\textwidth]{figures/consistency/zh.pdf}
    \includegraphics[width=0.24\textwidth]{figures/consistency/el.pdf}
    \includegraphics[width=0.24\textwidth]{figures/consistency/fr.pdf}
    \includegraphics[width=0.24\textwidth]{figures/consistency/ja.pdf}
    \includegraphics[width=0.24\textwidth]{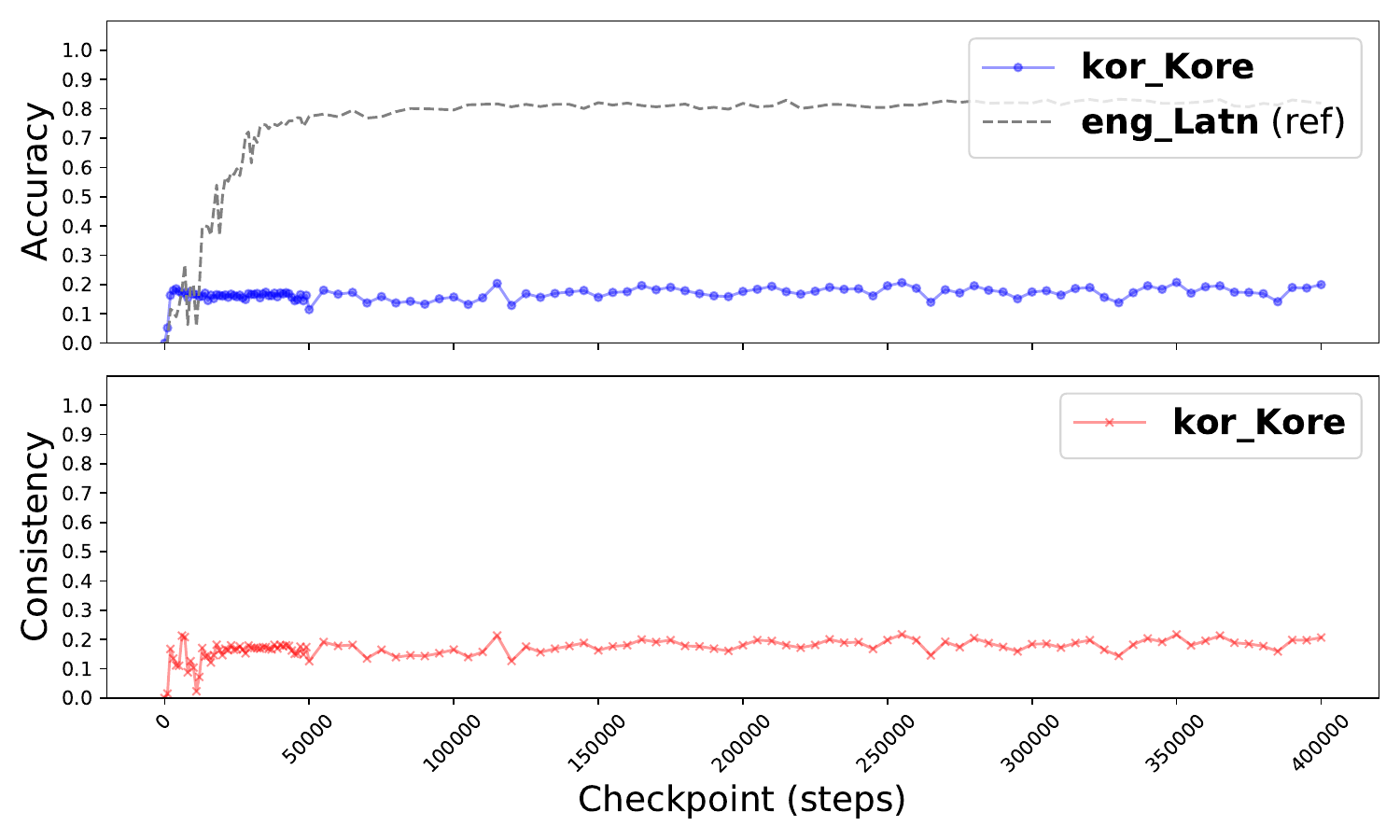}
    \includegraphics[width=0.24\textwidth]{figures/consistency/ru.pdf}
    \includegraphics[width=0.24\textwidth]{figures/consistency/es.pdf}
    \includegraphics[width=0.24\textwidth]{figures/consistency/tr.pdf}
    \includegraphics[width=0.24\textwidth]{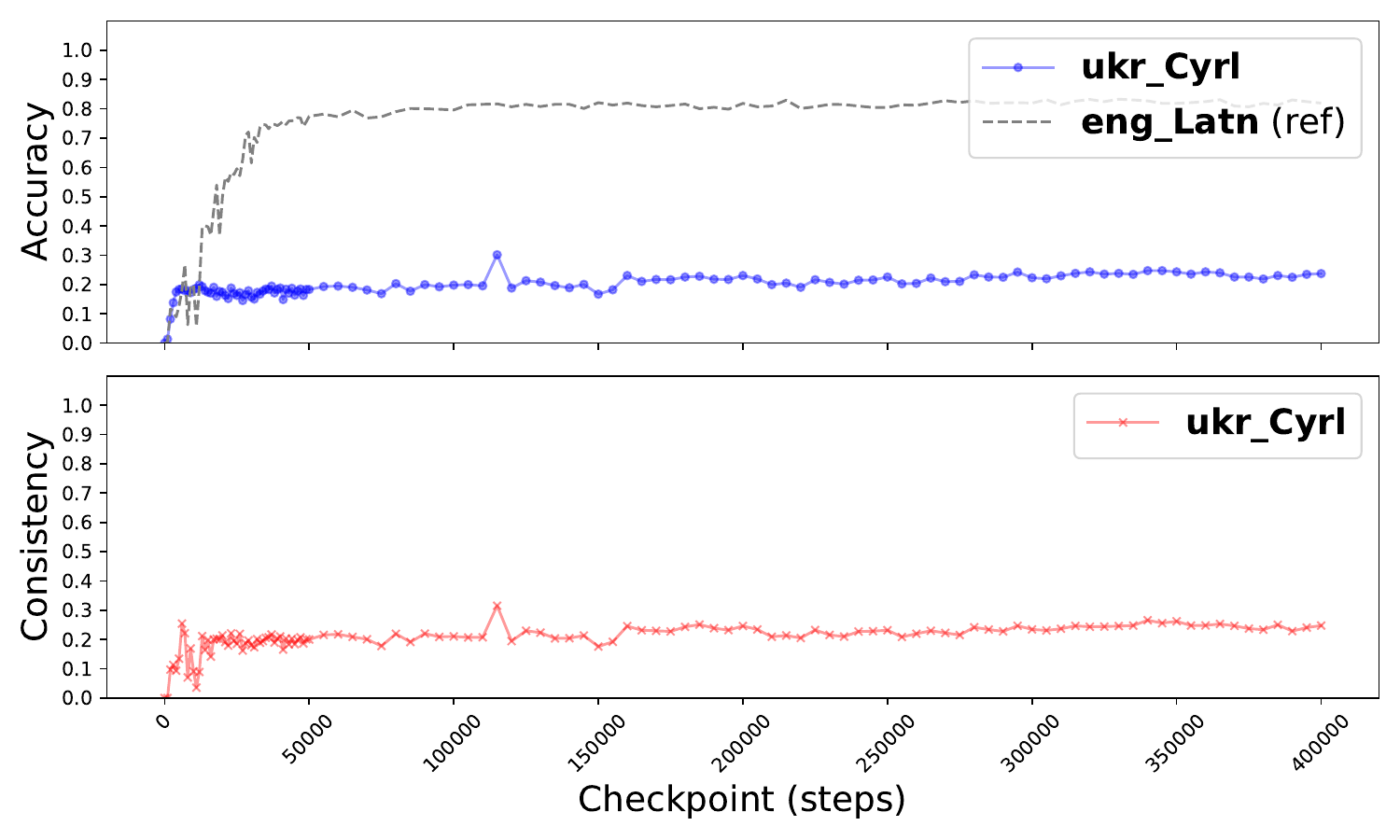}
    \caption{Factual accuracy (ACC) and crosslingual consistency (CO) for all languages.}
    \label{fig:performance_over_checkpoints_complete}
\end{figure*}

We present the complete factual recall dynamics in terms of \emph{accuracy} and \emph{crosslingual consistency} at each checkpoint of OLMo in Figure~\ref{fig:performance_over_checkpoints_complete}.

\section{Holistic Crosslingual Consistency}\seclabel{holistic}

To complement the English-centric consistency analysis in the main text, we investigate \textbf{holistic crosslingual consistency}, which quantifies the agreement of correct factual predictions across \textbf{all language pairs}. 
Similar to \secref{evaluation}, we compute the overlapping ratio of correct predictions in any two languages $l$ and $l'$:
$$
\text{CO}(l,l') = \frac{\sum_{i}^{|Q|}\mathbf{1}(\mathcal{M}(q_i^l) = o_i^l \land \mathcal{M}(q_i^{l'}) = o_i^{l'})}{\sum_{i}^{|Q|}\mathbf{1}(\mathcal{M}(q_i^l) = o_i^l \lor \mathcal{M}(q_i^{l'}) = o_i^{l'})}
$$
where $q_i^{l'}$ and $o_i^{l'}$ are the query and expected answer for the $i$th query in $l$ and $l'$, respectively, $\mathbf{1}(\cdot)$ is the indicator function, and $\mathcal{M}(\cdot)$ is the LLM's prediction function.

\begin{figure}
    \centering
    \includegraphics[width=0.48\textwidth]{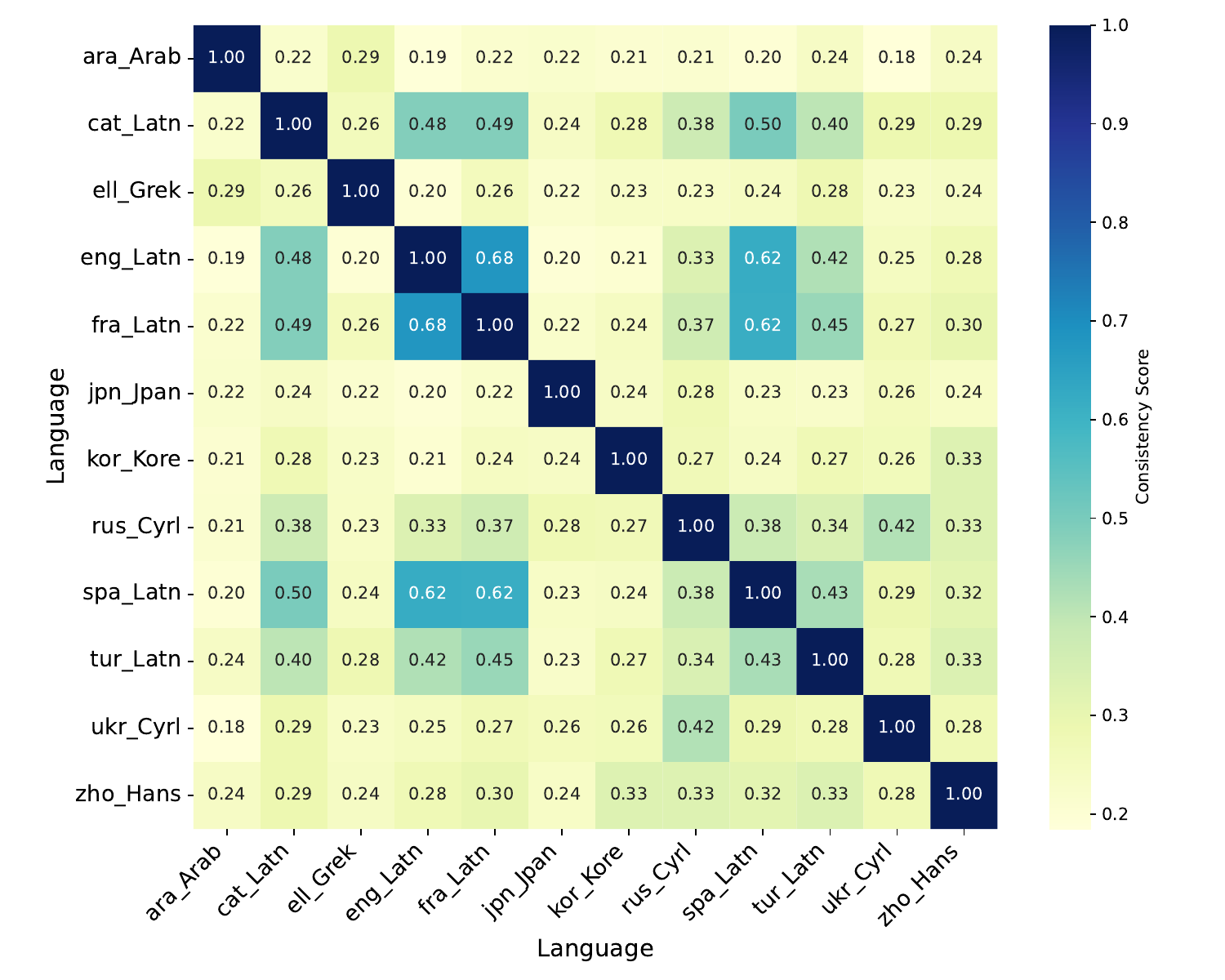}
    \caption{Crosslingual consistency of the model when it is pretrained for 400K steps. 
    The model exhibits stronger consistency among languages that share the same script. 
    In particular, Latin-script languages maintain consistently higher mutual consistency, while languages with distinct scripts -- such as jpn\_Jpan -- show lower consistency with others.}
    \label{fig:holistic_consistency}
\end{figure}

We first show the crosslingual consistency between any language pairs when the model is pretrained for 400K steps.
Figure~\ref{fig:holistic_consistency} presents the results.
We can observe that the consistency is generally low for most language pairs when the two involved languages do not share the same script, which is aligned with findings in the main text (cf. \secref{dynamics}) that most non-Latin script languages have low consistency when compared with the predominant language, English.
On the other hand, languages sharing the same script demonstrate higher similarity, for instance, Latin-script languages (fra\_Latn, span\_Latn, cat\_Latn, tur\_Latn, and eng\_Latn) and Cyrillic-script languages (rus\_Cyrl and ukr\_Cyrl).
This finding also aligns with \secref{dynamics}, indicating that shared script has a positive effect in improving the crosslingual transfer and crosslingual consistency.

\begin{figure}[ht]
    \centering
    \includegraphics[width=0.48\textwidth]{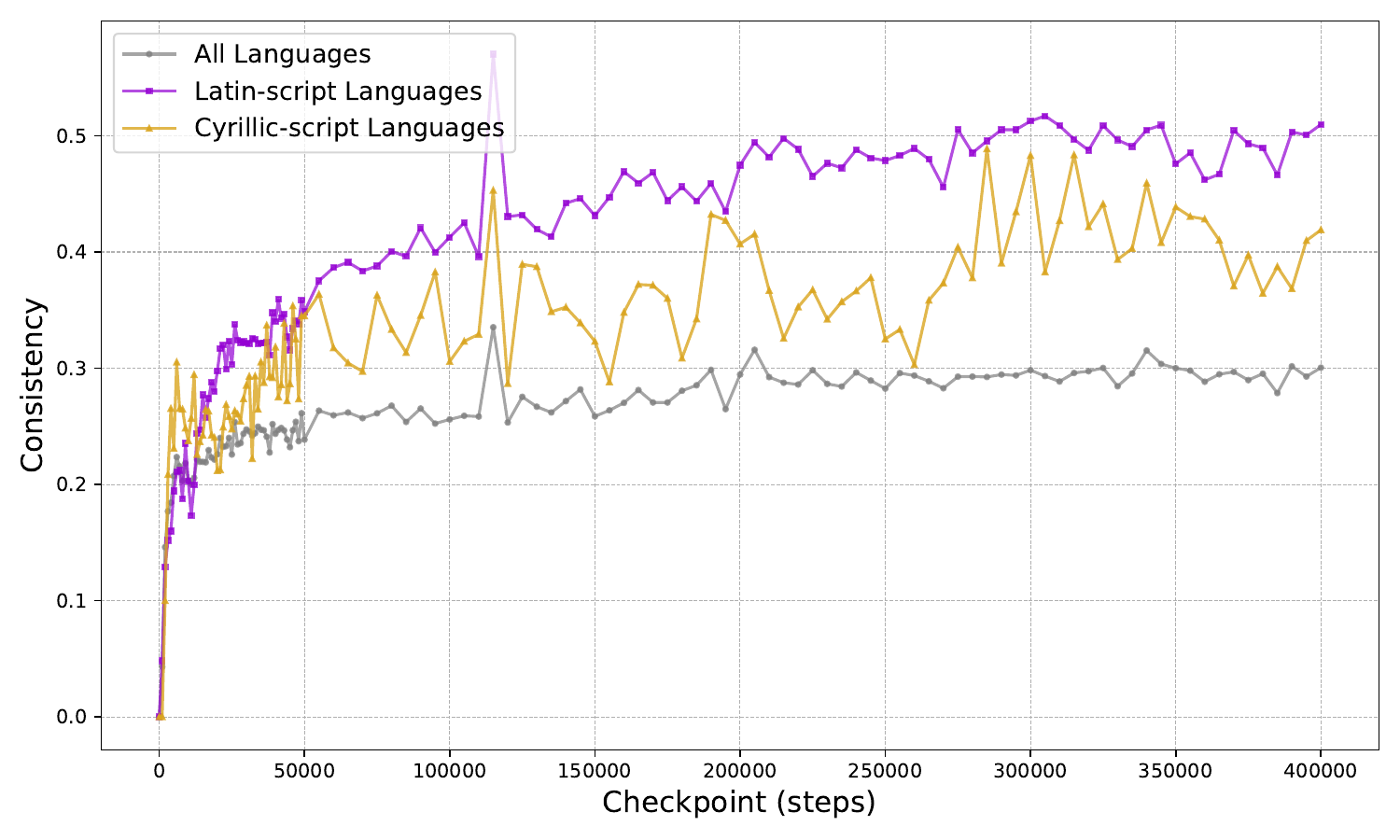}
    \caption{Dynamics of crosslingual consistency throughout pretraining. 
    We report the average consistency among Latin-script languages, Cyrillic-script languages, and all language pairs.
    While consistency continues to improve among Latin-script languages and Cyrillic-script languages, the overall consistency plateaus in the early stages, which is similar to the English-centric trends observed in Figure~\ref{fig:performance_over_checkpoints}.
    }
    \label{fig:holistic_consistency_checkpoint}
\end{figure}

We further analyze the dynamics of crosslingual consistency within script-specific language groups, namely, Latin-script and Cyrillic-script languages, to reveal how script similarity influences consistency during pretraining.
We average the consistency scores of each language pair to compute the per-group consistency.
Figure~\ref{fig:holistic_consistency_checkpoint} presents the results.
We observe that consistency improves as pretraining progresses, particularly among Latin-script languages, which maintain higher mutual consistency throughout pretraining. 
Similarly, Cyrillic-script languages show slower but noticeable gains, but with fluctuations -- possibly because only one pair of languages in this group. 
The overall consistency across all languages plateaus earlier. 
The results also align with the English-centric evaluation presented in \secref{dynamics}.
In summary, the supplementary analysis indicates that shared script and likely shared lexical structures contribute to greater alignment in factual recall across languages.

%% file: frequency_correctness.tex
\section{Fact Recall and Frequencies}

\subsection{Overall Results}\seclabel{appendix:overall_result}
\begin{figure*}
    \centering
    \includegraphics[width=0.19\textwidth]{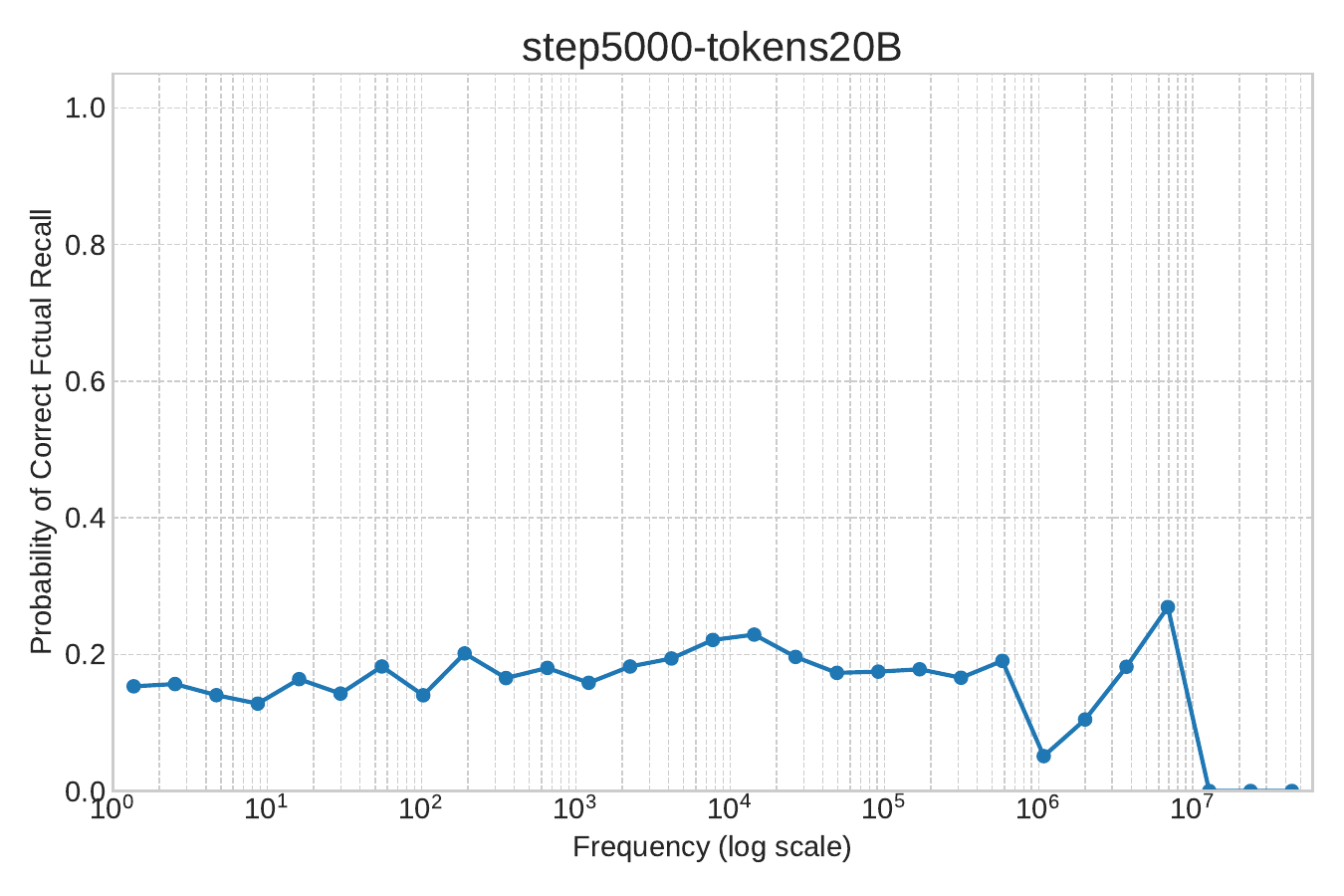}
    \includegraphics[width=0.19\textwidth]{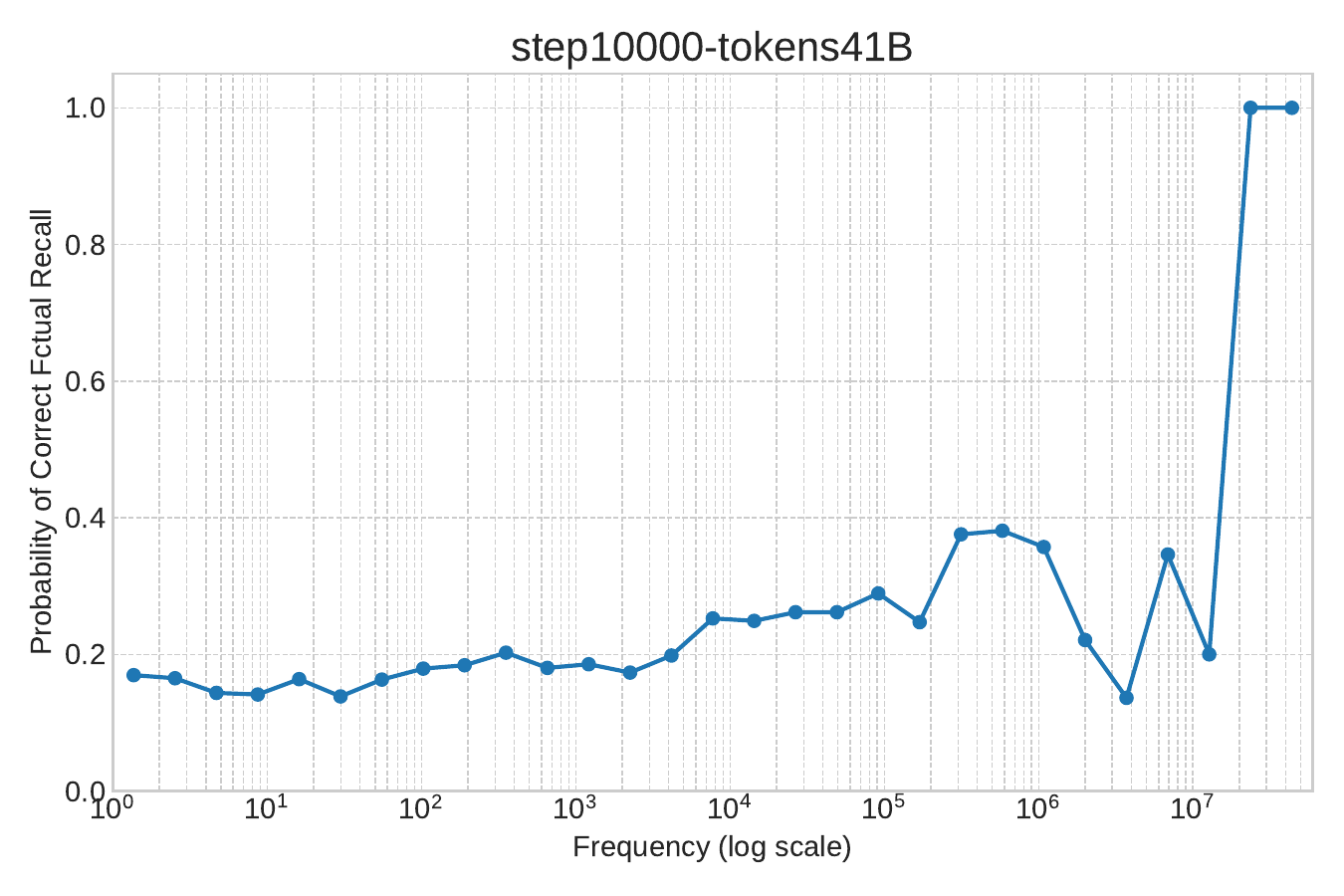}
    \includegraphics[width=0.19\textwidth]{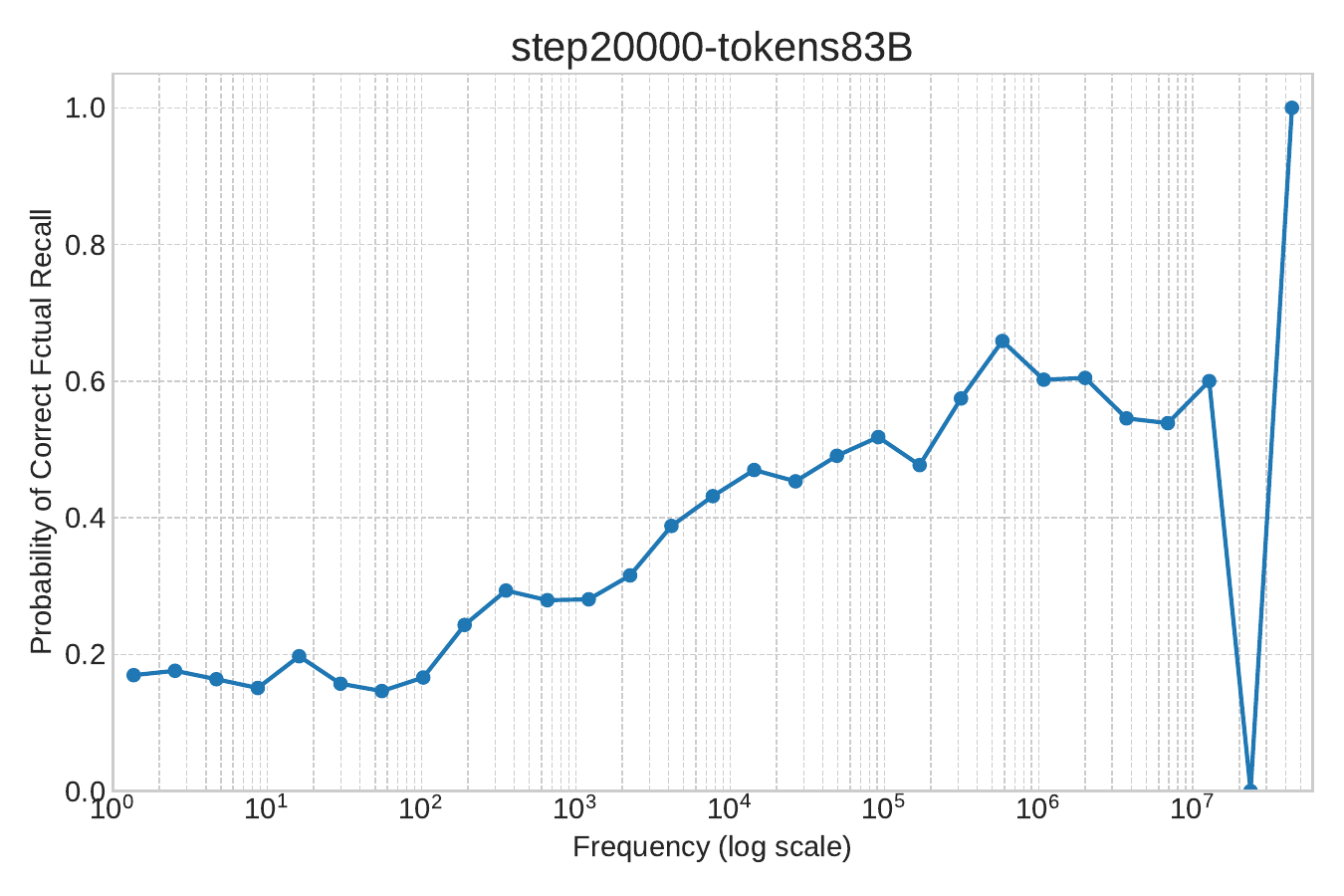}
    \includegraphics[width=0.19\textwidth]{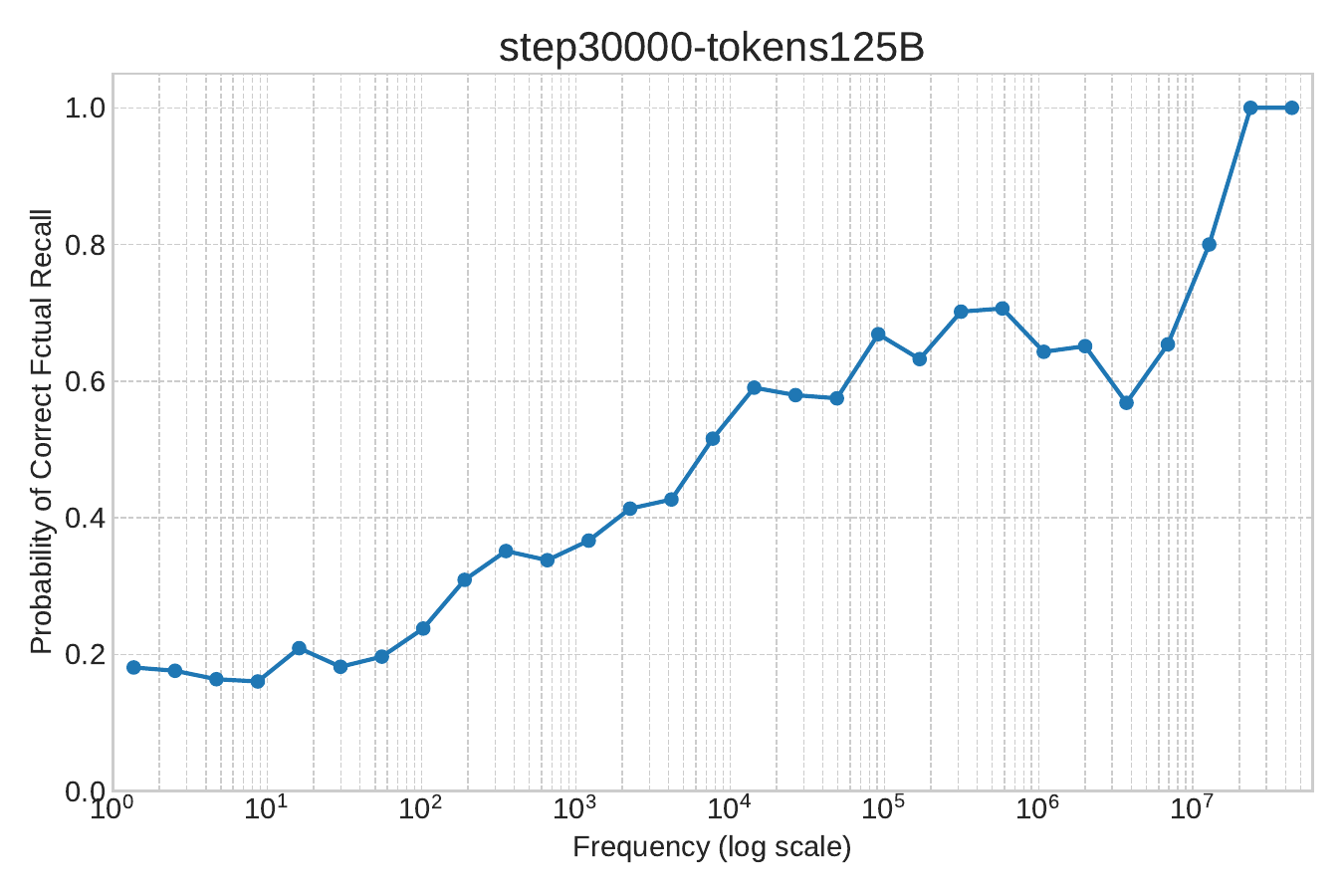}
    \includegraphics[width=0.19\textwidth]{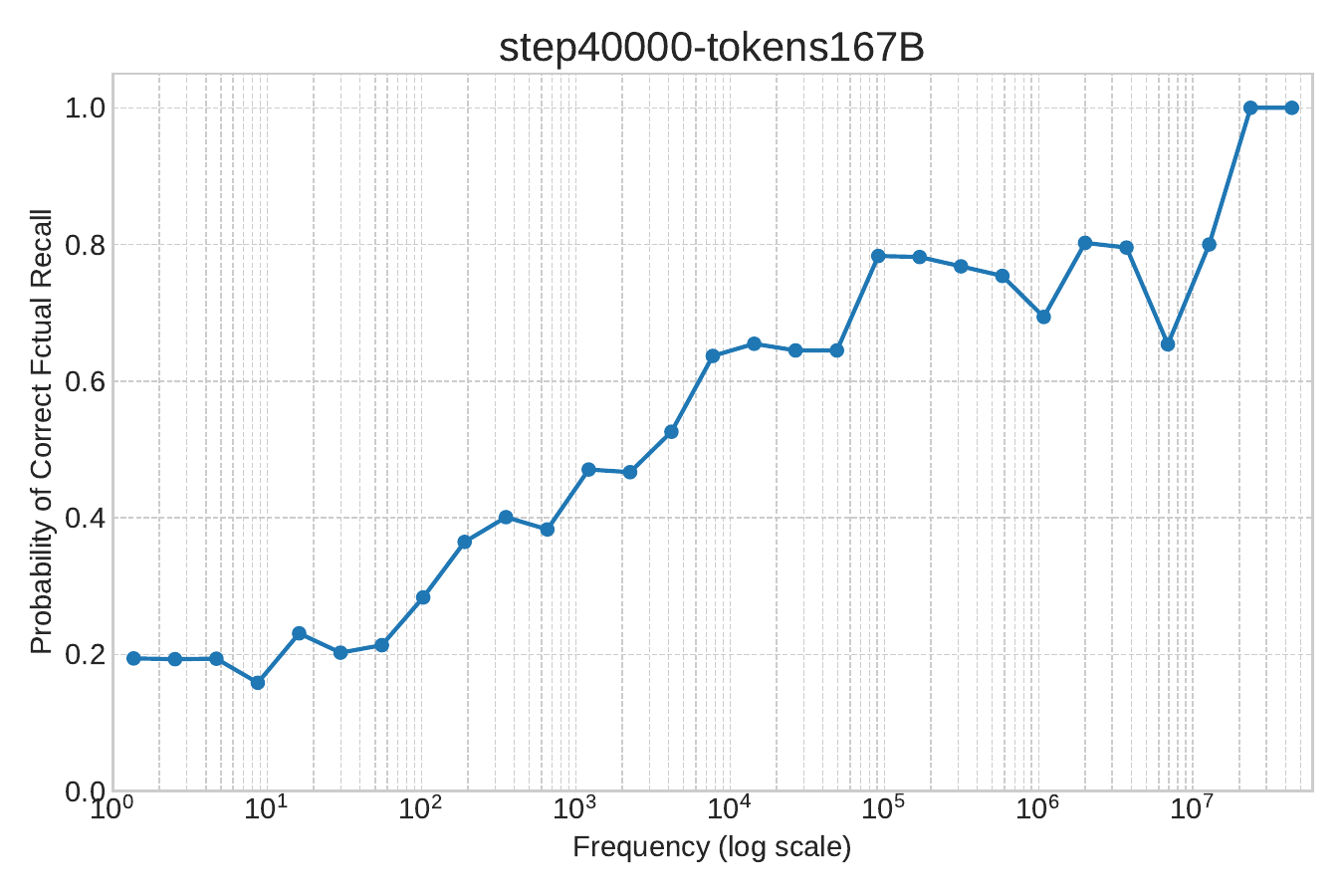}
    \includegraphics[width=0.19\textwidth]{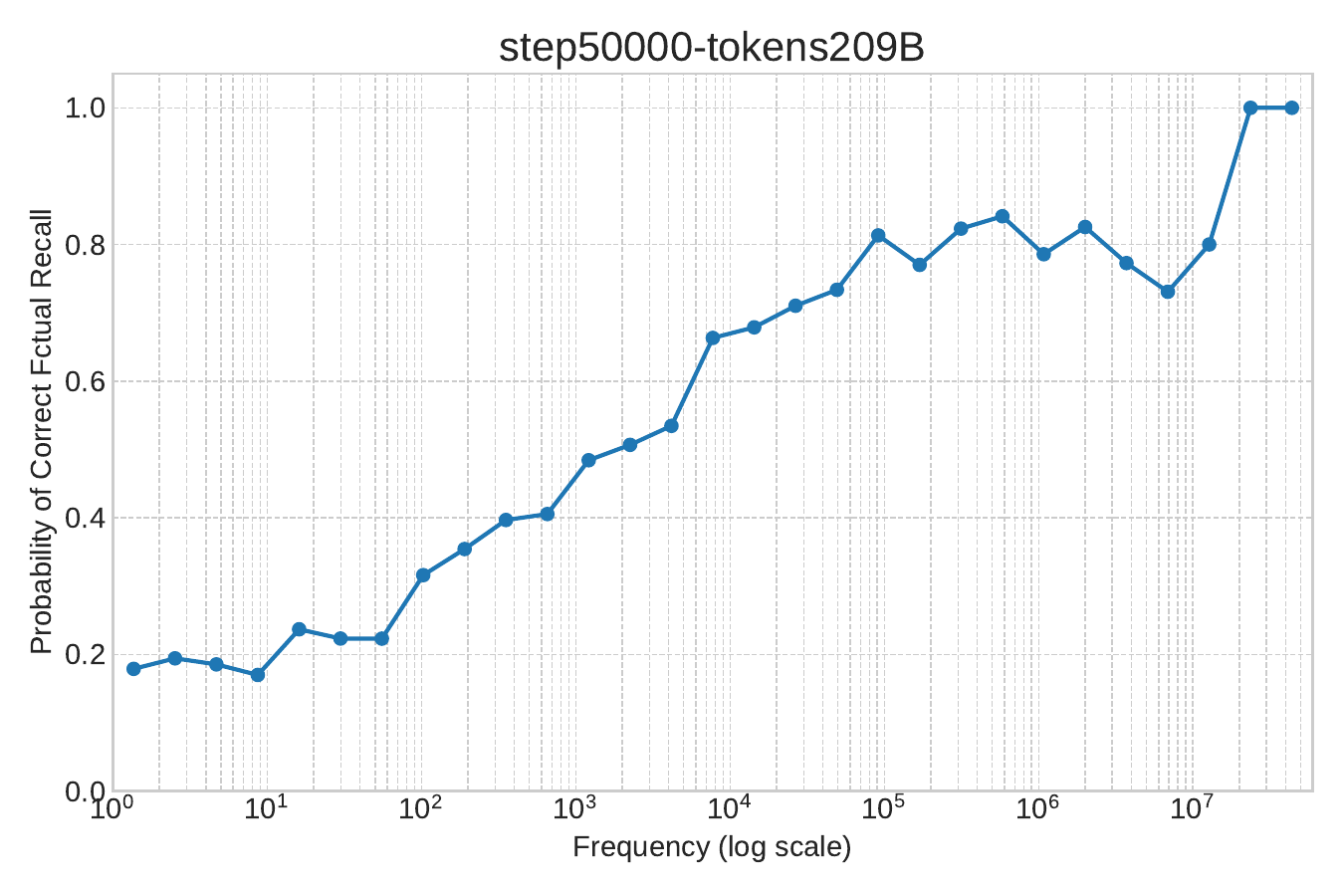}
    \includegraphics[width=0.19\textwidth]{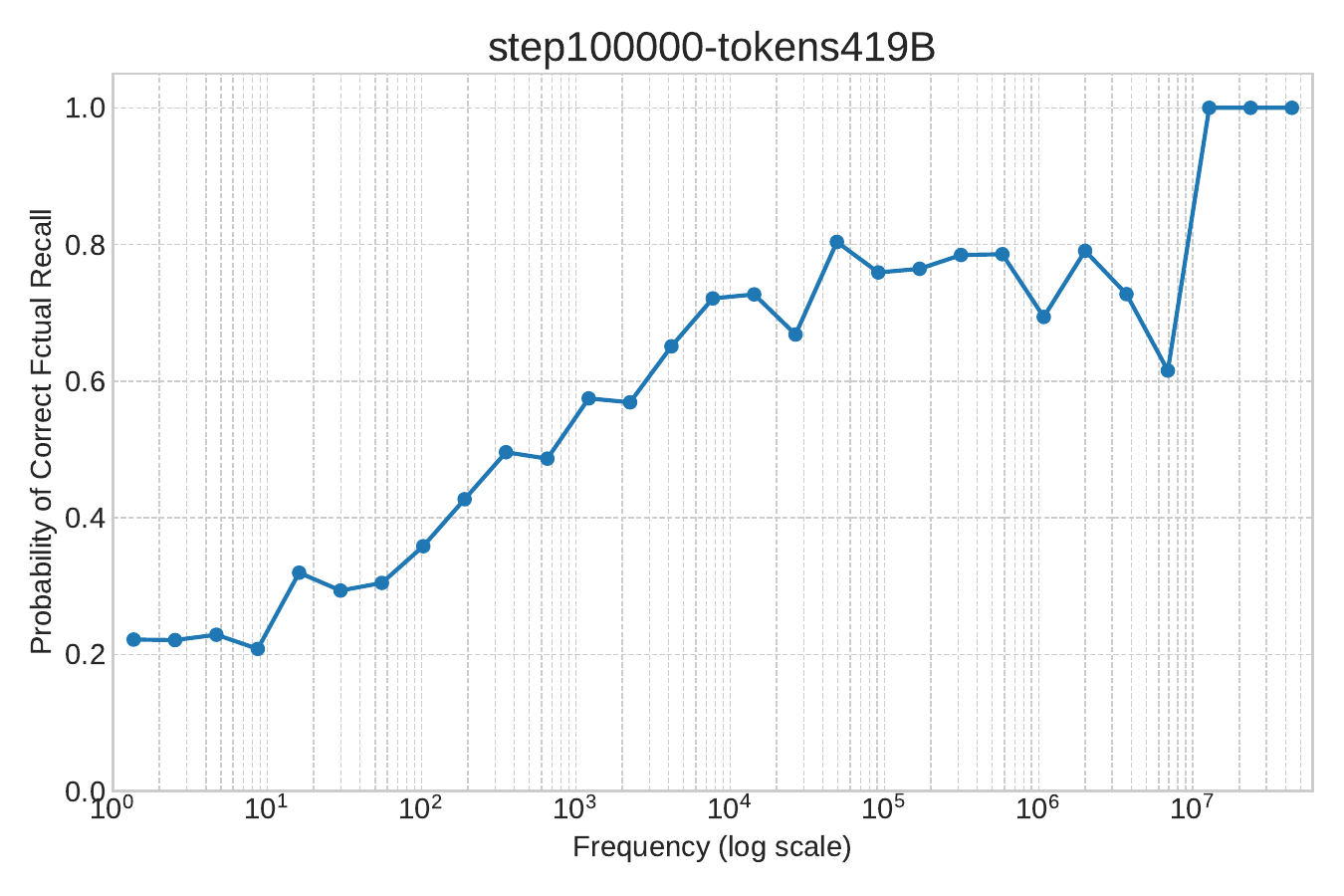}
    \includegraphics[width=0.19\textwidth]{figures/frequency_checkpoints/all_languages_probability_correct_vs_frequency_step100000.pdf}
    \includegraphics[width=0.19\textwidth]{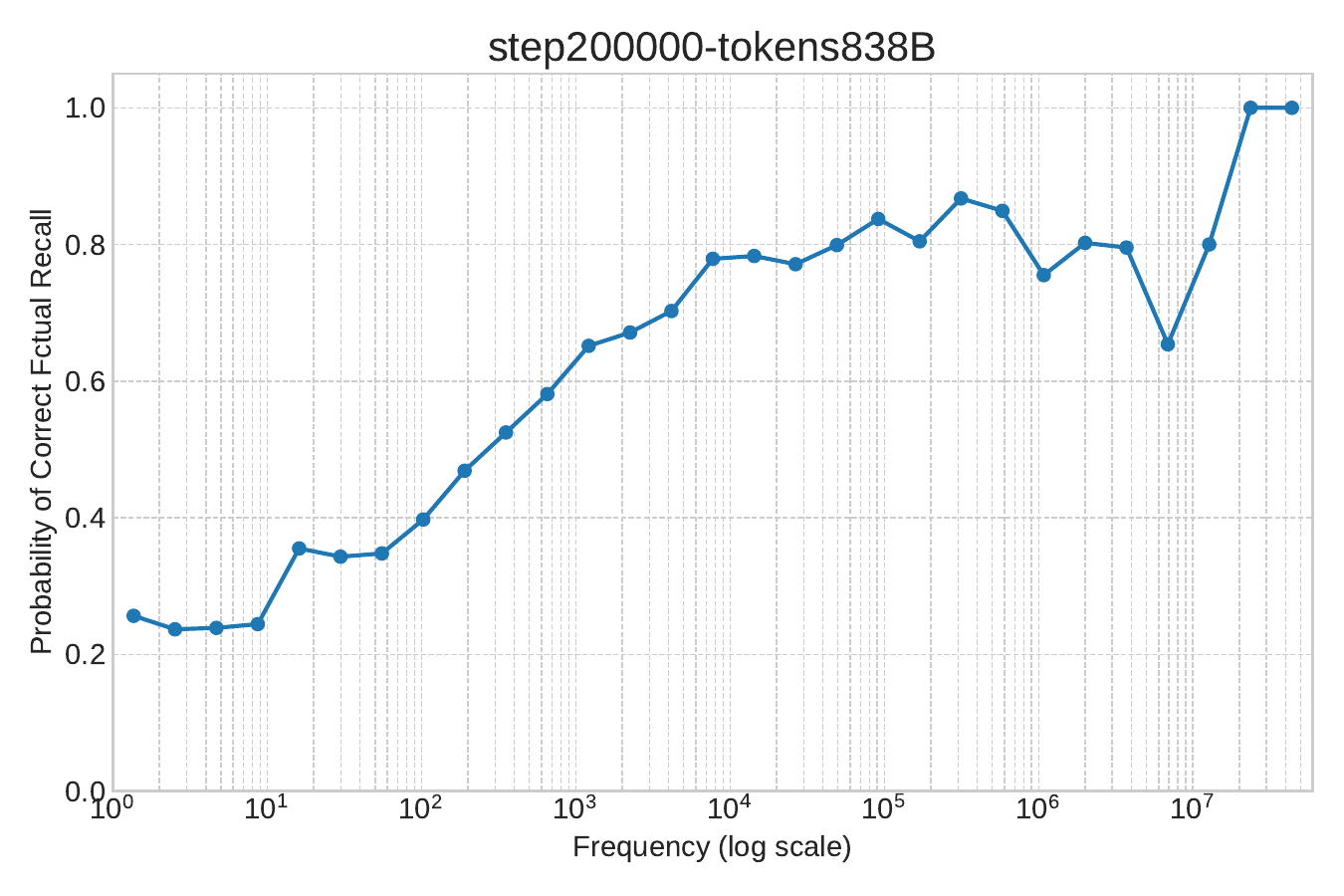}
    \includegraphics[width=0.19\textwidth]{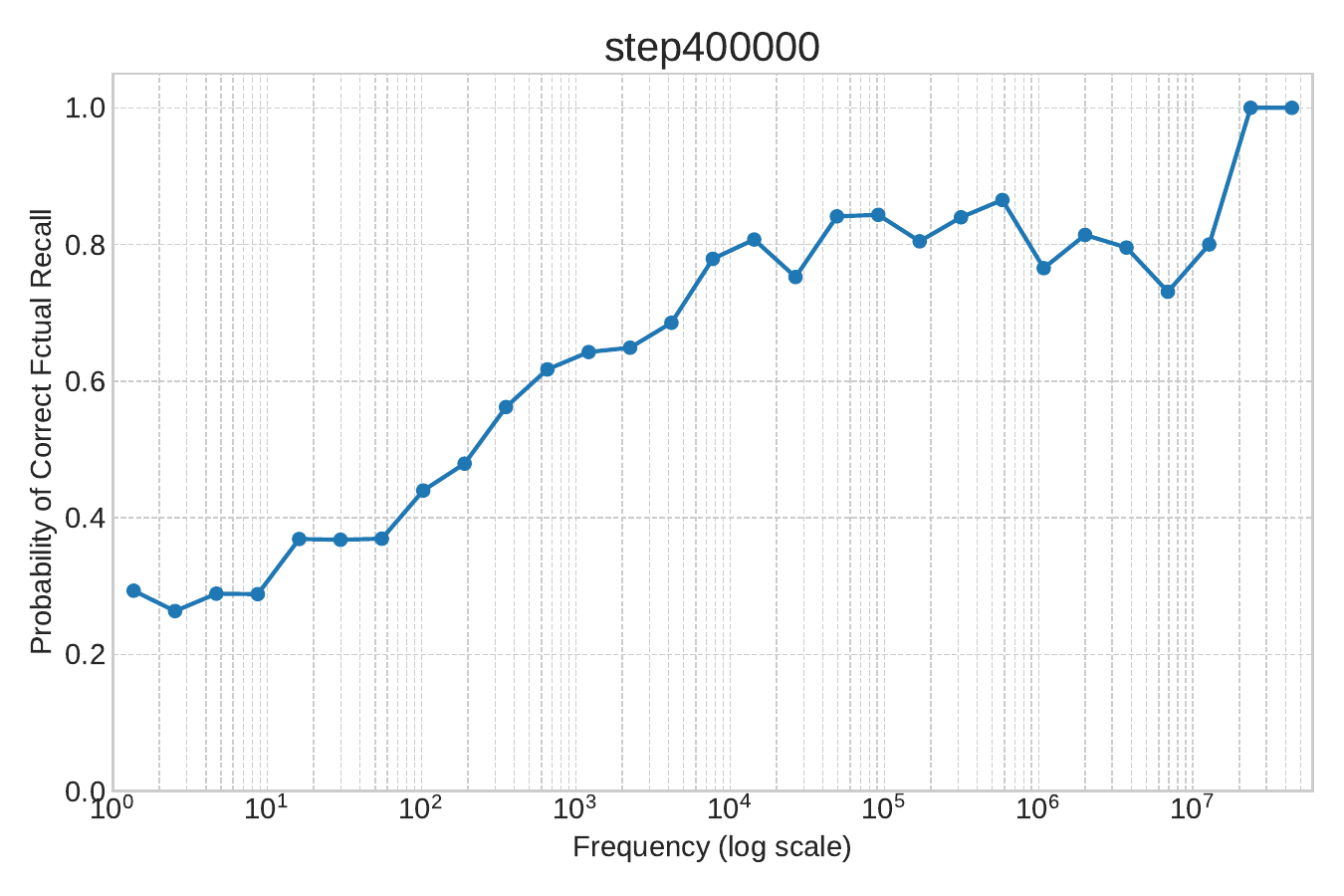}
\caption{Relationship between fact frequency and factual recall for all languages in 10 checkpoints. High-frequency facts are more likely to be correctly recalled than rare ones. This frequency–correctness correlation emerges very early in pretraining (roughly 30K steps) and becomes more pronounced over time.}
\label{fig:correctness_frequency_global_complete}
\end{figure*}

Figure~\ref{fig:correctness_frequency_global_complete} presents the evolution of the relationship between fact frequency and correctness across 10 checkpoints during pretraining. 
We observe that a linear relationship is gradually formed in the early stages (i.e., 5K to 30K steps).
This linear relationship indicates that high-frequency facts are more likely to be correctly recalled than low-frequency ones.
This trend stabilizes and sharpens as training progresses.
This emergent frequency–correctness correlation underscores the model's bias toward memorizing frequently encountered facts. 
The rapid formation of this pattern indicates that pretraining quickly internalizes statistical regularities in the data, which in turn guide factual recall.

\subsection{Per-Language Results}
\seclabel{appendix:per_language_result}

\begin{figure*}
    \centering
    \includegraphics[width=0.32\textwidth]{figures/frequency_distrubution/en/merged_correctness_frequency_plot.pdf}
    \includegraphics[width=0.32\textwidth]{figures/frequency_distrubution/es/merged_correctness_frequency_plot.pdf}
    \includegraphics[width=0.32\textwidth]{figures/frequency_distrubution/zh/merged_correctness_frequency_plot.pdf}
    \includegraphics[width=0.32\textwidth]{figures/frequency_distrubution/ca/merged_correctness_frequency_plot.pdf}
    \includegraphics[width=0.32\textwidth]{figures/frequency_distrubution/ru/merged_correctness_frequency_plot.pdf}
    \includegraphics[width=0.32\textwidth]{figures/frequency_distrubution/ar/merged_correctness_frequency_plot.pdf}
    \includegraphics[width=0.32\textwidth]{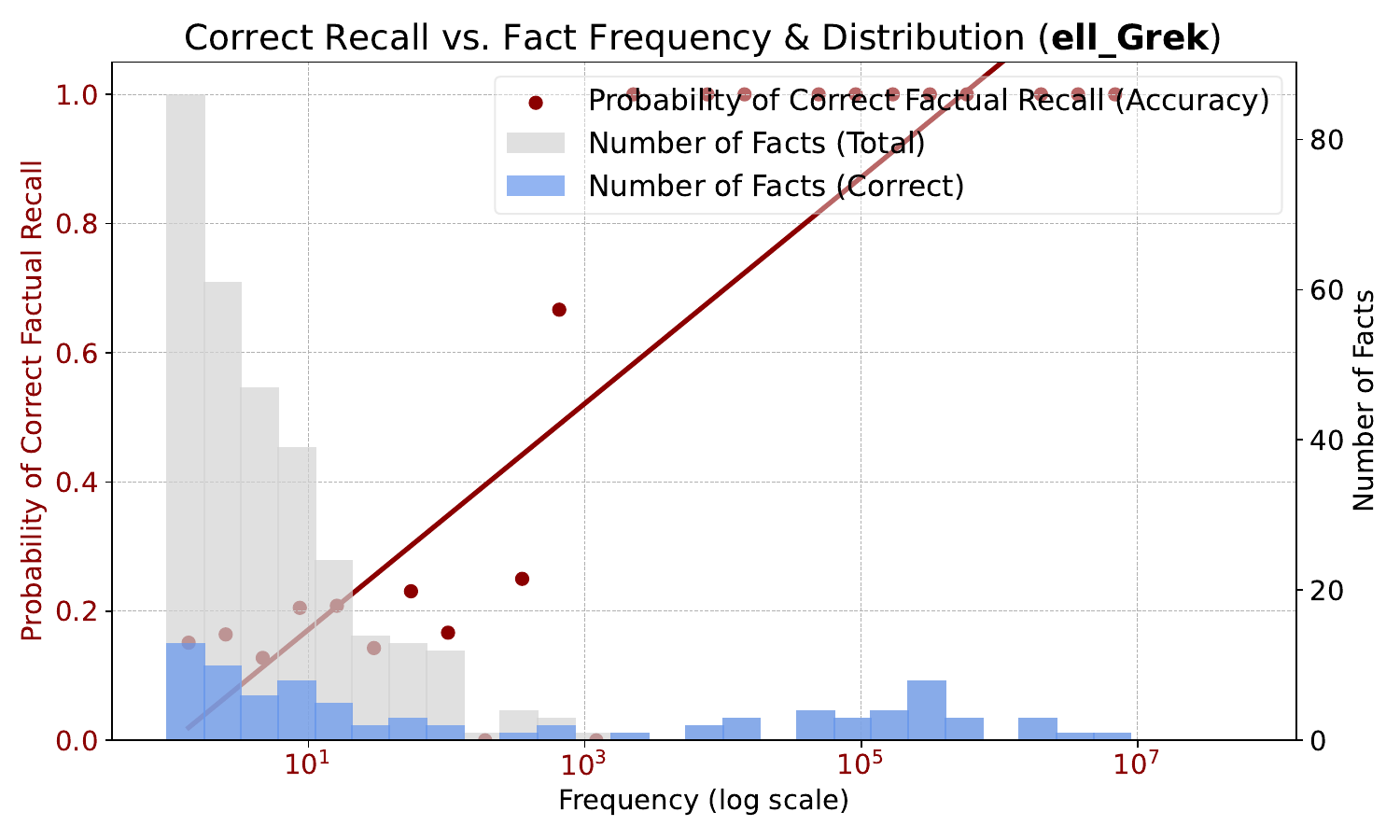}
    \includegraphics[width=0.32\textwidth]{figures/frequency_distrubution/fr/merged_correctness_frequency_plot.pdf}
   \includegraphics[width=0.32\textwidth]{figures/frequency_distrubution/ja/merged_correctness_frequency_plot.pdf}
   \includegraphics[width=0.32\textwidth]{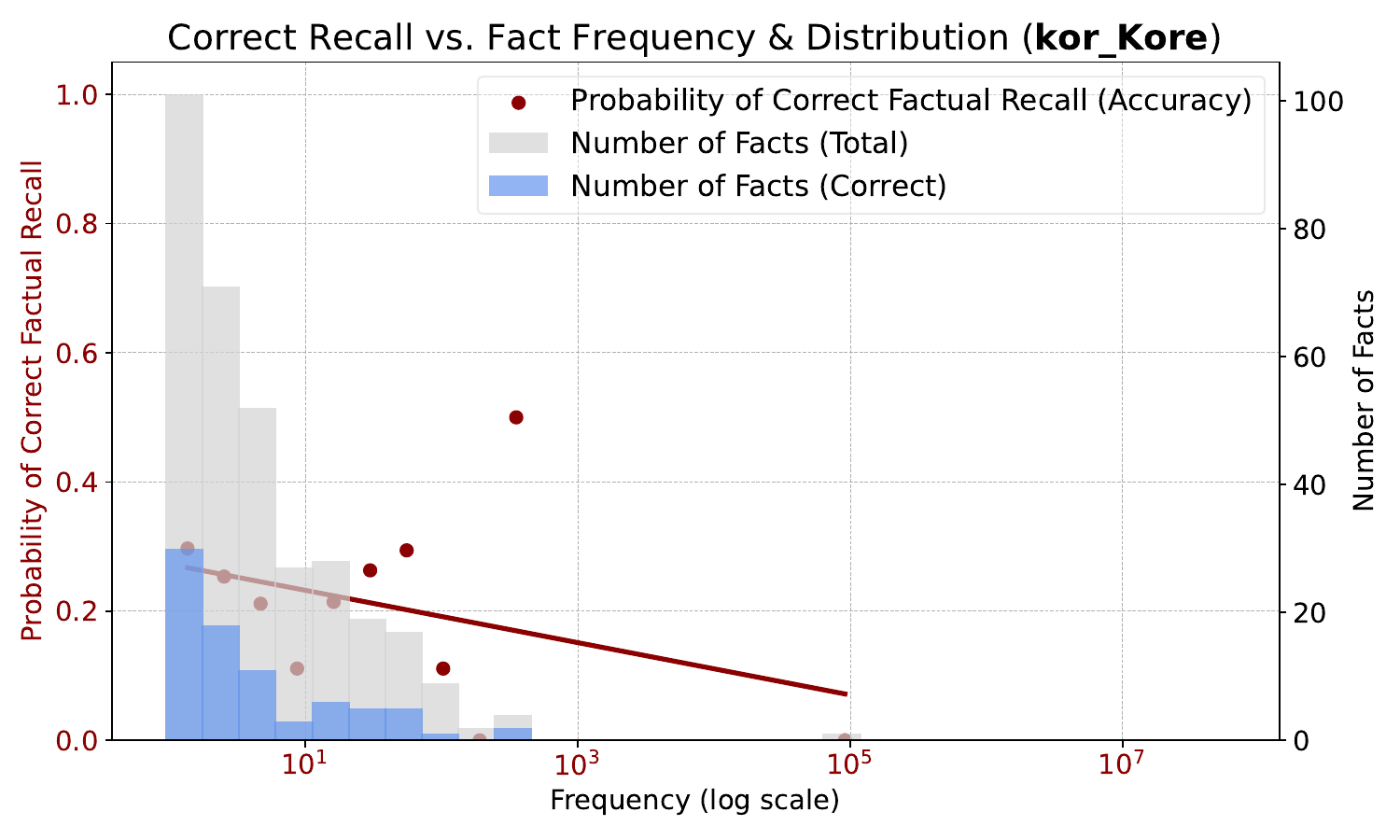}
   \includegraphics[width=0.32\textwidth]{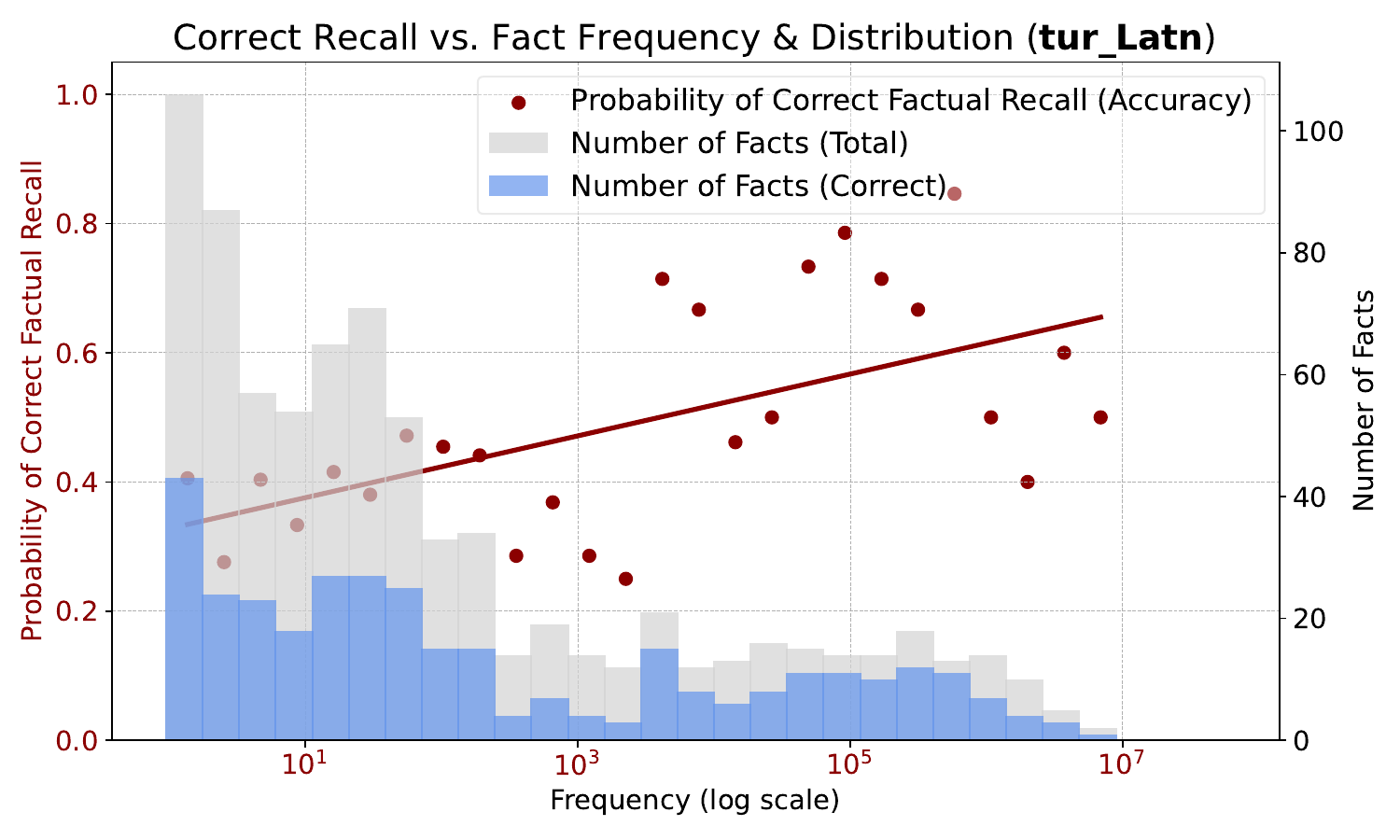}
   \includegraphics[width=0.32\textwidth]{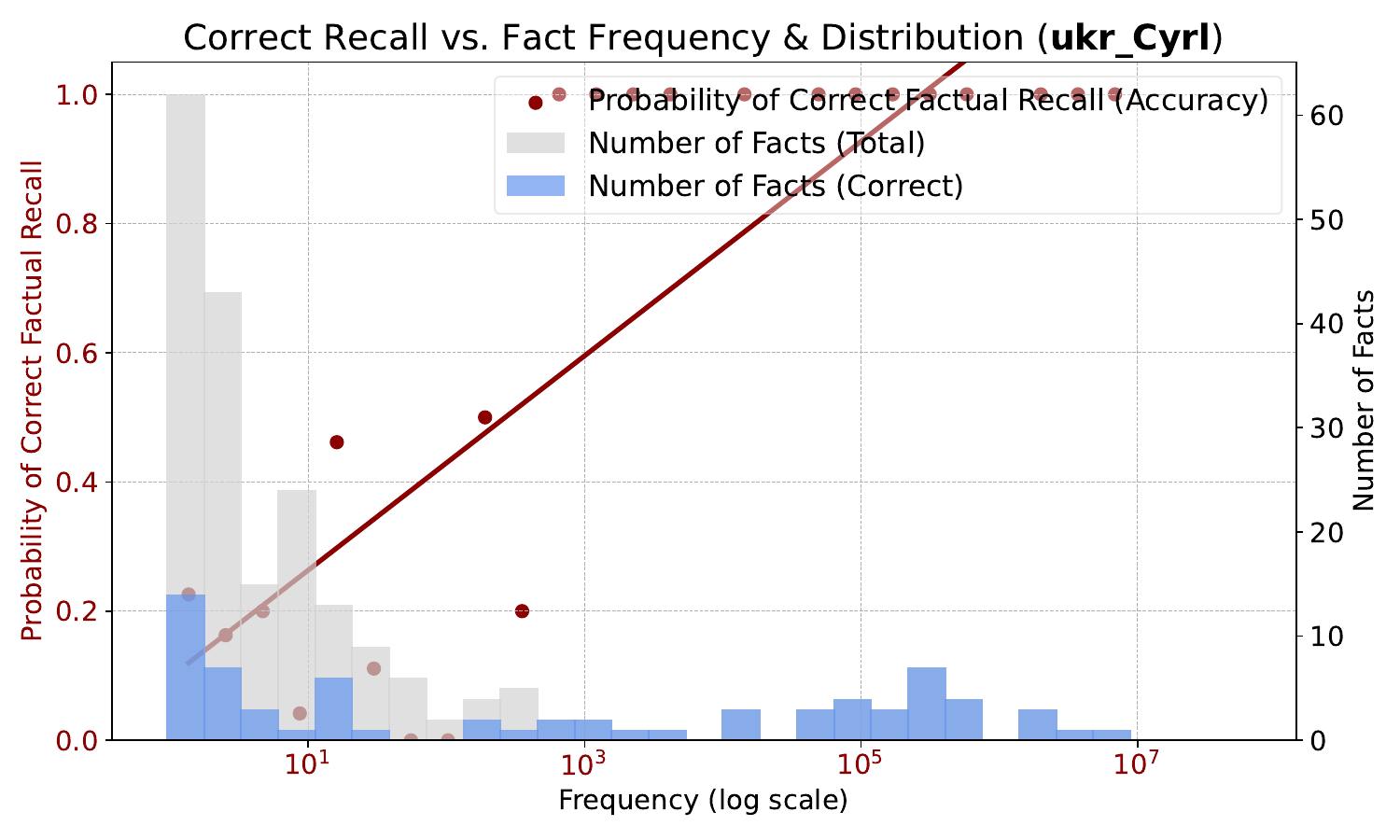}
\caption{Complete results of the relationship between fact frequency and the probability of correct factual recall in each language.}
    \label{fig:correctness_frequency_local_complete}
\end{figure*}

Figure~\ref{fig:correctness_frequency_local_complete} further breaks down the same frequency–correctness analysis by language, showing the distribution of fact frequencies and recall accuracy in each of the 12 languages. 
Because Dolma \citep{soldaini-etal-2024-dolma} is an English-centric dataset, the fact frequencies for Latin-based languages are more properly distributed.
In contrast, languages of other scripts have more uneven distributions -- with most facts occurring very few times or even not occurring at all (not shown in the figure).
However, the overall frequency–correctness correlation holds across languages, which is aligned with the global trend in \secref{appendix:overall_result}.
Notably, many languages have a substantial number of facts that are correctly predicted at low frequencies -- mainly due to crosslingual transfer, for which we investigate in \secref{transferred_facts}.

%% file: sensitivity.tex
\section{Threshold Classifier Sensitivity}\seclabel{sensitivity}
\begin{figure*}
    \centering
\includegraphics[width=0.98\textwidth]{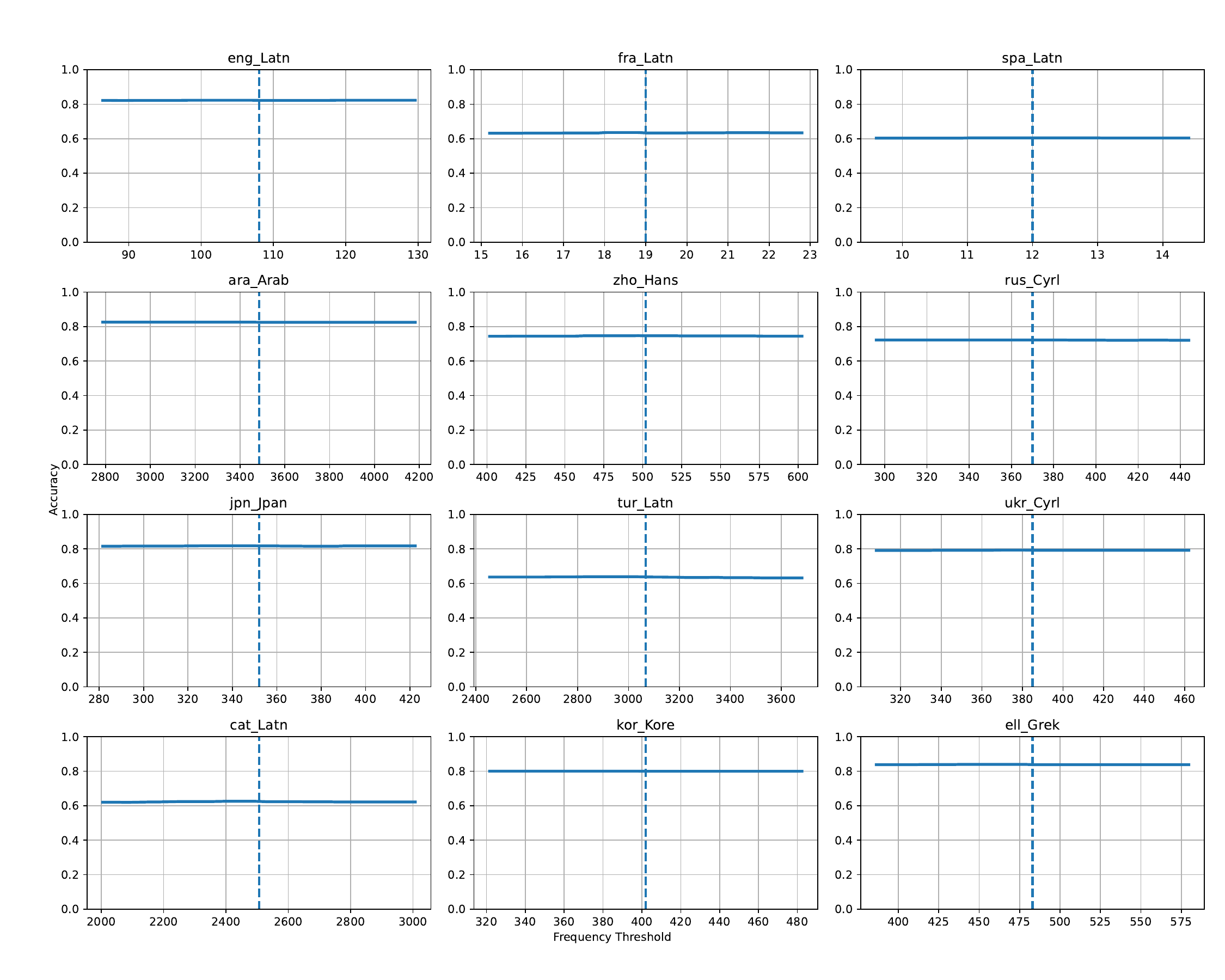}
    \caption{Classifier accuracy versus selected frequency threshold within a range of $\pm20\%$ of $t_l^*$, the chosen threshold. The dotted line shows the actual chosen threshold.
    }
    \label{fig:acc_vs_threshold}
\end{figure*}
In order to analyze the sensitivity of the threshold-based classifier from Section \secref{prediction} to the chosen threshold, we first plot the classifier accuracy for a range of thresholds within $t_l^*\pm20\%$, for a step size of 1\%, shown in Figure \ref{fig:acc_vs_threshold}. We observe that the curves across languages are mostly flat, suggesting that the classifier accuracy is robust to the chosen threshold. 

To further confirm the classifier's robustness, we randomly sample 90\% of the original dataset per language and select a new $t_l^*$ based on this subsample. We evaluate the classifier on the full dataset. The results for 5000 runs are shown in Table \ref{tab:threshold_bootstrap}. We note that though the confidence intervals for some thresholds vary widely, the resulting accuracy is very stable. Furthermore, the confidence intervals for the FP and FN counts, which are the focus of the analysis in Section \secref{transferred_facts} are narrow for most languages, with the exception of fra\_Latn and spa\_Latn.

We hypothesize that frequency-based prediction for these languages is confounded by two factors, both of which boost transfer from other languages: first, as we also noted in Section \secref{prediction}, fra\_Latn and spa\_Latn benefit strongly from transfer from English and other Latin-script languages, second, our analysis in Section \secref{dolma} indicates that fra\_Latn and spa\_Latn are well-represented in the pre-training data (cf. \secref{transferred_facts}).

\begin{table*}[t]
\centering
\small
\setlength{\tabcolsep}{3pt}
\begin{tabular}{lrrr rrr rrr rrr}
\toprule
 & \multicolumn{3}{c}{\textbf{Threshold}} & \multicolumn{3}{c}{\textbf{Accuracy}} & \multicolumn{3}{c}{\textbf{FP}} & \multicolumn{3}{c}{\textbf{FN}} \\
 \midrule
\textbf{Lang} & \textbf{Orig.} & \textbf{Mean} & \textbf{95\% CI} & \textbf{Orig.} & \textbf{Mean} & \textbf{95\% CI} & \textbf{Orig.} & \textbf{Mean} & \textbf{95\% CI} & \textbf{Orig.} & \textbf{Mean} & \textbf{95\% CI} \\
\midrule
ara\_Arab & 3485 & 3247 & {[}953, 3485{]} & 0.83 & 0.83 & {[}0.82, 0.83{]} & 0 & 0 & {[}0, 3{]} & 209 & 209 & {[}208, 209{]} \\
cat\_Latn & 2506 & 2462 & {[}2389, 2506{]} & 0.63 & 0.63 & {[}0.62, 0.63{]} & 64 & 65 & {[}64, 65{]} & 384 & 383 & {[}384, 384{]} \\
ell\_Grek & 483 & 641 & {[}268, 1692{]} & 0.84 & 0.84 & {[}0.84, 0.84{]} & 2 & 2 & {[}0, 4{]} & 190 & 190 & {[}189, 192{]} \\
eng\_Latn & 108 & 83 & {[}1, 146{]} & 0.82 & 0.82 & {[}0.82, 0.82{]} & 205 & 207 & {[}203, 213{]} & 7 & 6 & {[}1, 9{]} \\
fra\_Latn & 19 & 16 & {[}5, 25{]} & 0.64 & 0.64 & {[}0.63, 0.64{]} & 302 & 318 & {[}290, 361{]} & 134 & 119 & {[}77, 146{]} \\
jpn\_Jpan & 352 & 378 & {[}326, 450{]} & 0.82 & 0.82 & {[}0.82, 0.82{]} & 6 & 6 & {[}4, 7{]} & 212 & 212 & {[}212, 215{]} \\
kor\_Kore & 402 & 376 & {[}262, 402{]} & 0.80 & 0.80 & {[}0.80, 0.80{]} & 1 & 1 & {[}1, 3{]} & 238 & 238 & {[}237, 238{]} \\
rus\_Cyrl & 370 & 305 & {[}201, 370{]} & 0.72 & 0.72 & {[}0.72, 0.72{]} & 2 & 4 & {[}2, 8{]} & 330 & 328 & {[}325, 330{]} \\
spa\_Latn & 12 & 11 & {[}5, 59{]} & 0.60 & 0.60 & {[}0.60, 0.60{]} & 304 & 325 & {[}194, 365{]} & 169 & 149 & {[}109, 287{]} \\
tur\_Latn & 3068 & 3048 & {[}2816, 3068{]} & 0.64 & 0.64 & {[}0.64, 0.64{]} & 60 & 60 & {[}60, 61{]} & 373 & 373 & {[}373, 373{]} \\
ukr\_Cyrl & 385 & 382 & {[}368, 385{]} & 0.79 & 0.79 & {[}0.79, 0.79{]} & 0 & 0 & {[}0, 1{]} & 248 & 248 & {[}248, 248{]} \\
zho\_Hans & 502 & 494 & {[}461, 502{]} & 0.75 & 0.75 & {[}0.75, 0.75{]} & 7 & 8 & {[}7, 10{]} & 296 & 296 & {[}295, 296{]} \\
\bottomrule
\end{tabular}
\caption{Mean threshold, accuracy, false positives, false negatives, over 5000 runs of selecting a threshold $t_l^*$ using a randomly subsampled dataset. We include the results from selecting a threshold on the full dataset for comparison, denoted ``Orig.''.}
\label{tab:threshold_bootstrap}
\end{table*}

%% file: similarity.tex
\section{Complete Similarity Progression}
\seclabel{appendix:similarity}

\begin{figure*}
    \centering
    \includegraphics[width=0.32\textwidth]{figures/similarity/ara_Arab_lasttoken_mean_similarity.pdf}
    \includegraphics[width=0.32\textwidth]{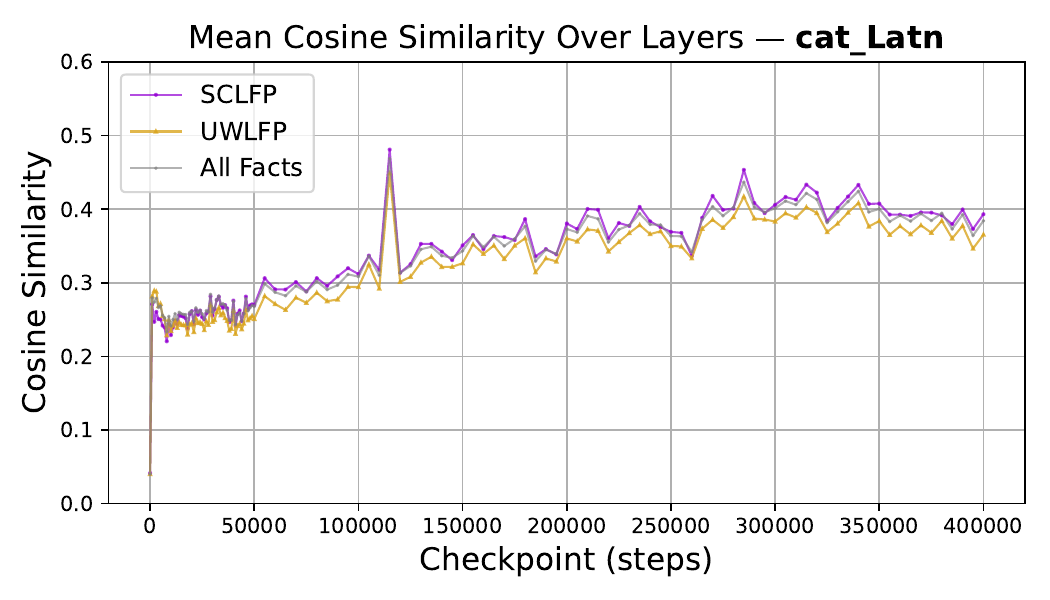}
    \includegraphics[width=0.32\textwidth]
    {figures/similarity/zho_Hans_lasttoken_mean_similarity.pdf}
    \includegraphics[width=0.32\textwidth]{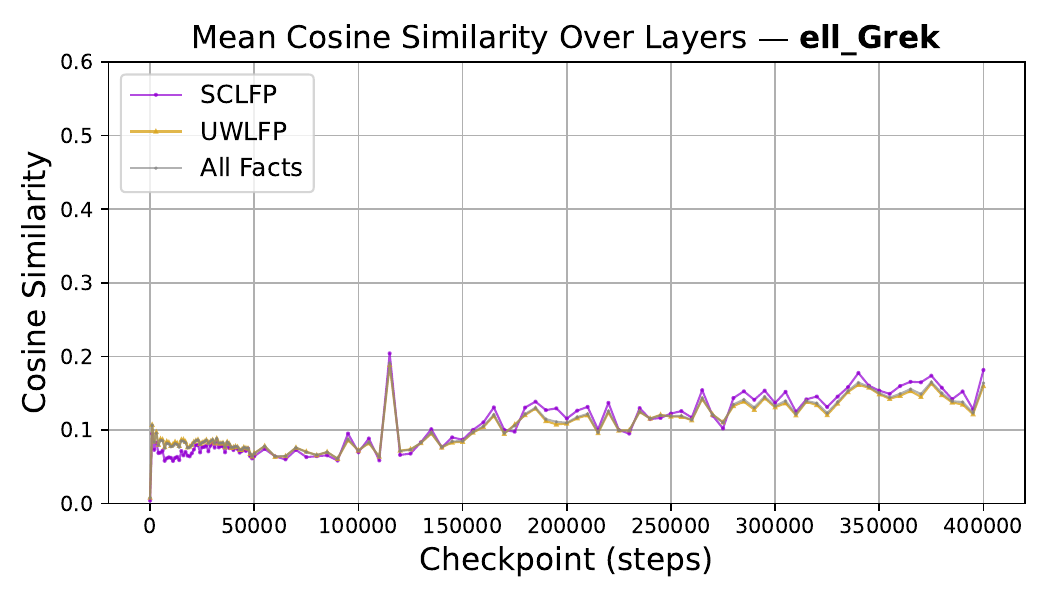}
    \includegraphics[width=0.32\textwidth]{figures/similarity/fra_Latn_lasttoken_mean_similarity.pdf}
    \includegraphics[width=0.32\textwidth]{figures/similarity/jpn_Jpan_lasttoken_mean_similarity.pdf}
    \includegraphics[width=0.32\textwidth]{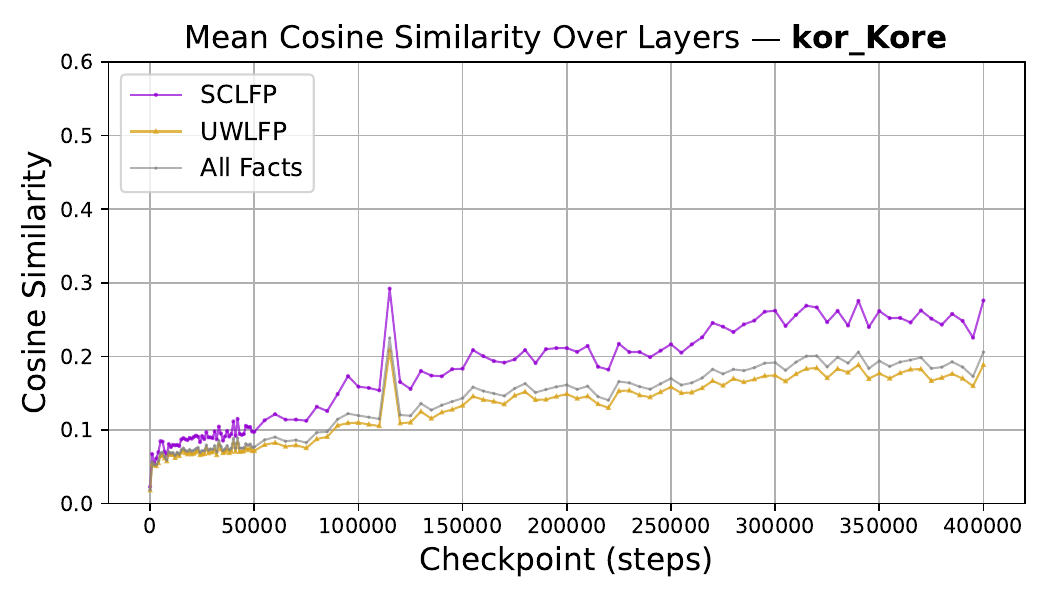}
    \includegraphics[width=0.32\textwidth]{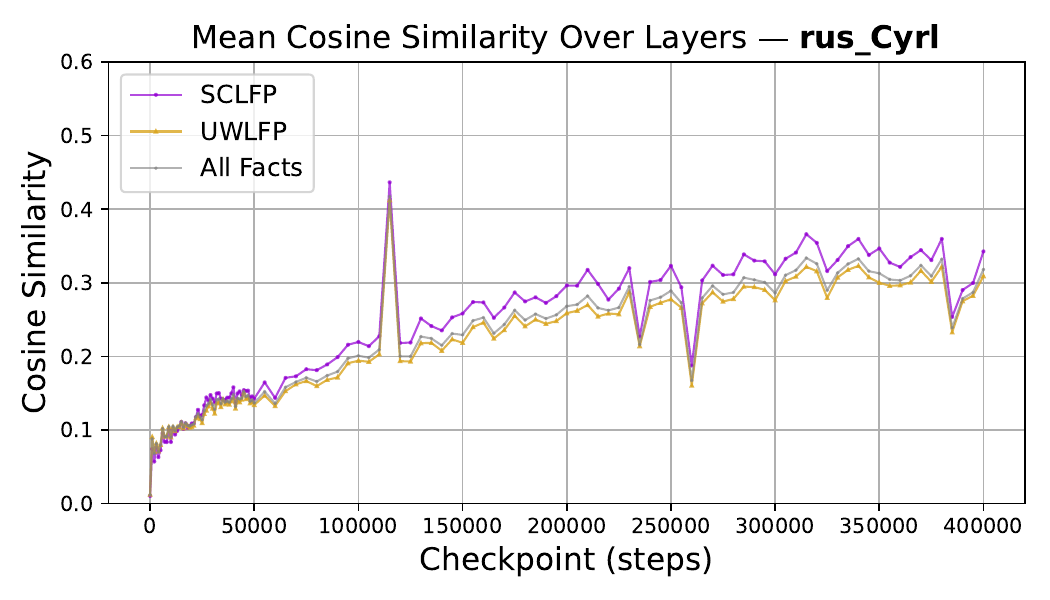}
    \includegraphics[width=0.32\textwidth]{figures/similarity/spa_Latn_lasttoken_mean_similarity.pdf}
    \includegraphics[width=0.32\textwidth]{figures/similarity/tur_Latn_lasttoken_mean_similarity.pdf}
    \includegraphics[width=0.32\textwidth]{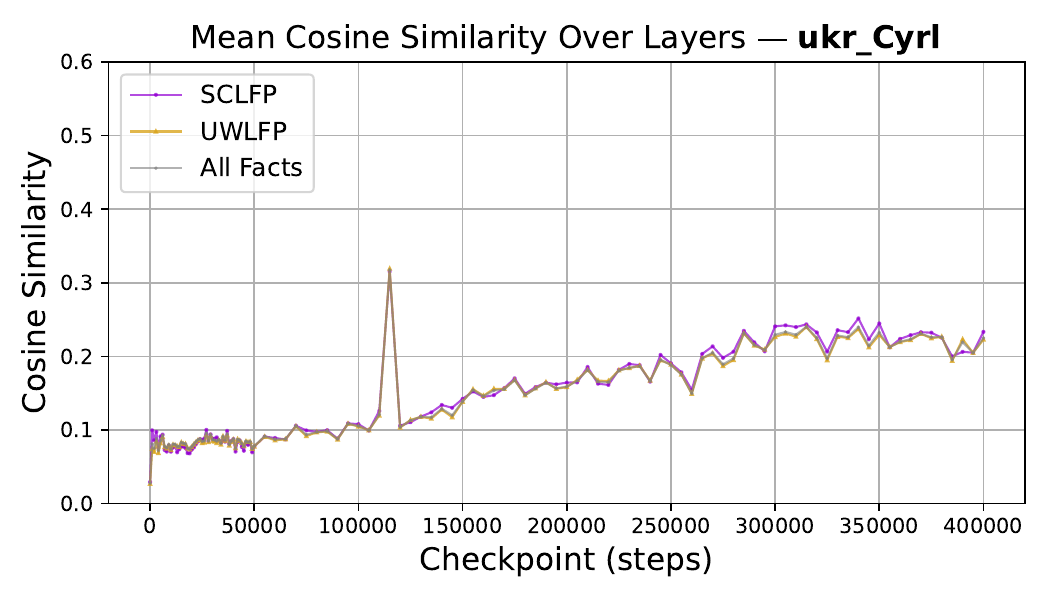}
    \caption{Complete results of mean cosine similarity for \falsenegatives, \truenegatives, and all facts between each language and English during pretraining.
    All languages exhibit higher similarity for \falsenegatives compared to \truenegatives, indicating crosslingual transfer based on better aligned representations.
    }
    \label{fig:similarity_over_checkpoints_complete}
\end{figure*}

To supplement the representative trends shown in Figure~\ref{fig:similarity_over_checkpoints}, we present the full set of similarity dynamics across all 12 languages, as show in Figure~\ref{fig:similarity_over_checkpoints_complete}. 
These plots track the mean cosine similarity between contextualized representations of fact pairs (one in English and one in the target language) across training checkpoints. 
We separately report trends for \falsenegatives, \truenegatives, and all fact pairs, enabling a detailed view into how representation alignment evolves throughout pretraining.

Across languages and scripts, we consistently observe that \falsenegatives exhibit greater similarity with English than \truenegatives.
Since both \falsenegatives and \truenegatives are low-frequency facts, the similarity gap indicates that \truenegatives are correctly recalled because their representations are better aligned with their English counterparts, while \truenegatives in each language are less similar compared to the English counterparts and thus fail to benefit from crosslingual transfer.
One interesting case is ukr\_Cryl, where the gap between \falsenegatives and \truenegatives is not pronounced.
We hypothesize that ukr\_Cryl benefits crosslingual transfer more from rus\_Cryl instead of English because of shared script. 
The higher crosslingual consistency in the 400K-step model (cf. Figure~\ref{fig:holistic_consistency}) and continuously improving consistency in pretraining (cf. Figure~\ref{fig:holistic_consistency_checkpoint}) support our hypothesis.
These full-language plots further strengthen our claim: pretraining on English benefits other languages not just through shared tokens or frequency-based priors, but also through crosslingual transfer from representational alignment, which goes beyond script boundaries. 

% These full-language plots further strengthen our central claim: 
% pretraining on English benefits other languages not just through shared lexical tokens or frequency-based priors, but also through alignment at the representational level. 
% The consistency of these trends across languages and scripts highlights the robustness of crosslingual transfer mechanisms in multilingual language models—even when surface-level features differ.

%% file: transferred_facts.tex
\section{Complete Learning Dynamics on \falsenegatives{}s}\seclabel{complete_fns_dynamics}

\begin{figure*}
    \centering
    \includegraphics[width=0.24\textwidth]{figures/transferred_facts/ar.pdf}
    \includegraphics[width=0.24\textwidth]{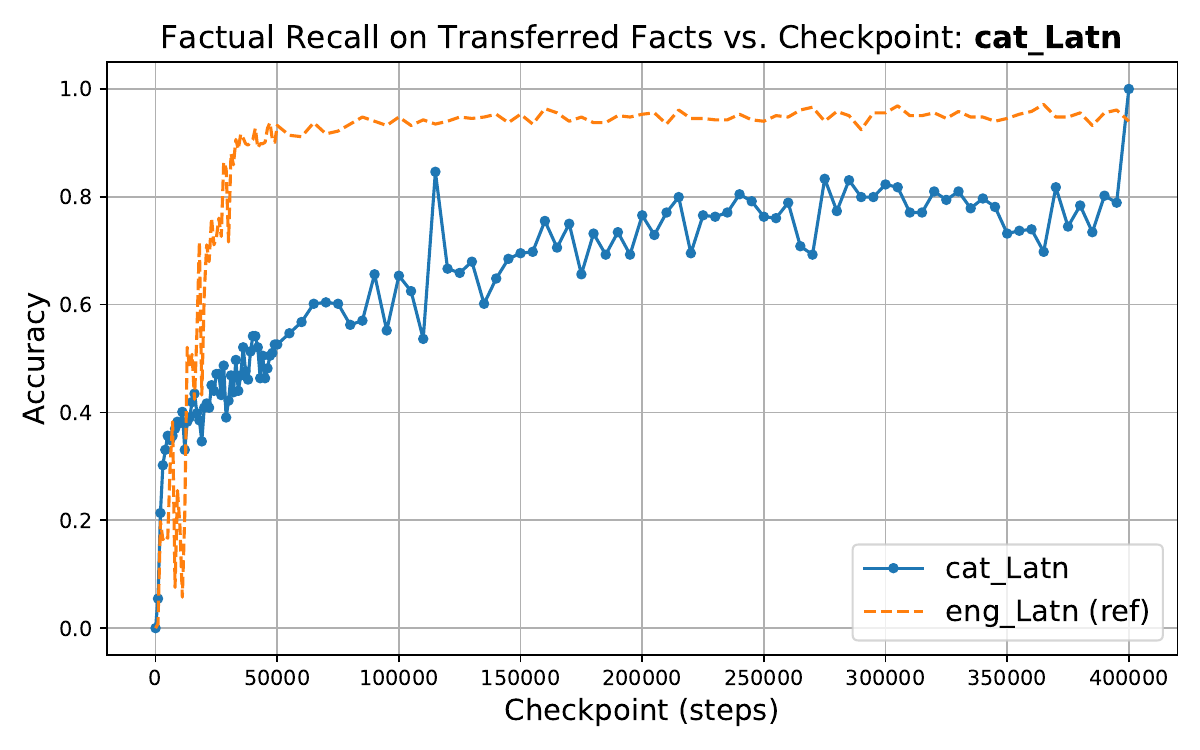}
    \includegraphics[width=0.24\textwidth]{figures/transferred_facts/el.pdf}
    \includegraphics[width=0.24\textwidth]{figures/transferred_facts/es.pdf}
    \includegraphics[width=0.24\textwidth]{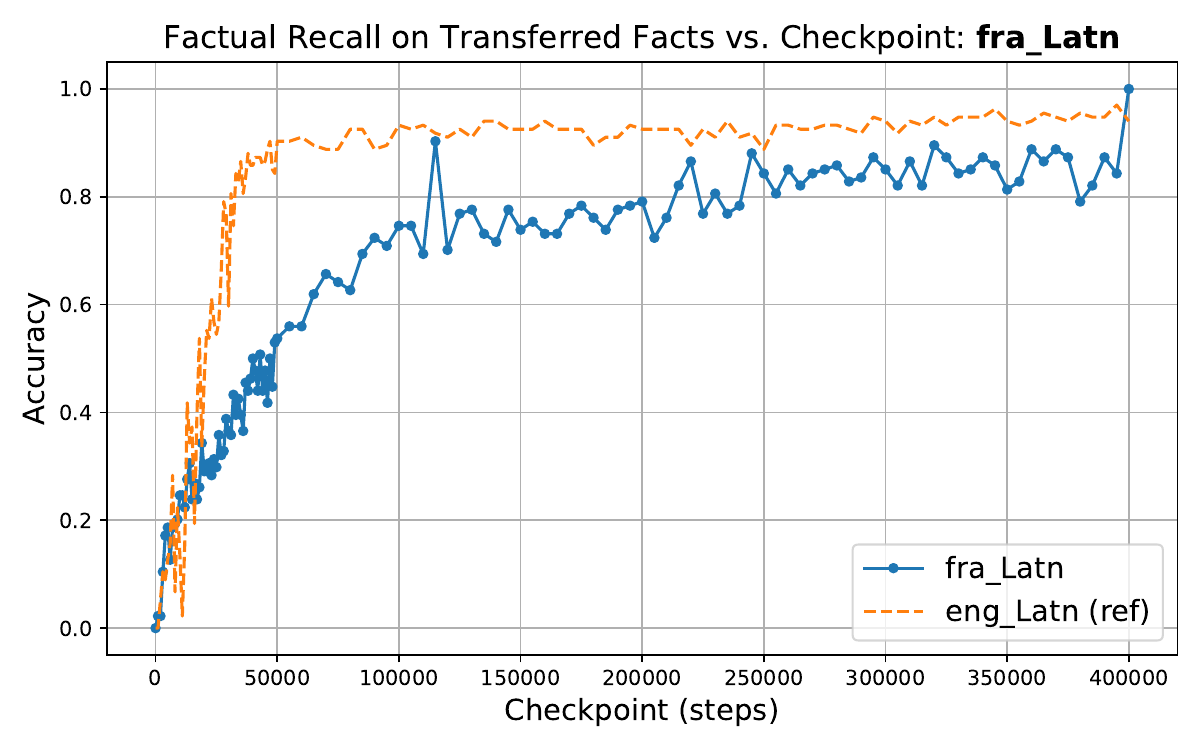}
    \includegraphics[width=0.24\textwidth]{figures/transferred_facts/ja.pdf}
    \includegraphics[width=0.24\textwidth]{figures/transferred_facts/ko.pdf}
    \includegraphics[width=0.24\textwidth]{figures/transferred_facts/ru.pdf}
    \includegraphics[width=0.24\textwidth]{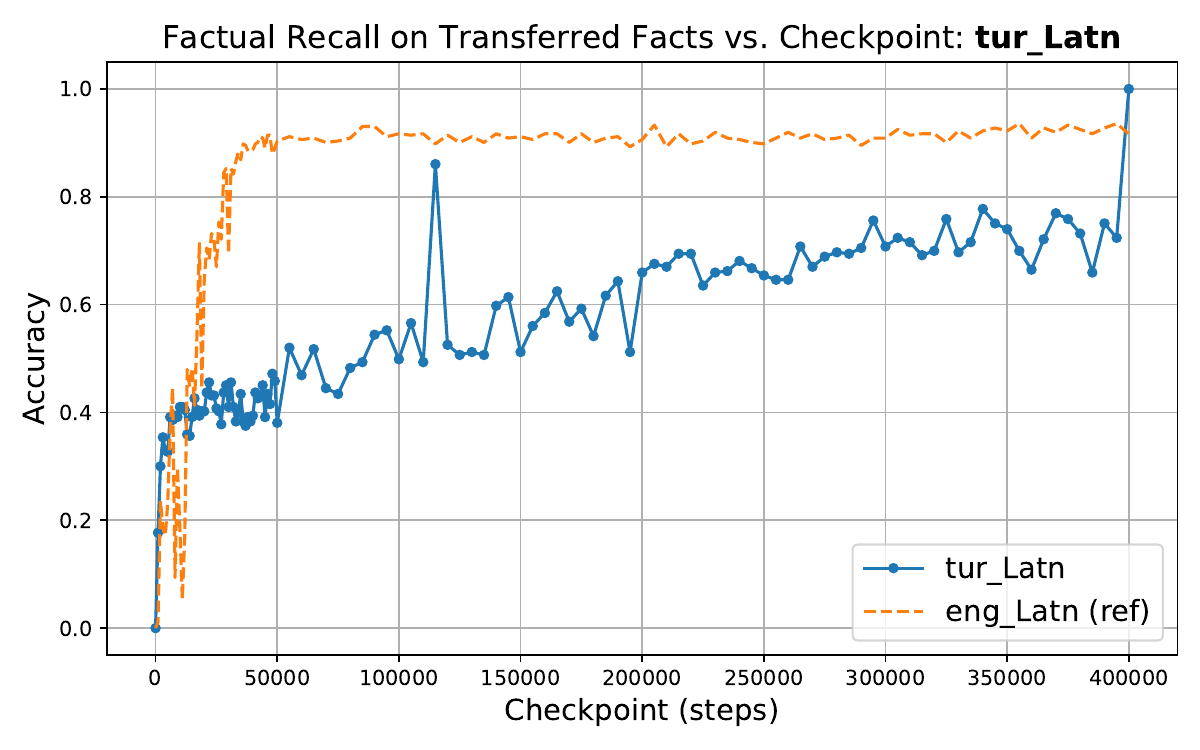}
    \includegraphics[width=0.24\textwidth]{figures/transferred_facts/uk.pdf}
    \includegraphics[width=0.24\textwidth]{figures/transferred_facts/zh.pdf}
    \caption{Dynamics of learning on the \falsenegatives{}s
(surprisingly correct low frequency predictions, i.e., FNs in Table \protect\ref{tab:frequency_threshold})
across all languages.
}
    \label{fig:transferred_facts_over_checkpoints_complete}
\end{figure*}

We present the learning trajectories of \falsenegatives{}s across all languages in Figure~\ref{fig:transferred_facts_over_checkpoints_complete}.

\section{Complementary Analysis of Facts}
\seclabel{appendix:transferred_facts}

To gain a deeper understanding of how factual knowledge in different languages benefits from English-centric pretraining, we conduct a complementary analysis focusing on surface-level features of facts, particularly the overlap in object strings across languages.

\subsection{Same Object Effect}

We hypothesize that facts in a language $l$ that share the \textbf{same object string} as their English counterparts are more likely to benefit from transfer during pretraining. 
To investigate this, we report in Table~\ref{tab:same_object_percentages} the proportion of facts in each language that share the same object with English, grouped by \falsenegatives and \textbf{non-\falsenegatives} according to our threshold-based classification (cf. \secref{prediction}).

We find that very few \falsenegatives share identical objects with English.
This is expected since \falsenegatives in each language have low frequencies.\footnote{If a fact in a language has low frequency, it is very unlikely that it shares the same object with its English counterpart.}
This finding, actually, further supports our claim that crosslingual transfer in \falsenegatives arises from deeper representational alignment (c.f. \secref{similarity}), not from trivial lexical overlap.
In contrast, a substantial number of non-\falsenegatives (which are mostly high-frequency facts) do share the same object string with English, especially in Latin-script languages.

\begin{figure}
    \centering
    \includegraphics[width=0.48\textwidth]{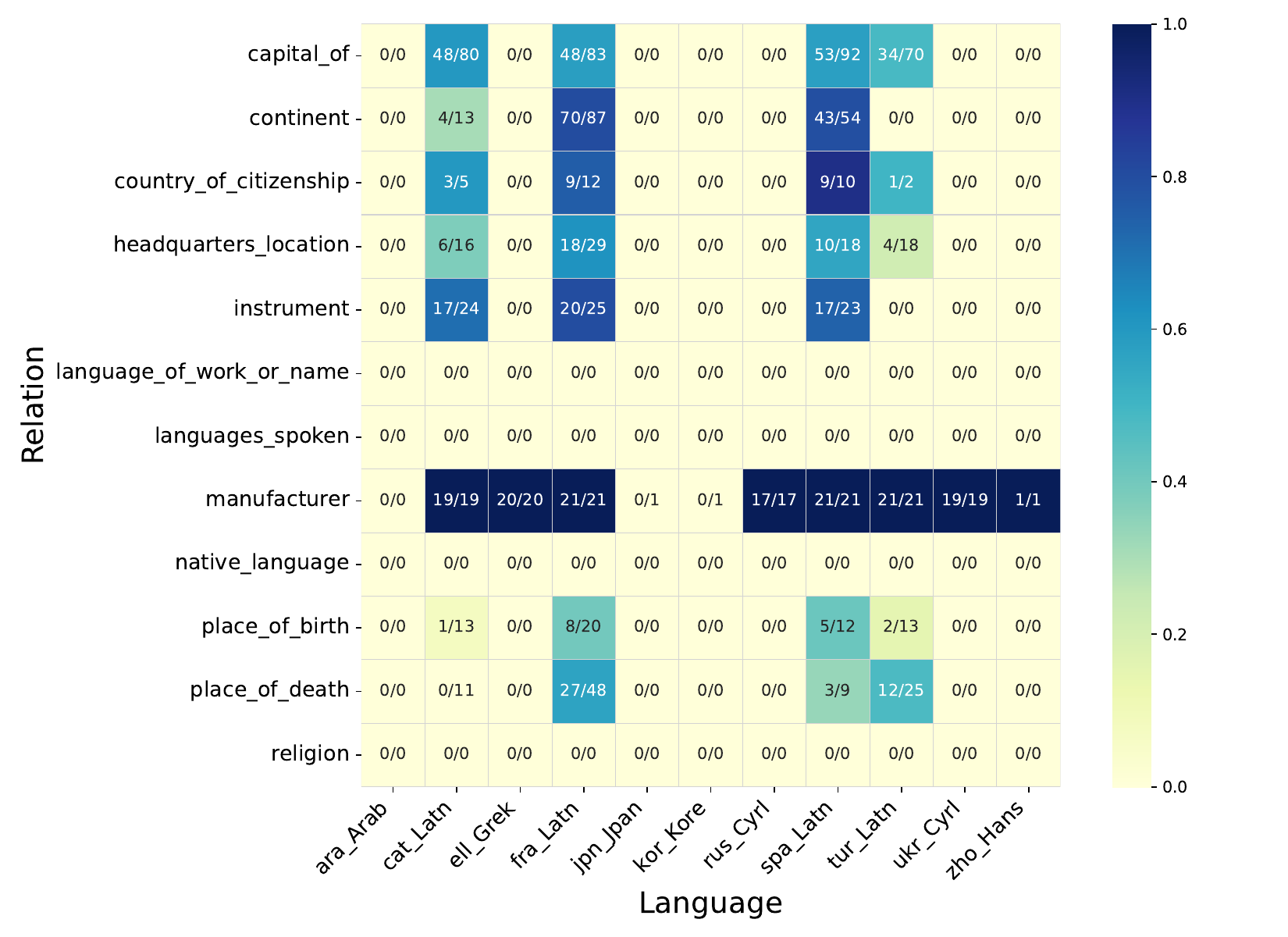}
    \caption{Distribution of \textbf{identical-object facts} across relation types for each language. A cell labeled ``$17/24$'' indicates that 17 out of 24 facts are correctly recalled, where the 24 facts are those whose English counterparts are also correctly predicted. 
    Cells marked ``$0/0$'' indicate that no such facts exist for that relation in the given language. 
    The results suggest that many languages, particularly those using the Latin script, benefit from sharing identical object strings with English.}

    \label{fig:identical_object_fact_dist}
\end{figure}

To further understand the influence of object overlap, we select the subset of facts in each language whose English counterpart (i.e., same fact index) is correctly recalled by the model. 
These \textbf{identical-object facts} are strong candidates for crosslingual transfer from English via lexical alignment.
Figure~\ref{fig:identical_object_fact_dist} shows the distribution of these facts across relation types, along with the proportion of them that are correctly recalled in each language. 
The results confirm our expectations: Latin-script languages show consistently high recall rates for identical-object facts across multiple relation types. 
We also observe meaningful gains in non-Latin-script languages, particularly in the \texttt{manufacturer} relation, where object strings often reference brand names directly borrowed from English (e.g., ``Apple''). 
These findings further highlight how both representational and lexical factors contribute to multilingual factual recall.

\begin{table*}
\centering
\begin{tabular}{lrrr|rrr}
\toprule
\textbf{Language} &  \#\textbf{\falsenegatives} &  \#object matched & ratio &  \#\textbf{non-\falsenegatives} & \#object matched & ratio \\
\midrule
\textbf{tur\_Latn} &       373 &        14 &          3.8\% &          824 &          149 &            \textbf{18.1\%} \\
\textbf{spa\_Latn} &       169 &         6 &          3.6\% &         1028 &          239 &            \textbf{23.2\%} \\
\textbf{cat\_Latn} &       384 &        11 &          2.9\% &          813 &          181 &            \textbf{22.3\%} \\
\textbf{fra\_Latn} &       134 &        16 &         11.9\% &         1063 &          325 &            \textbf{30.6\%} \\
\textbf{ara\_Arab} &       209 &         0 &          0.0\% &          988 &            0 &            0.0\% \\
\textbf{zho\_Hans} &       296 &         0 &          0.0\% &          901 &            1 &            \textbf{0.1\%} \\
\textbf{rus\_Cyrl} &       330 &         0 &          0.0\% &          867 &           17 &            \textbf{2.0\%} \\
\textbf{jpn\_Jpan} &       212 &         0 &          0.0\% &          985 &            1 &            \textbf{0.1\%} \\
\textbf{ukr\_Cyrl} &       248 &         0 &          0.0\% &          949 &           19 &            \textbf{2.0\%} \\
\textbf{kor\_Kore} &       238 &         0 &          0.0\% &          959 &            1 &            \textbf{0.1\%} \\
\textbf{ell\_Grek} &       190 &         0 &          0.0\% &         1007 &           20 &            \textbf{2.0\%} \\
\bottomrule
\end{tabular}
\caption{Statistics of object agreement with English in \falsenegatives and \textbf{non-\falsenegatives} across languages. 
Many Latin-script languages tend to have a higher proportion of identical objects in \textbf{non-\falsenegatives} compared to \falsenegatives.
}
\label{tab:same_object_percentages}
\end{table*}

%% file: removing_identical_facts.tex
\section{Effects of Excluding Identical Facts Across Languages}\seclabel{exclude}

% in this section, we exclude all facts in each language that have an identical subject-object pair in any other language, and see how this actually changes the distribution and our previous results 

In \secref{frequency}, we show that fact frequency can reliably predict the factual recall accuracy.
The frequency of each fact is approximated by counting the number of documents where the subject and object strings of a fact co-occur.
Although this measure has been widely used in previous research \citep{elazar2023measuring,merullo2025linear}, there might be a further underlying confounding variable in the multilingual context.
If two languages use the same subject/object strings for a fact, then the frequency of that fact will be the same in the two languages.
This is particularly the case for Latin-script languages.
For example, both French and English use ``France'' and ``Paris'', so the subject-object pair will be identical and the two languages will have the same frequency for this fact, even if sometimes the fact occurs in French text while sometimes in English text.
In other words, many fact frequencies will be \textbf{aggregated statistics} over multiple script-sharing languages.\footnote{Of course, due to the shared tokens, every occurrence of subject/object strings will affect the recallability of the fact shared by multiple languages. Therefore, we simply use the aggregated statistics for each language in the main text.}
Therefore, we want to investigate how the results will be affected if this confounding variable is excluded.

We exclude facts in each language whose subject-object pairs match those in any other language (via string matching).
This results in fewer facts in each language, but the remaining facts in each language are not affected by other languages (at least the languages considered in this study).
Then we re-conduct the same investigation presented \secref{local_results} and \secref{prediction}.

\begin{table*}[ht]
\centering
\small
\setlength{\tabcolsep}{15pt}
\begin{tabular}{lrrrrrrrr}
\toprule
\textbf{Lang} & \textbf{Threshold} & \textbf{Accuracy} & \textbf{FP} & \textbf{FN} & \textbf{TP} & \textbf{TN} & \textbf{Total} \\
\midrule
ara\_Arab & 3485 & 0.83 & 0   & 209 & 1   & 987 & 1197 \\
cat\_Latn & 2506 & 0.60 & 17  & 359 & 25  & 549 & 950  \\
ell\_Grek & 483  & 0.84 & 2   & 190 & 4   & 970 & 1166 \\
eng\_Latn & 108  & 0.82 & 156 & 7   & 740 & 11  & 914  \\
fra\_Latn & 19   & 0.62 & 221 & 134 & 436 & 152 & 943  \\
jpn\_Jpan & 352  & 0.82 & 5   & 212 & 8   & 968 & 1193 \\
kor\_Kore & 402  & 0.80 & 0   & 238 & 1   & 957 & 1196 \\
rus\_Cyrl & 201  & 0.70 & 4   & 319 & 17  & 744 & 1084 \\
spa\_Latn & 5    & 0.59 & 281 & 106 & 391 & 155 & 933  \\
tur\_Latn & 3068 & 0.64 & 16  & 369 & 27  & 645 & 1057 \\
ukr\_Cyrl & 219  & 0.78 & 2   & 238 & 3   & 838 & 1081 \\
zho\_Hans & 502  & 0.75 & 7   & 296 & 17  & 872 & 1192 \\
\bottomrule
\end{tabular}
\caption{Best threshold, accuracy, and error breakdown (false positives, false negatives, true positives, and true negatives) for predicting factual recall correctness using fact frequency. For each language, we exclude facts whose subject-object pairs match those in any other language (via string matching). The results closely mirror those in Table~\ref{tab:frequency_threshold}, suggesting that identical subject-object facts across languages have minimal influence on the robustness of frequency predicting factual recall correctness, even for Latin-based languages and Cyrillic-based languages, which share many identical subject/objects for named entities.}
\label{tab:frequency_threshold_excluded}
\end{table*}

\begin{figure*}
    \centering
    \includegraphics[width=0.32\textwidth]{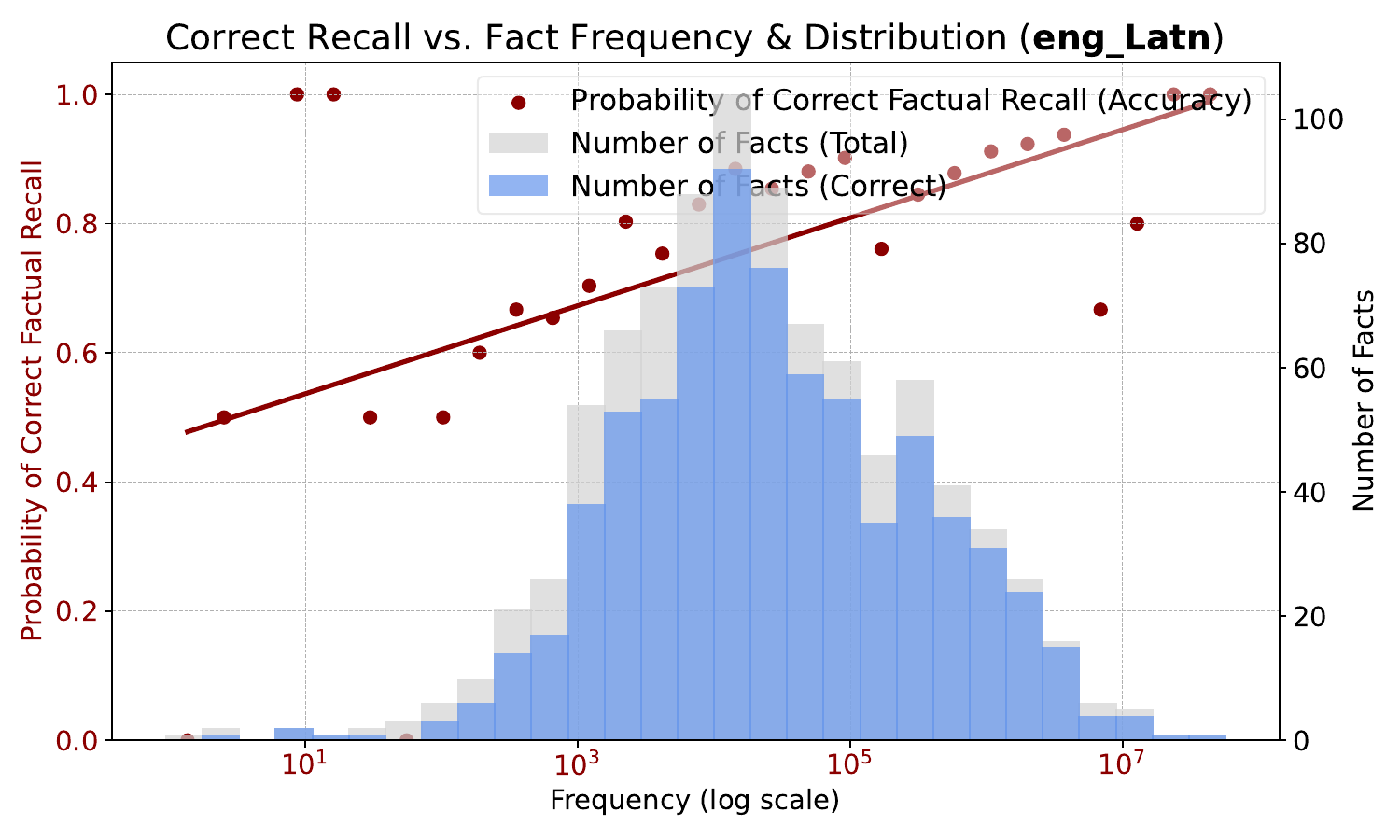}
    \includegraphics[width=0.32\textwidth]{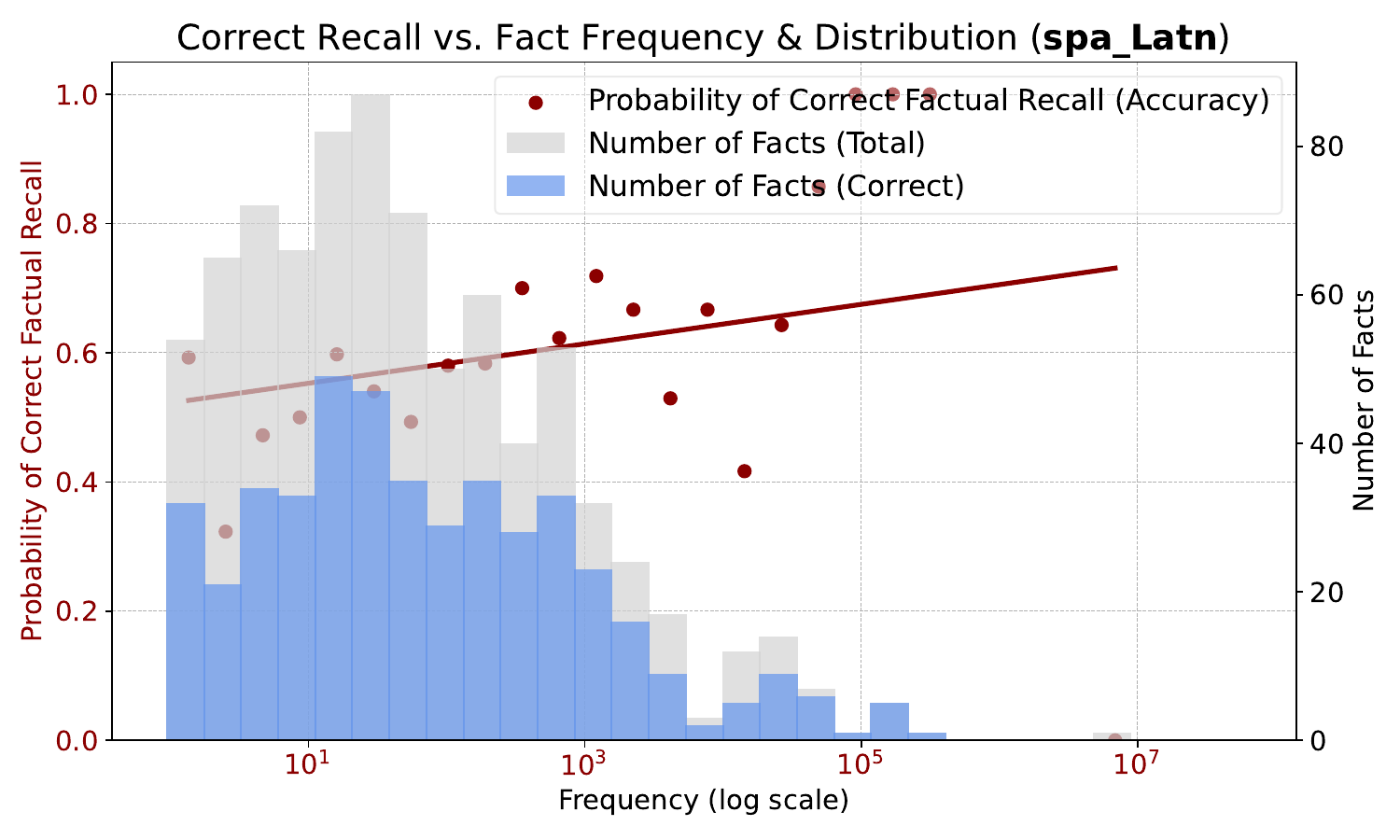}
    \includegraphics[width=0.32\textwidth]{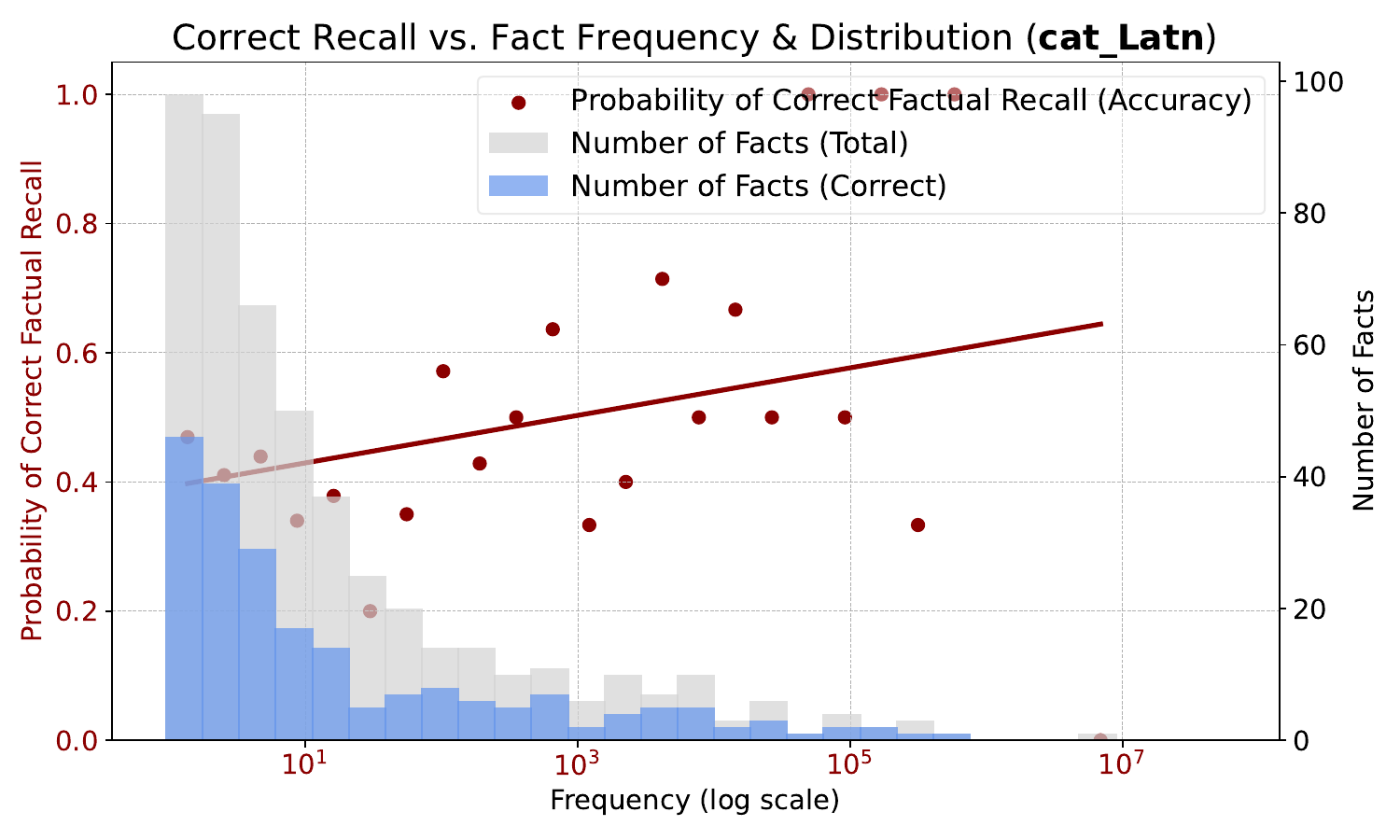}
    \includegraphics[width=0.32\textwidth]{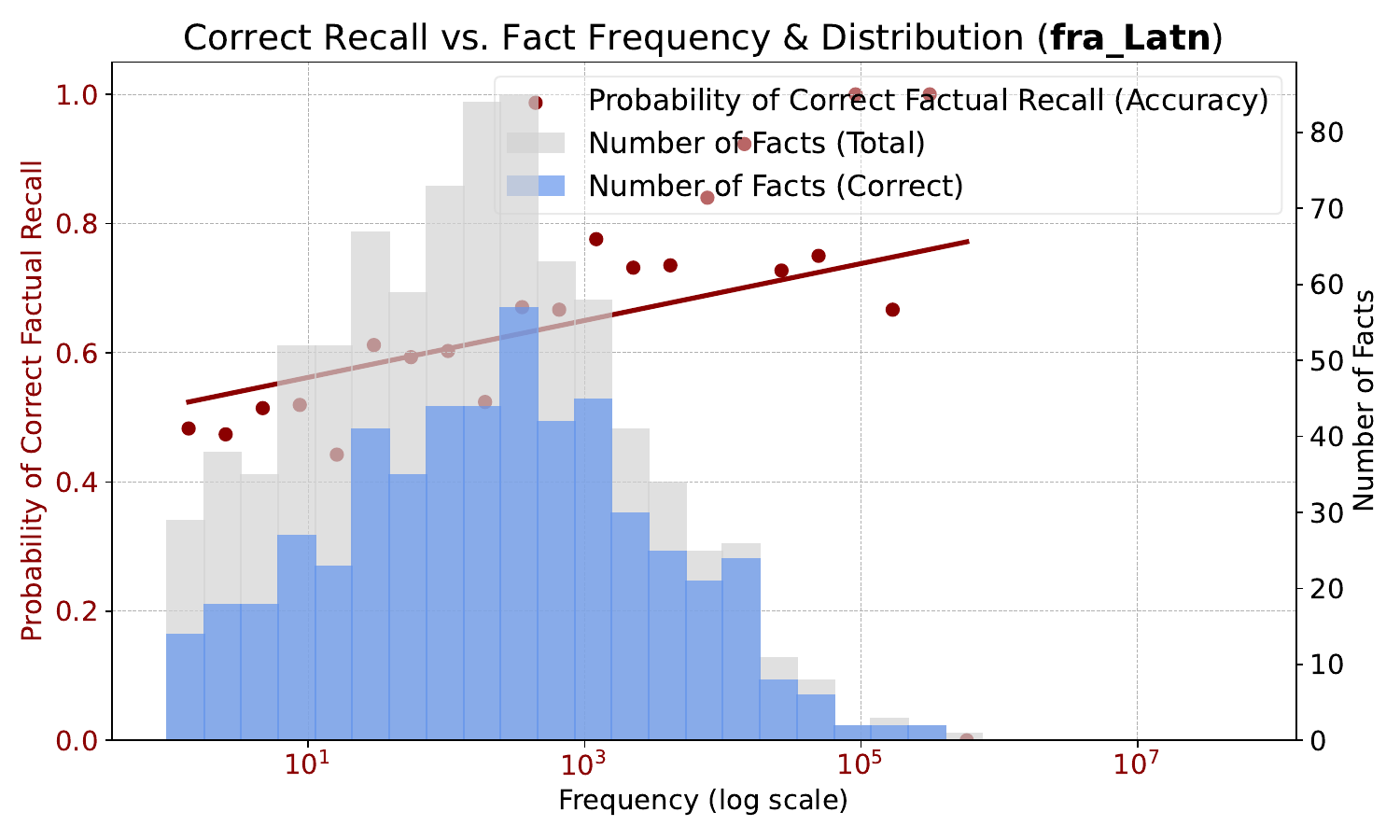}
    \includegraphics[width=0.32\textwidth]{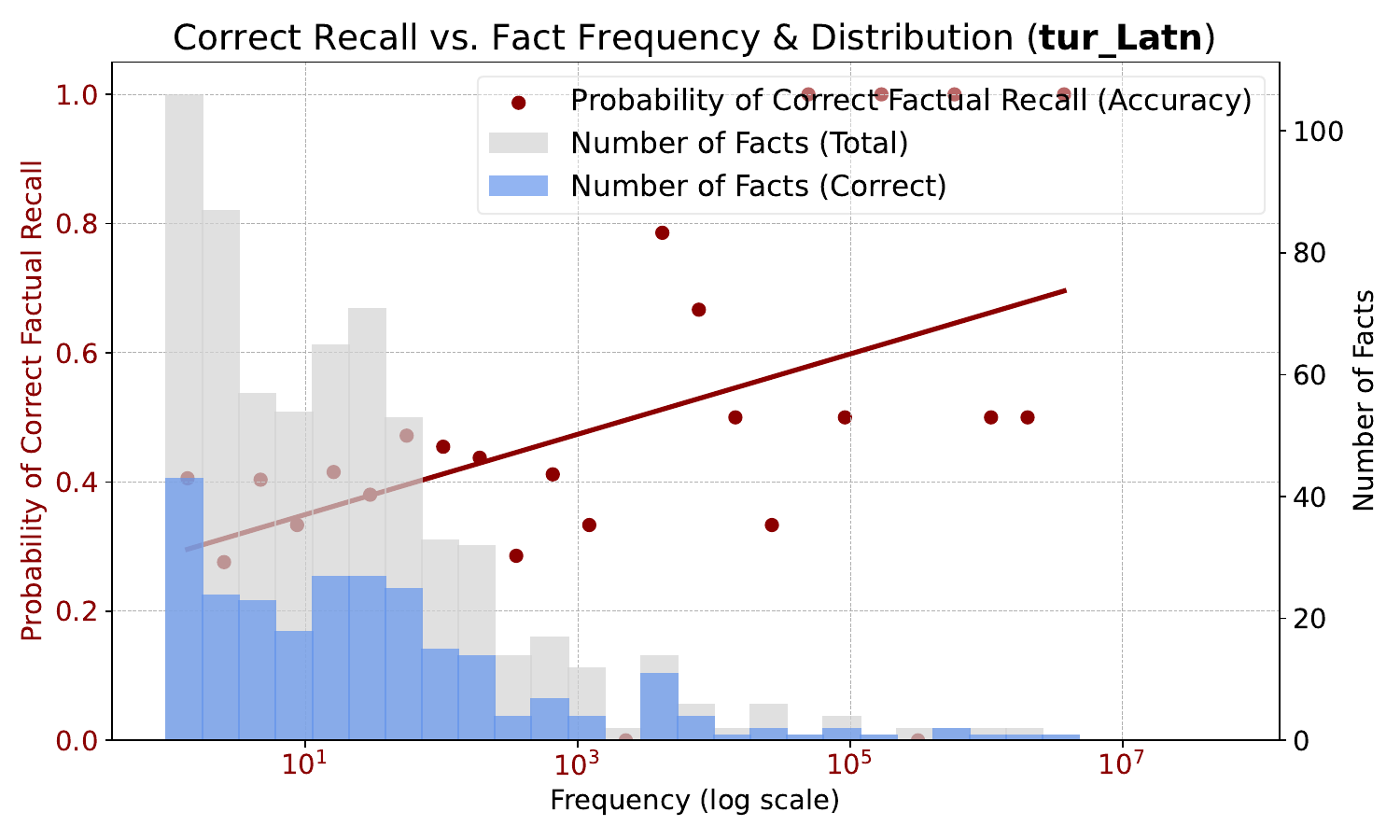}\\
    \includegraphics[width=0.32\textwidth]{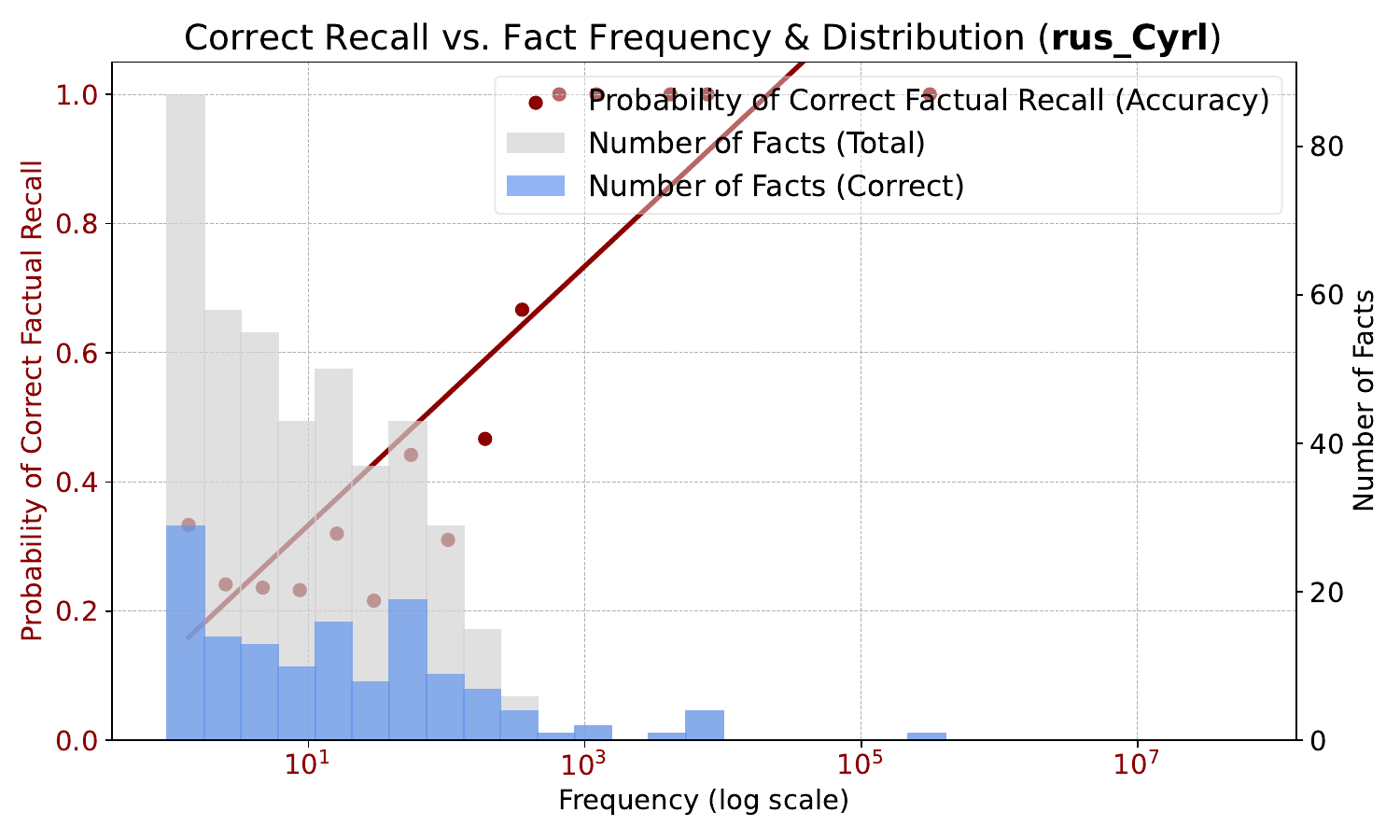}
    \includegraphics[width=0.32\textwidth]{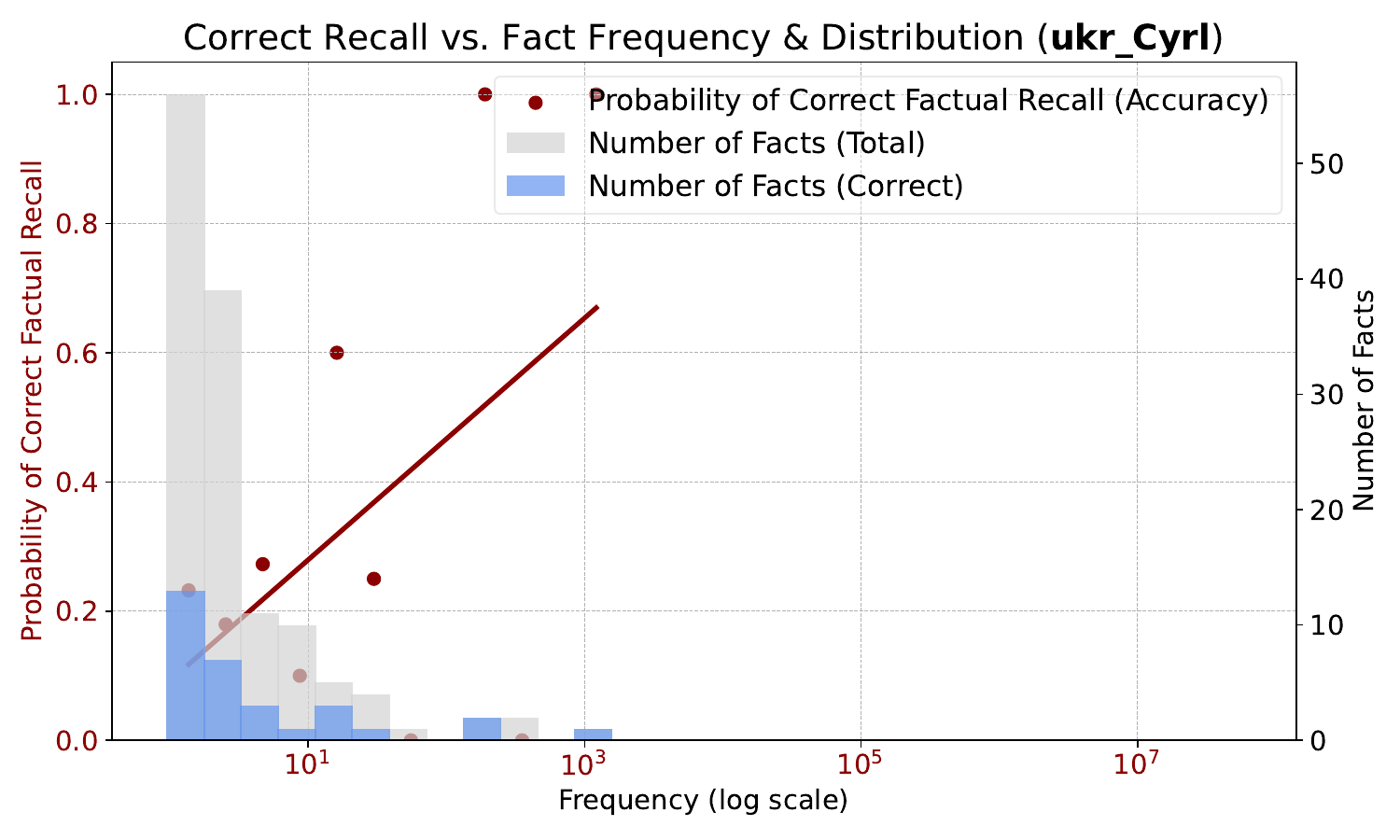}
\caption{Relationship between fact frequency and the probability of correct factual recall for \textbf{five Latin-script languages} (eng\_Latn, spa\_Latn, cat\_Latn, fra\_Latn, tur\_Latn) and \textbf{two Cyrillic-script languages} (ukr\_Cyrl, rus\_Cyrl) when excluding facts with subject-object pairs that exactly match those in any other languages. While shared script appears to influence the distribution of fact frequencies, a consistent trend remains across languages: higher fact frequency is associated with a higher possibility of correct factual recall.}\label{fig:correctness_frequency_local_excluded}
\end{figure*}

We first present the per-language relationship between fact frequency and factual recall for \textbf{five Latin-script languages} (eng\_Latn, spa\_Latn, cat\_Latn, fra\_Latn, tur\_Latn) and \textbf{two Cyrillic-script languages} (ukr\_Cyrl, rus\_Cyrl) in Figure~\ref{fig:correctness_frequency_local_excluded}.
We observe that, even though there are fewer facts in some languages compared with Figure~\ref{fig:correctness_frequency_local_complete}, where identical facts are not excluded, the trend still remains in each language: higher-frequency facts are more likely to be correctly predicted.

We then present the frequency-based classification for each language.
Similar to the setting in \secref{prediction}, the best threshold is selected by maximizing the overall accuracy.
Table~\ref{tab:frequency_threshold_excluded} shows the results.
We observe that there are almost no changes for languages that neither use Latin script nor Cyrillic script compared to Table~\ref{tab:frequency_threshold}.
This is expected since only a very tiny number of facts are removed from these languages.
On the other hand, we observe that there are some minor changes in Latin-script and Cyrillic-script languages.
These changes are mainly in the absolute number of FP, FN, TP, TN, and Total.
The best threshold has almost not changed at all except for spa\_Latn, rus\_Cyrl, and ukl\_Cyrl, indicating the robustness of classification and similar frequency distribution before and after removing the identical facts.
Since we are interested in false negatives -- facts with low frequencies that are correctly predicted, we also compute the agreement between false negatives before and after the identical facts are removed.
The overlapping rate is more than 98\% averaged across languages, indicating that the identical facts have almost no influence on the analysis presented in the main text.

%% file: dolma_documents.tex
\section{Multilingual Coverage in Dolma}\seclabel{dolma}

\begin{figure*}[ht]
    \centering
    \includegraphics[width=0.65\textwidth]{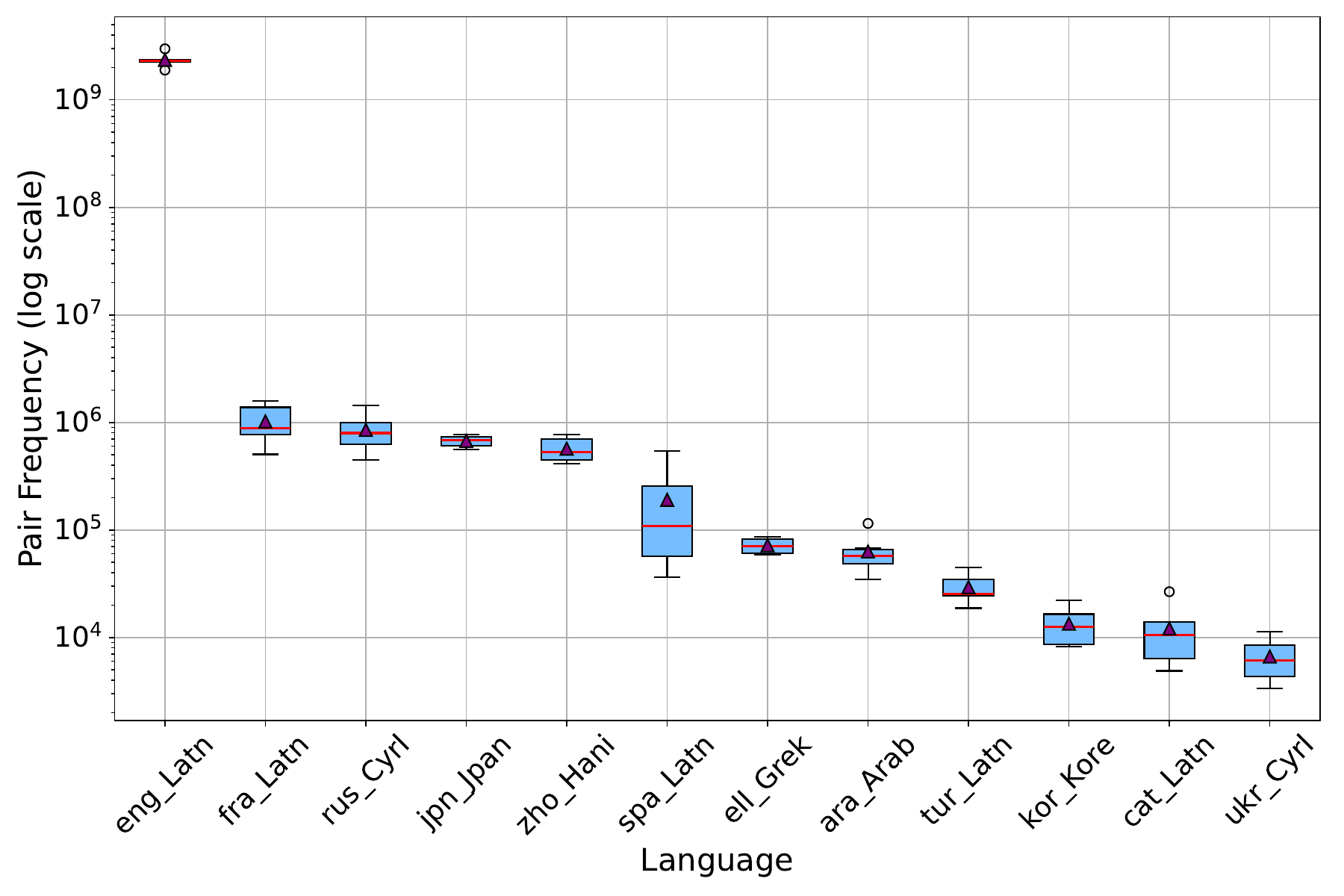}
    \caption{Pair frequency distribution (log scale) for the top four most frequent language-specific tokens in the Dolma corpus, measured across 12 languages.}
    \label{fig:dolma-lang-stat}
\end{figure*}

We estimate the coverage of Dolma for each language based on the frequency of token pairs. We tokenize the GlotLID Corpus~\citep{kargaran-etal-2023-glotlid}, a multilingual corpus comprising texts from diverse sources, using DataTrove tokenizers~\citep{penedo2024datatrove} specific to each language. From the tokenized output, we select the top four most frequent tokens that predominantly occur in one target language but not in the others. We then compute the frequencies of all unique, non-repetitive token pairs formed from these top tokens within the Dolma corpus. The results are presented in Figure~\ref{fig:dolma-lang-stat}. The low variance within each language’s boxplot indicates that the method offers a stable and reliable comparative measure of multilingual coverage. The figure reveals a substantial disparity in pair frequency across languages, ranging from high-resource languages such as French (fra\_Latn) to low-resource ones like Ukrainian (ukr\_Cyrl).

%% file: per_lang_per_relation.tex
\section{Per-Relation Dynamics Across Languages}\seclabel{per_relation}

\begin{figure*}
    \centering
    \includegraphics[width=0.24\textwidth]{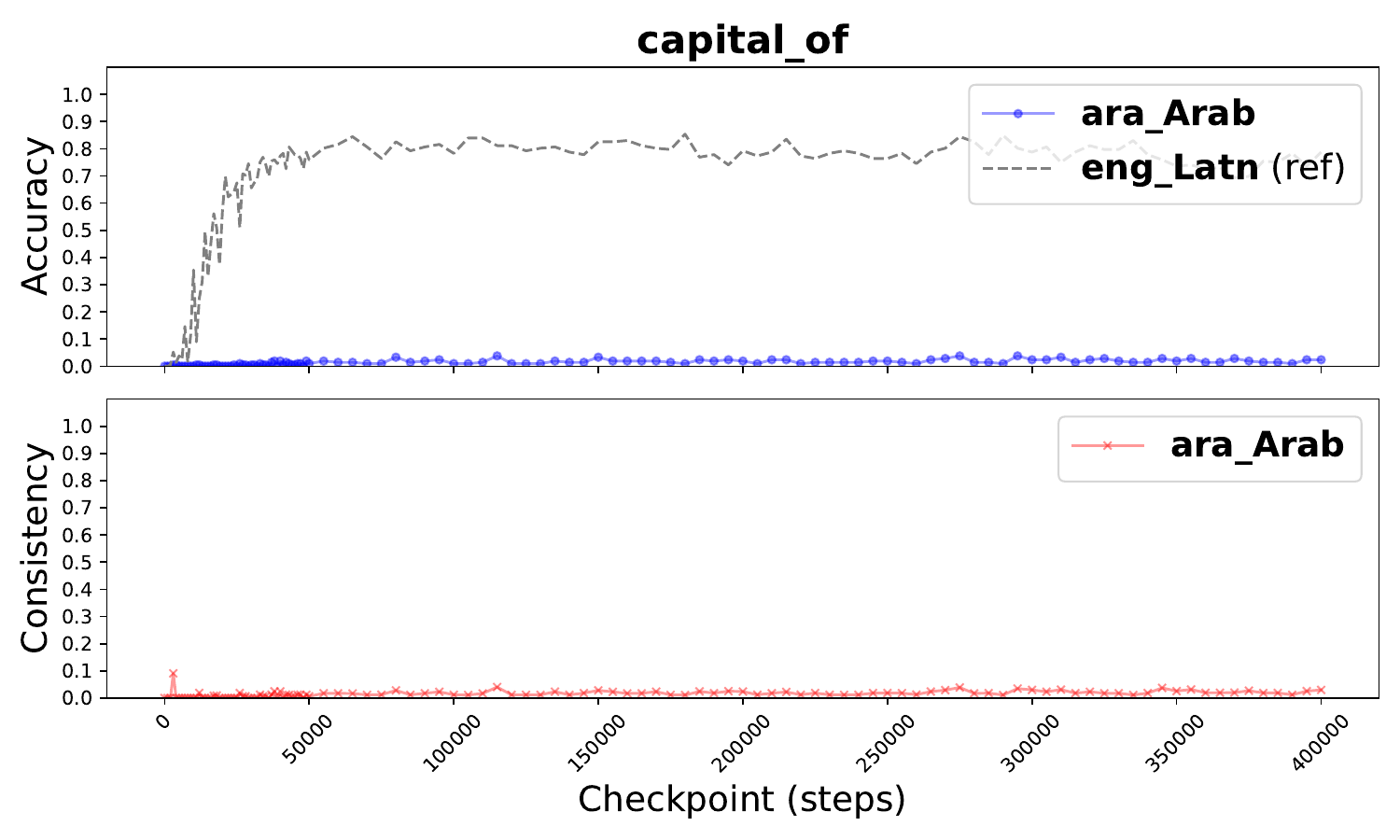}
    \includegraphics[width=0.24\textwidth]{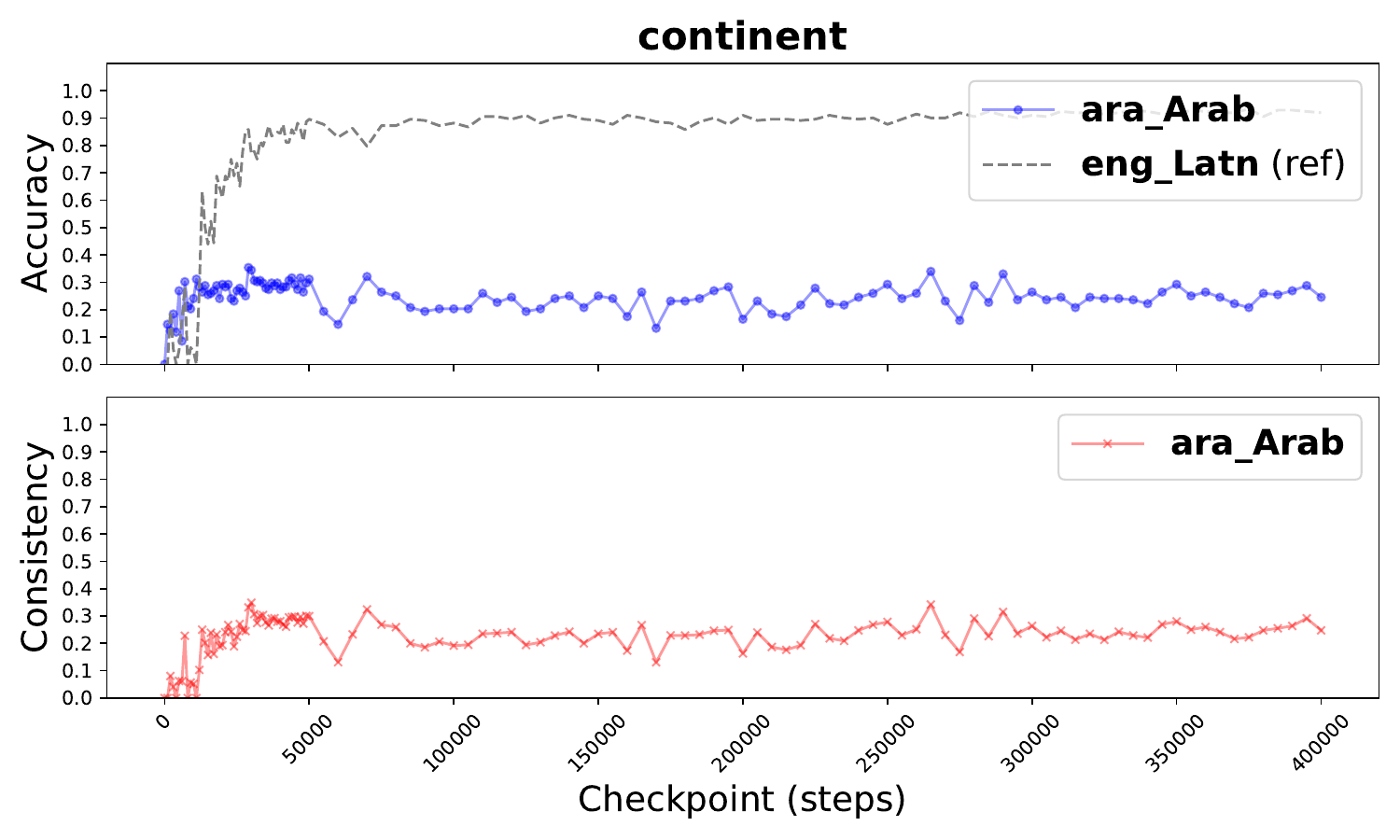}
    \includegraphics[width=0.24\textwidth]{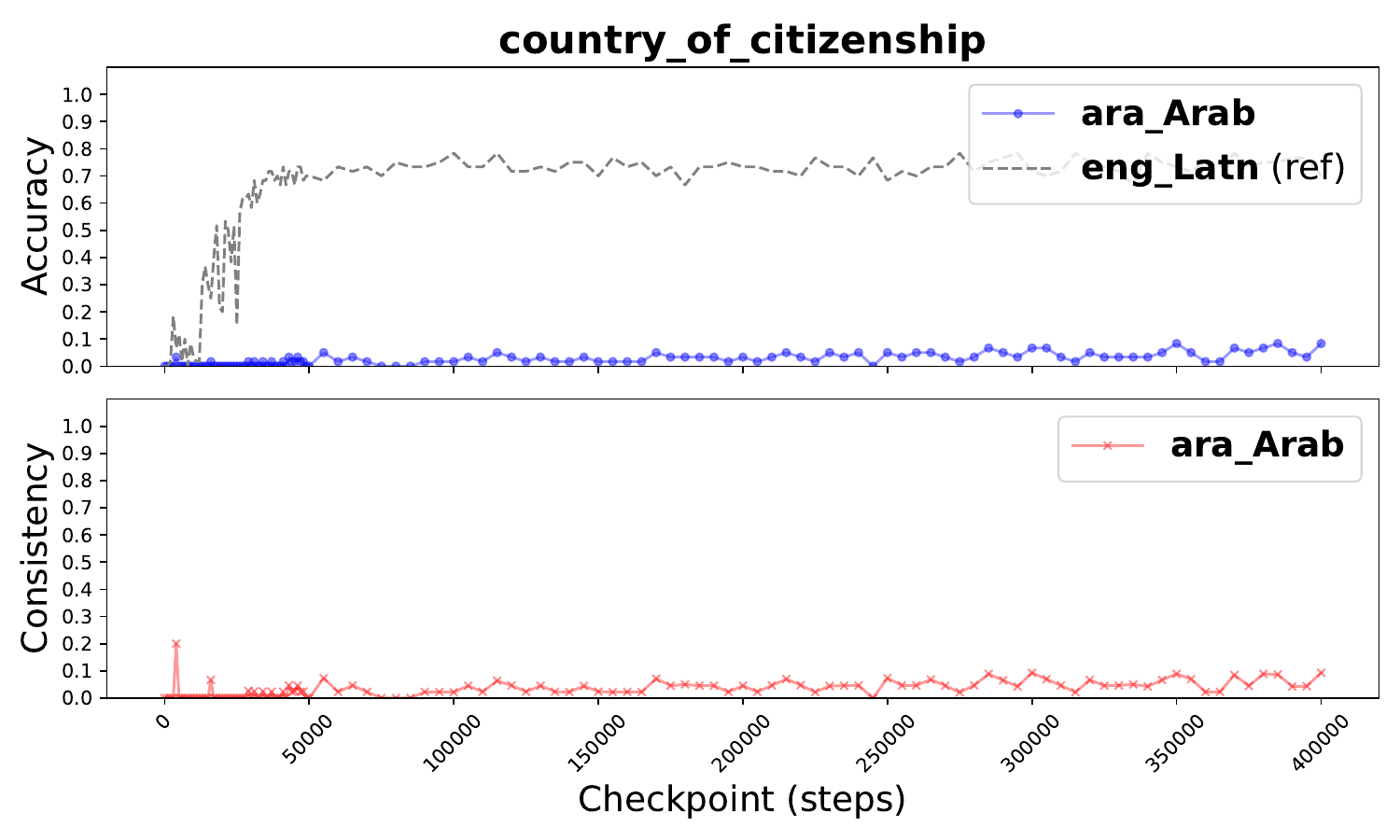}
    \includegraphics[width=0.24\textwidth]{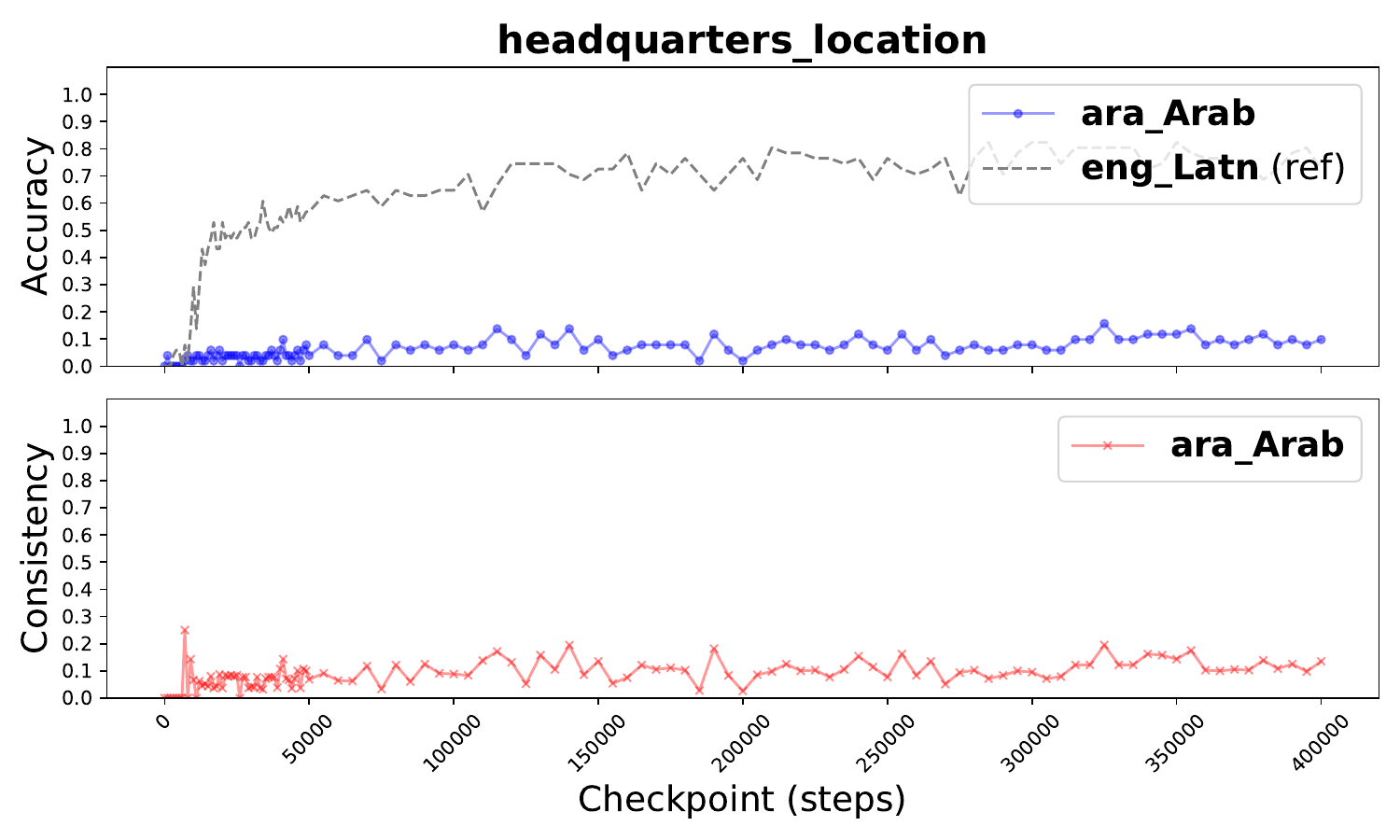}
    \includegraphics[width=0.24\textwidth]{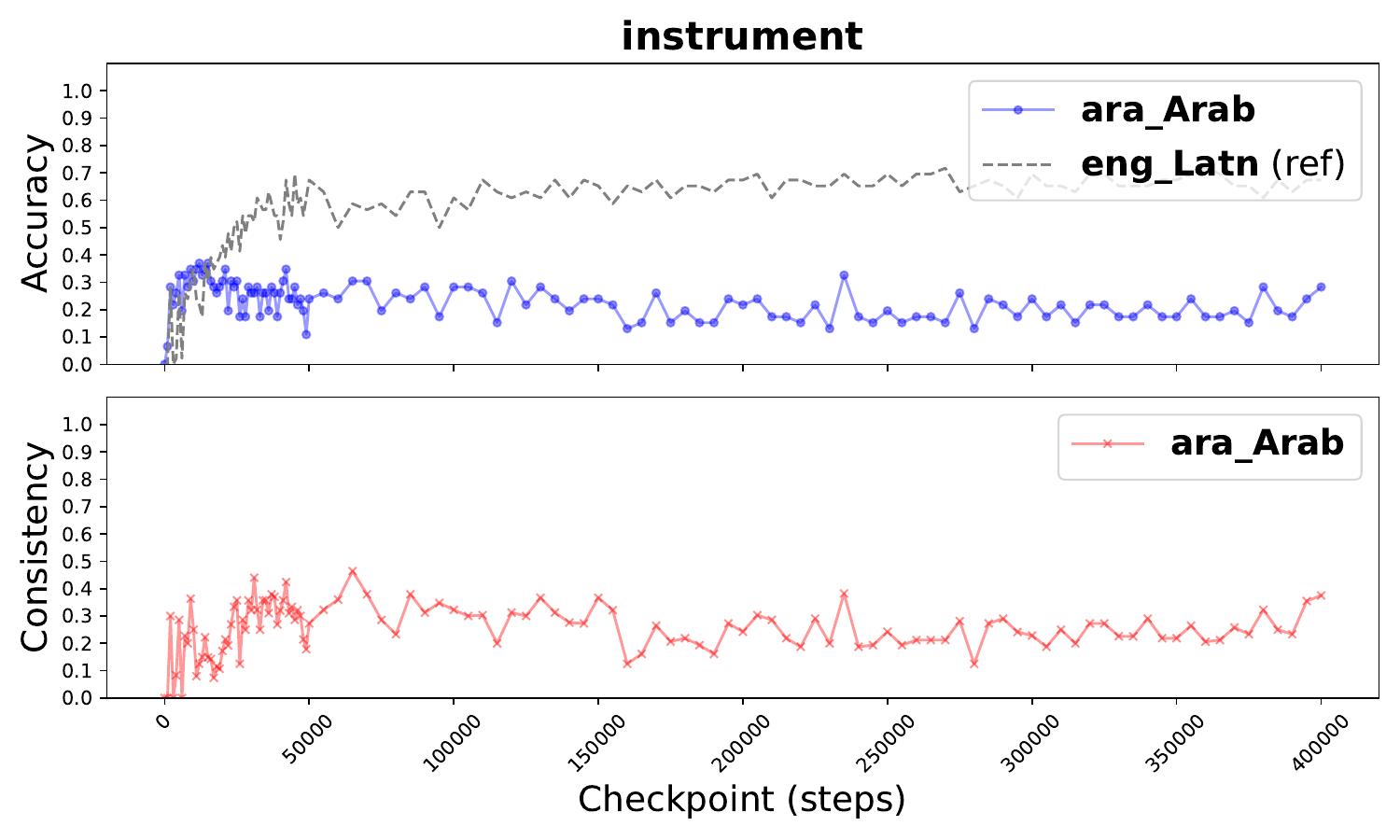}
    \includegraphics[width=0.24\textwidth]{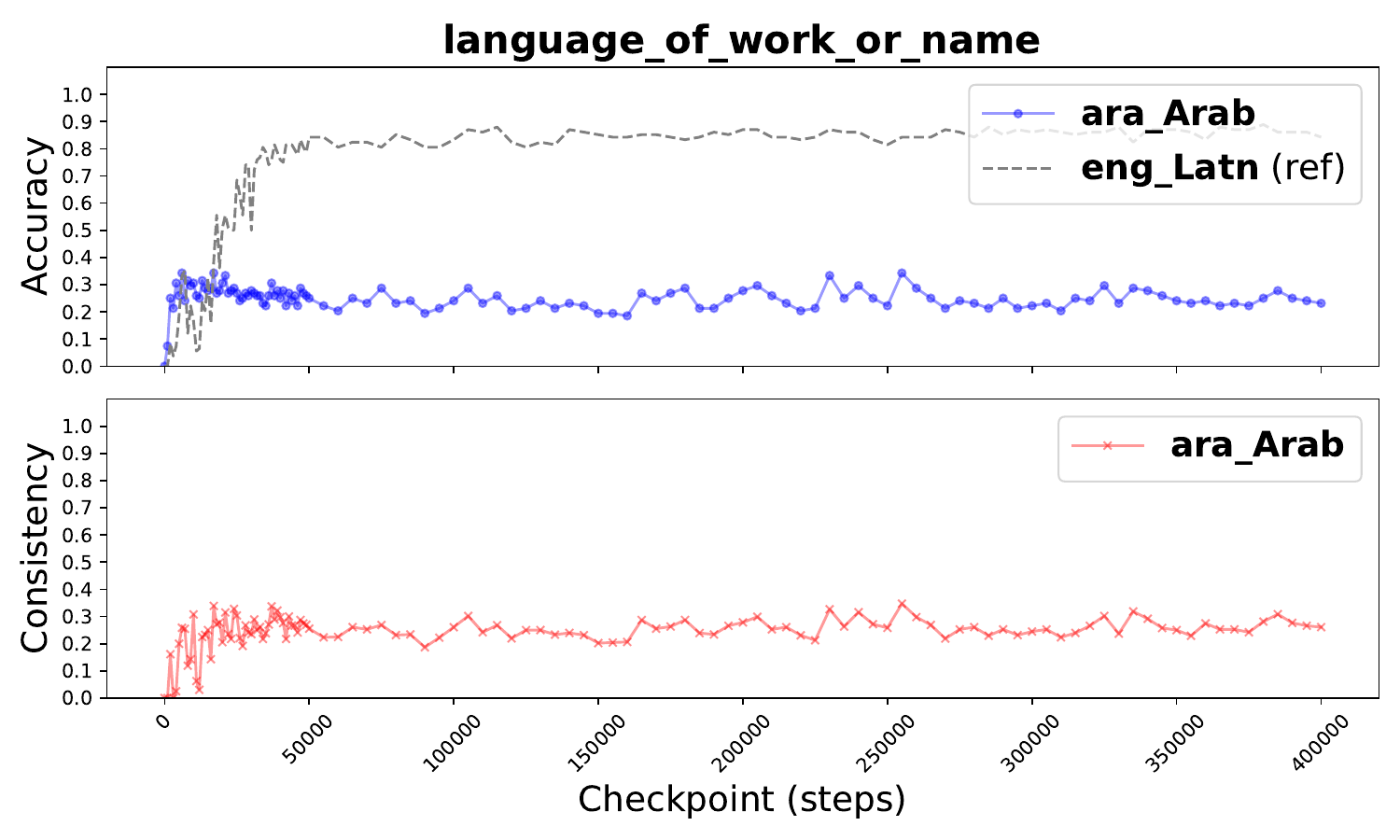}
    \includegraphics[width=0.24\textwidth]{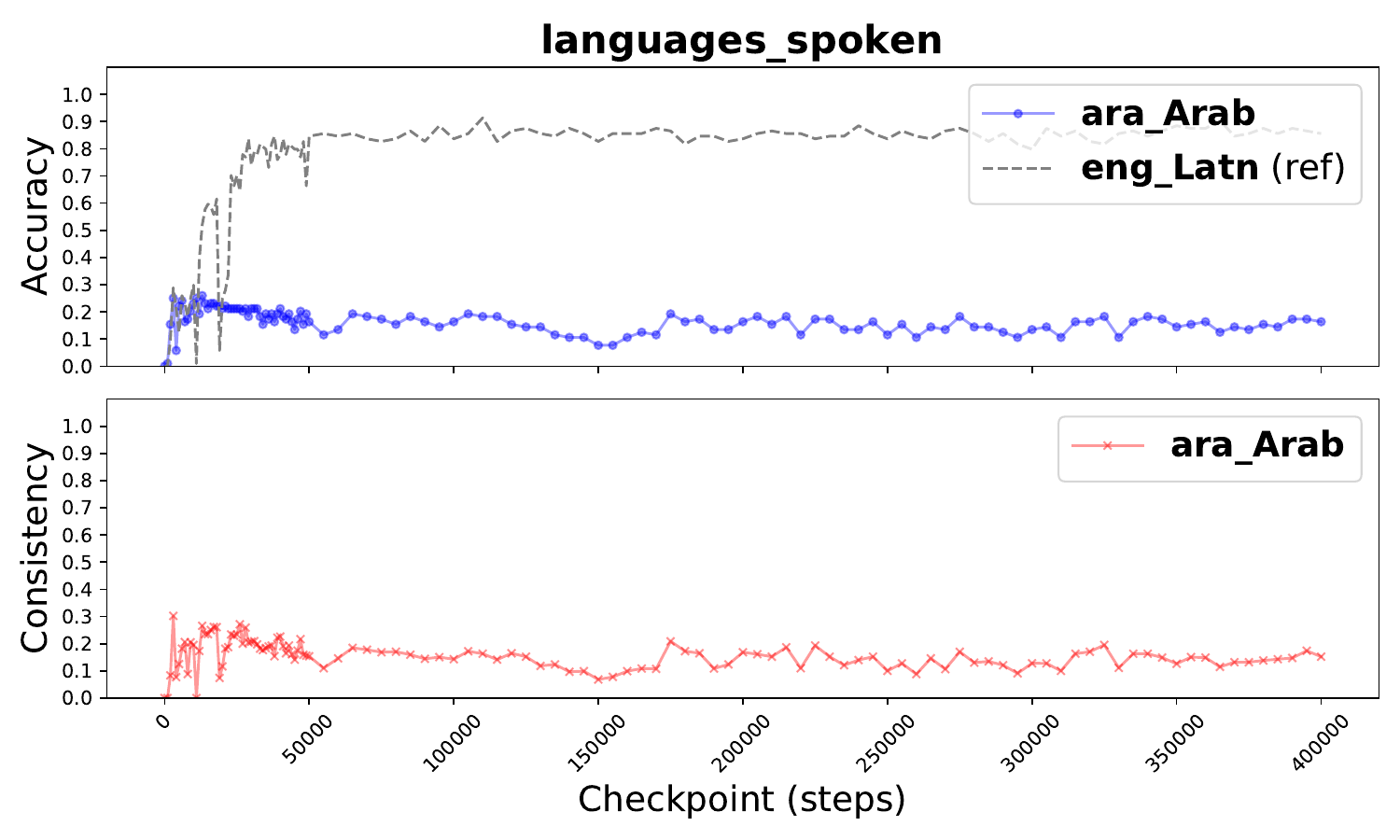}
    \includegraphics[width=0.24\textwidth]{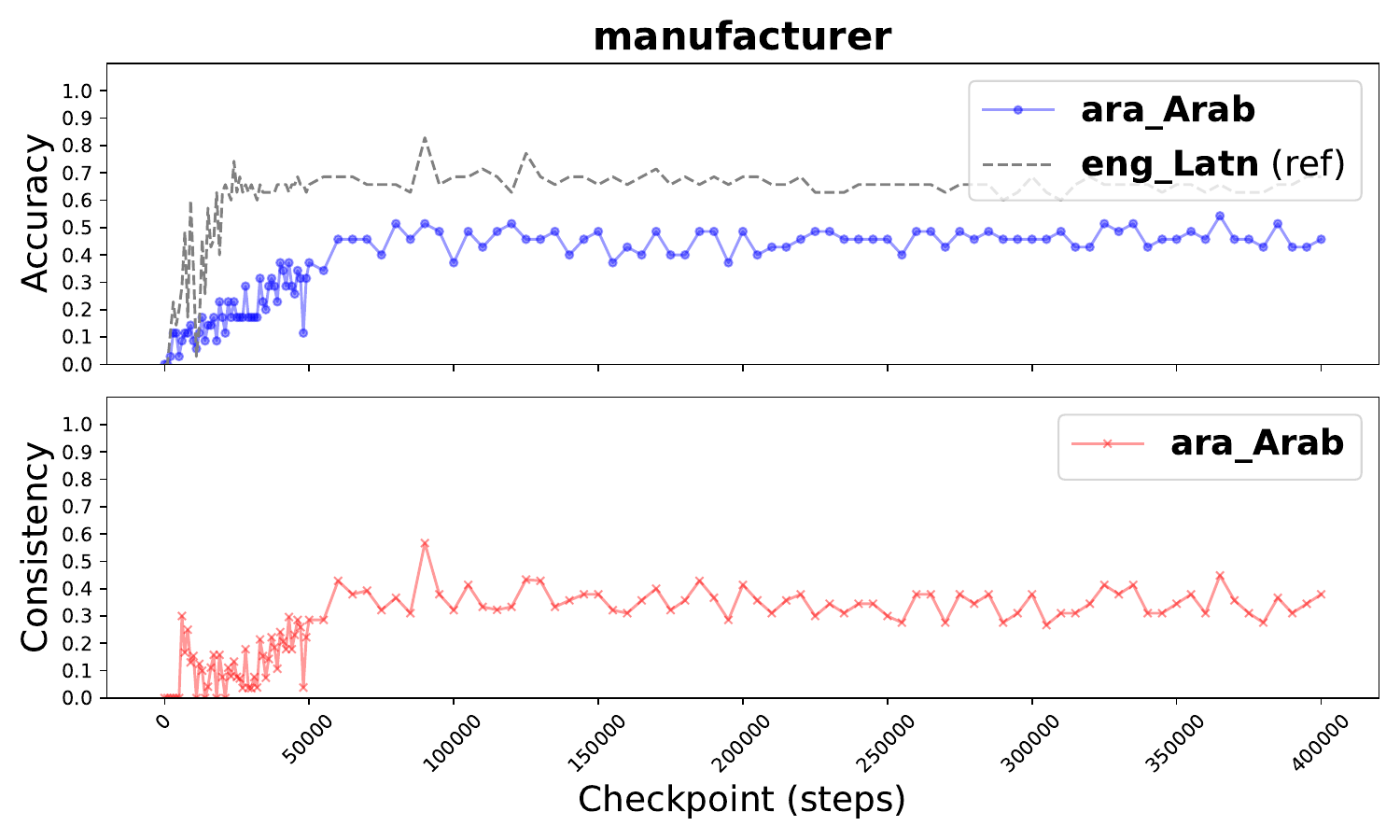}
    \includegraphics[width=0.24\textwidth]{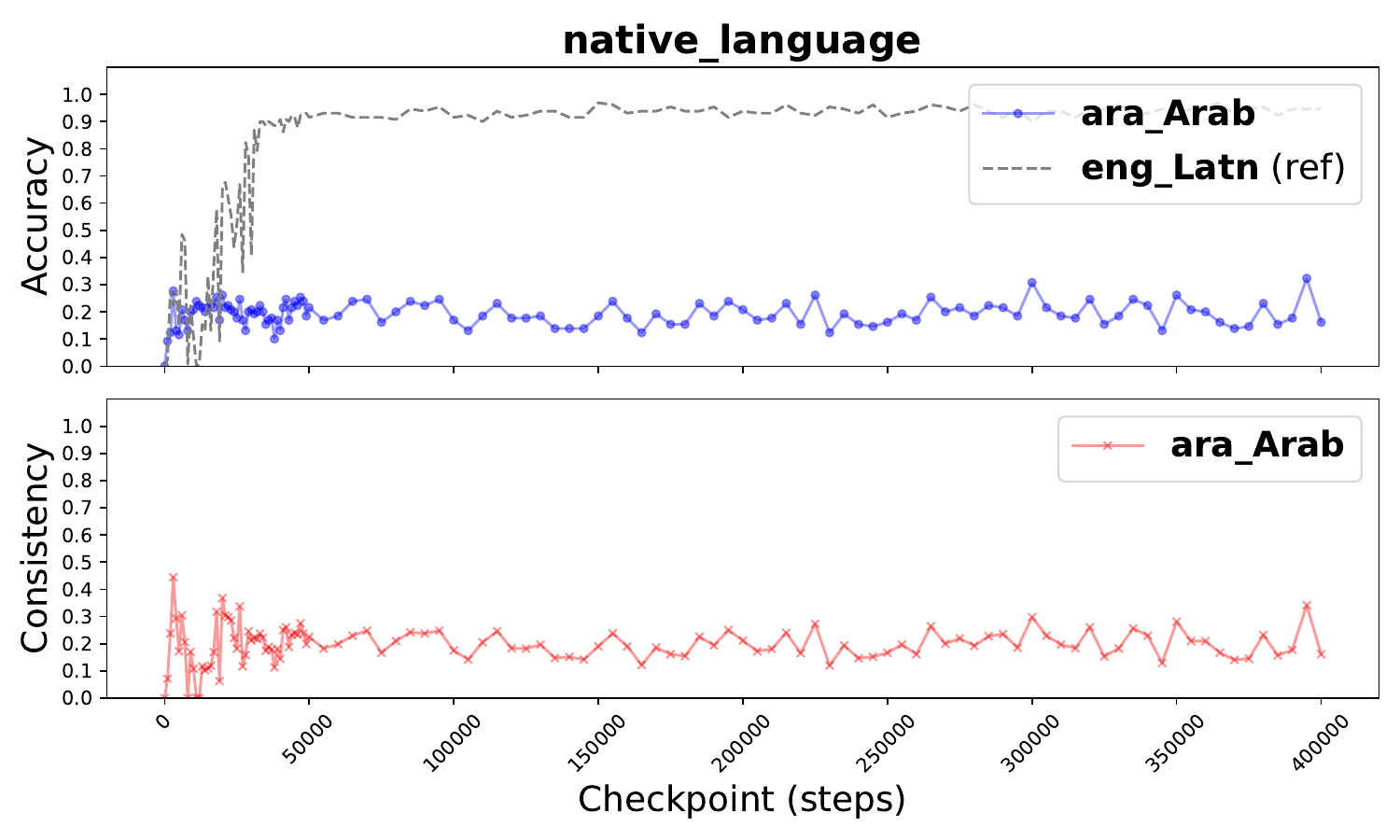}
    \includegraphics[width=0.24\textwidth]{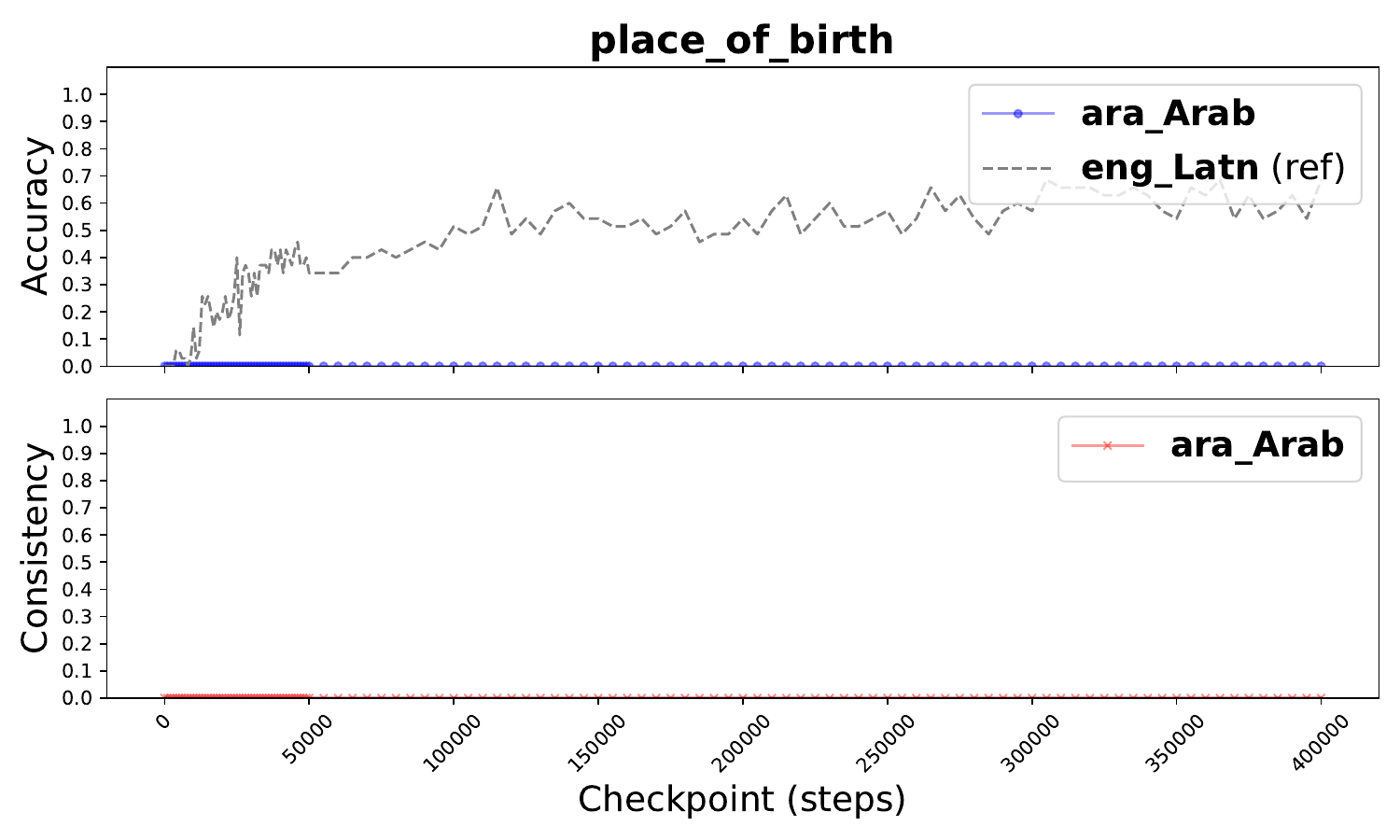}
    \includegraphics[width=0.24\textwidth]{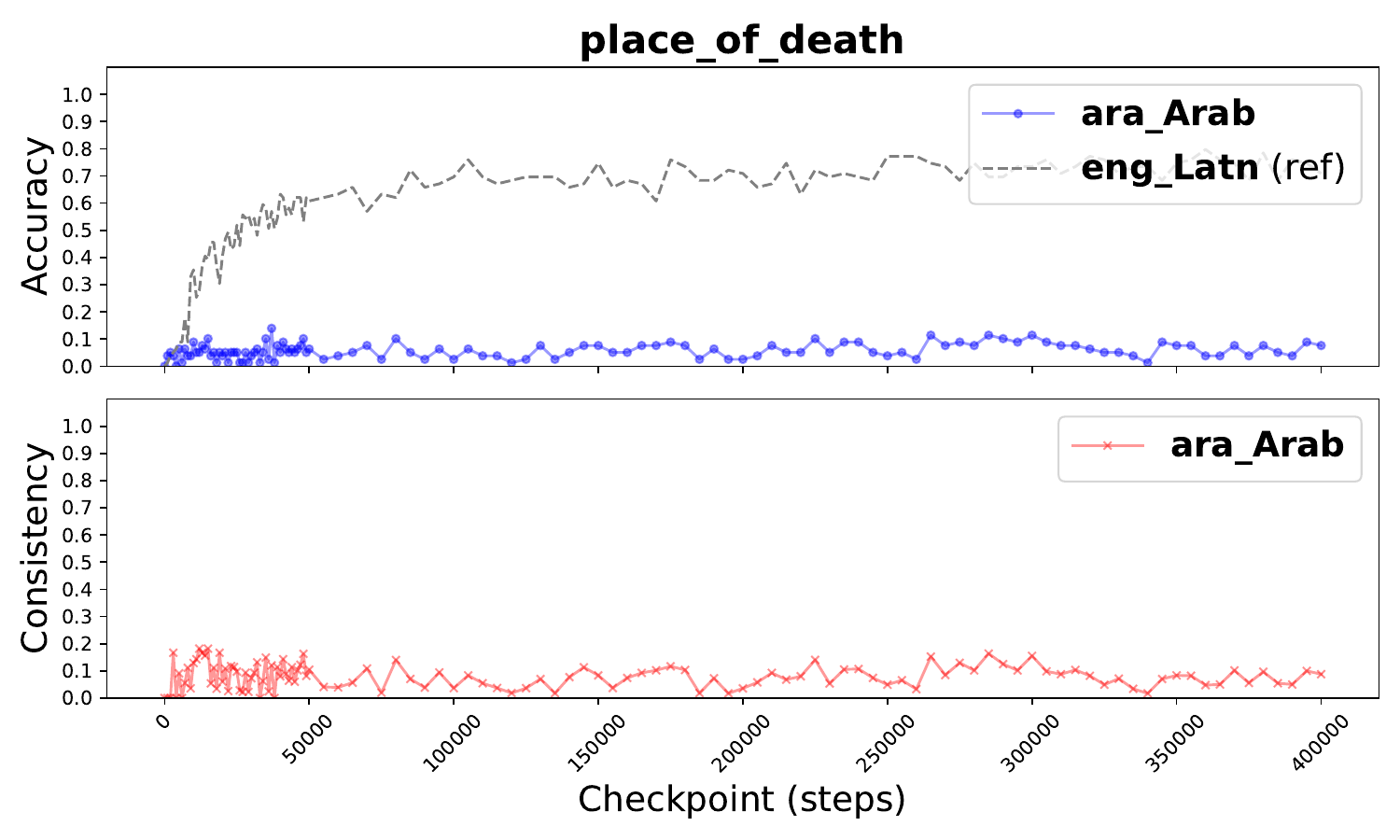}
    \includegraphics[width=0.24\textwidth]{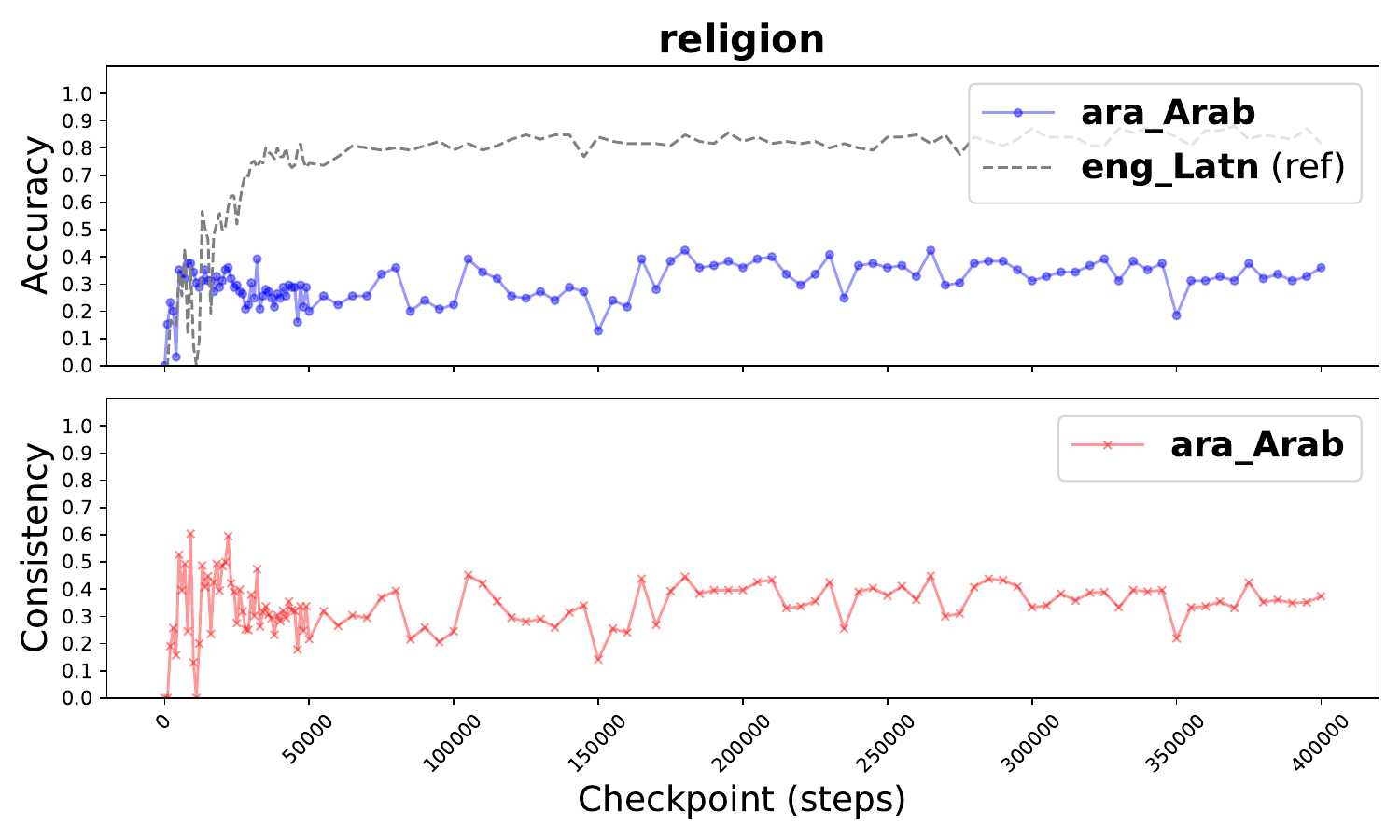}
    \caption{Factual accuracy (ACC) and crosslingual consistency (CO) for each relation type in \textbf{ara\_Arab}.}
    \label{fig:performance_over_checkpoints_ar}
\end{figure*}

\begin{figure*}
    \centering
    \includegraphics[width=0.24\textwidth]{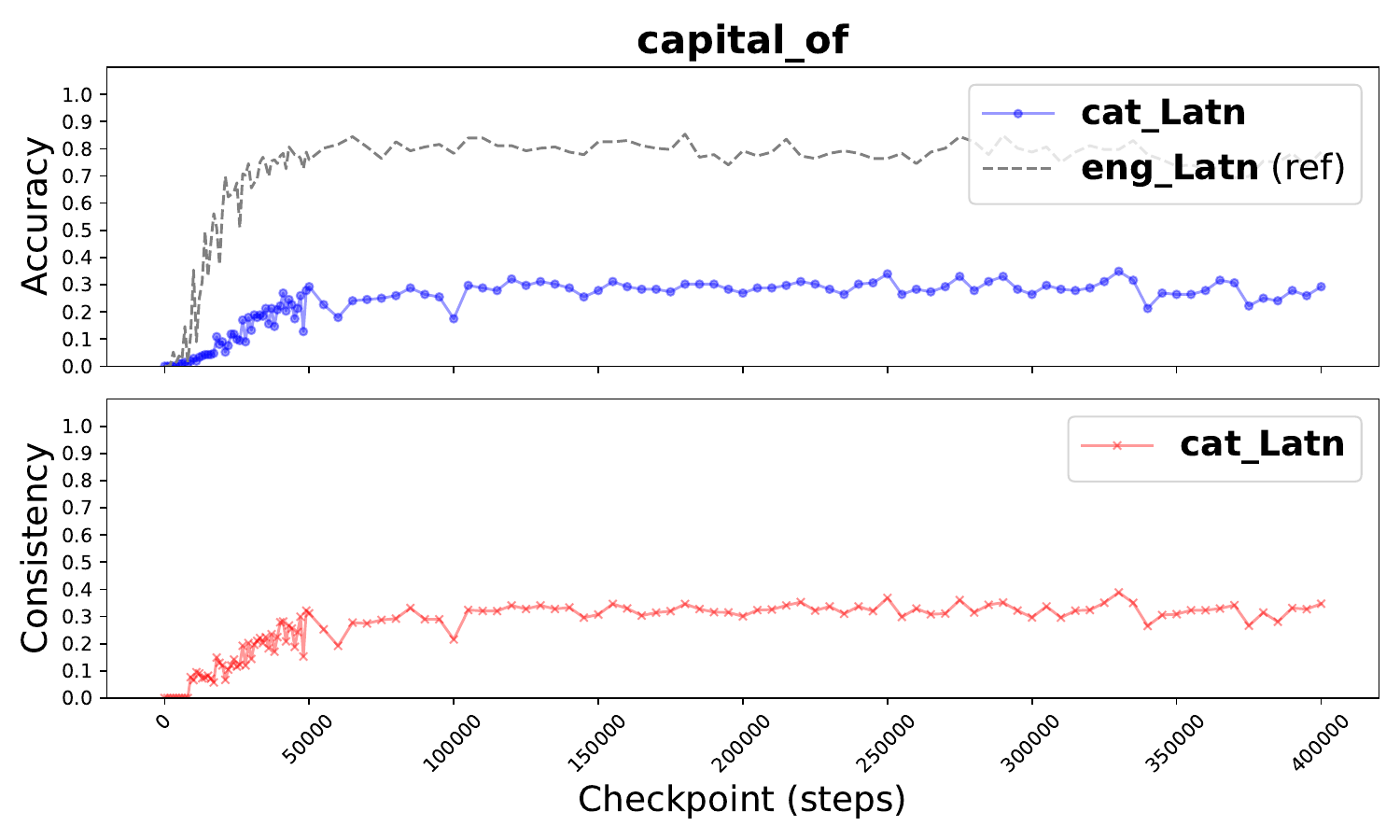}
    \includegraphics[width=0.24\textwidth]{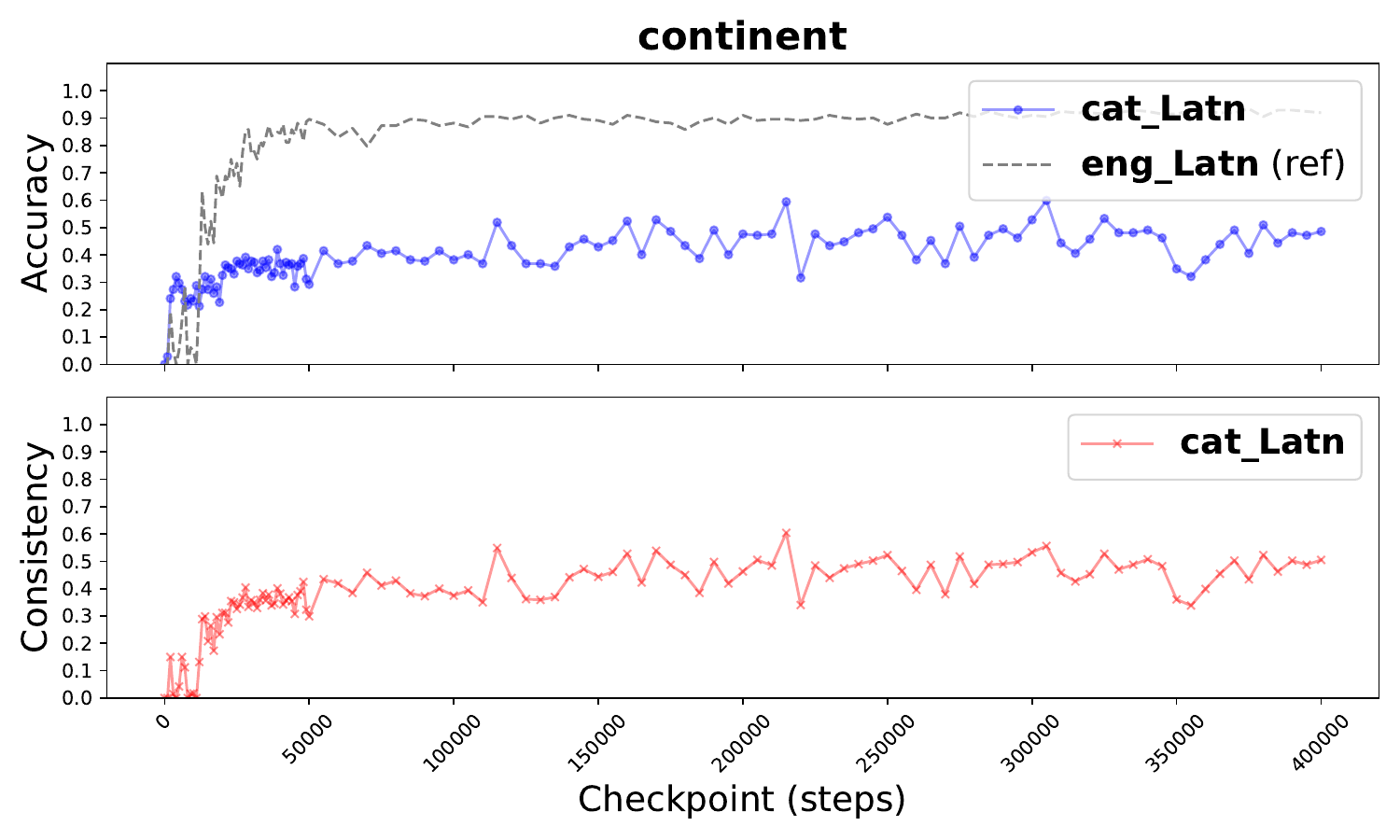}
    \includegraphics[width=0.24\textwidth]{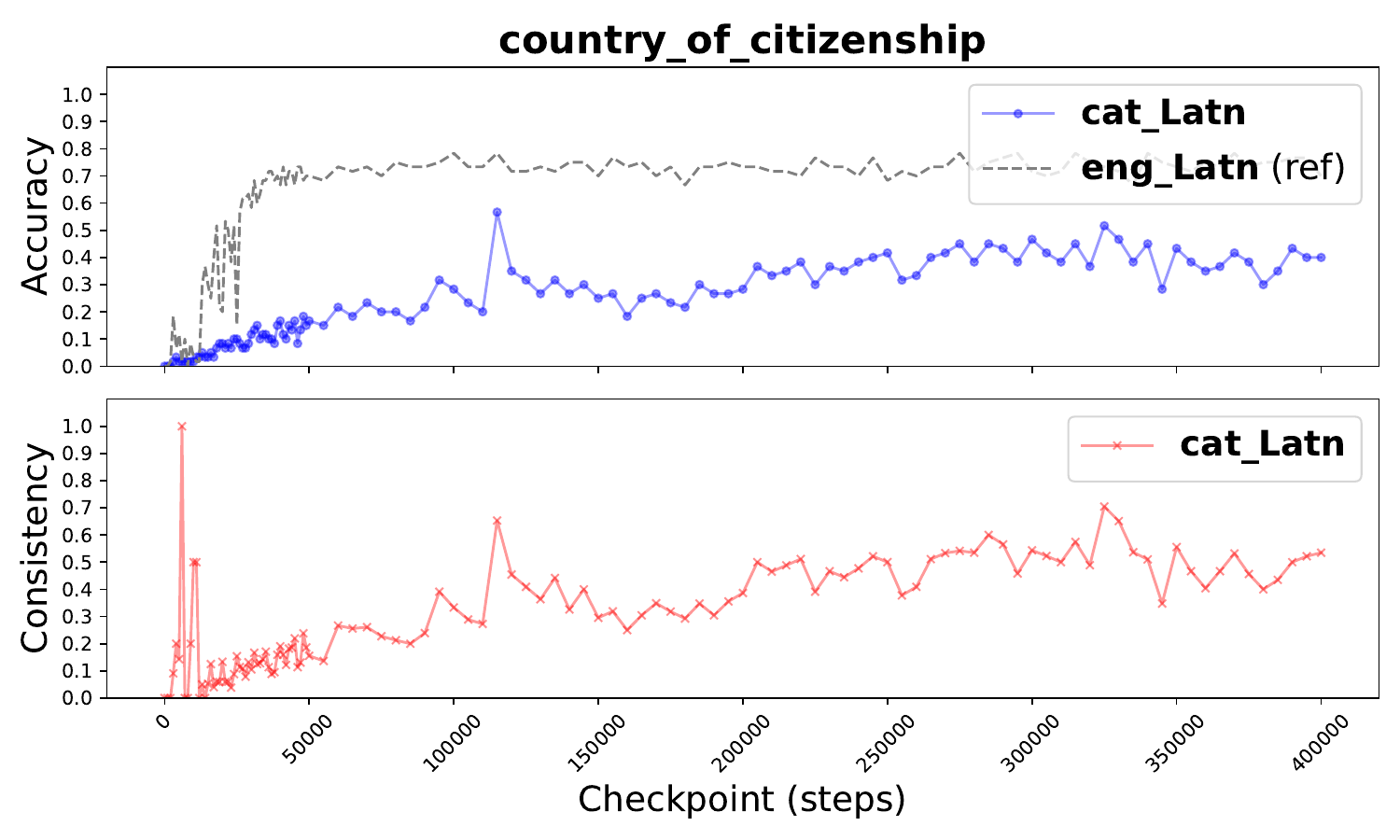}
    \includegraphics[width=0.24\textwidth]{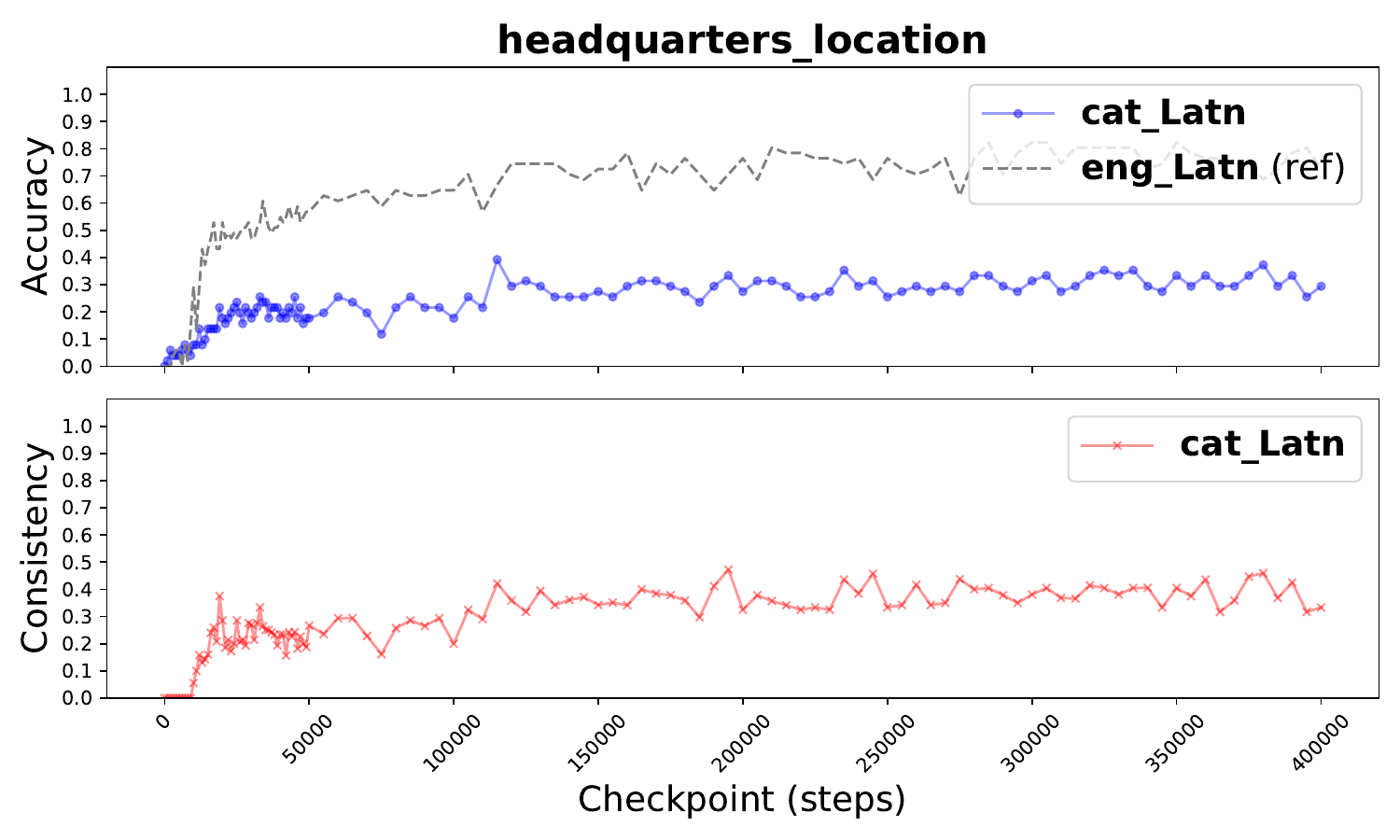}
    \includegraphics[width=0.24\textwidth]{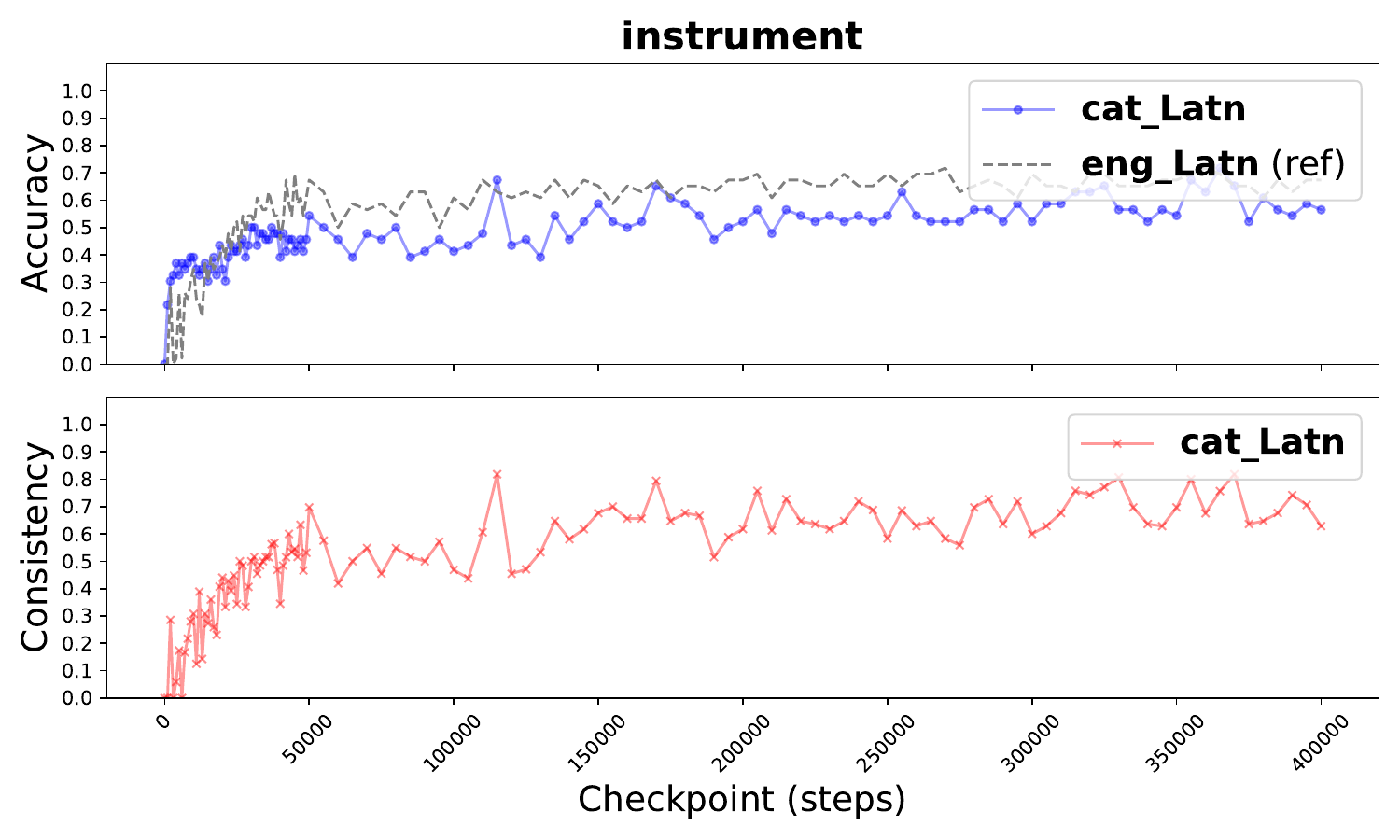}
    \includegraphics[width=0.24\textwidth]{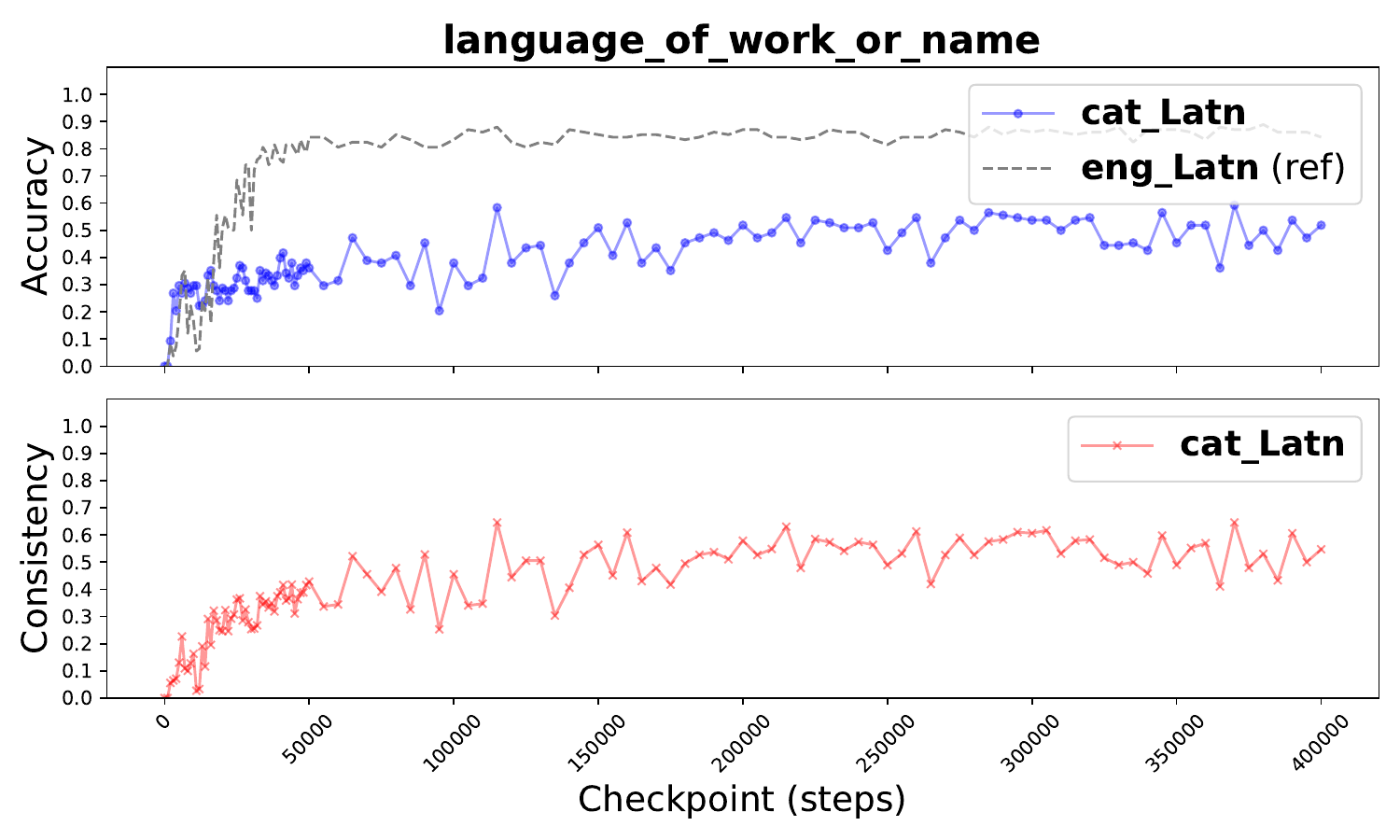}
    \includegraphics[width=0.24\textwidth]{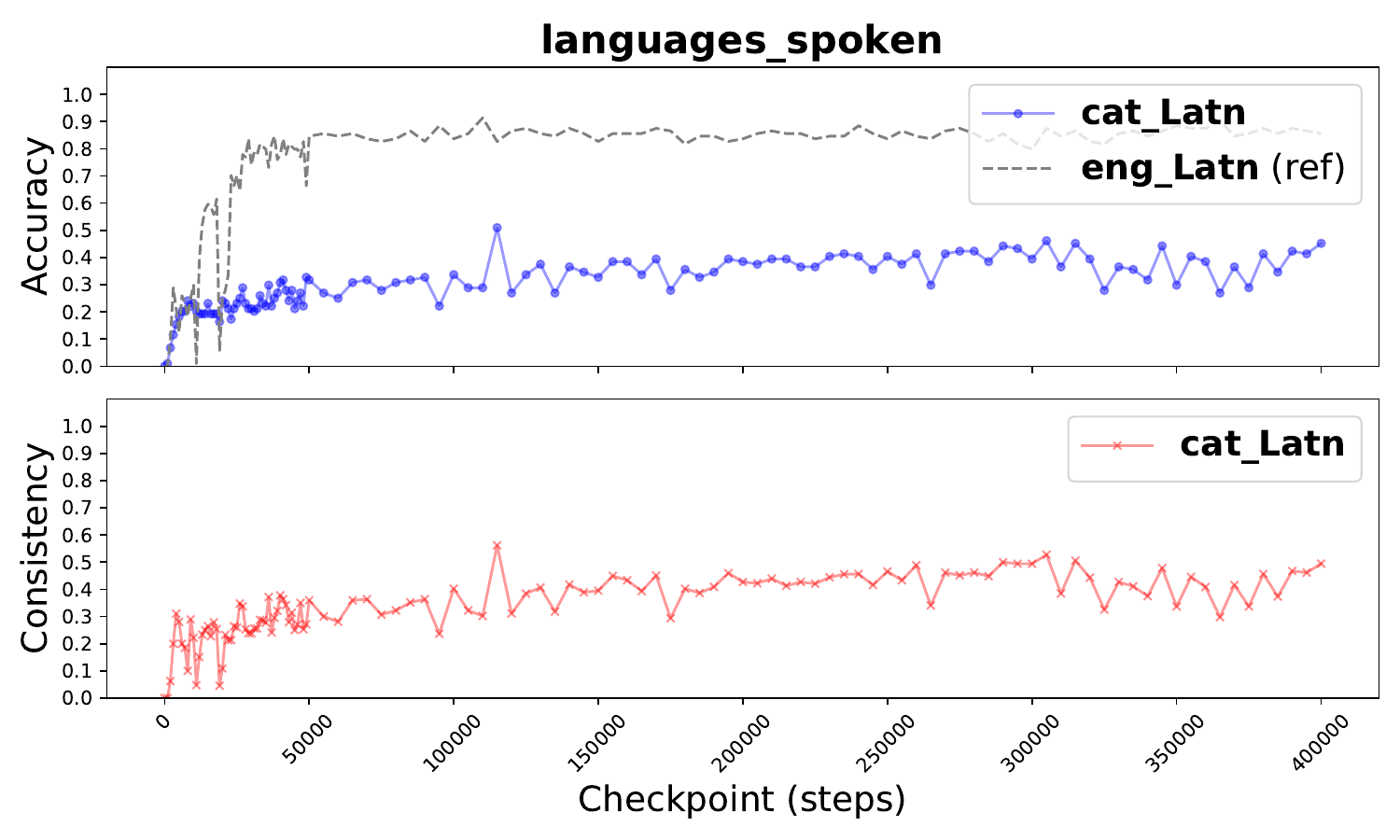}
    \includegraphics[width=0.24\textwidth]{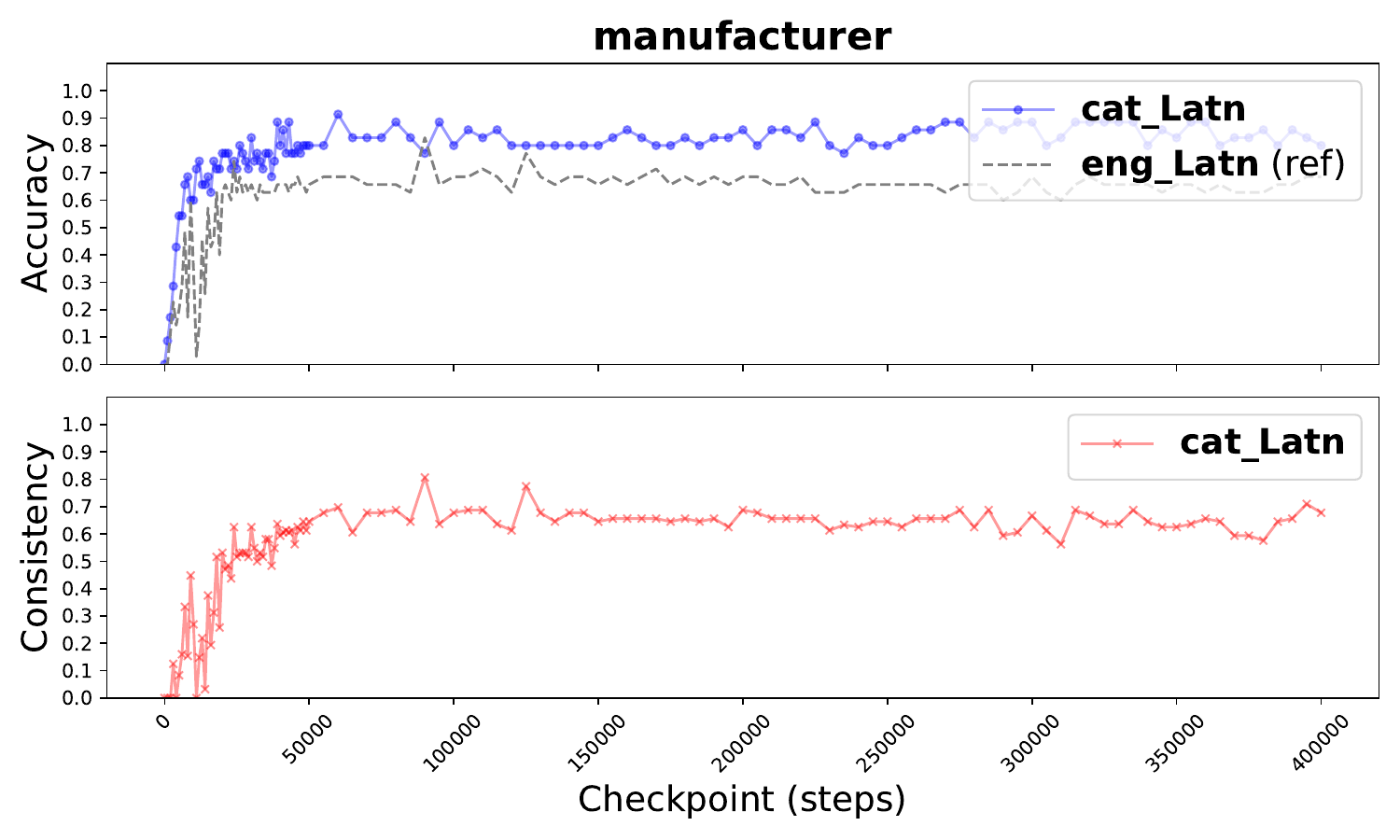}
    \includegraphics[width=0.24\textwidth]{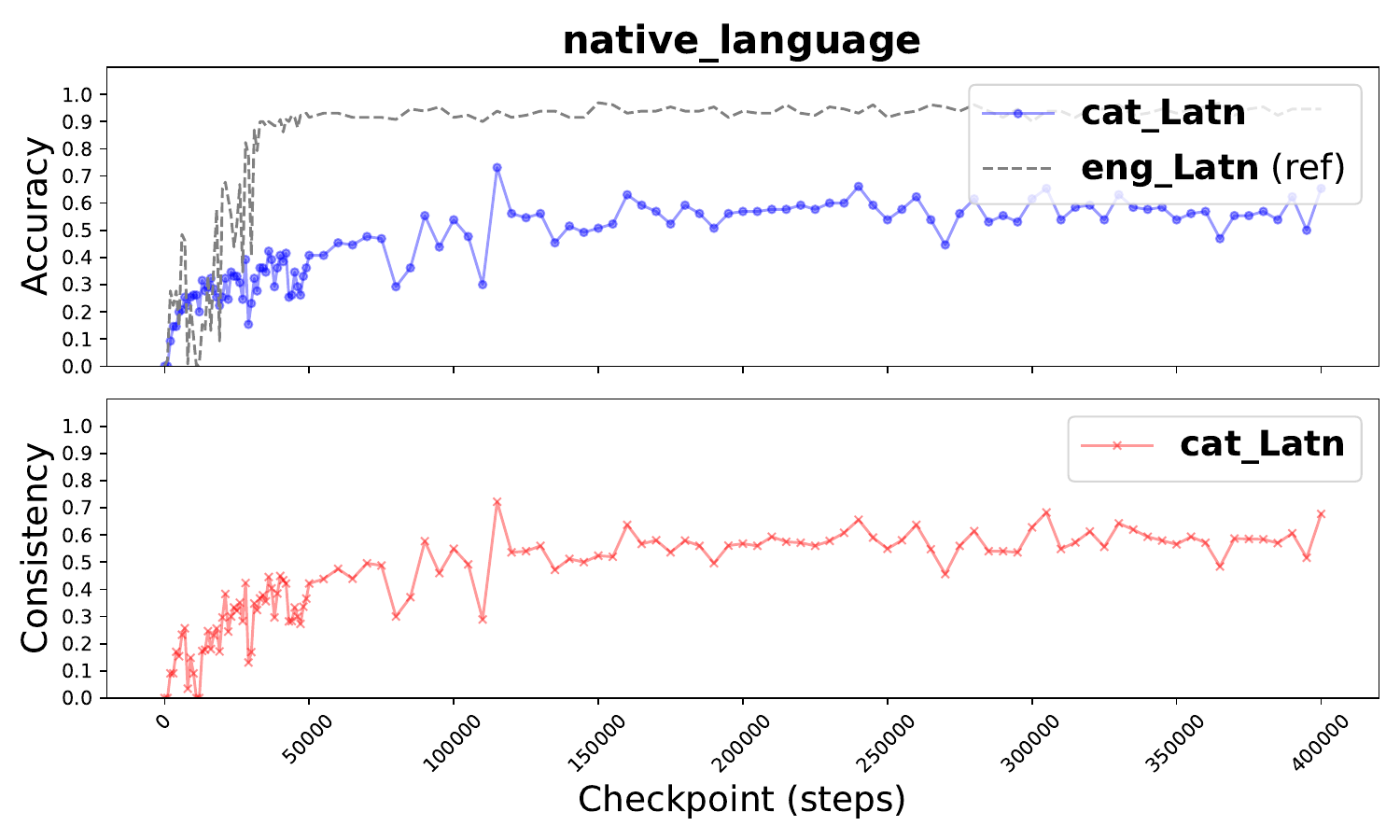}
    \includegraphics[width=0.24\textwidth]{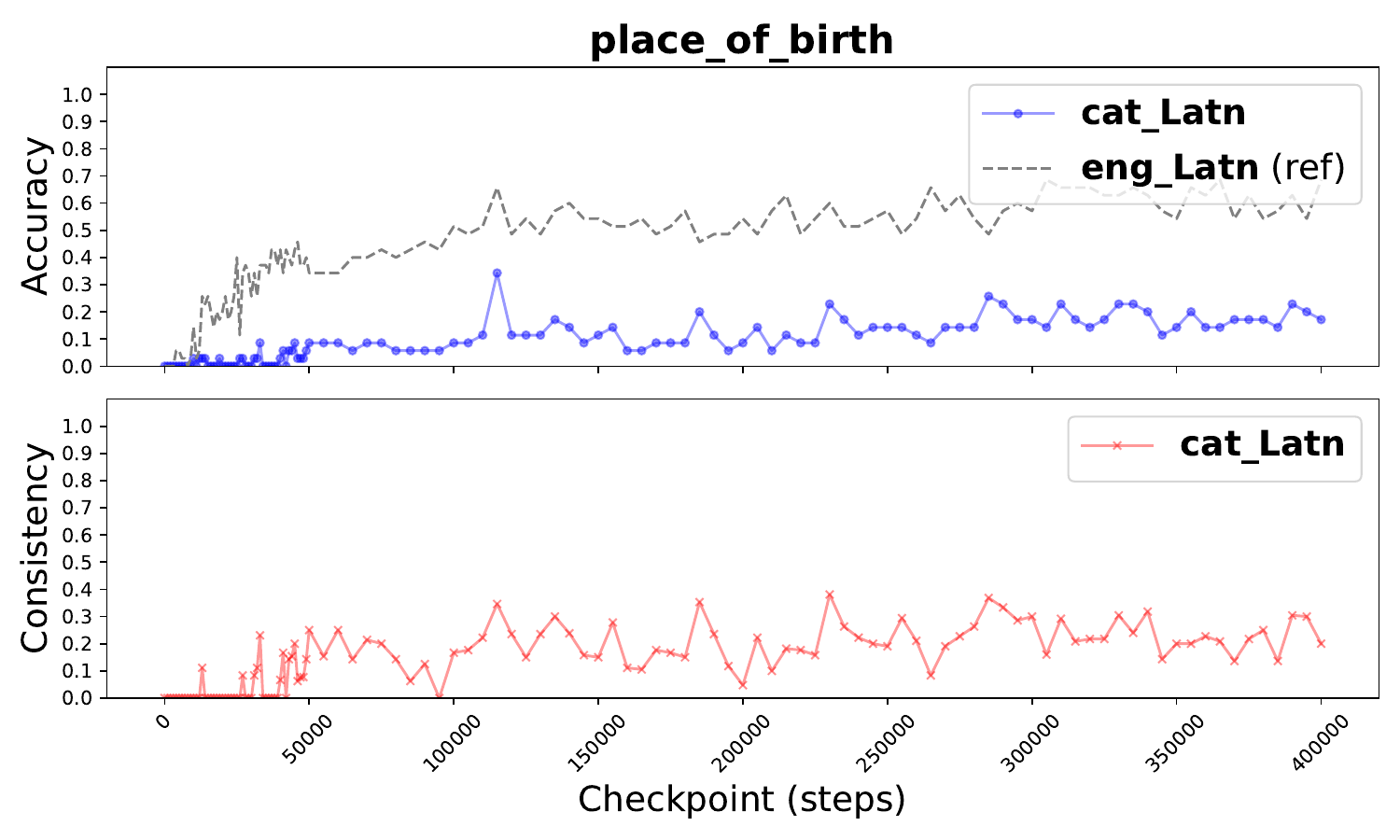}
    \includegraphics[width=0.24\textwidth]{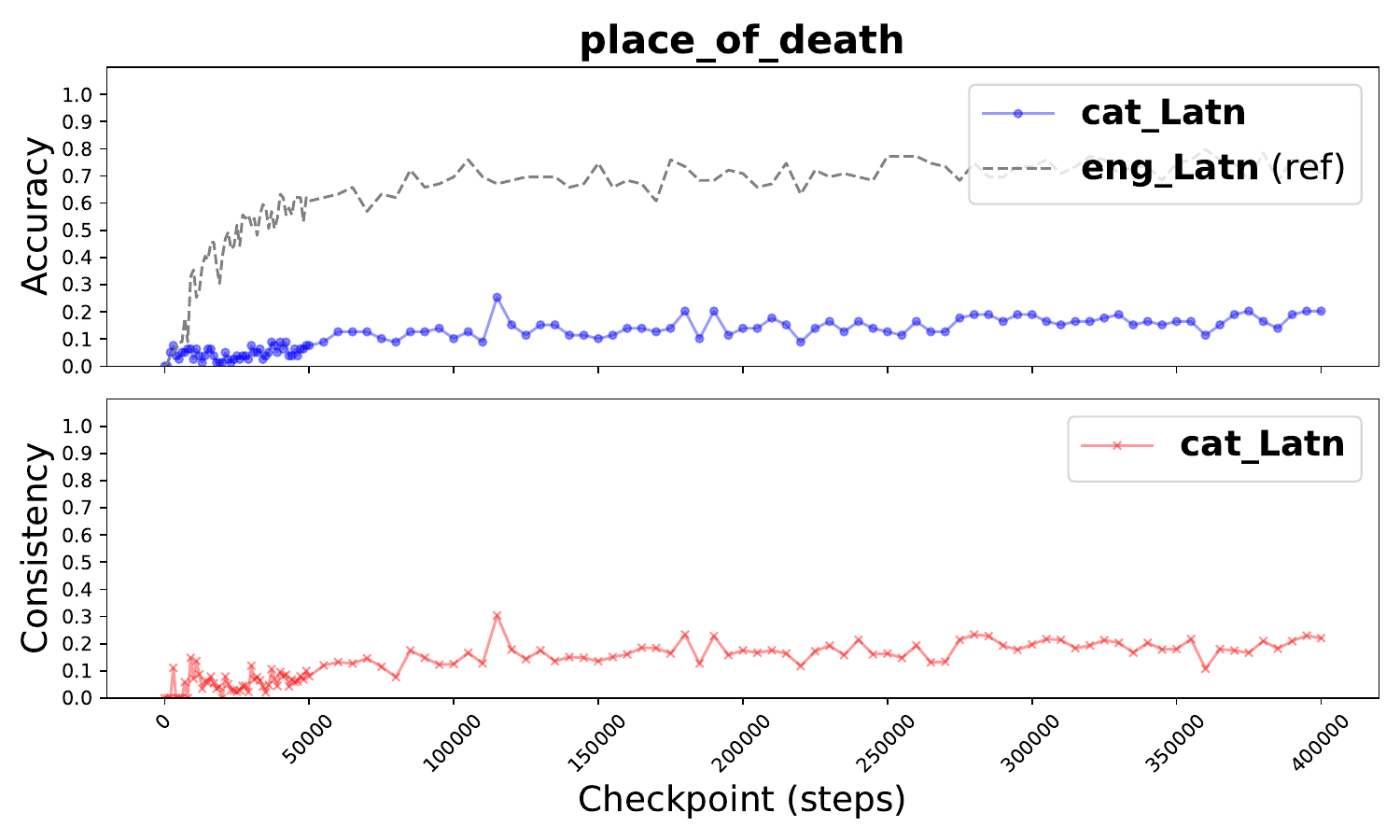}
    \includegraphics[width=0.24\textwidth]{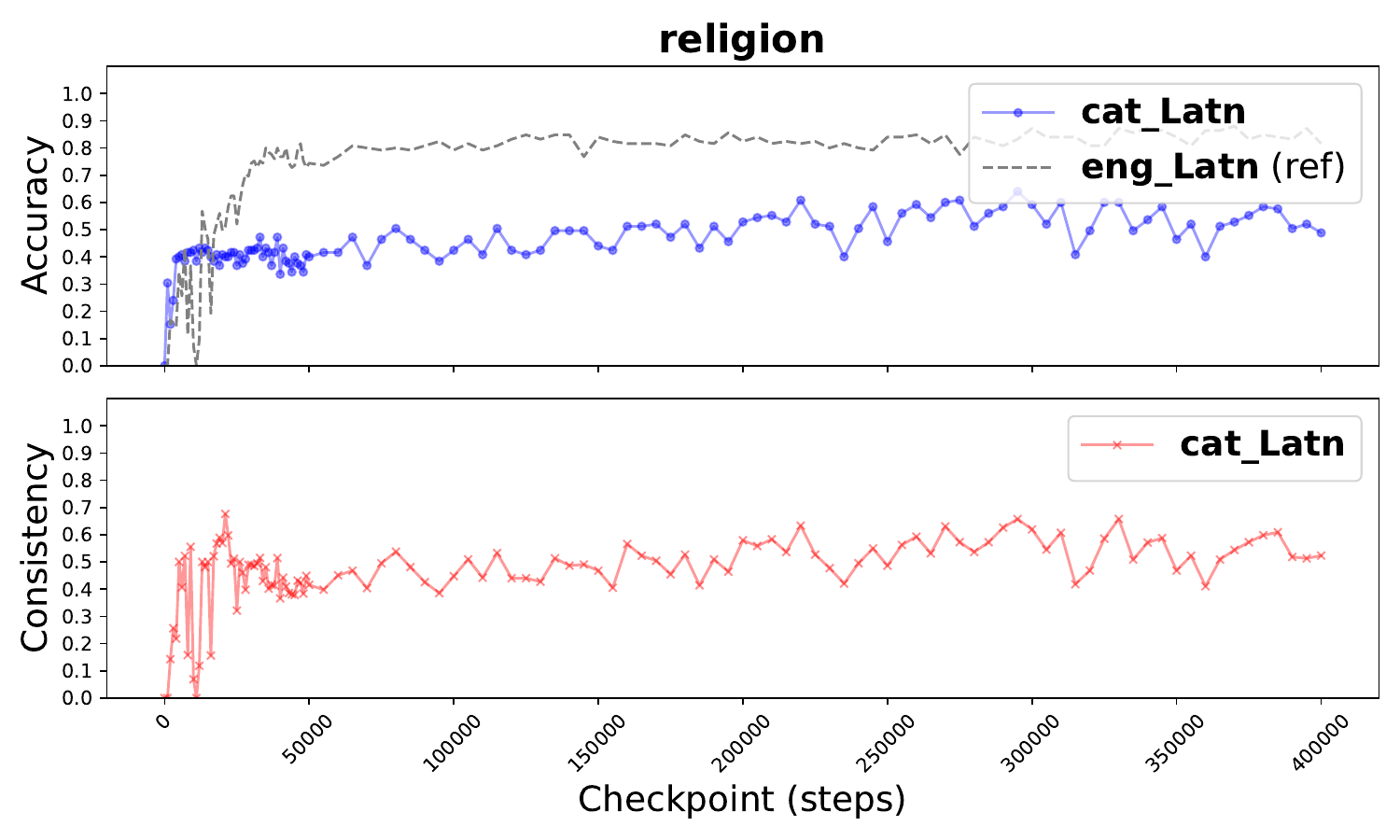}
    \caption{Factual accuracy (ACC) and crosslingual consistency (CO) for each relation type in \textbf{cat\_Latn}.}
    \label{fig:performance_over_checkpoints_ca}
\end{figure*}

\begin{figure*}
    \centering
    \includegraphics[width=0.24\textwidth]{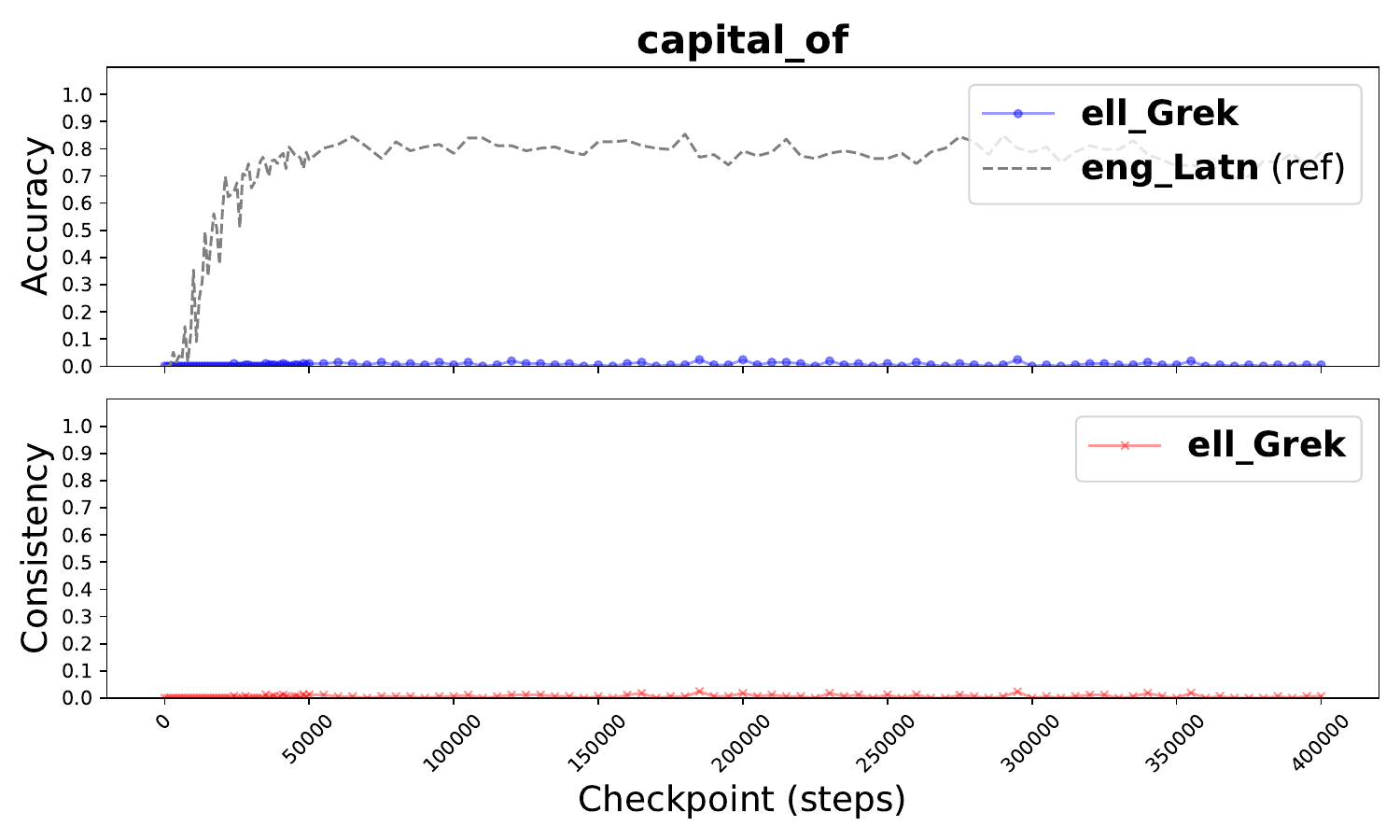}
    \includegraphics[width=0.24\textwidth]{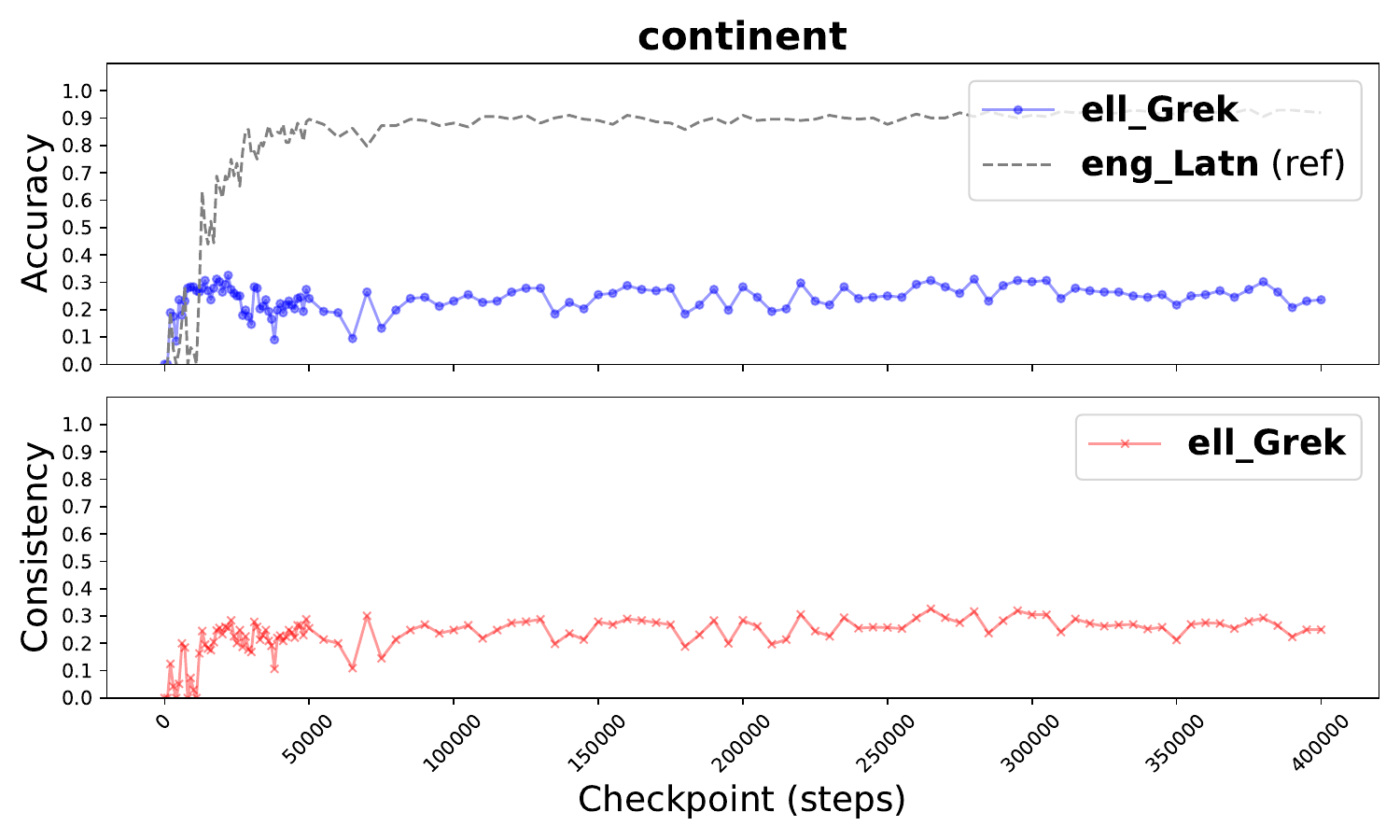}
    \includegraphics[width=0.24\textwidth]{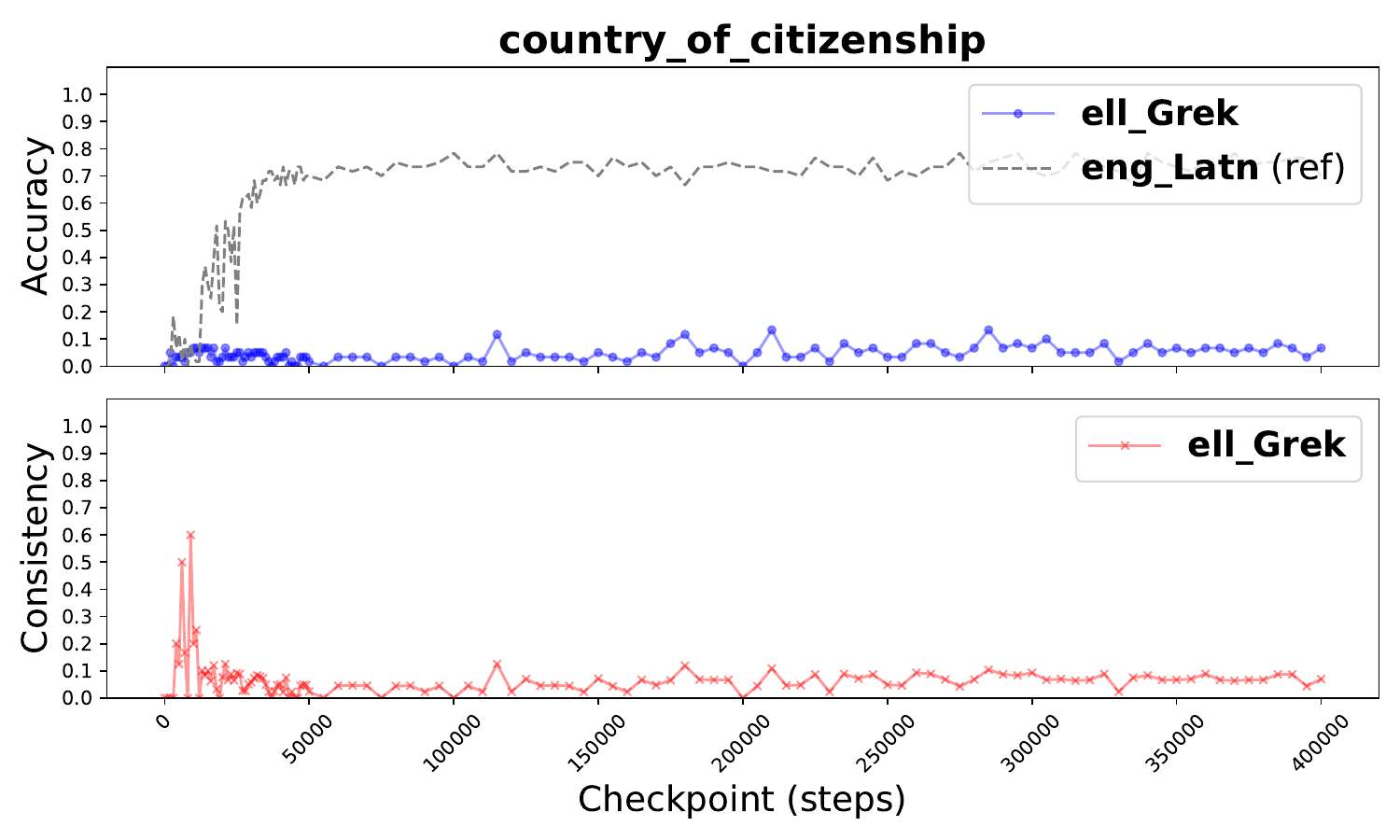}
    \includegraphics[width=0.24\textwidth]{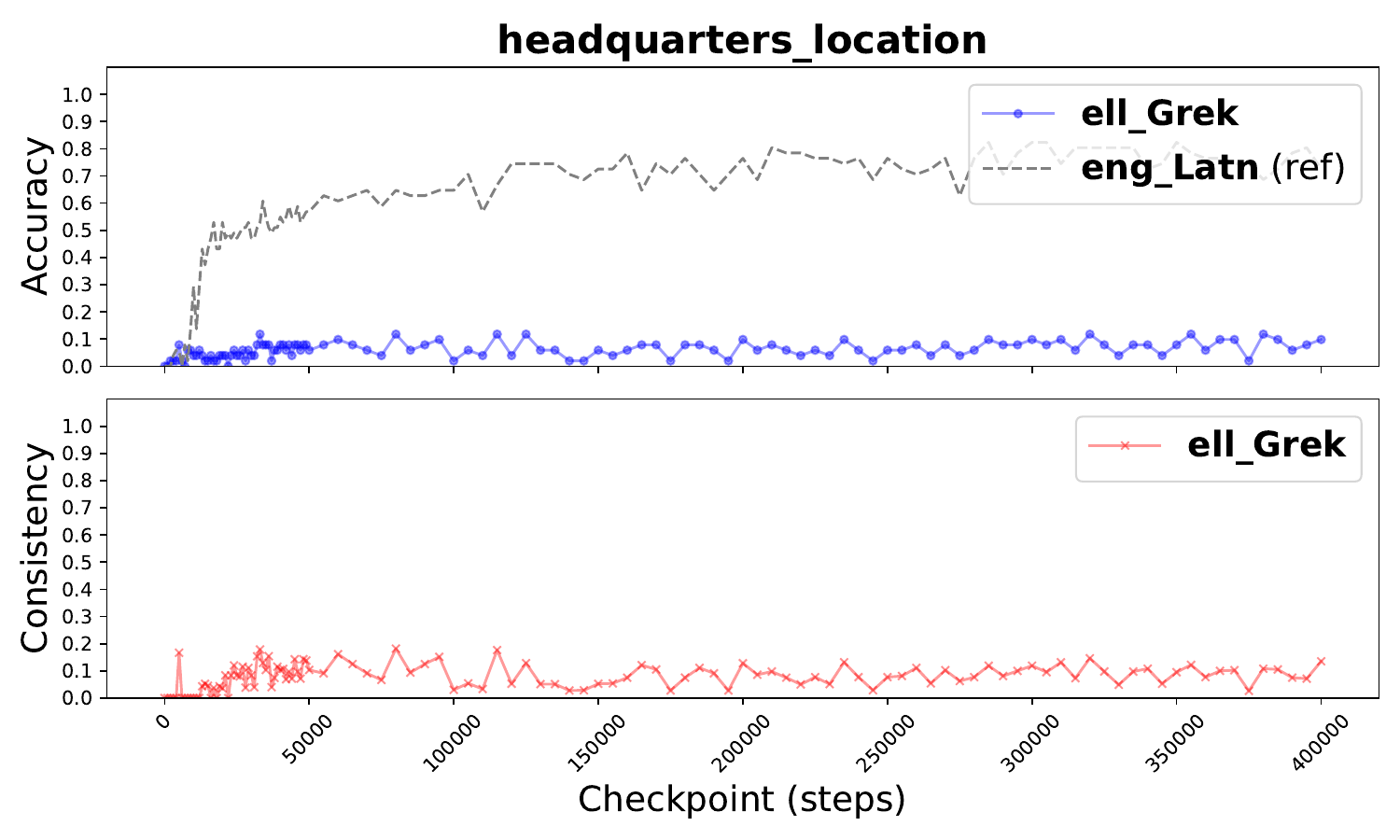}
    \includegraphics[width=0.24\textwidth]{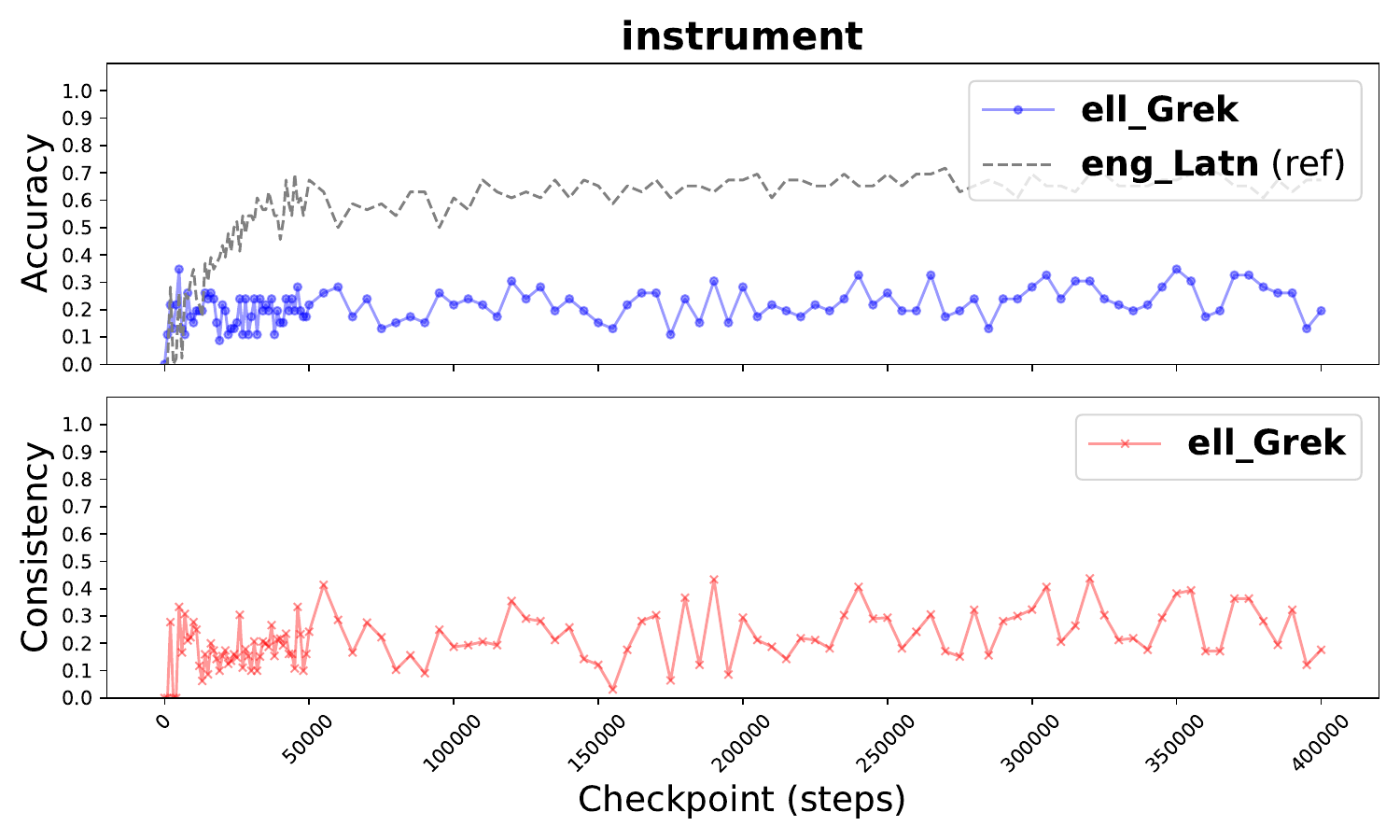}
    \includegraphics[width=0.24\textwidth]{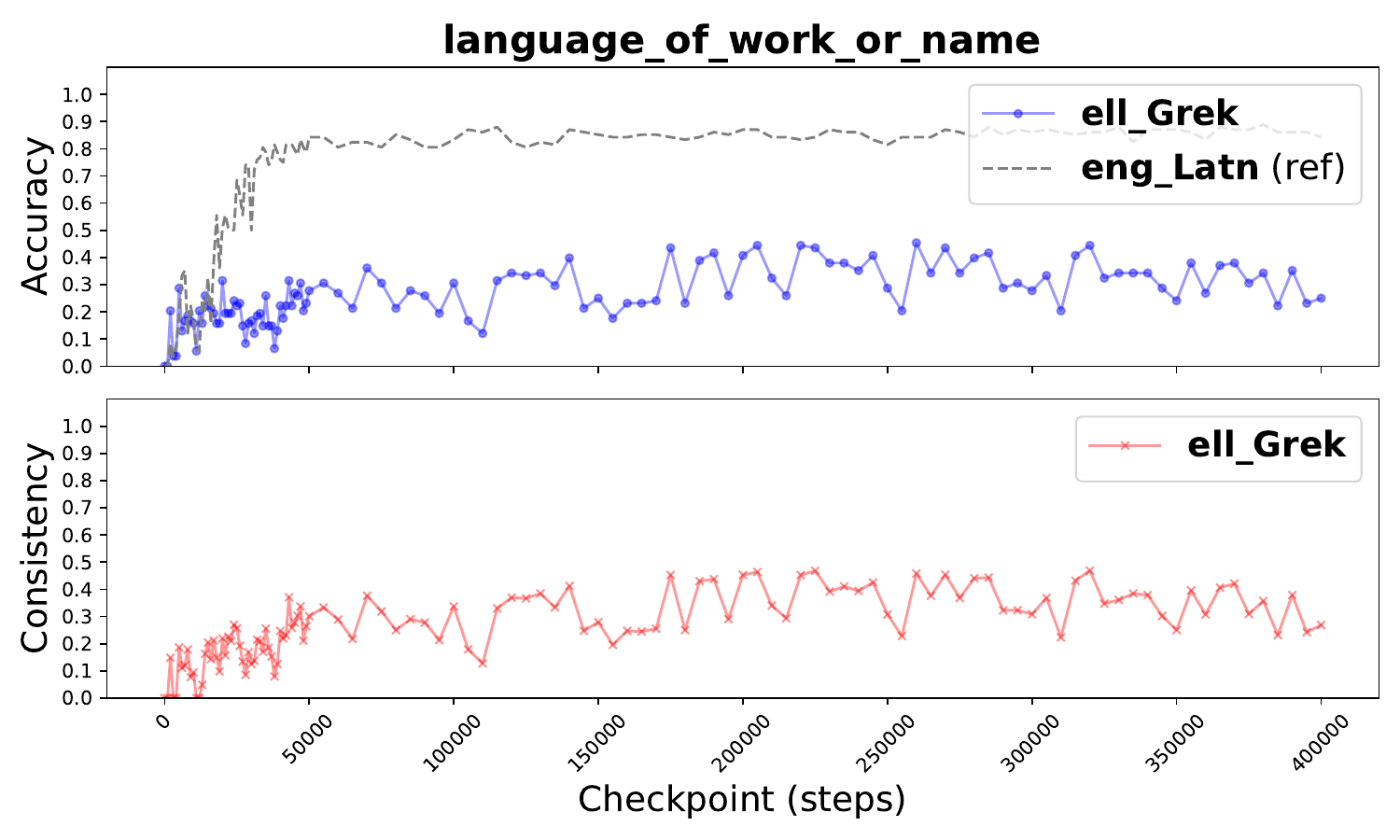}
    \includegraphics[width=0.24\textwidth]{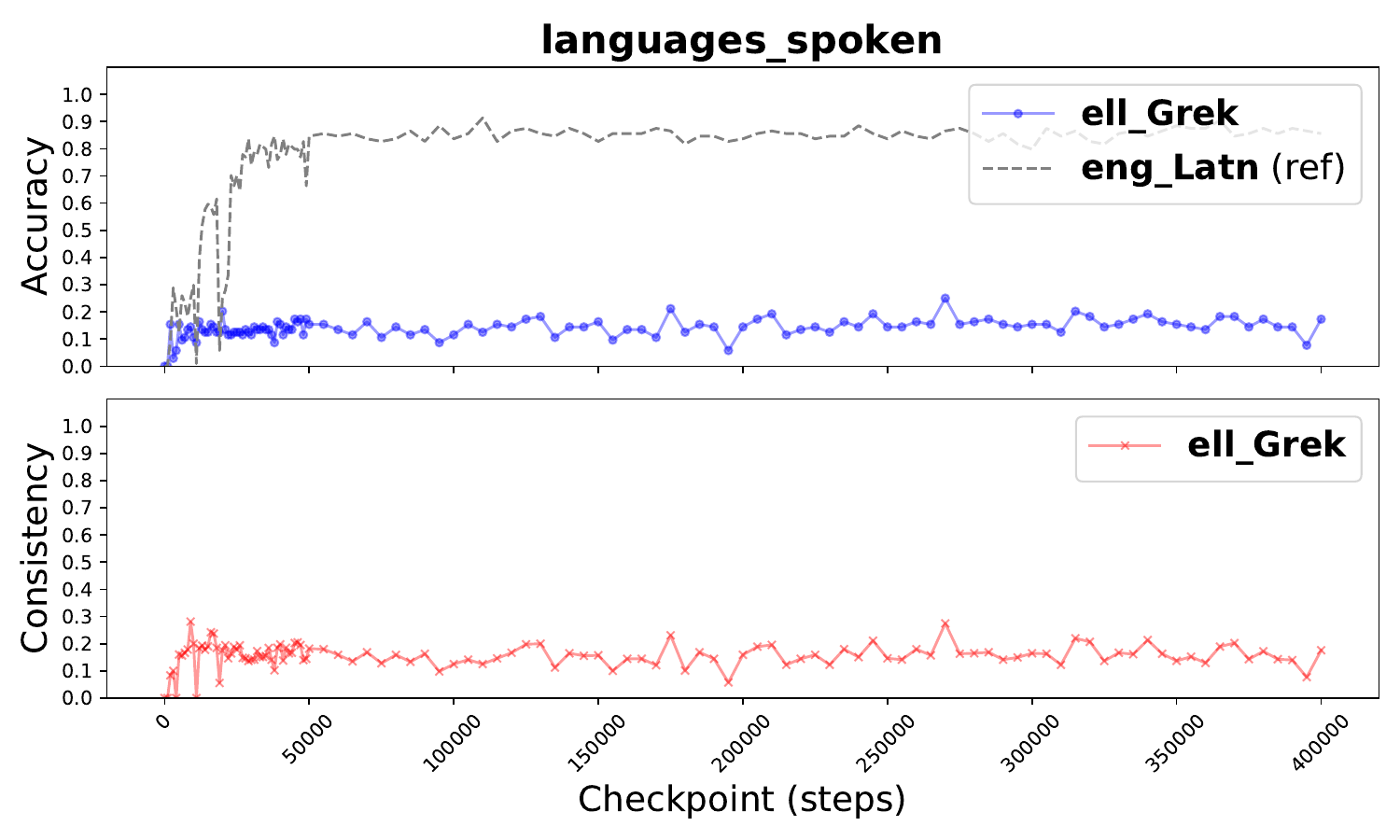}
    \includegraphics[width=0.24\textwidth]{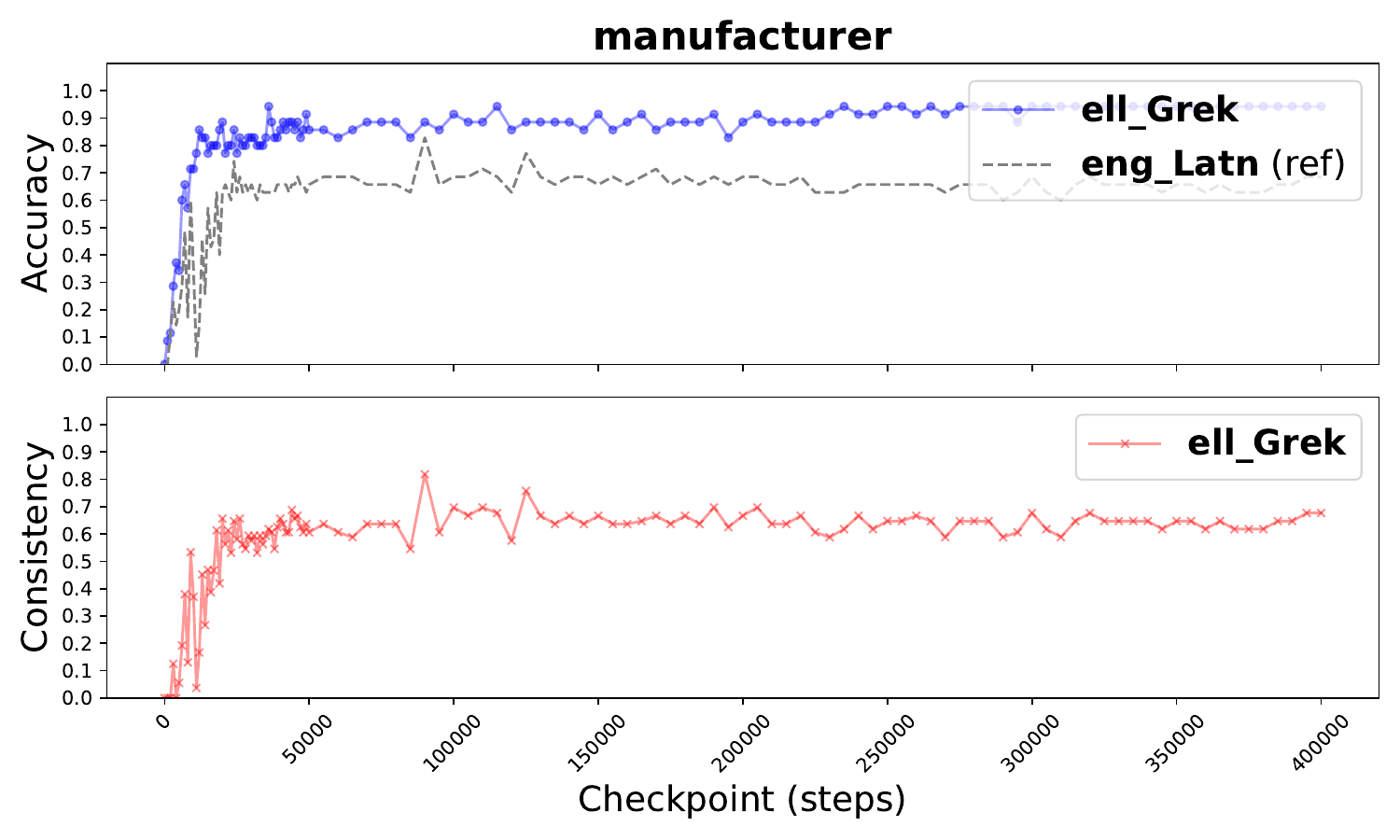}
    \includegraphics[width=0.24\textwidth]{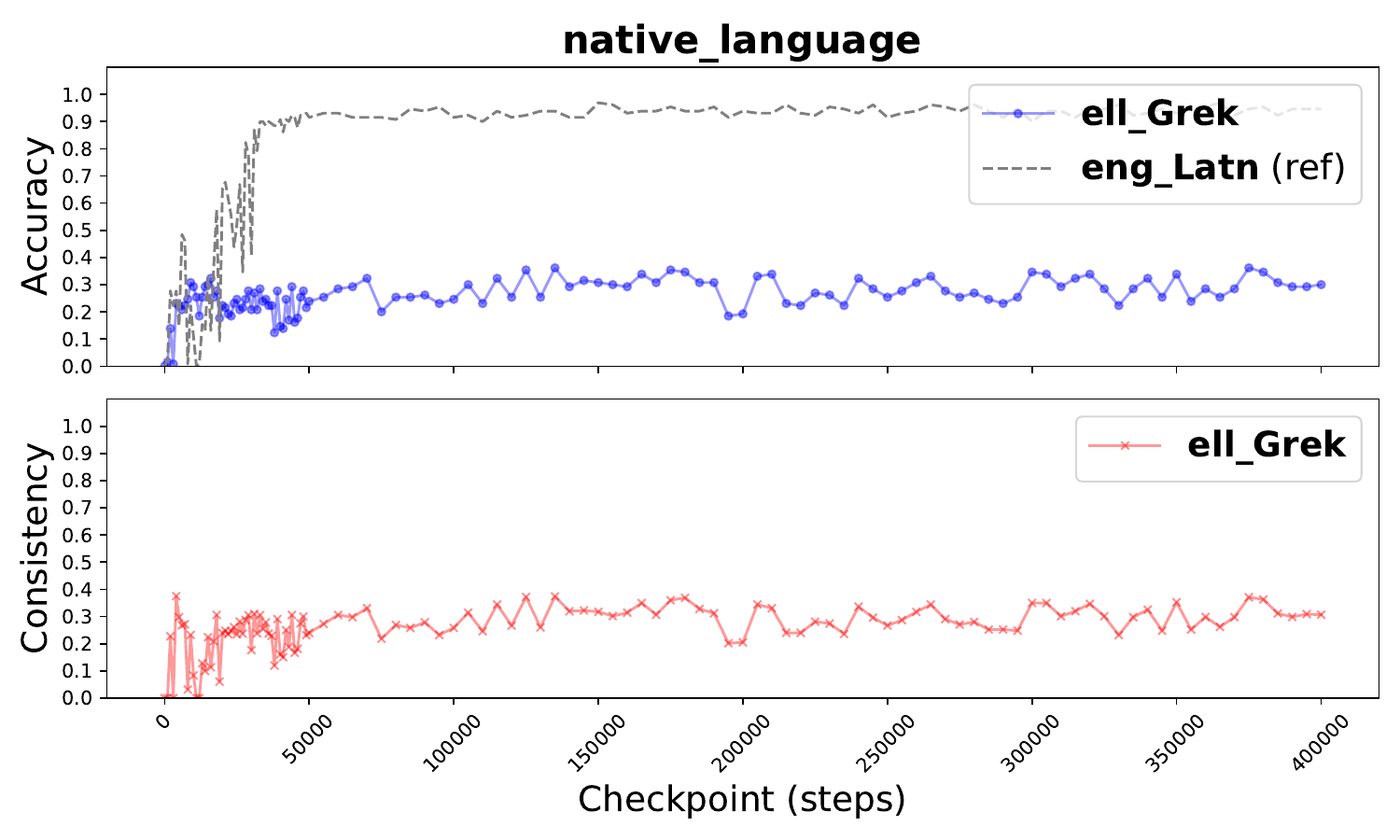}
    \includegraphics[width=0.24\textwidth]{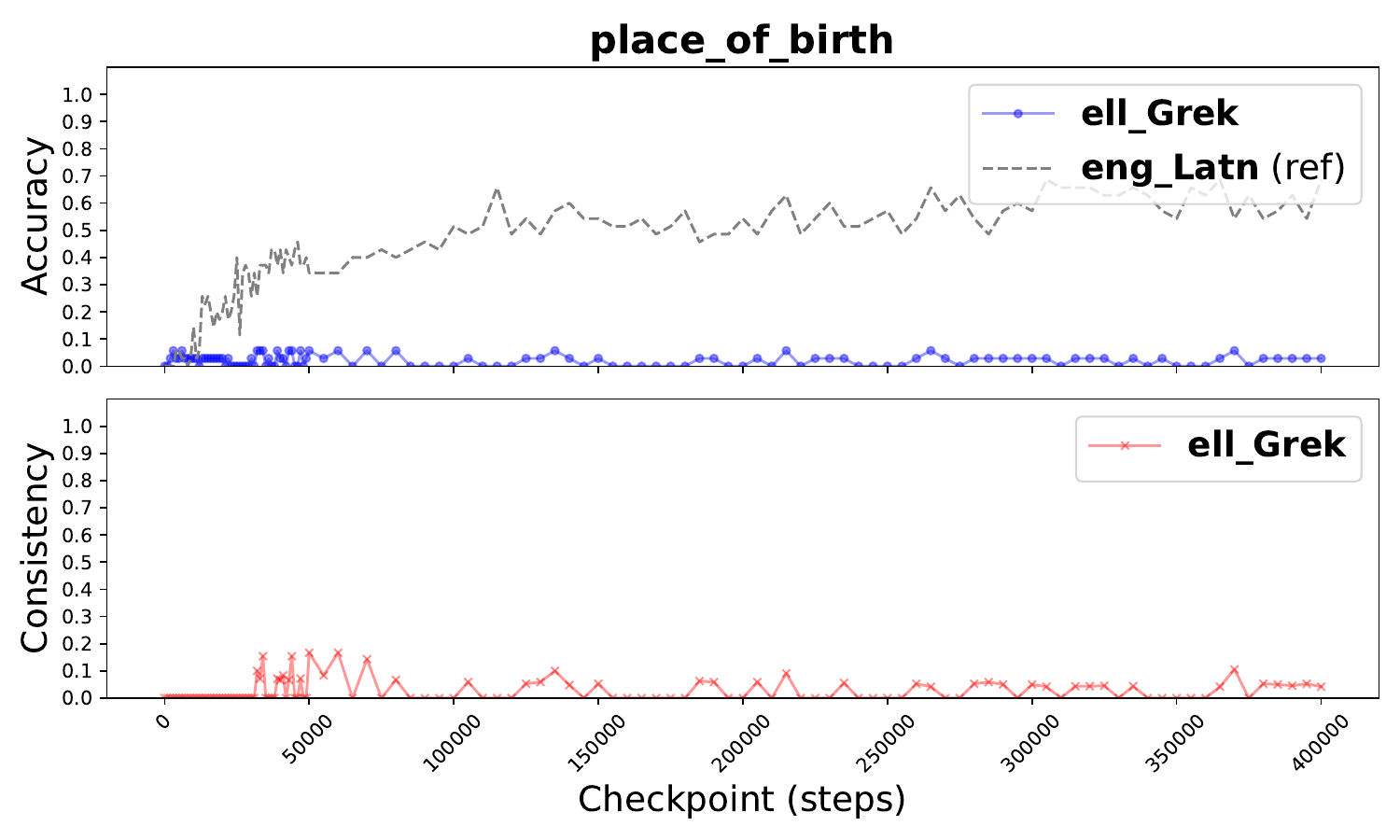}
    \includegraphics[width=0.24\textwidth]{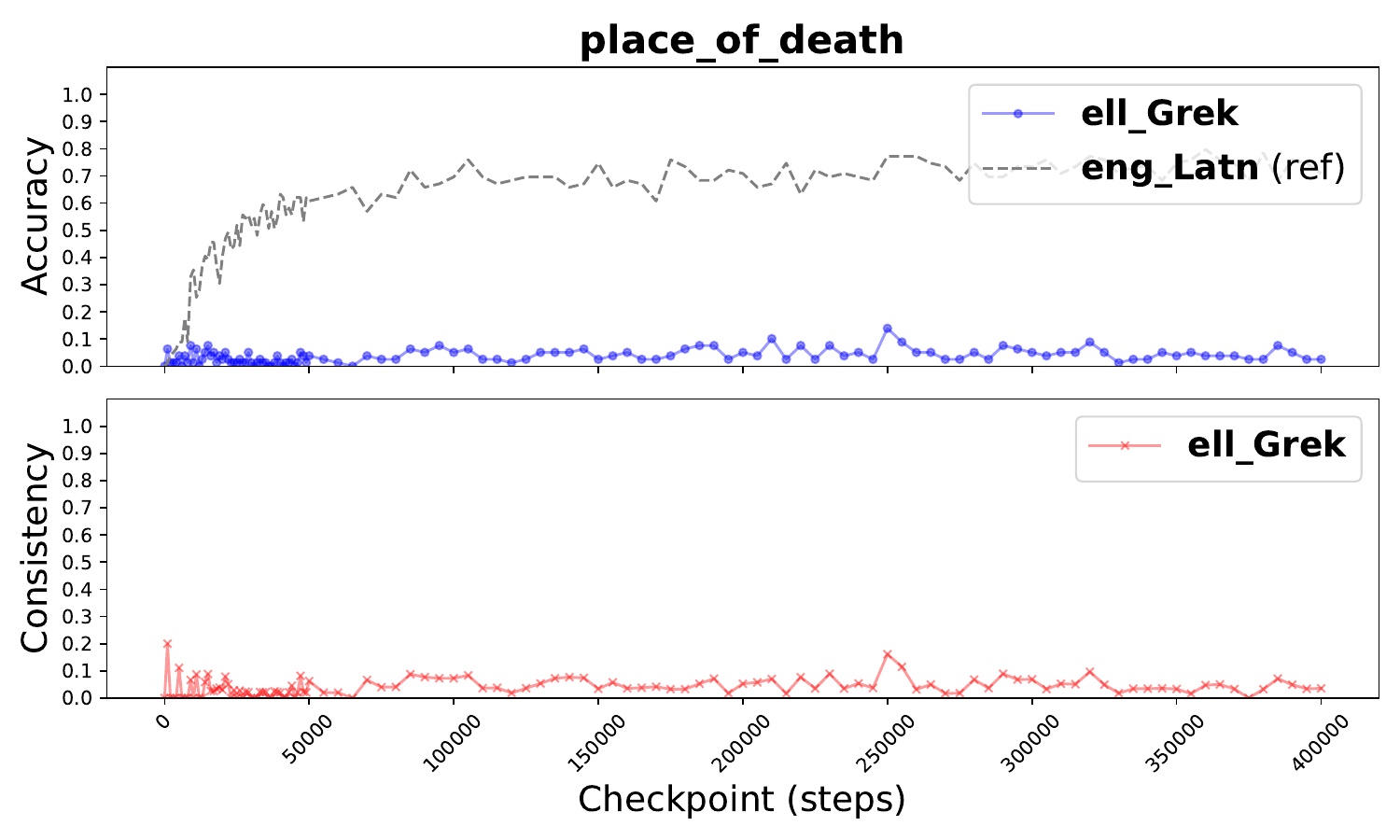}
    \includegraphics[width=0.24\textwidth]{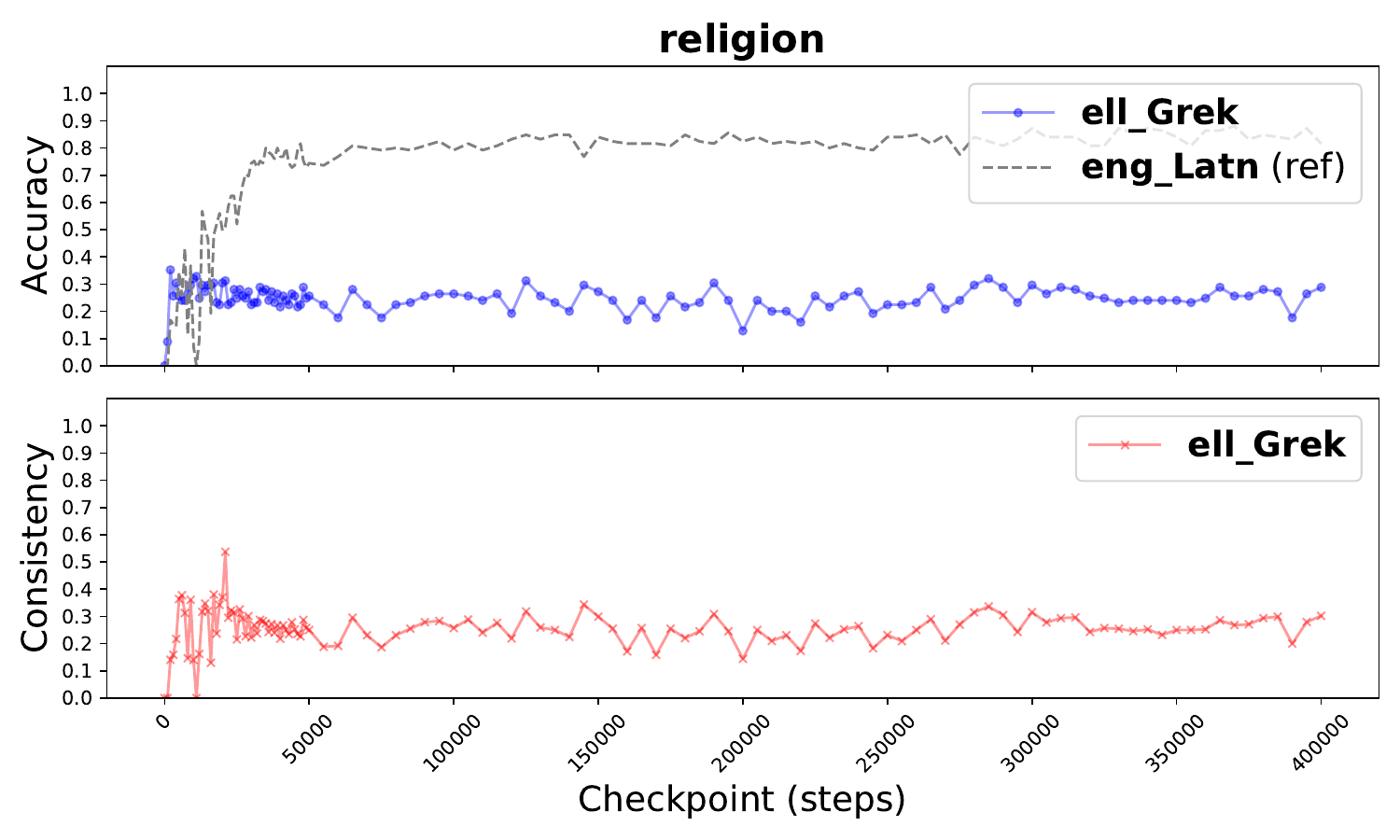}
    \caption{Factual accuracy (ACC) and crosslingual consistency (CO) for each relation type in \textbf{ell\_Grek}.}
    \label{fig:performance_over_checkpoints_el}
\end{figure*}

\begin{figure*}
    \centering
    \includegraphics[width=0.24\textwidth]{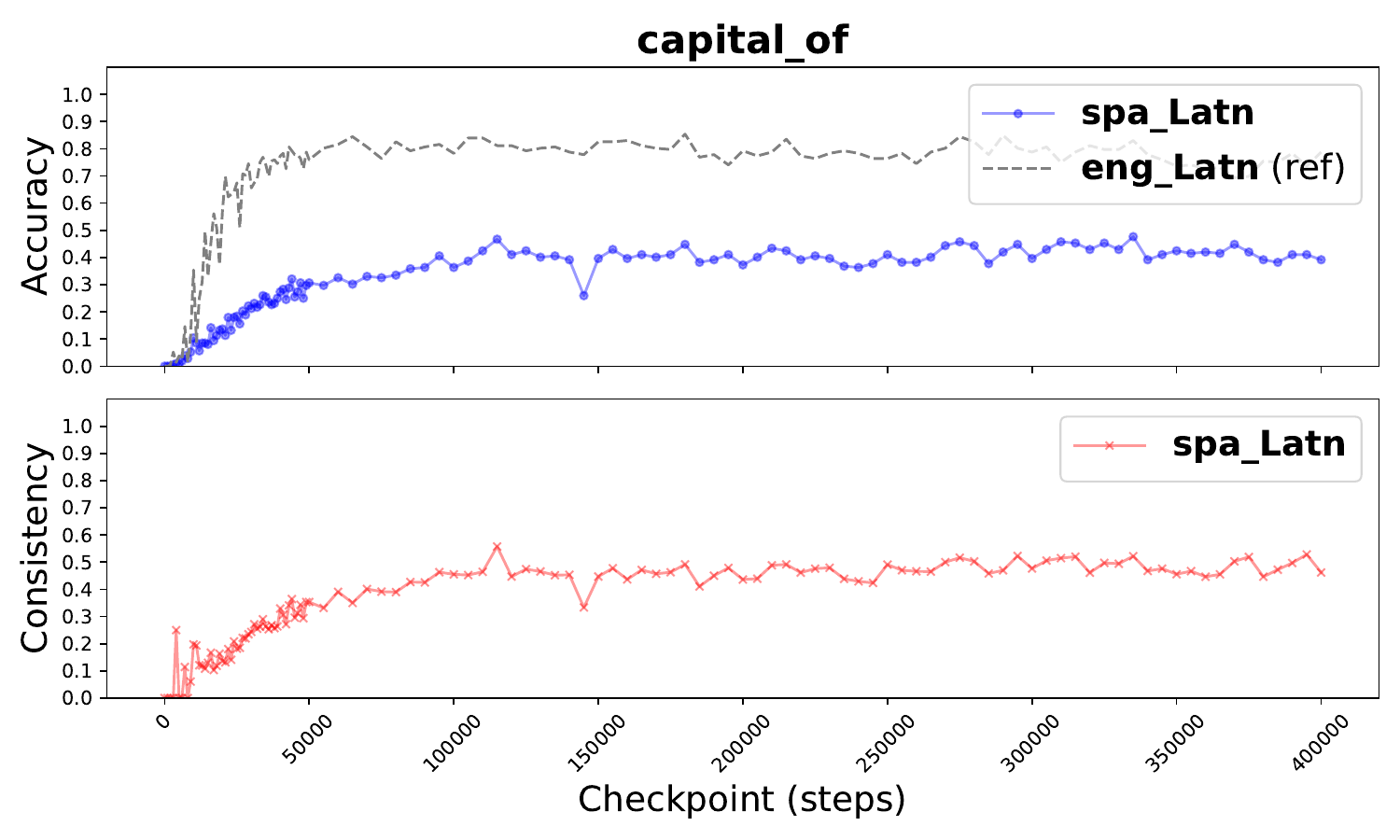}
    \includegraphics[width=0.24\textwidth]{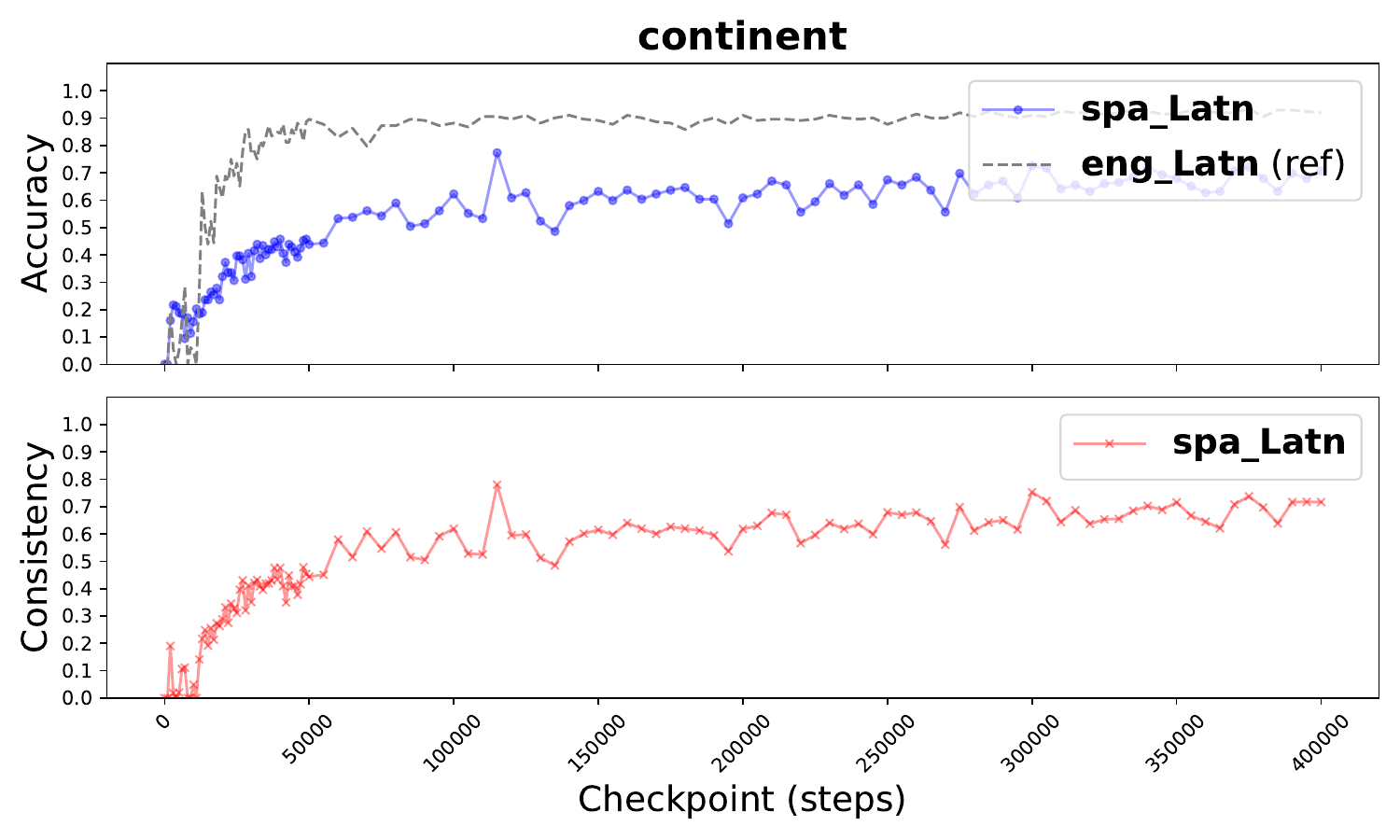}
    \includegraphics[width=0.24\textwidth]{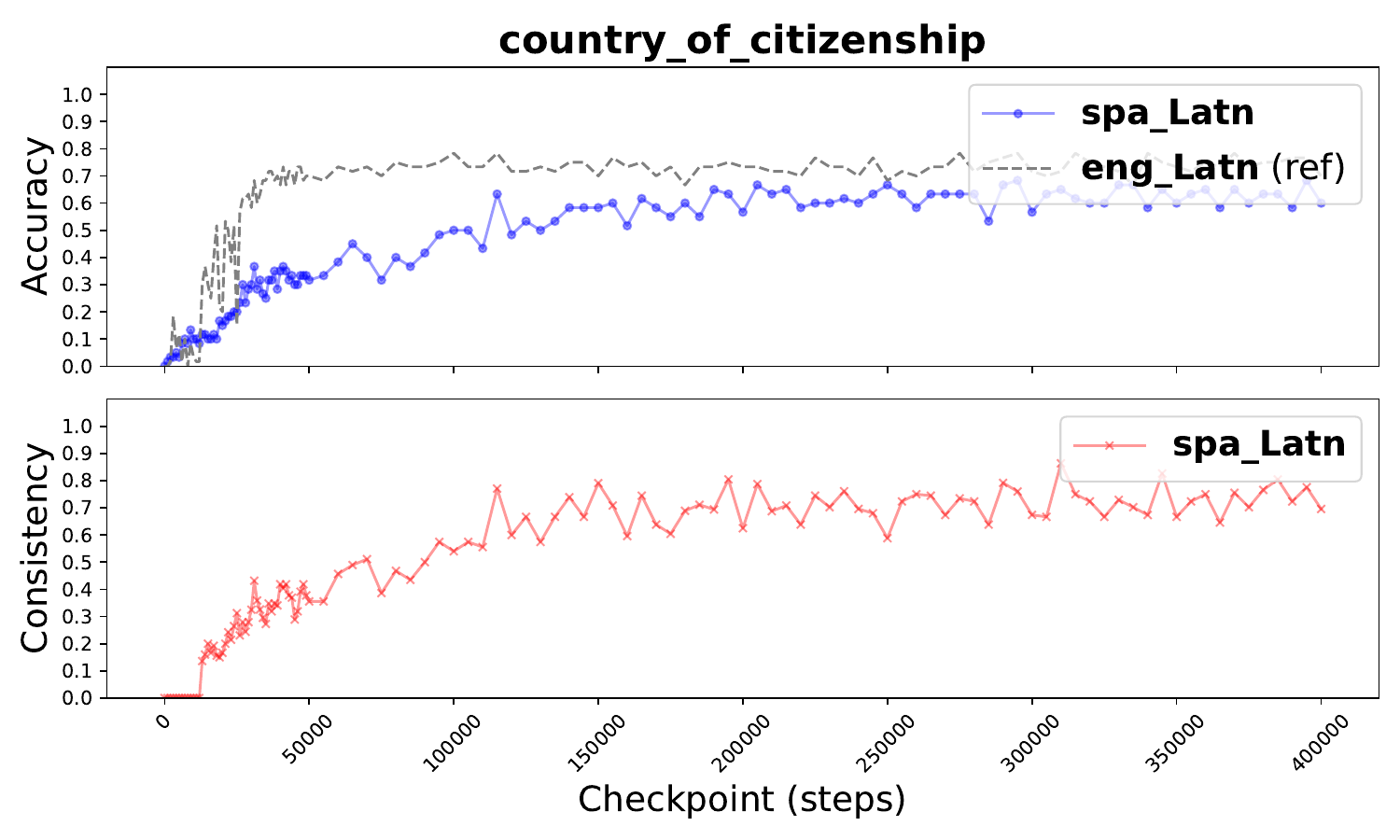}
    \includegraphics[width=0.24\textwidth]{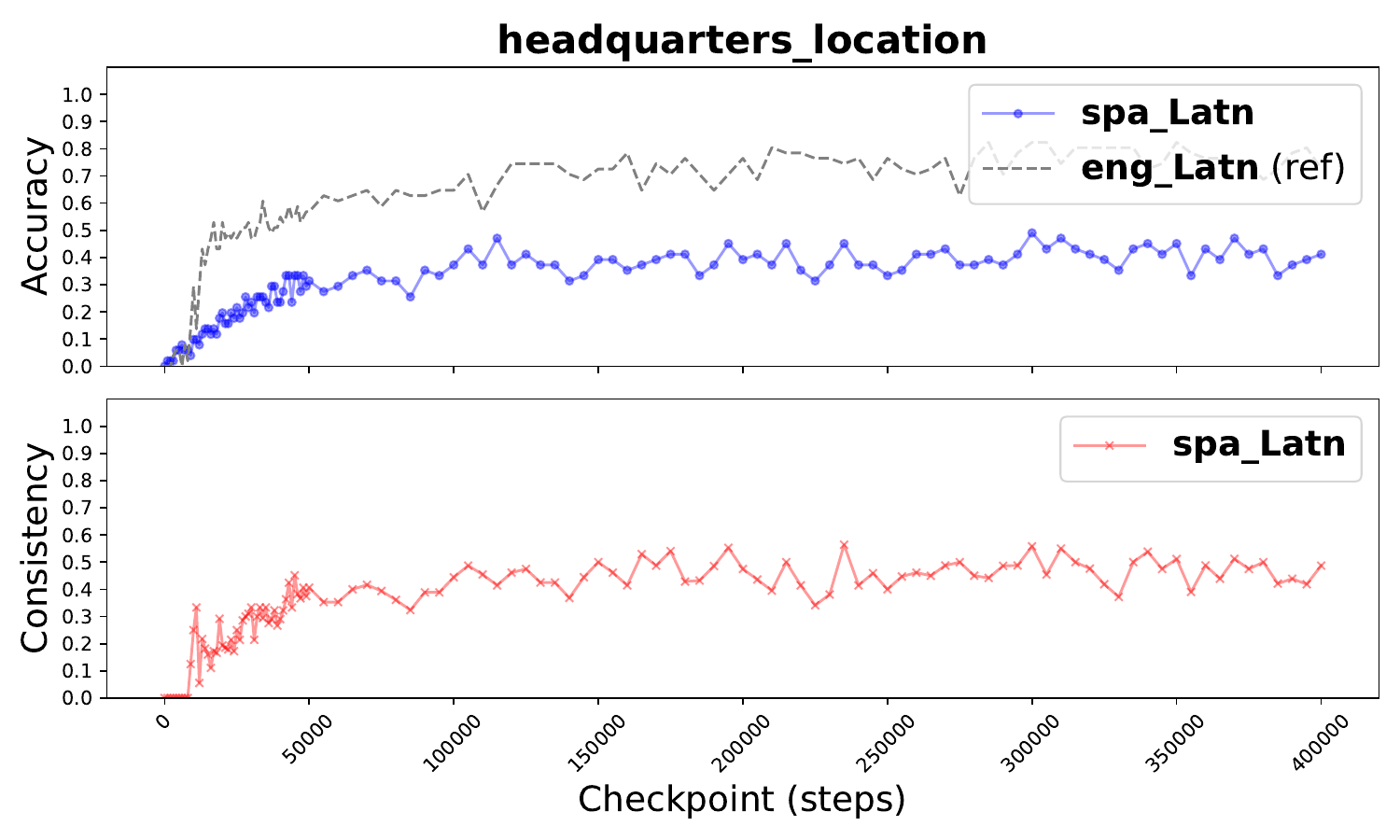}
    \includegraphics[width=0.24\textwidth]{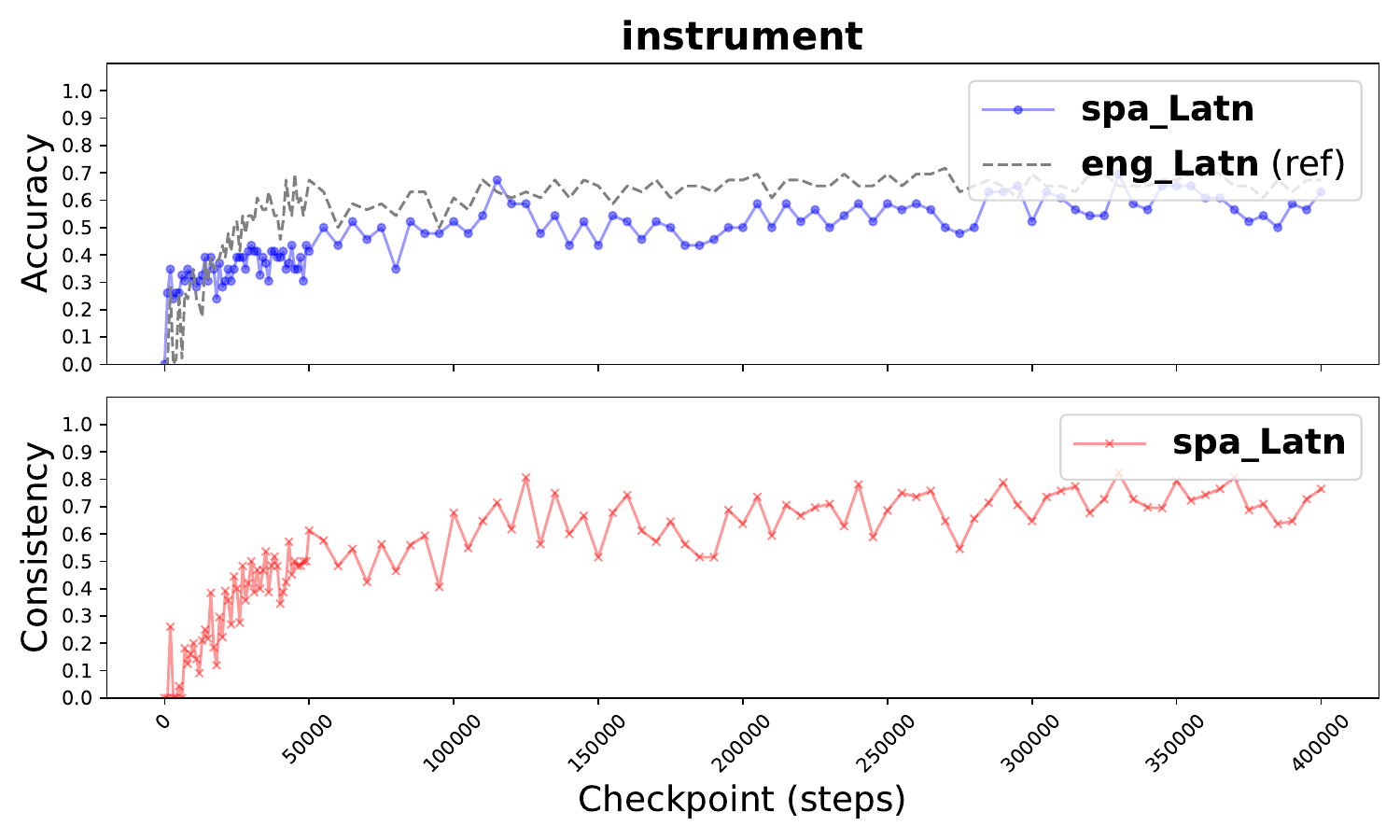}
    \includegraphics[width=0.24\textwidth]{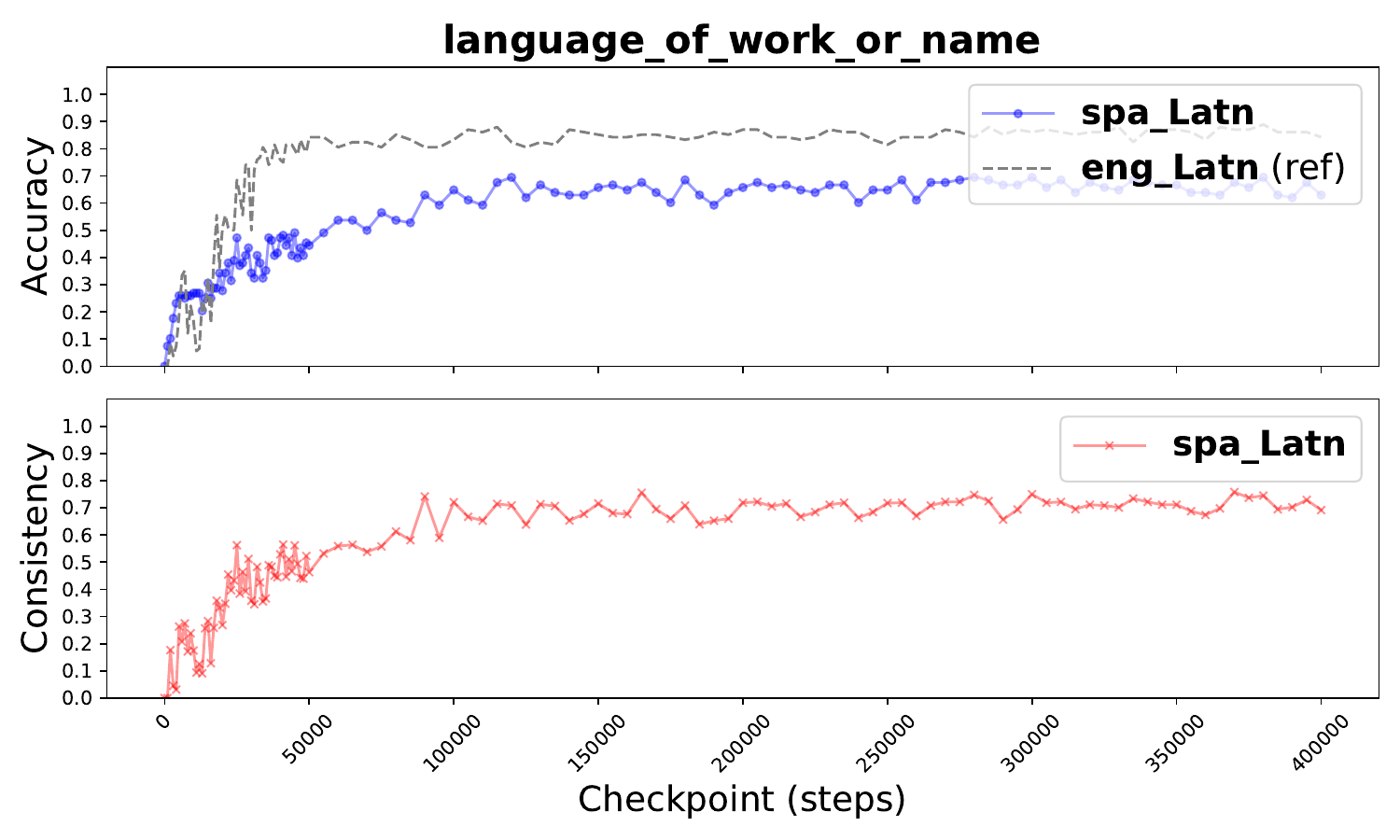}
    \includegraphics[width=0.24\textwidth]{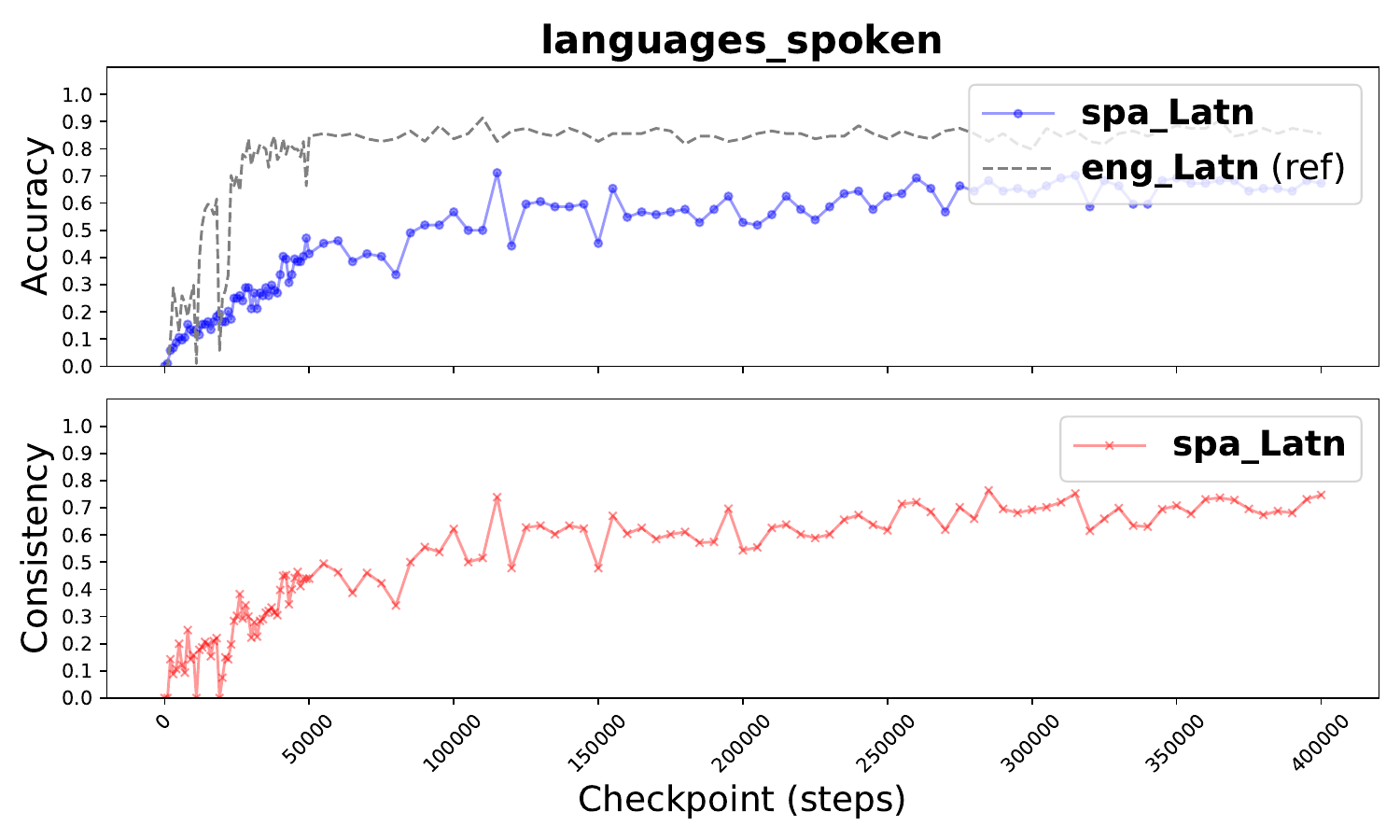}
    \includegraphics[width=0.24\textwidth]{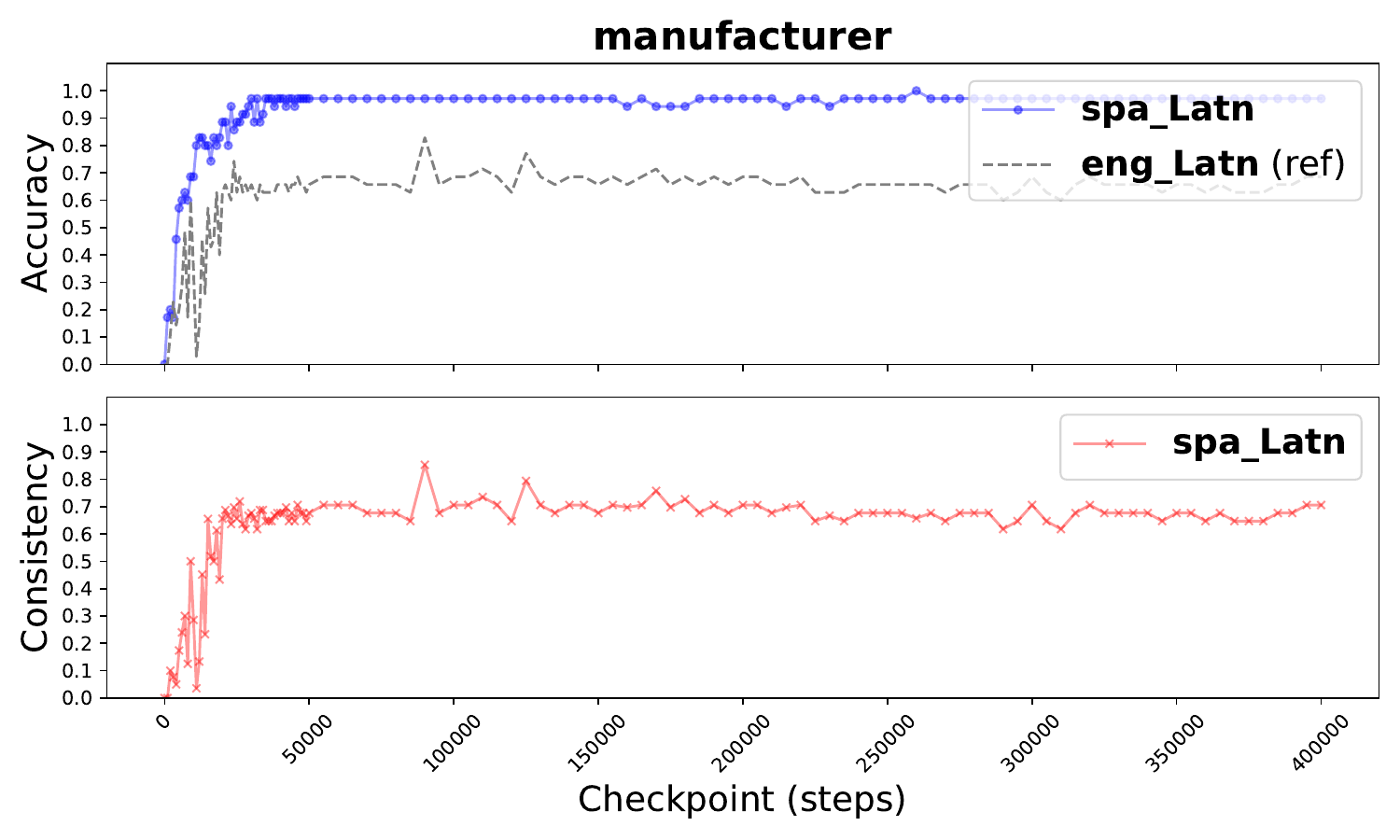}
    \includegraphics[width=0.24\textwidth]{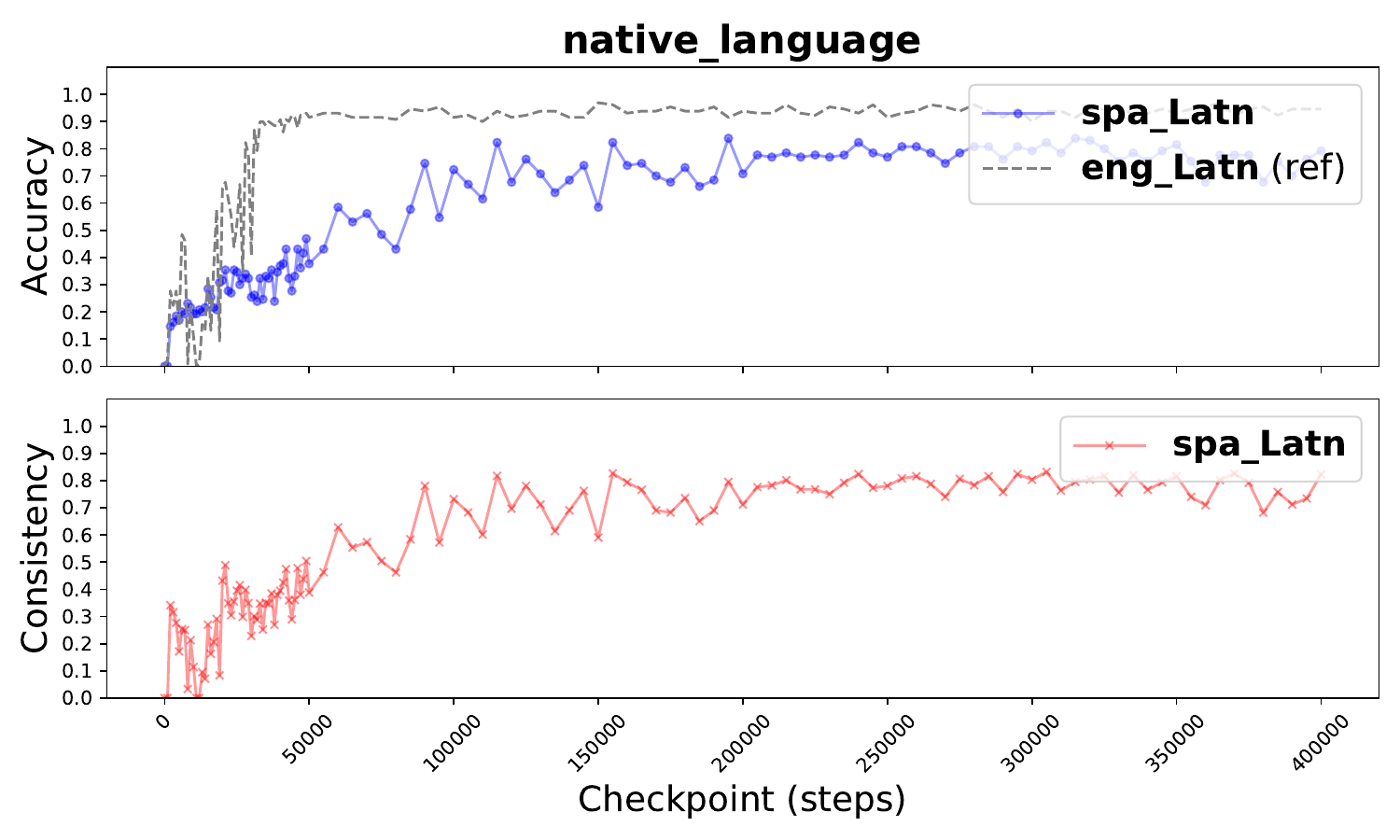}
    \includegraphics[width=0.24\textwidth]{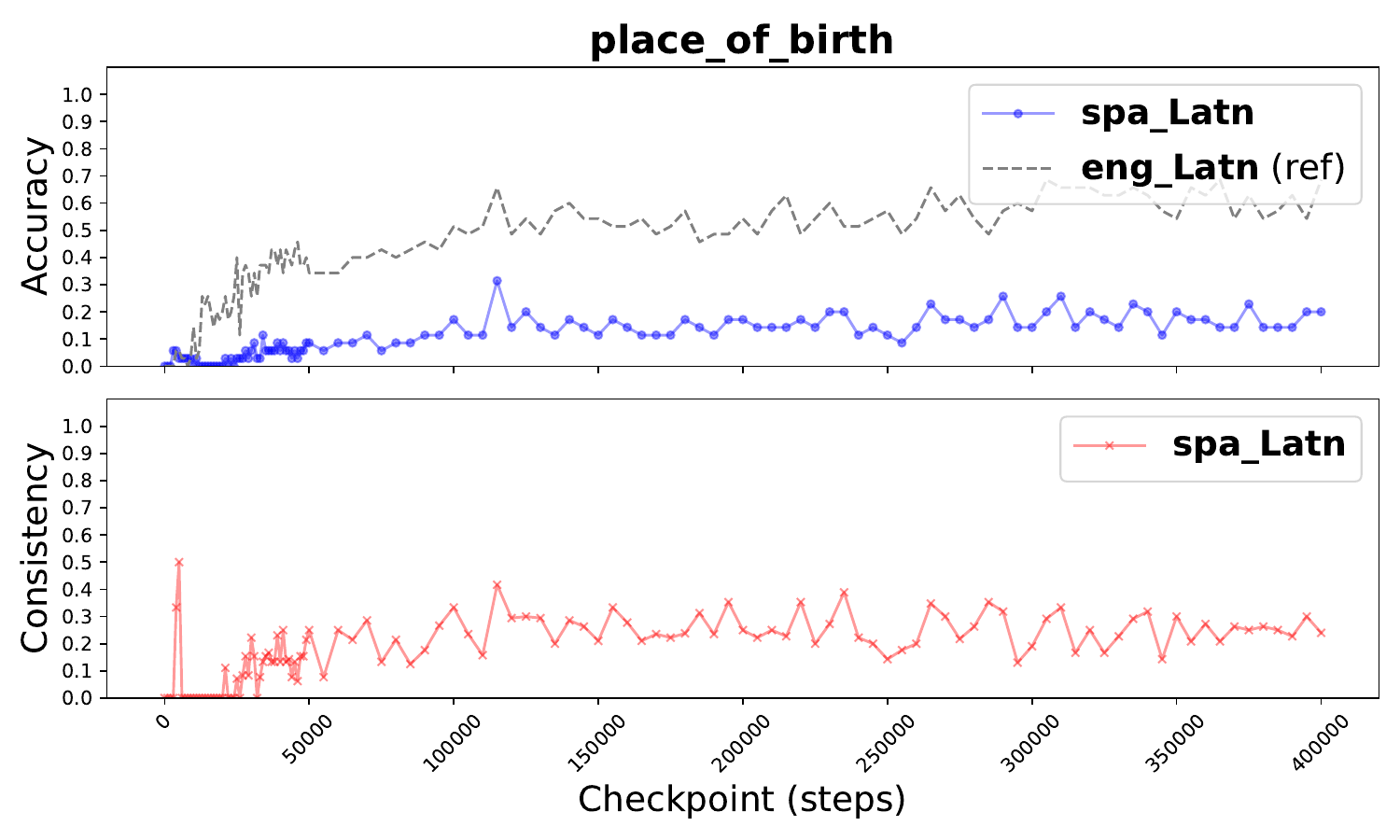}
    \includegraphics[width=0.24\textwidth]{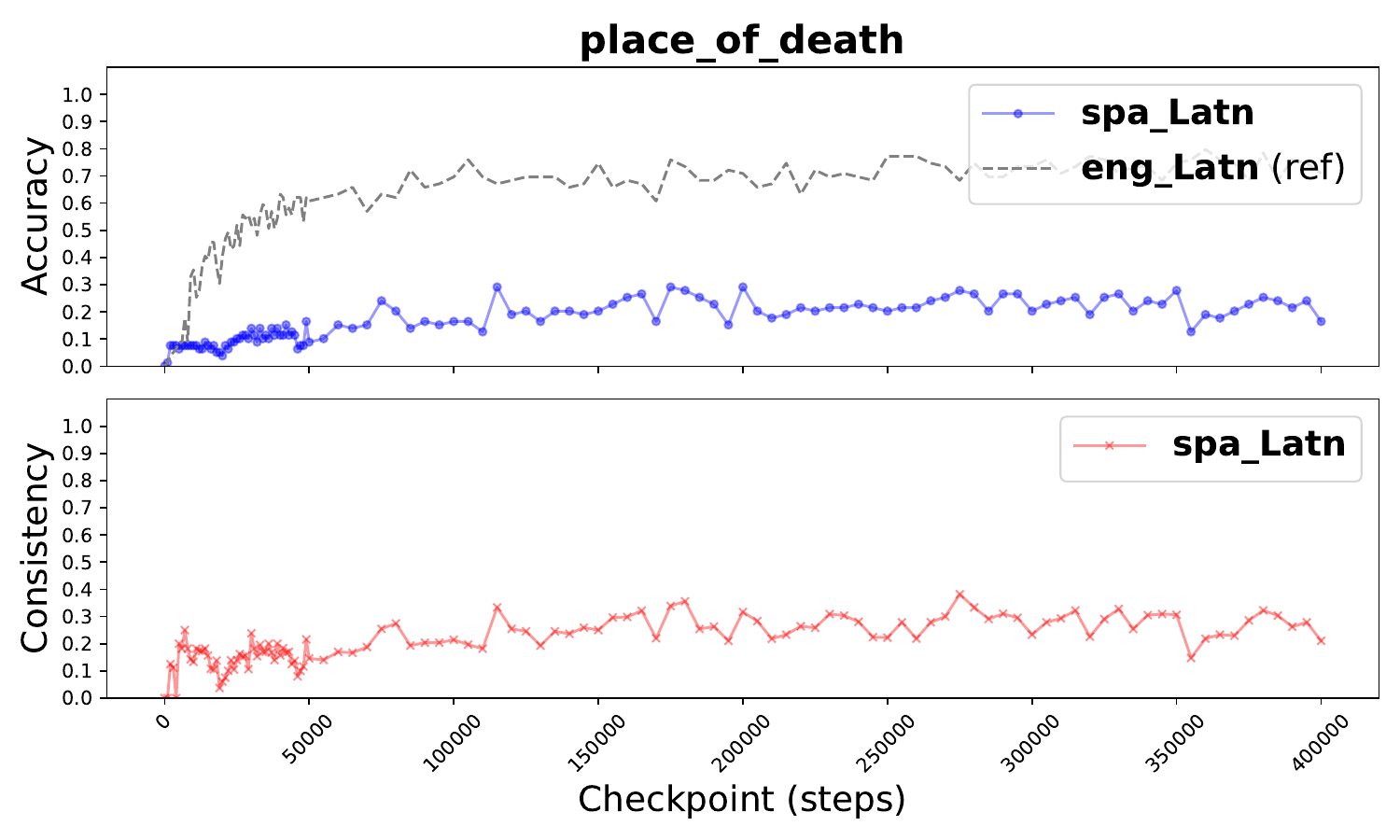}
    \includegraphics[width=0.24\textwidth]{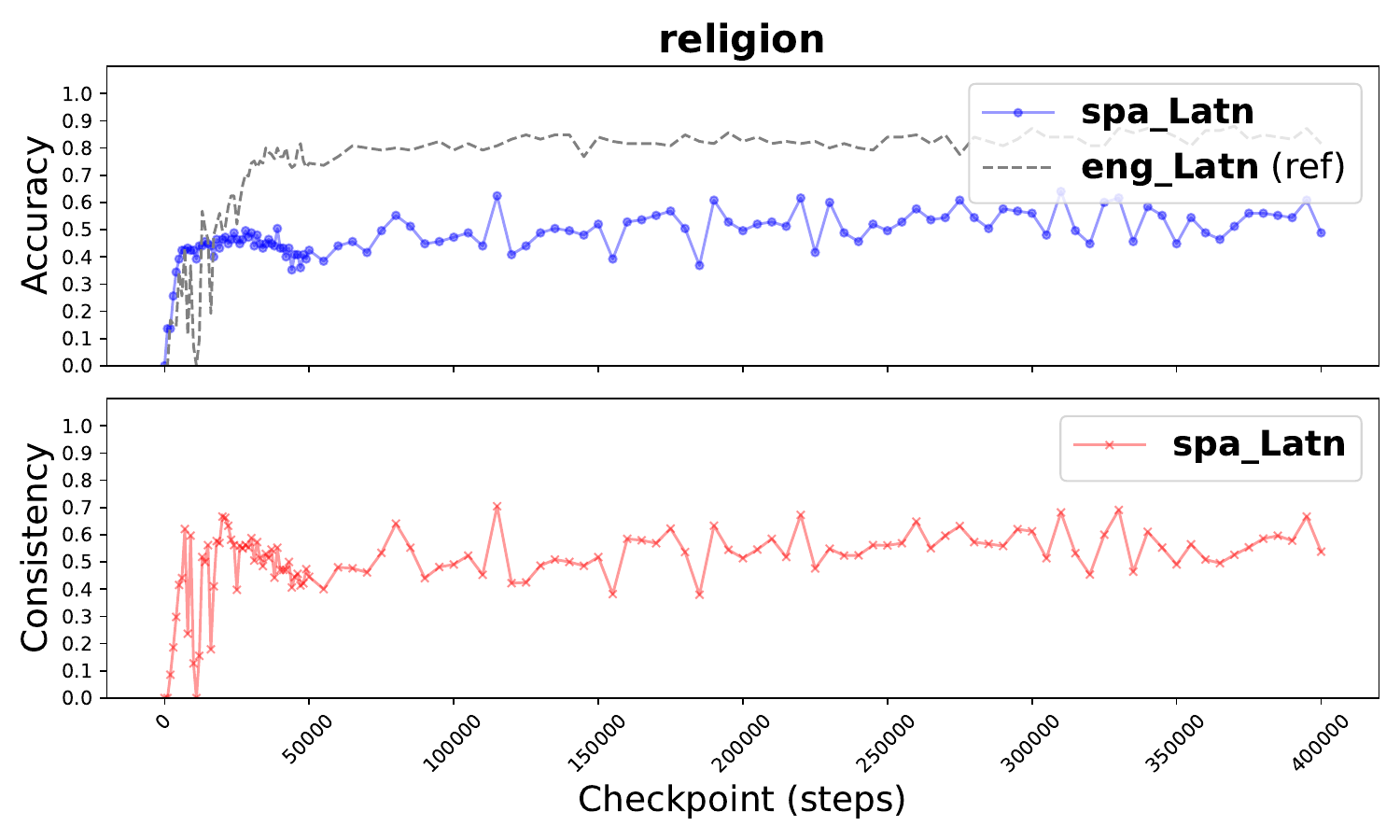}
    \caption{Factual accuracy (ACC) and crosslingual consistency (CO) for each relation type in \textbf{spa\_Latn}.}
    \label{fig:performance_over_checkpoints_es}
\end{figure*}

\begin{figure*}
    \centering
    \includegraphics[width=0.24\textwidth]{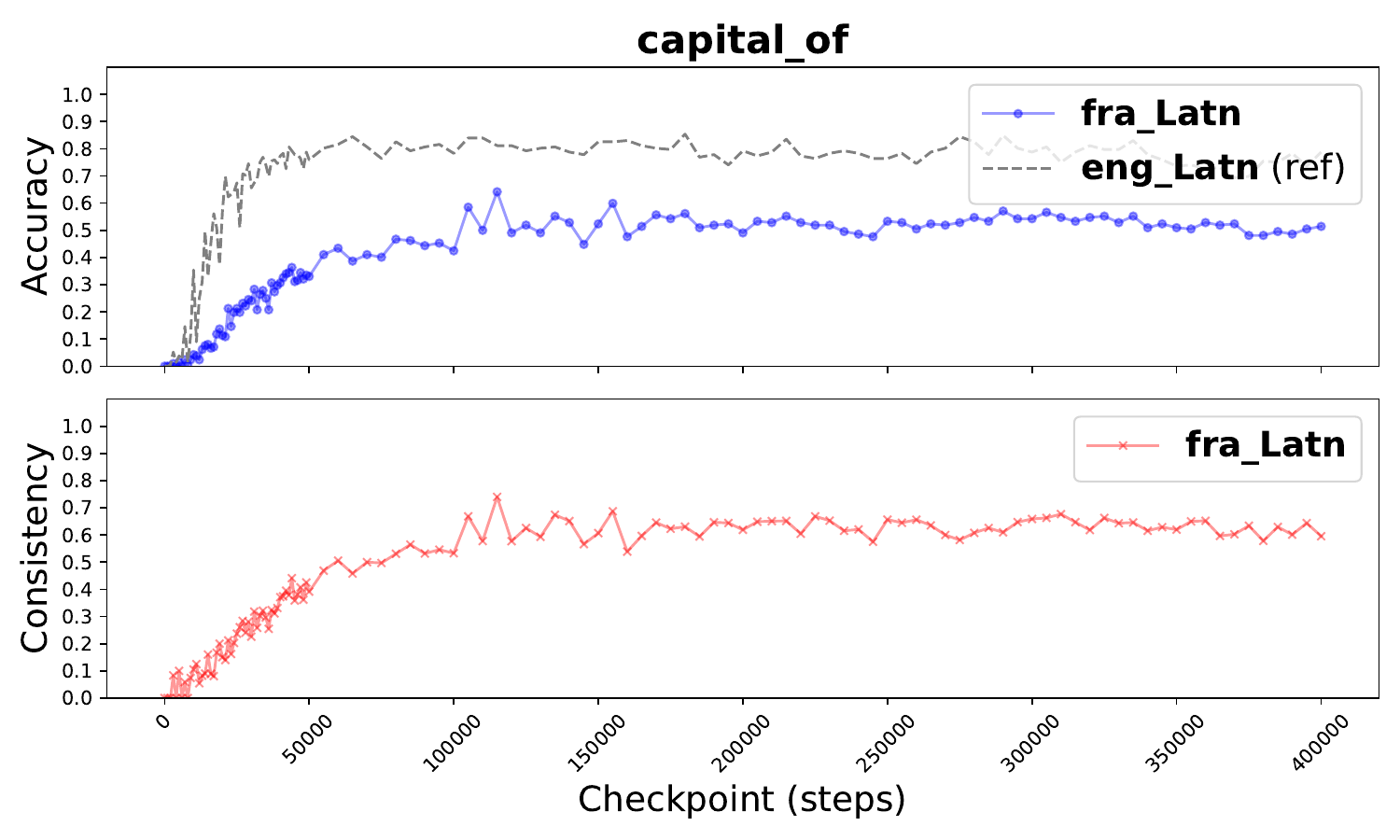}
    \includegraphics[width=0.24\textwidth]{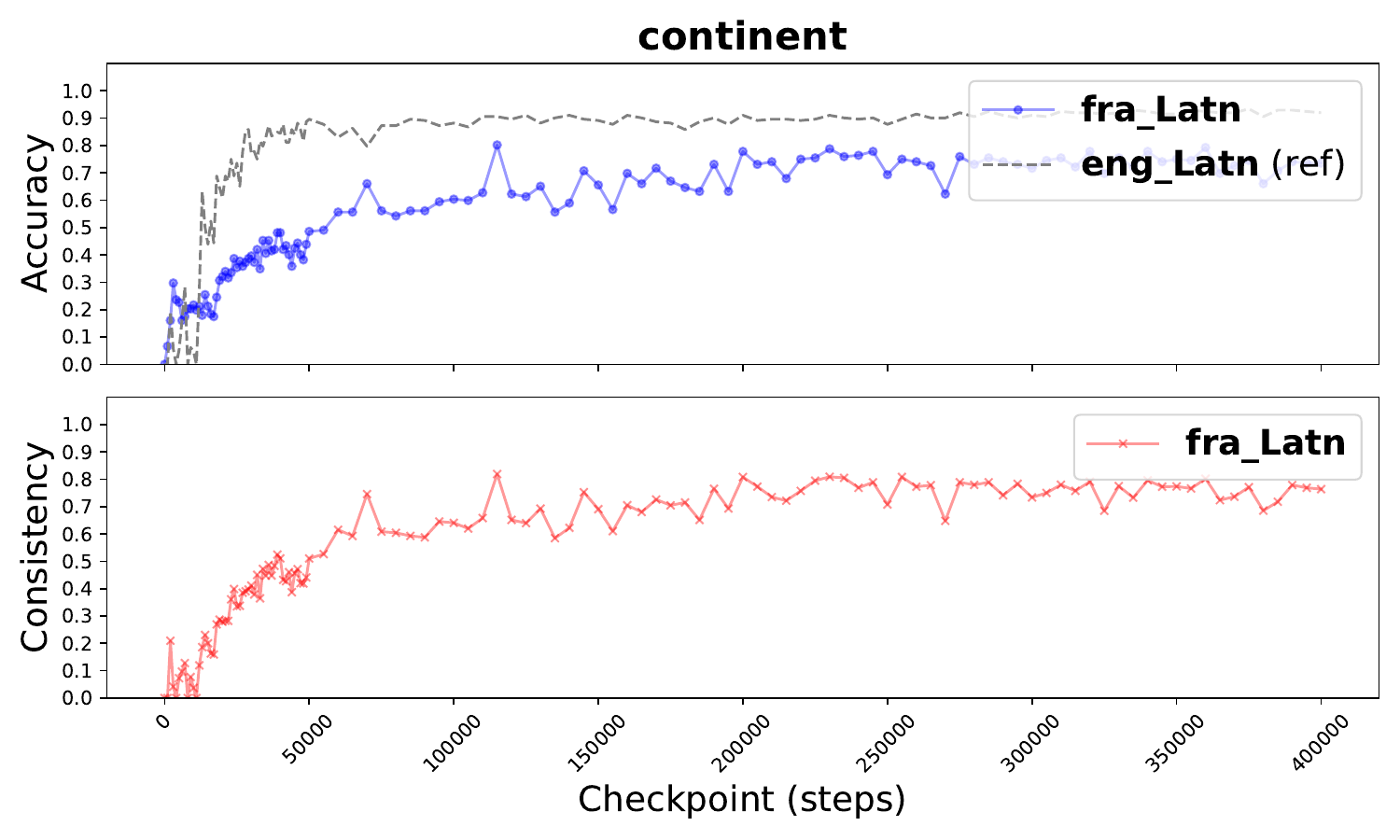}
    \includegraphics[width=0.24\textwidth]{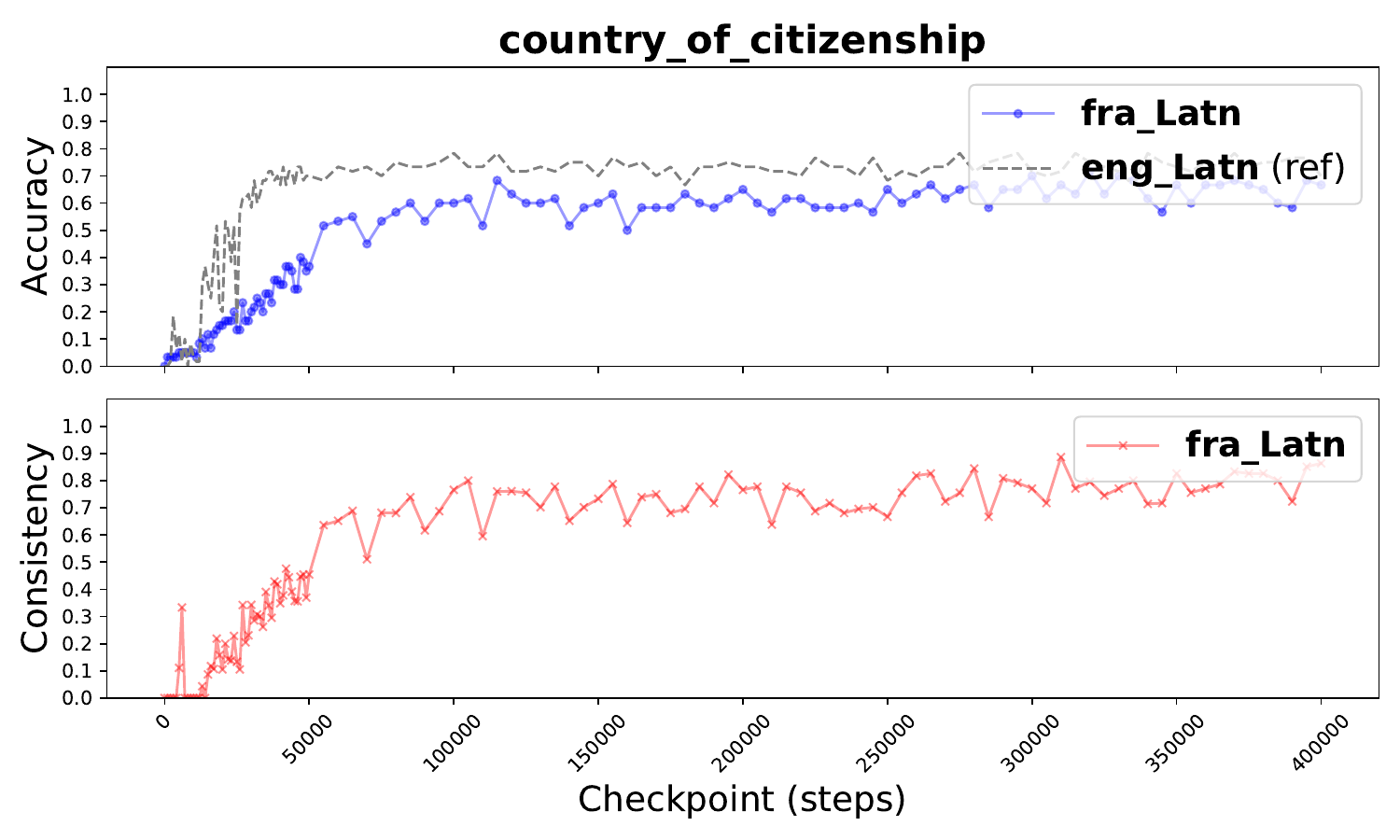}
    \includegraphics[width=0.24\textwidth]{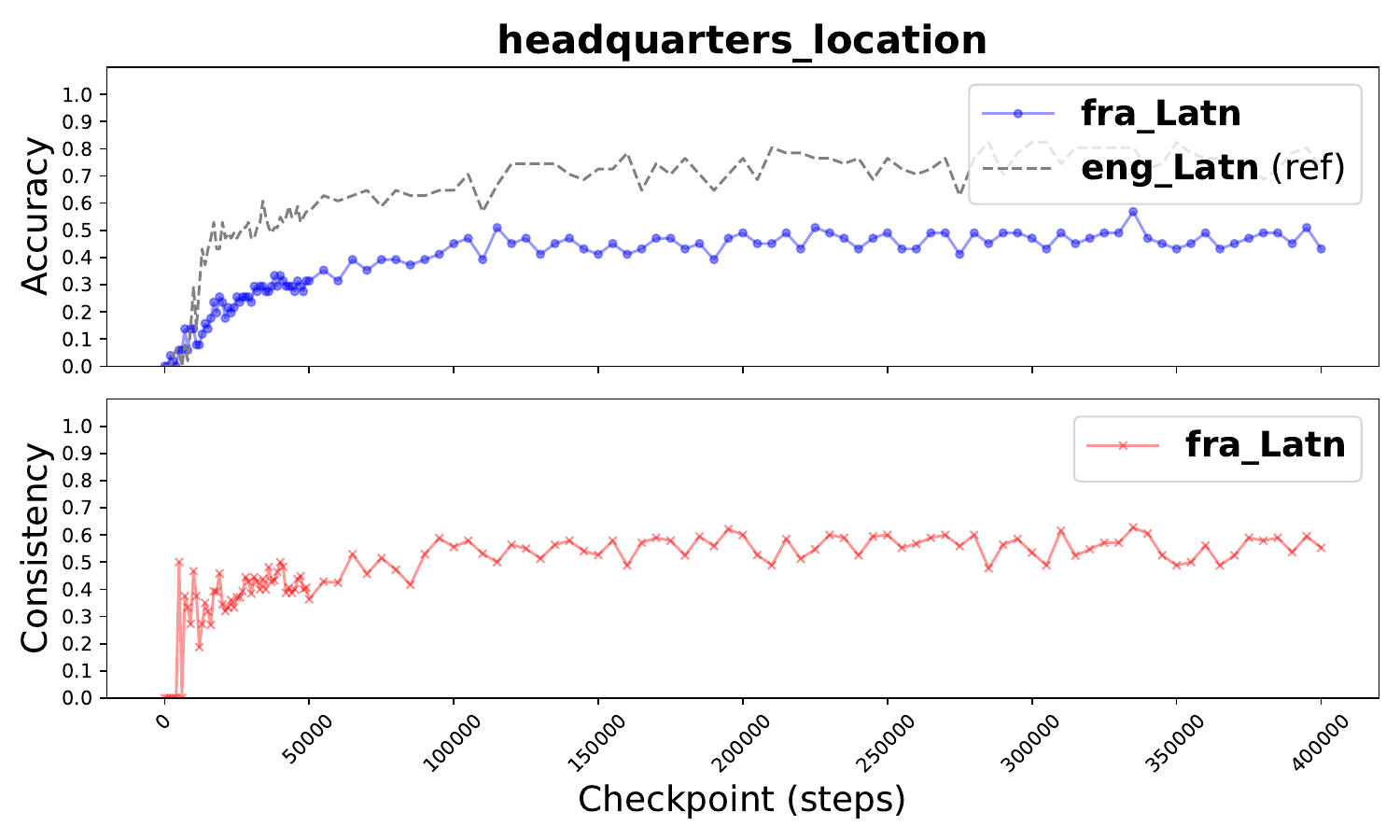}
    \includegraphics[width=0.24\textwidth]{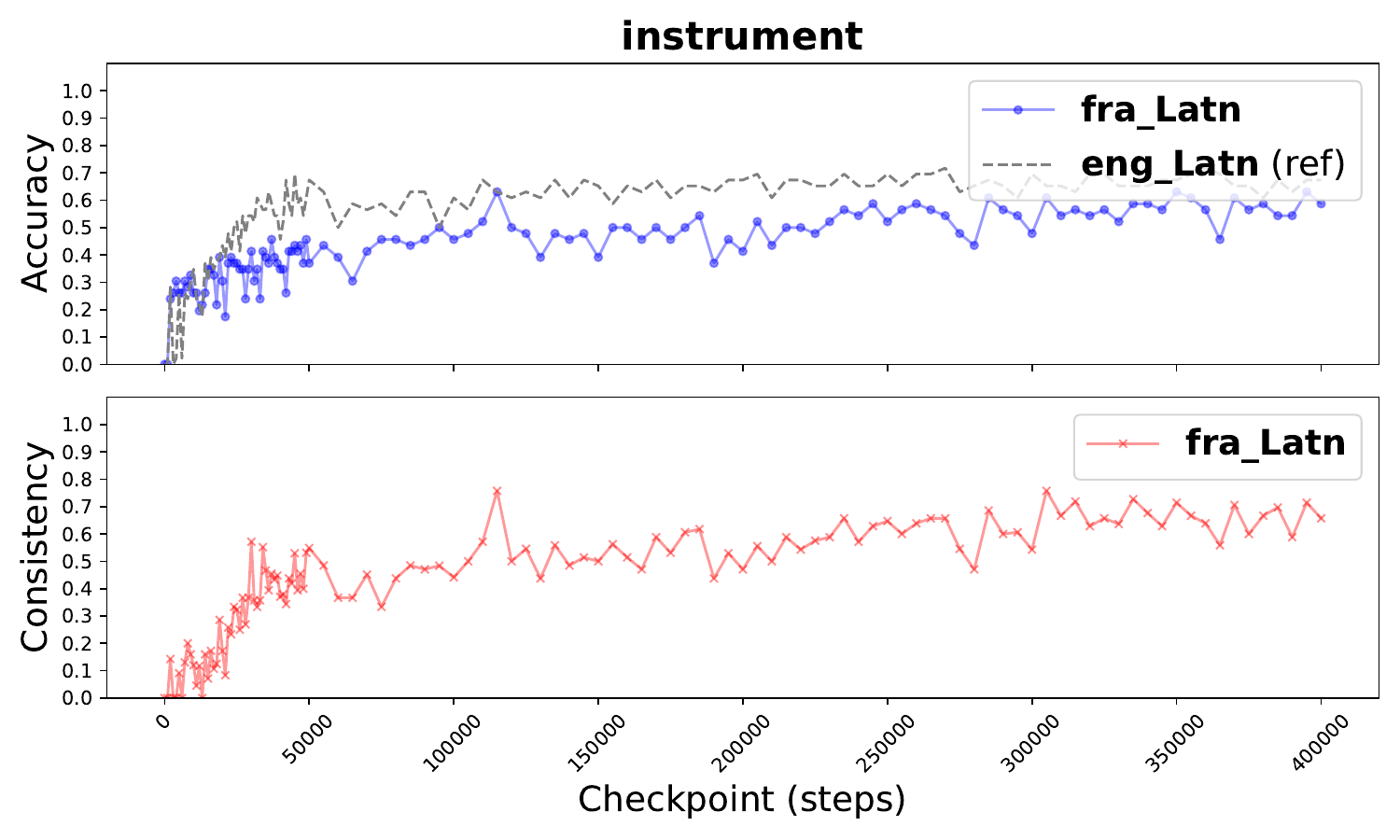}
    \includegraphics[width=0.24\textwidth]{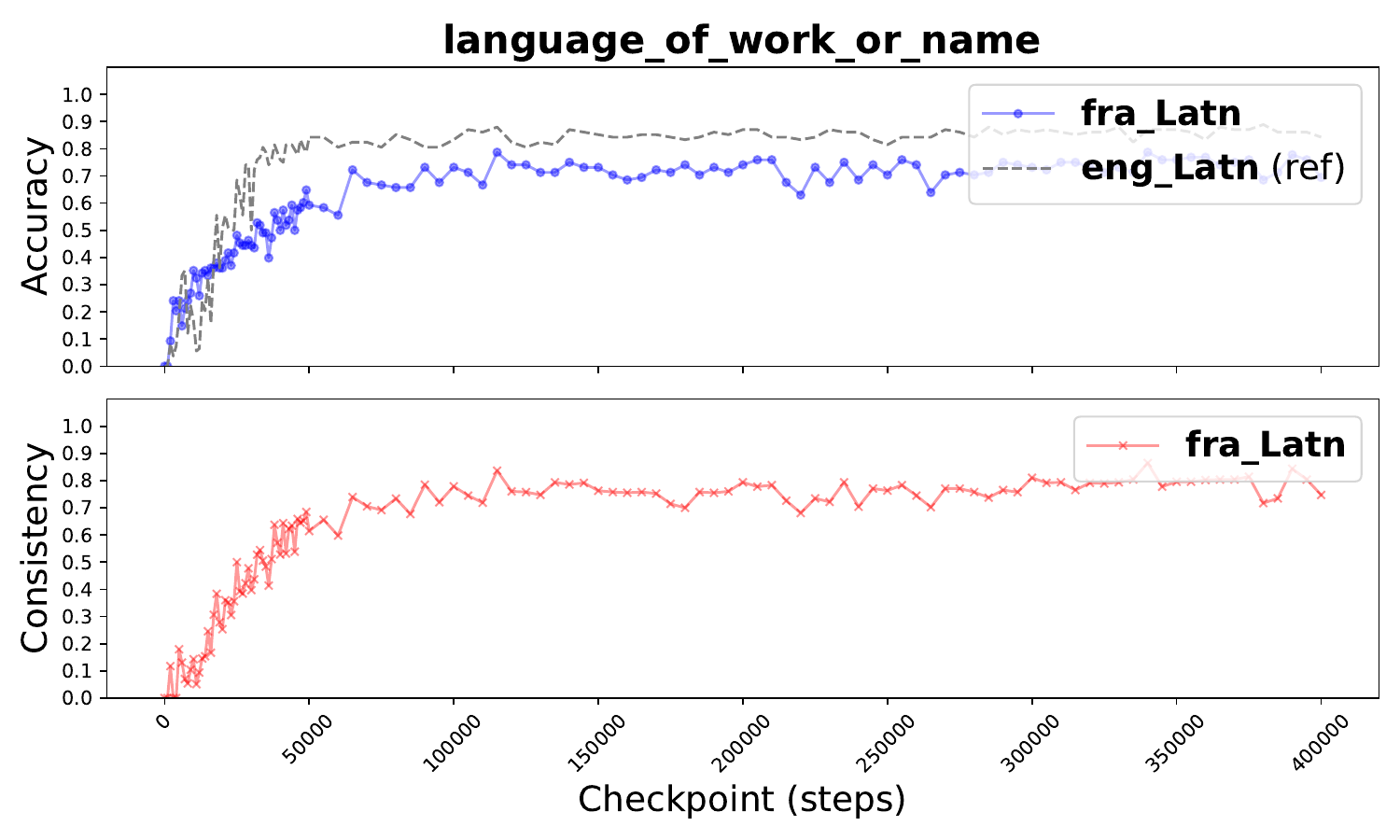}
    \includegraphics[width=0.24\textwidth]{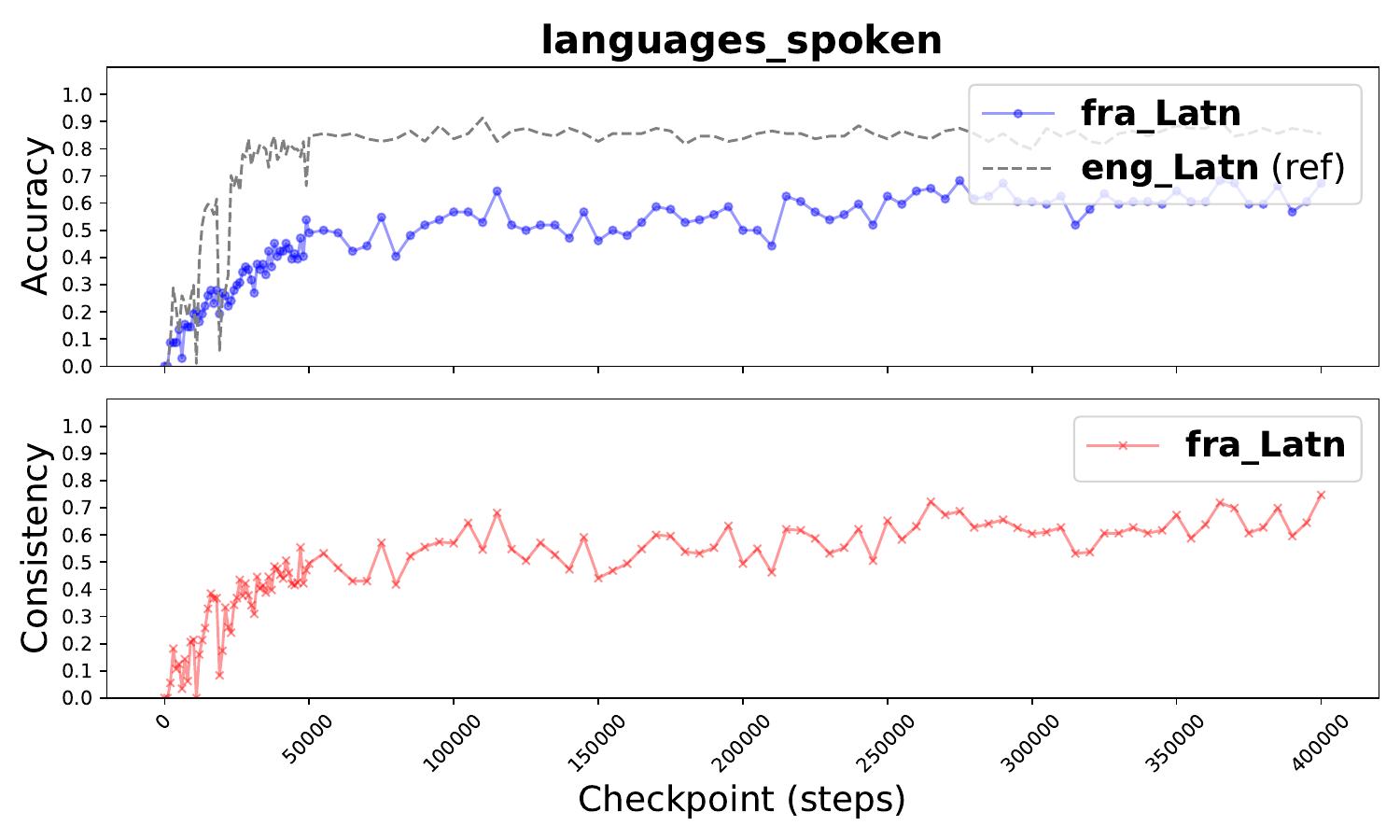}
    \includegraphics[width=0.24\textwidth]{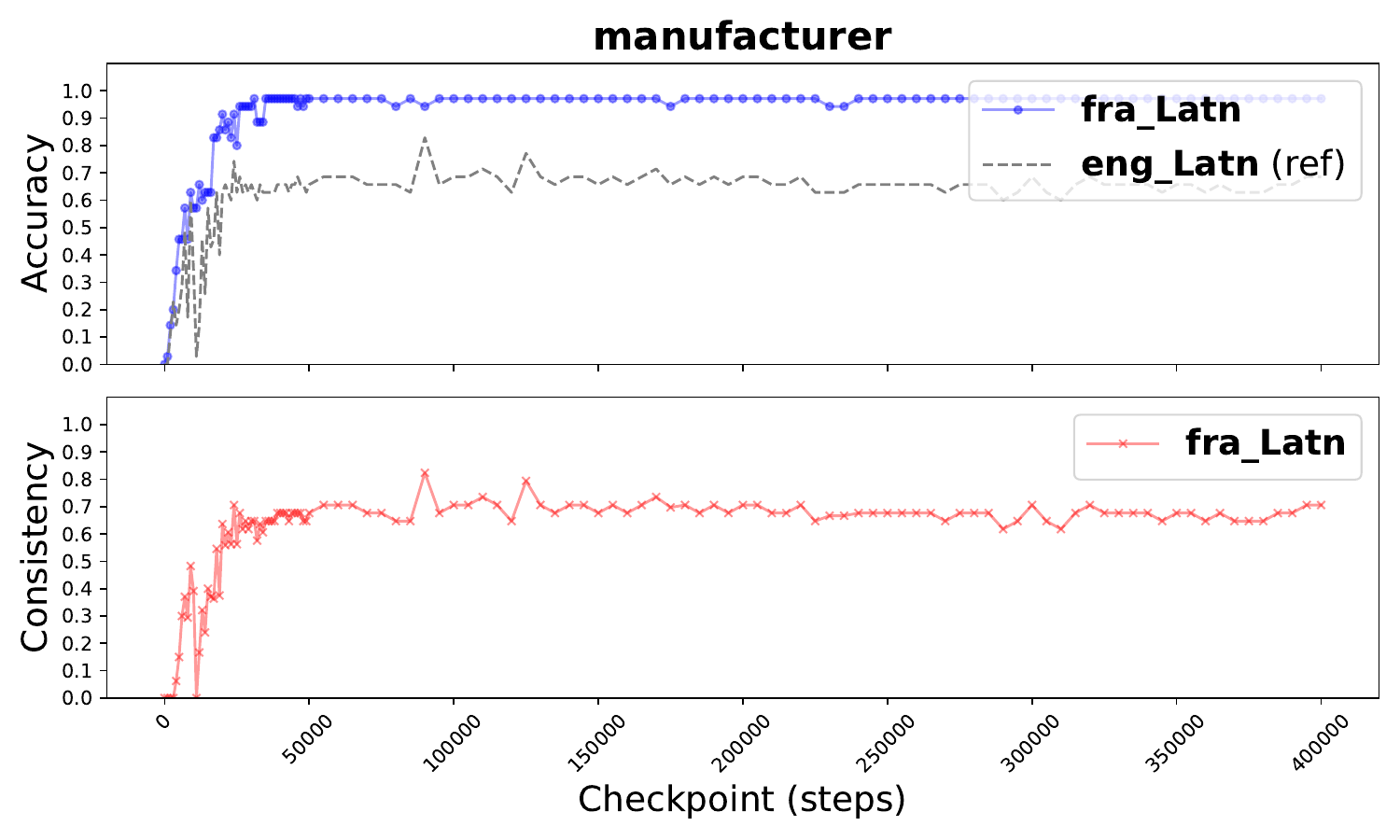}
    \includegraphics[width=0.24\textwidth]{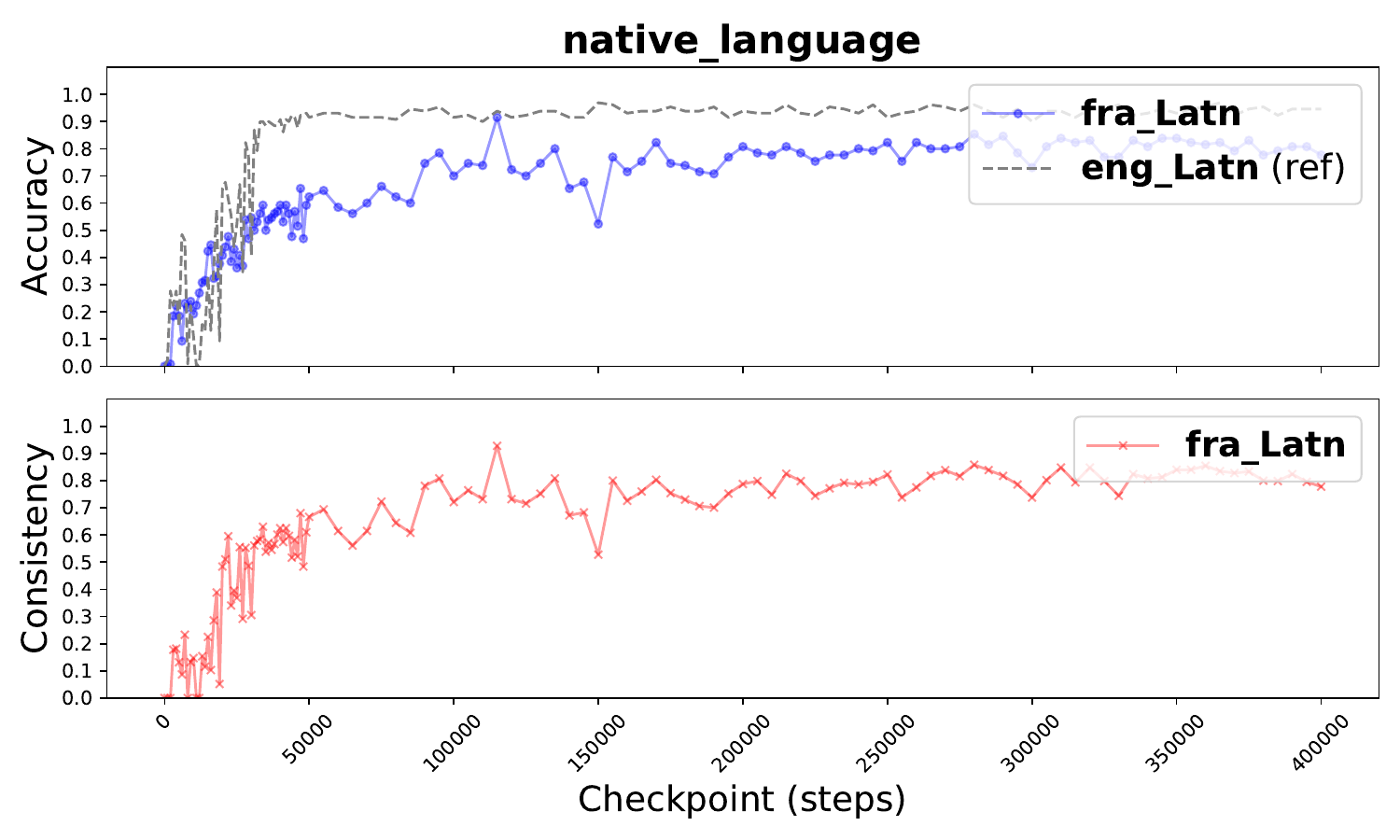}
    \includegraphics[width=0.24\textwidth]{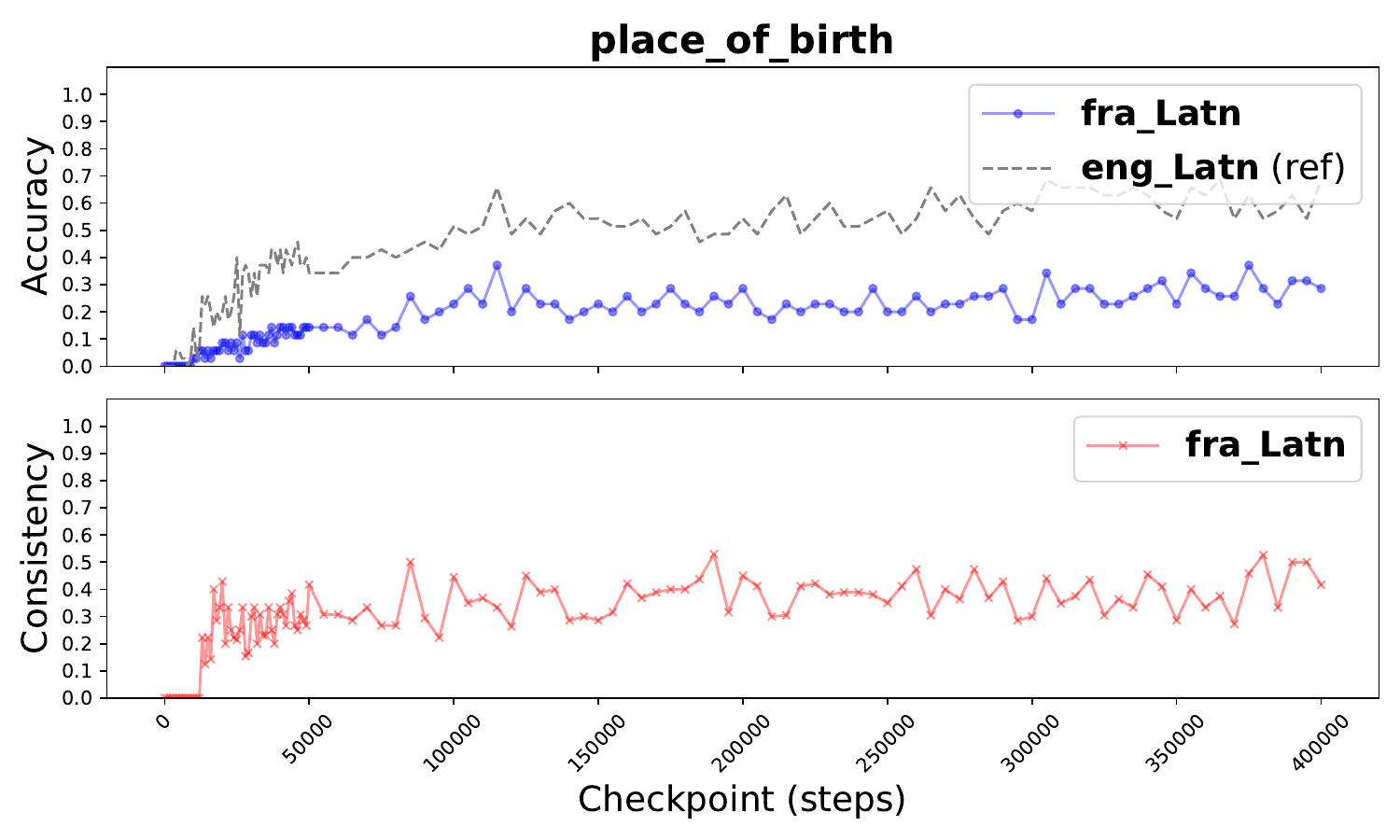}
    \includegraphics[width=0.24\textwidth]{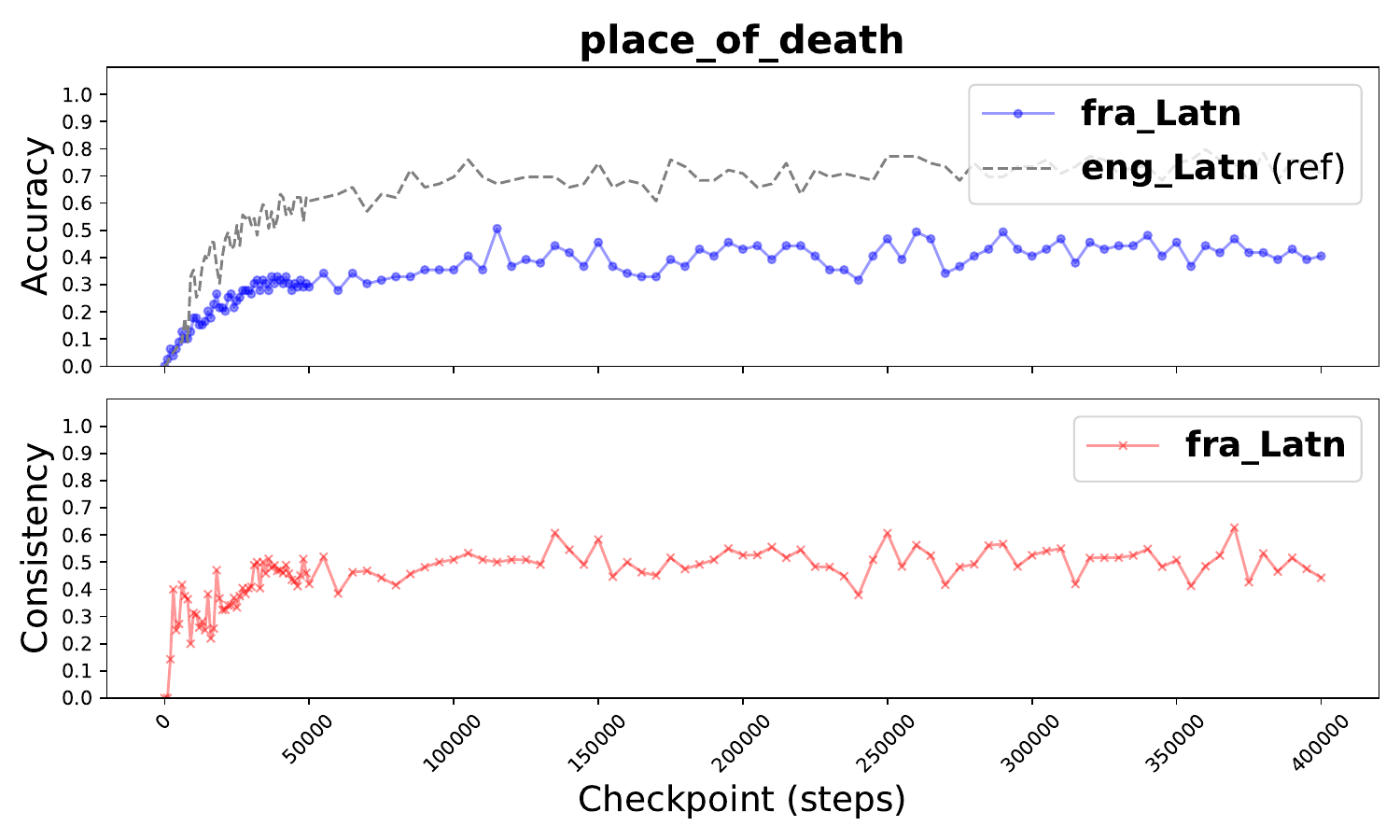}
    \includegraphics[width=0.24\textwidth]{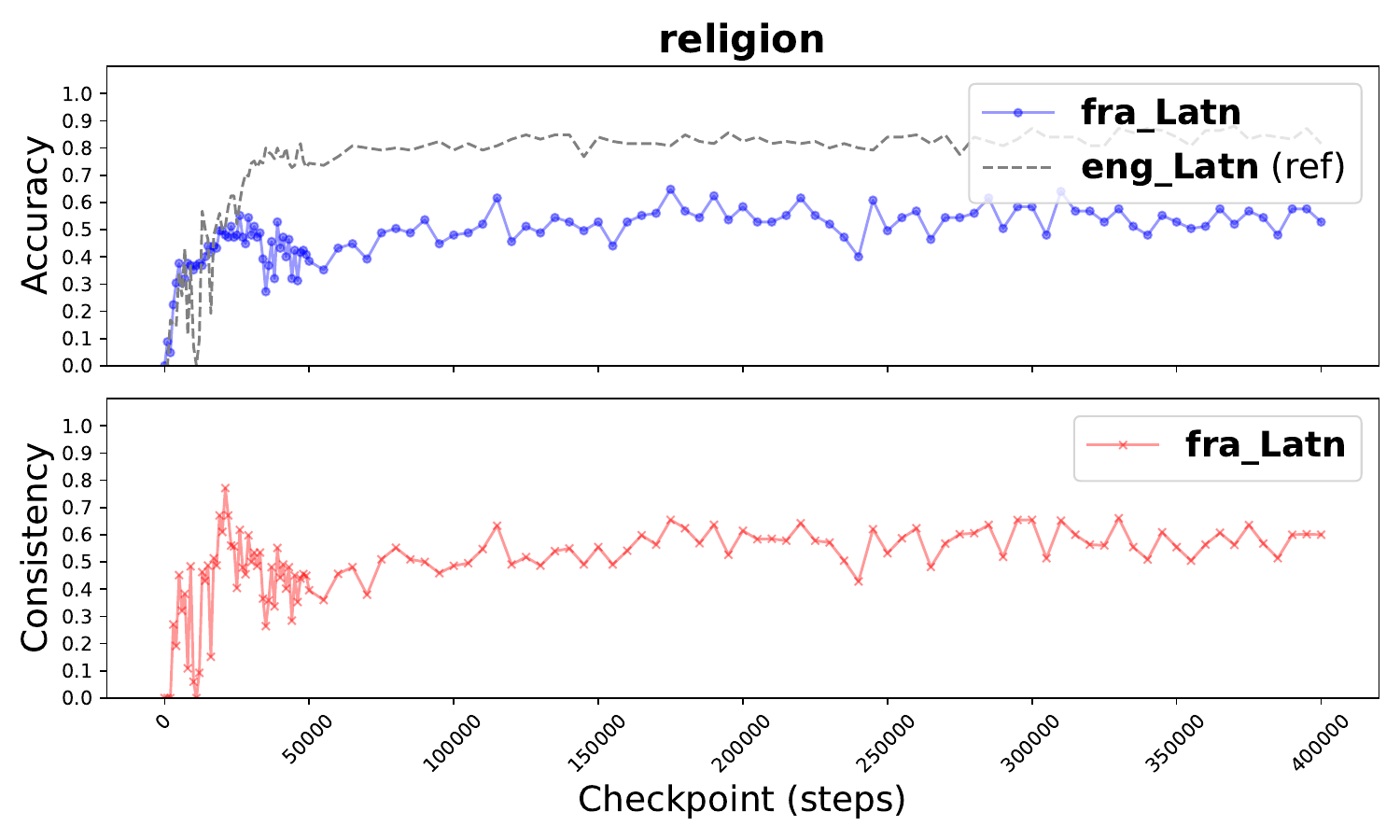}
    \caption{Factual accuracy (ACC) and crosslingual consistency (CO) for each relation type in \textbf{fra\_Latn}.}
    \label{fig:performance_over_checkpoints_fr}
\end{figure*}

\begin{figure*}
    \centering
    \includegraphics[width=0.24\textwidth]{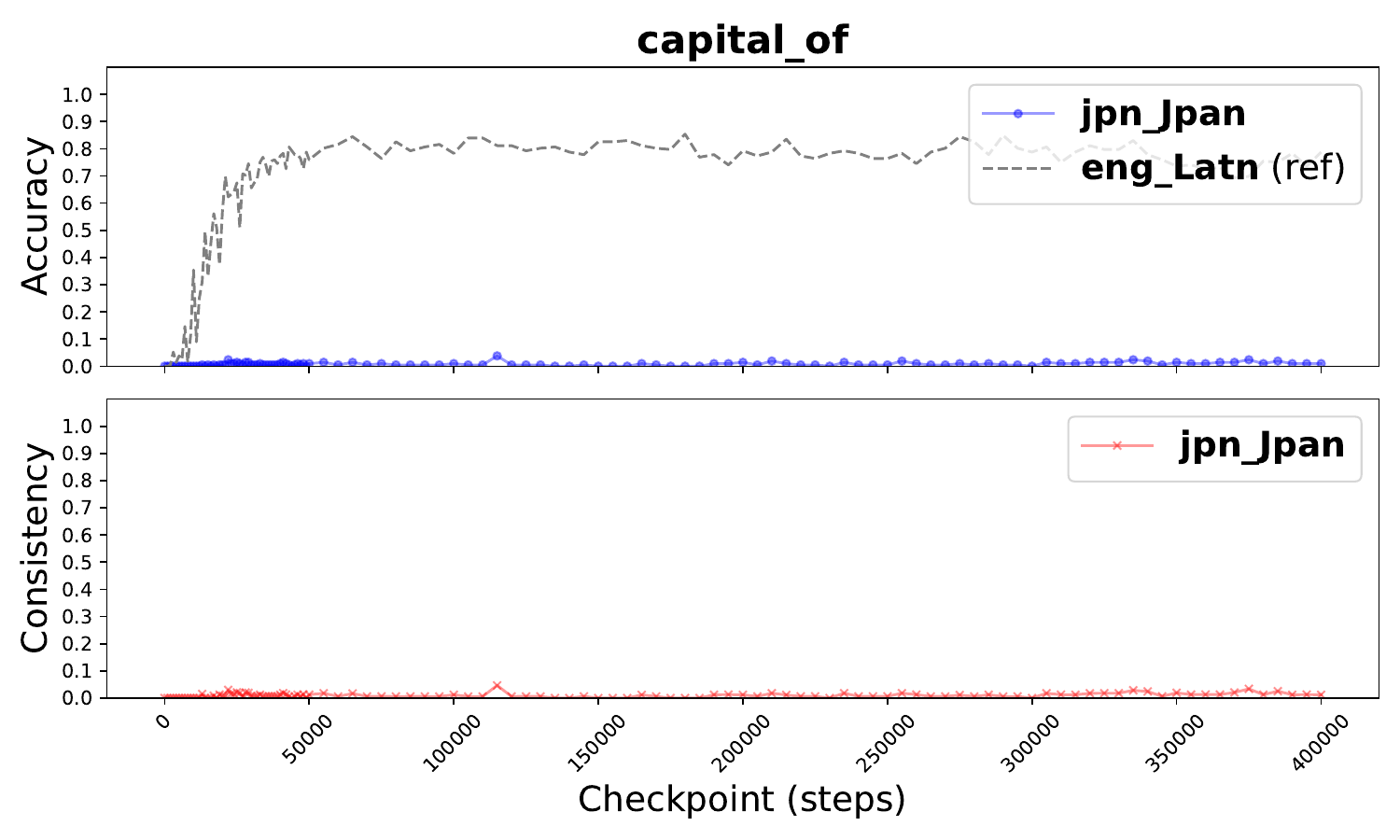}
    \includegraphics[width=0.24\textwidth]{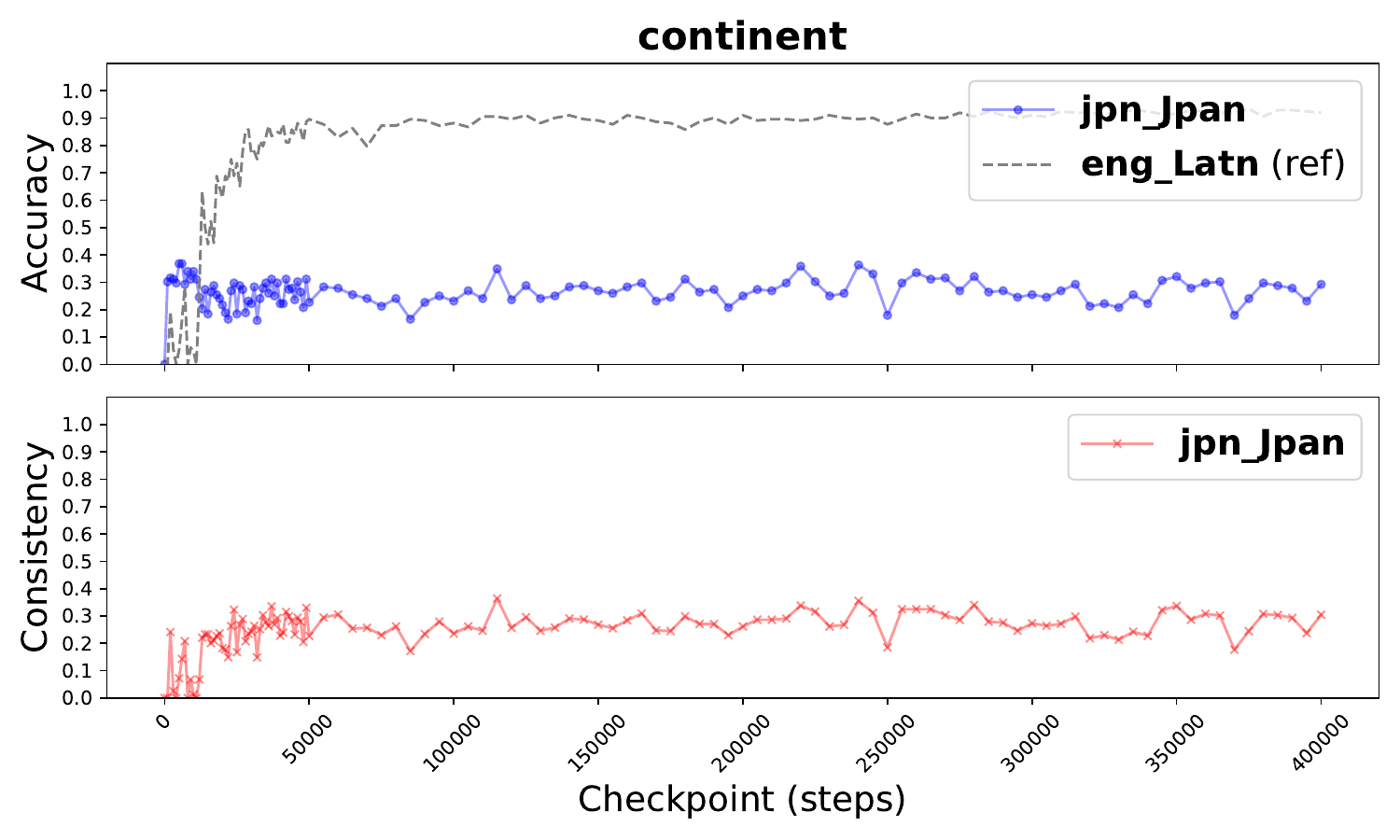}
    \includegraphics[width=0.24\textwidth]{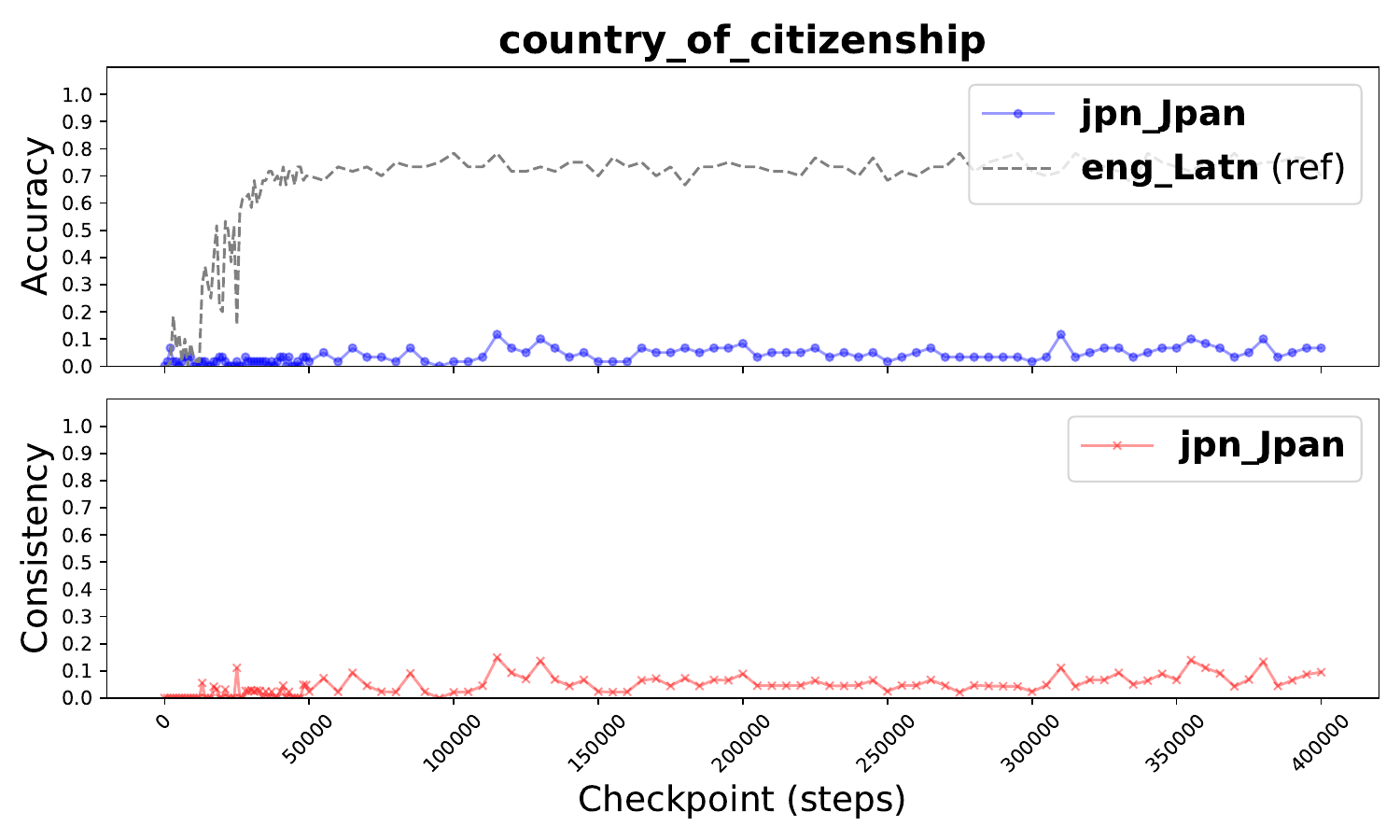}
    \includegraphics[width=0.24\textwidth]{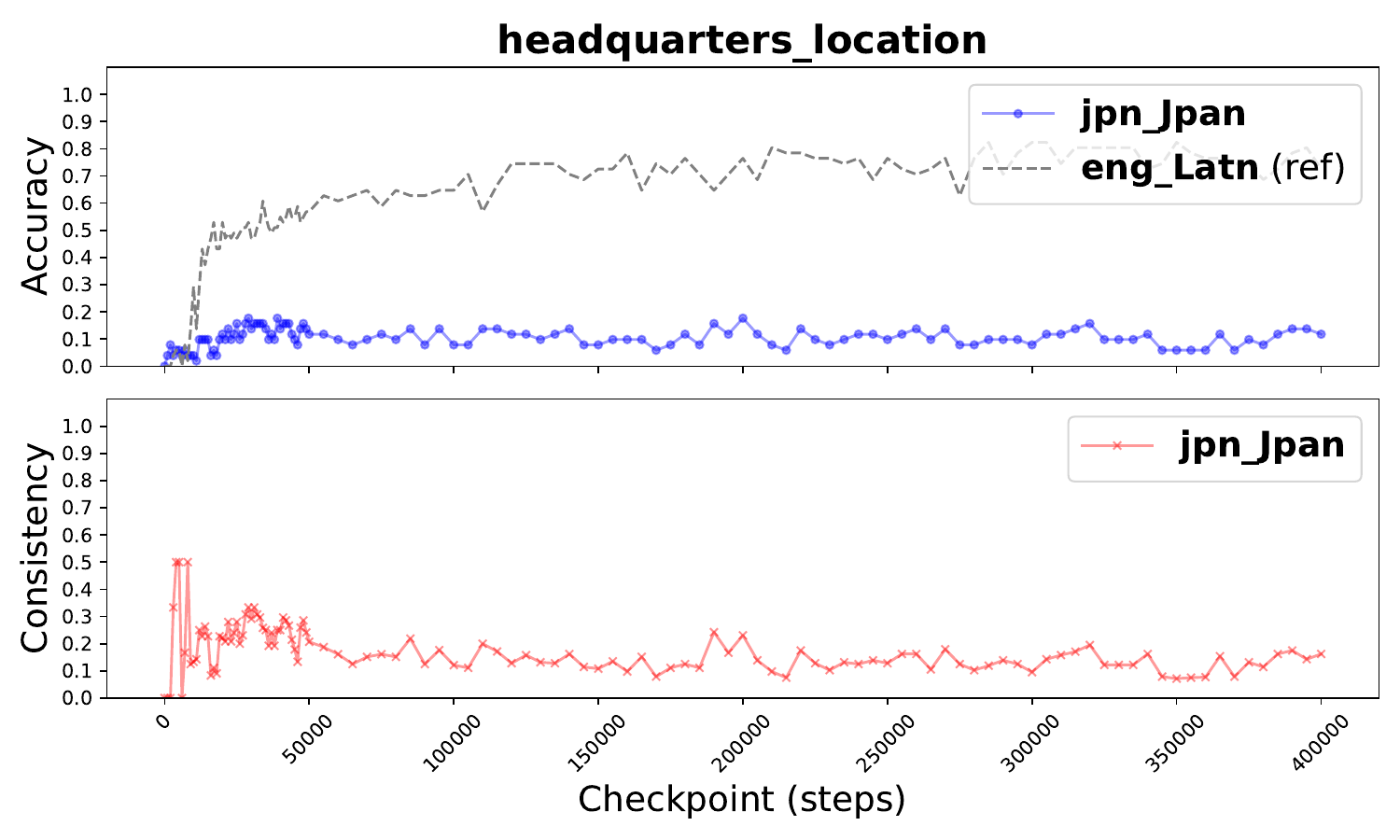}
    \includegraphics[width=0.24\textwidth]{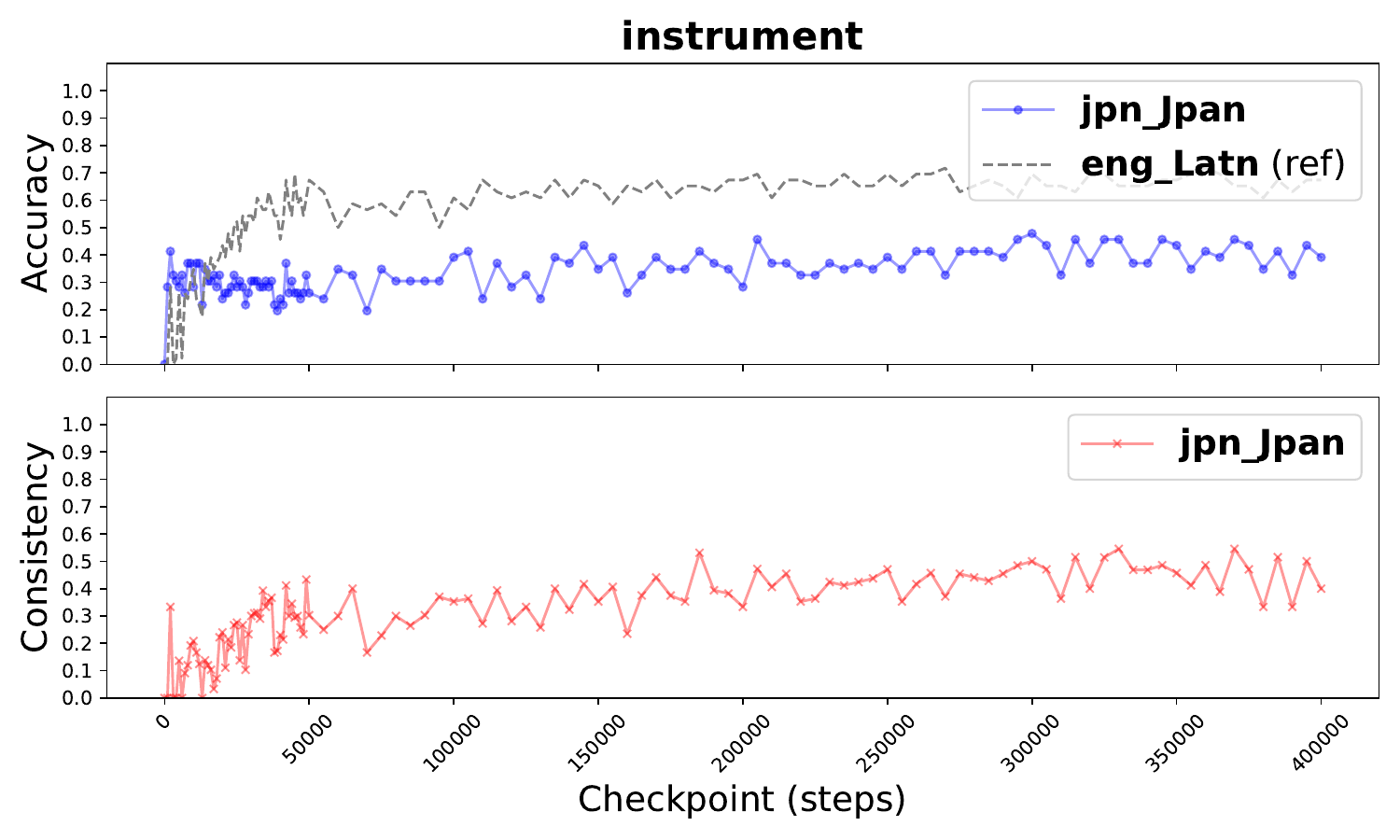}
    \includegraphics[width=0.24\textwidth]{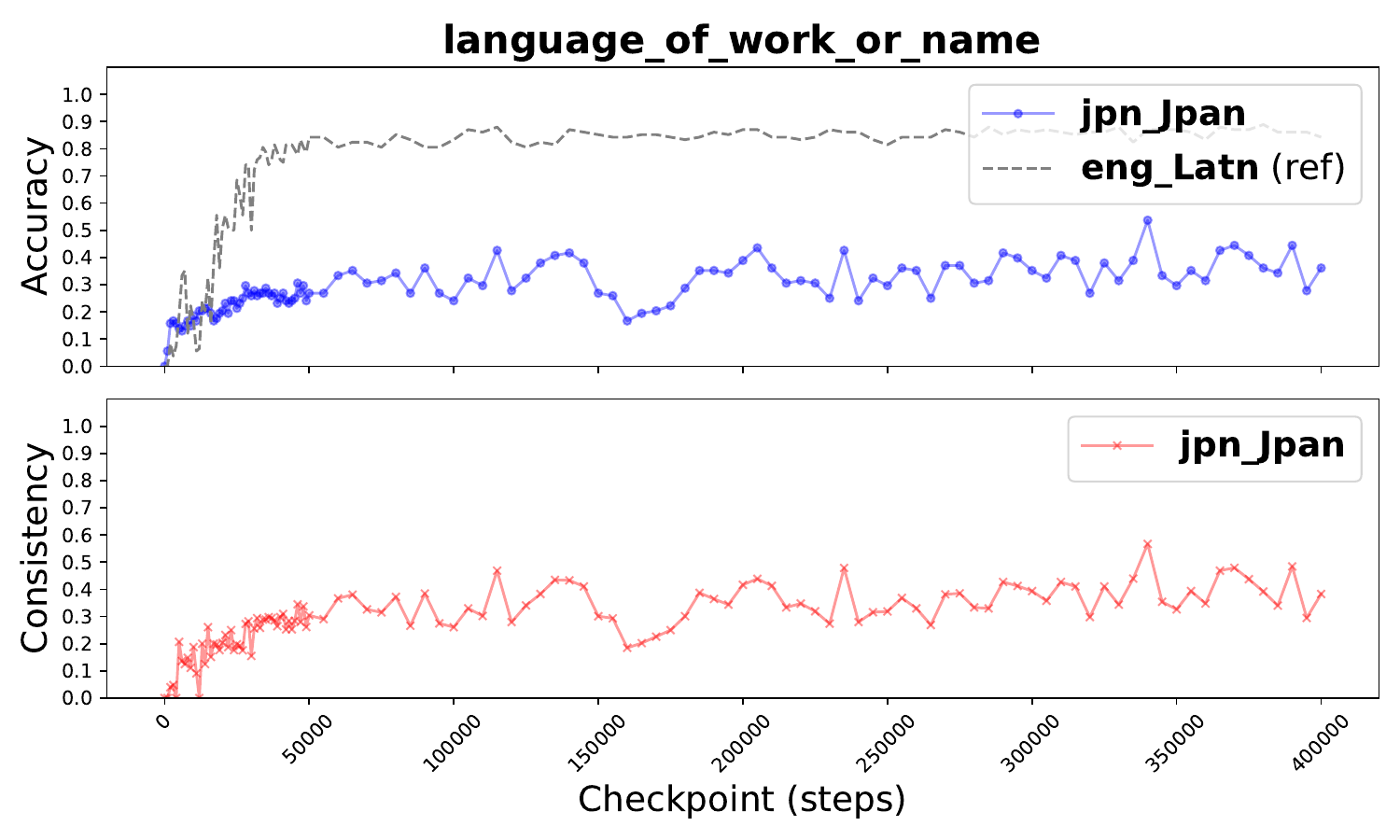}
    \includegraphics[width=0.24\textwidth]{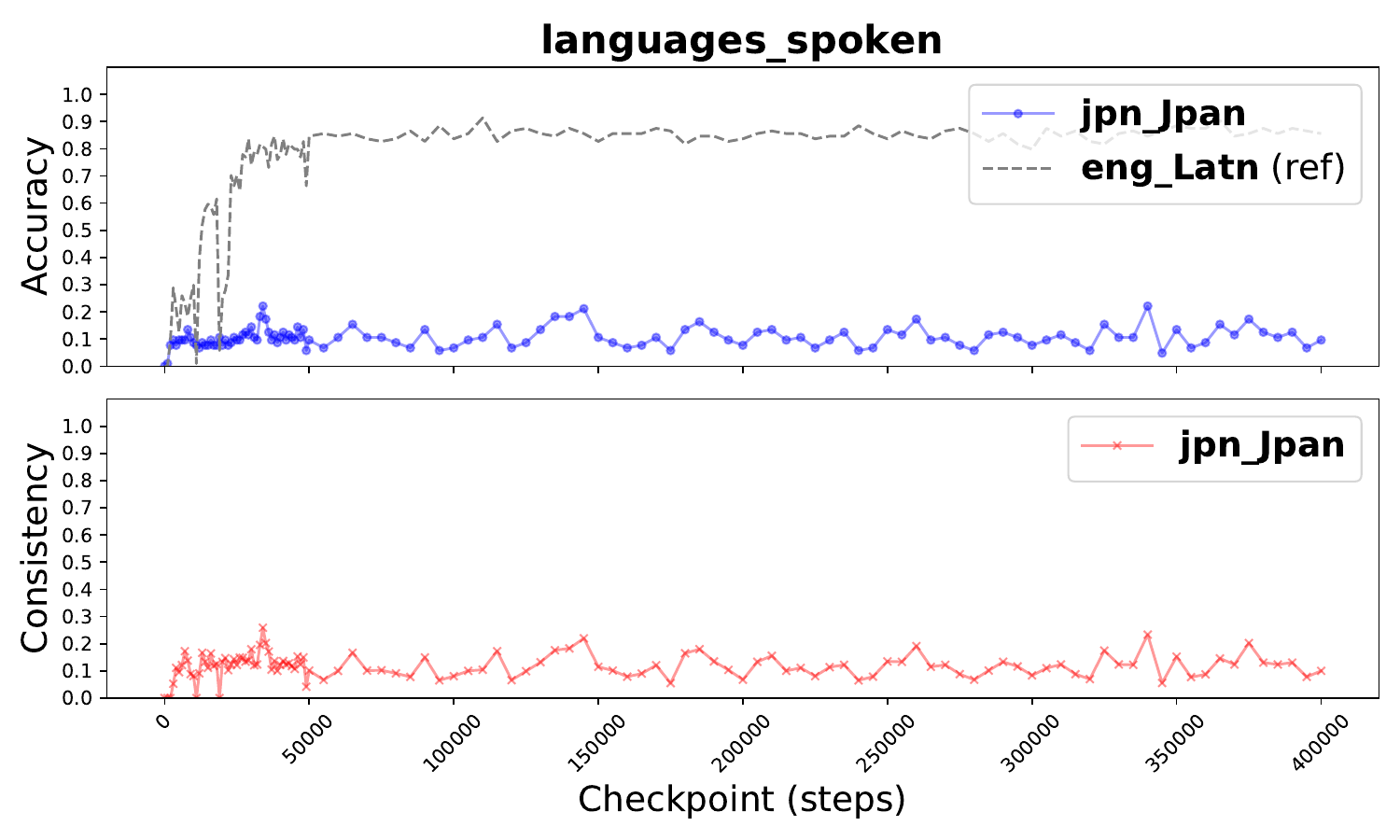}
    \includegraphics[width=0.24\textwidth]{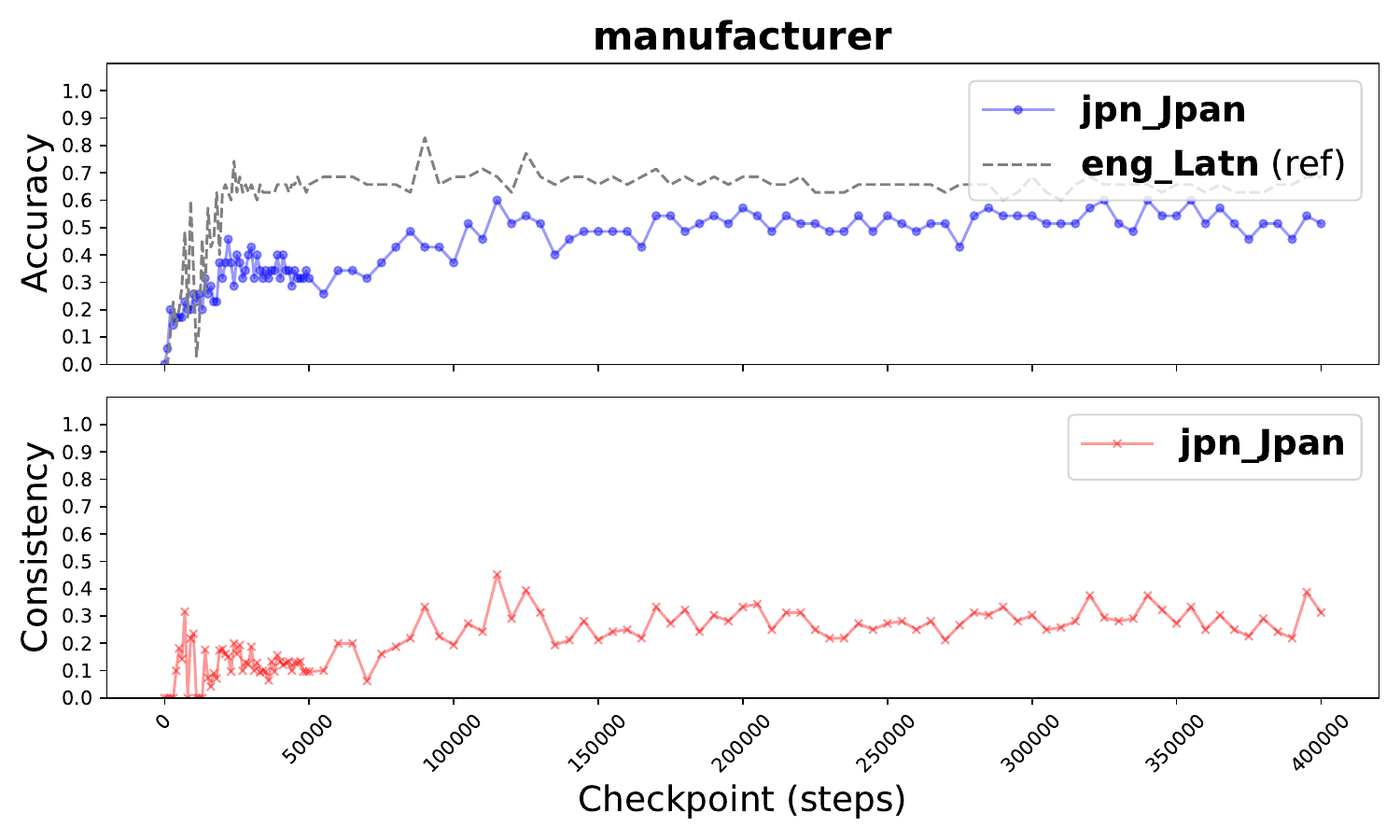}
    \includegraphics[width=0.24\textwidth]{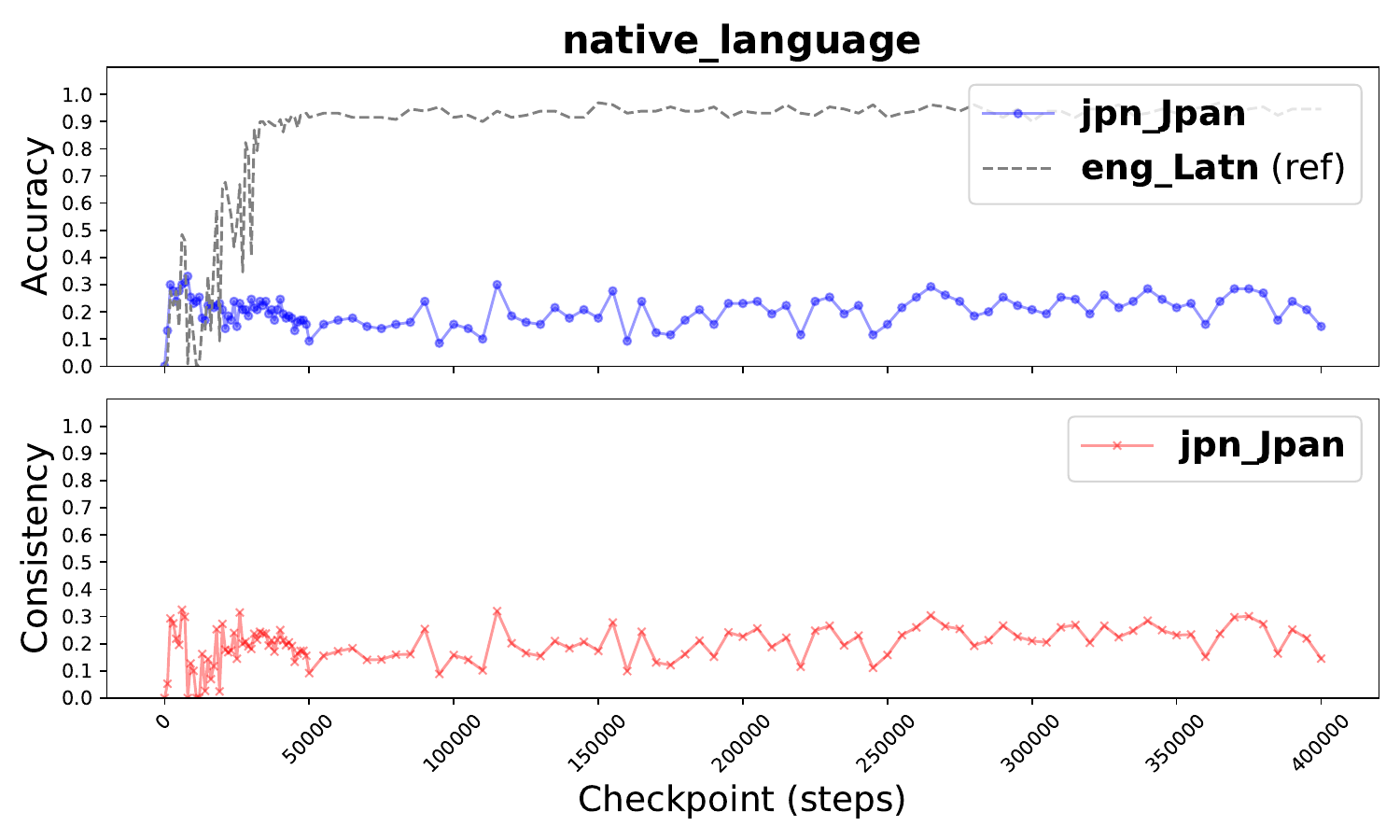}
    \includegraphics[width=0.24\textwidth]{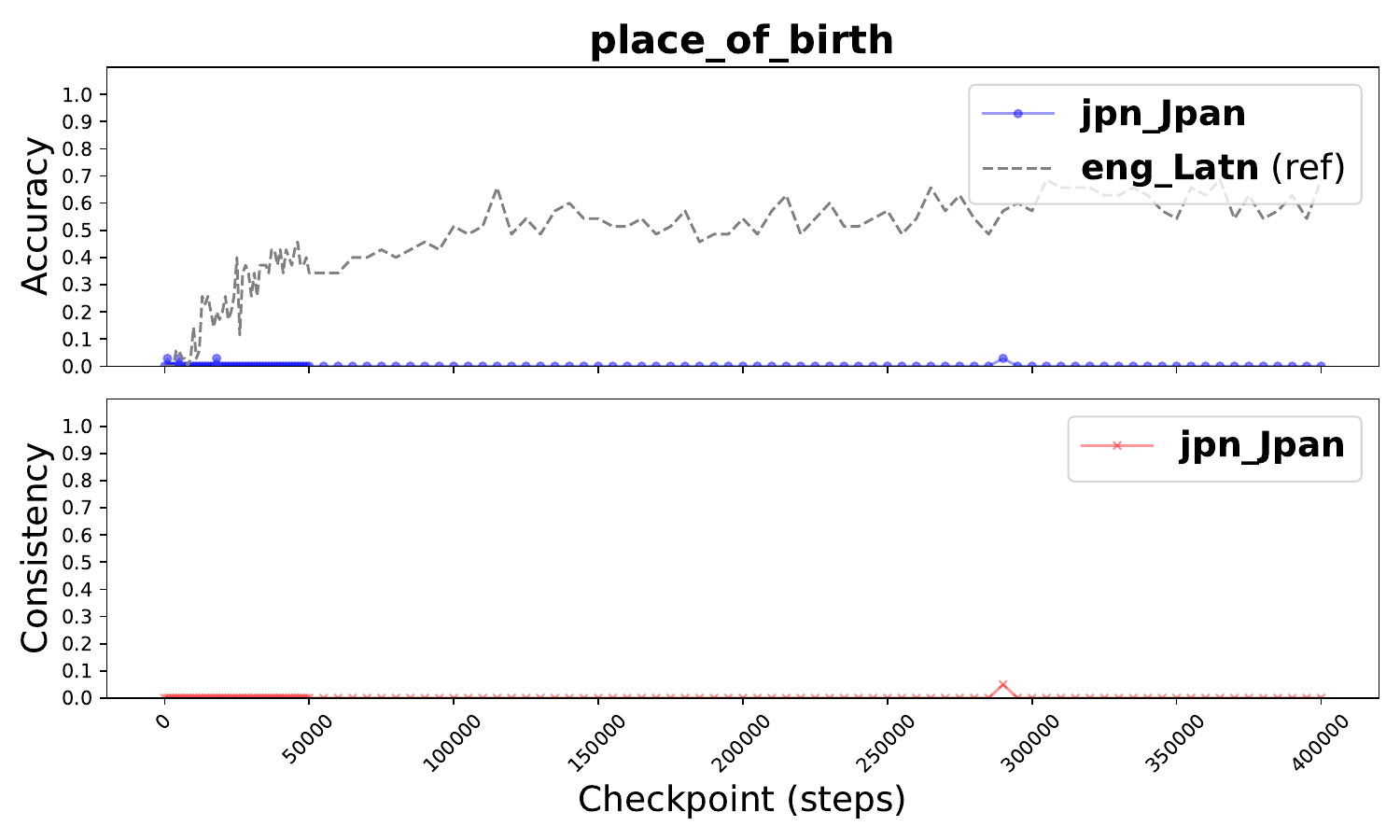}
    \includegraphics[width=0.24\textwidth]{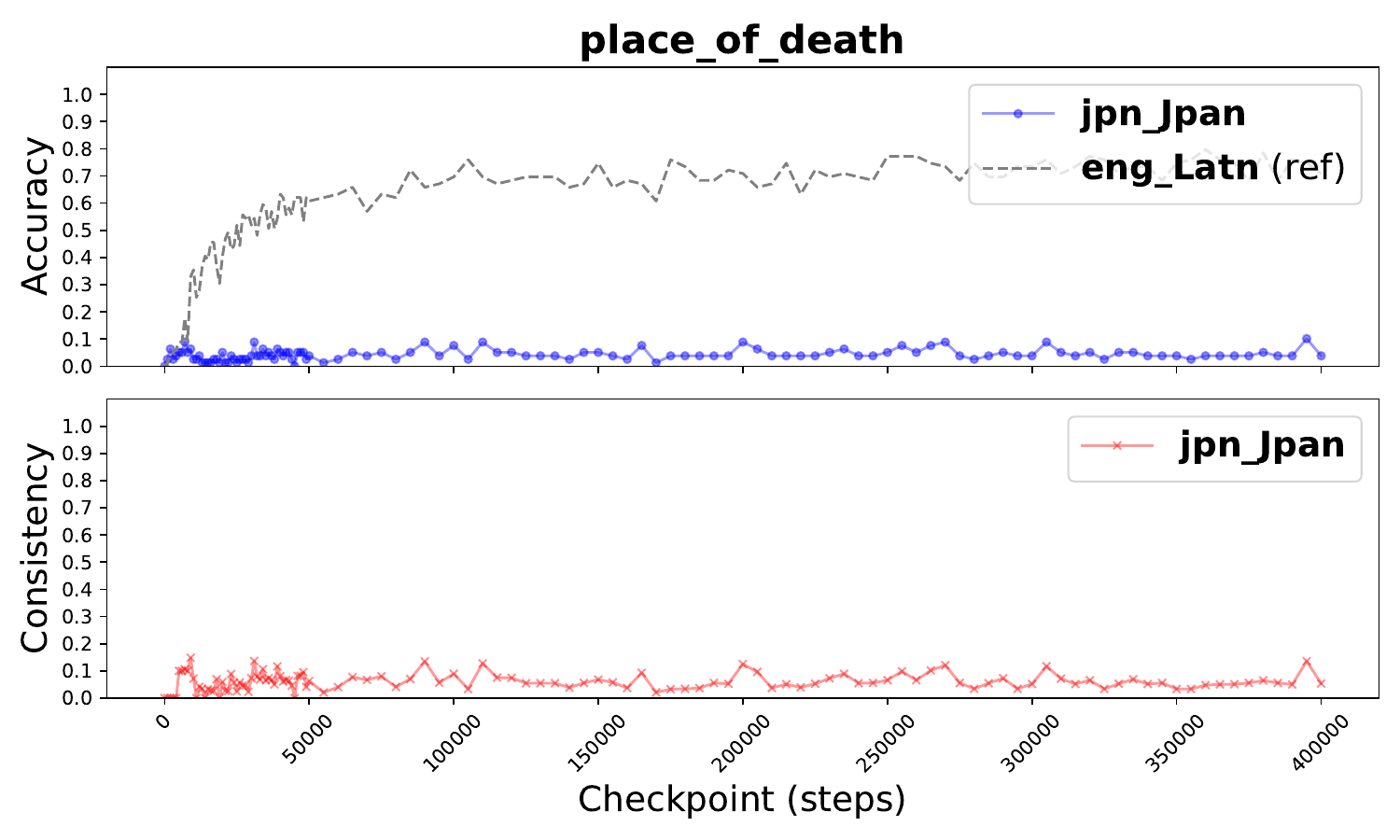}
    \includegraphics[width=0.24\textwidth]{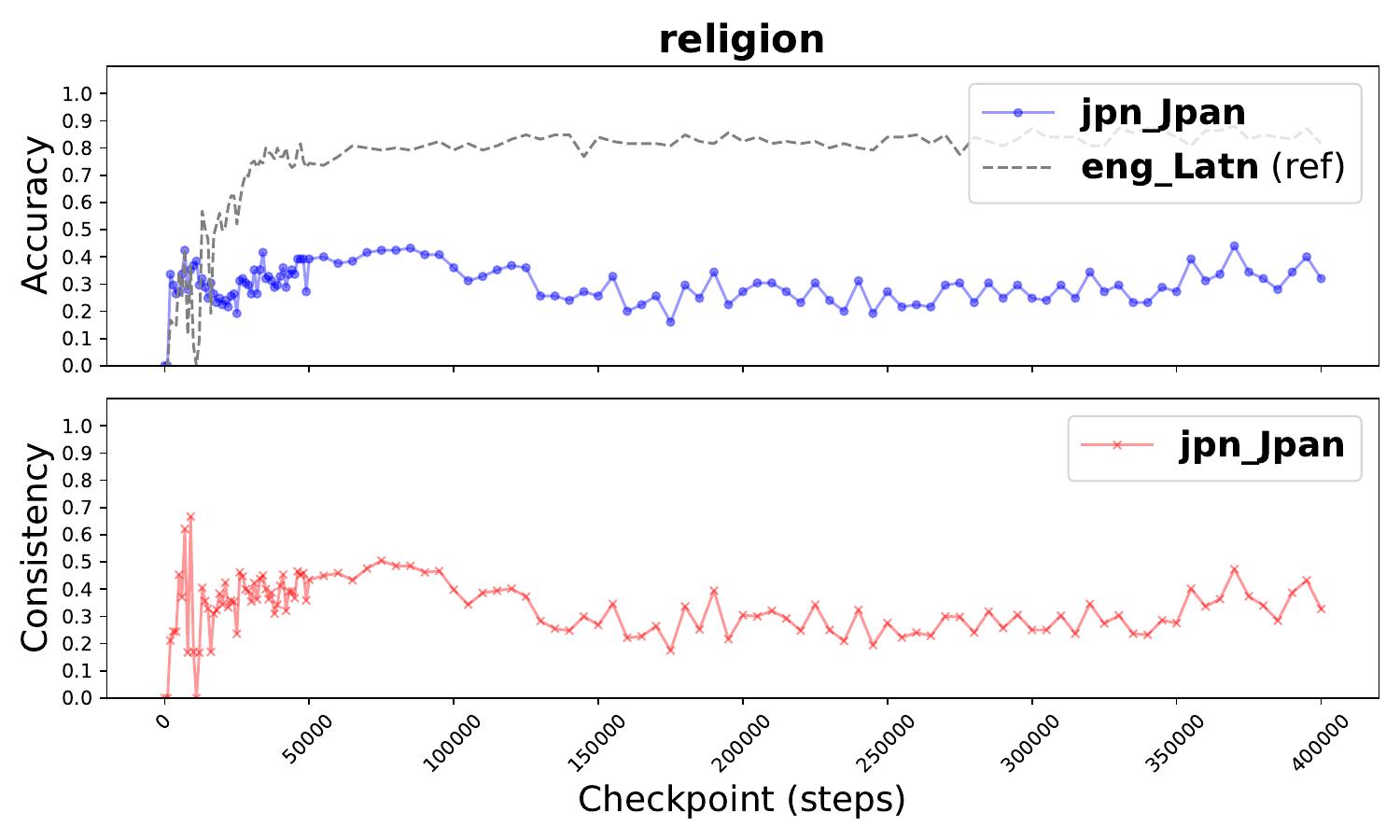}
    \caption{Factual accuracy (ACC) and crosslingual consistency (CO) for each relation type in \textbf{jpn\_Jpan}.}
    \label{fig:performance_over_checkpoints_ja}
\end{figure*}

\begin{figure*}
    \centering
    \includegraphics[width=0.24\textwidth]{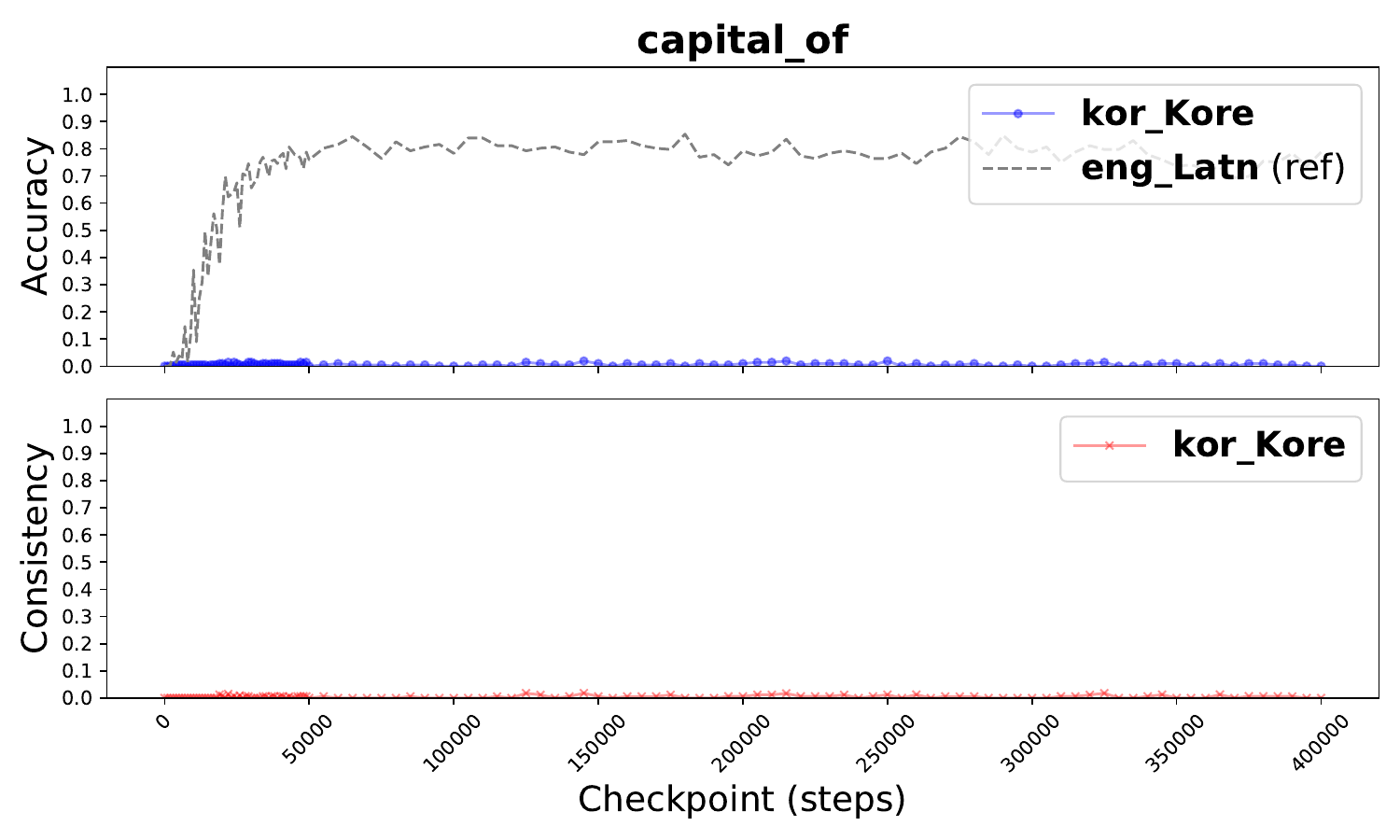}
    \includegraphics[width=0.24\textwidth]{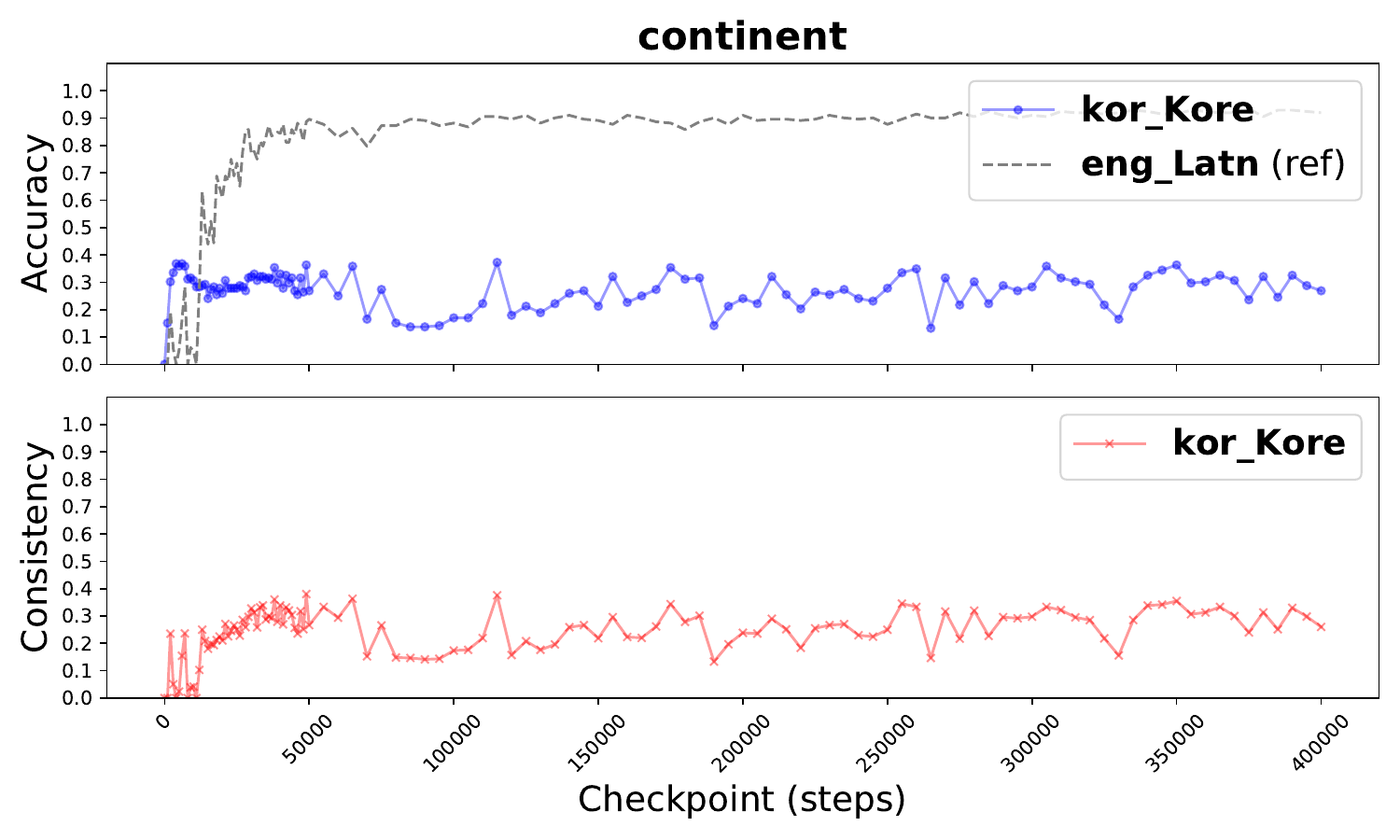}
    \includegraphics[width=0.24\textwidth]{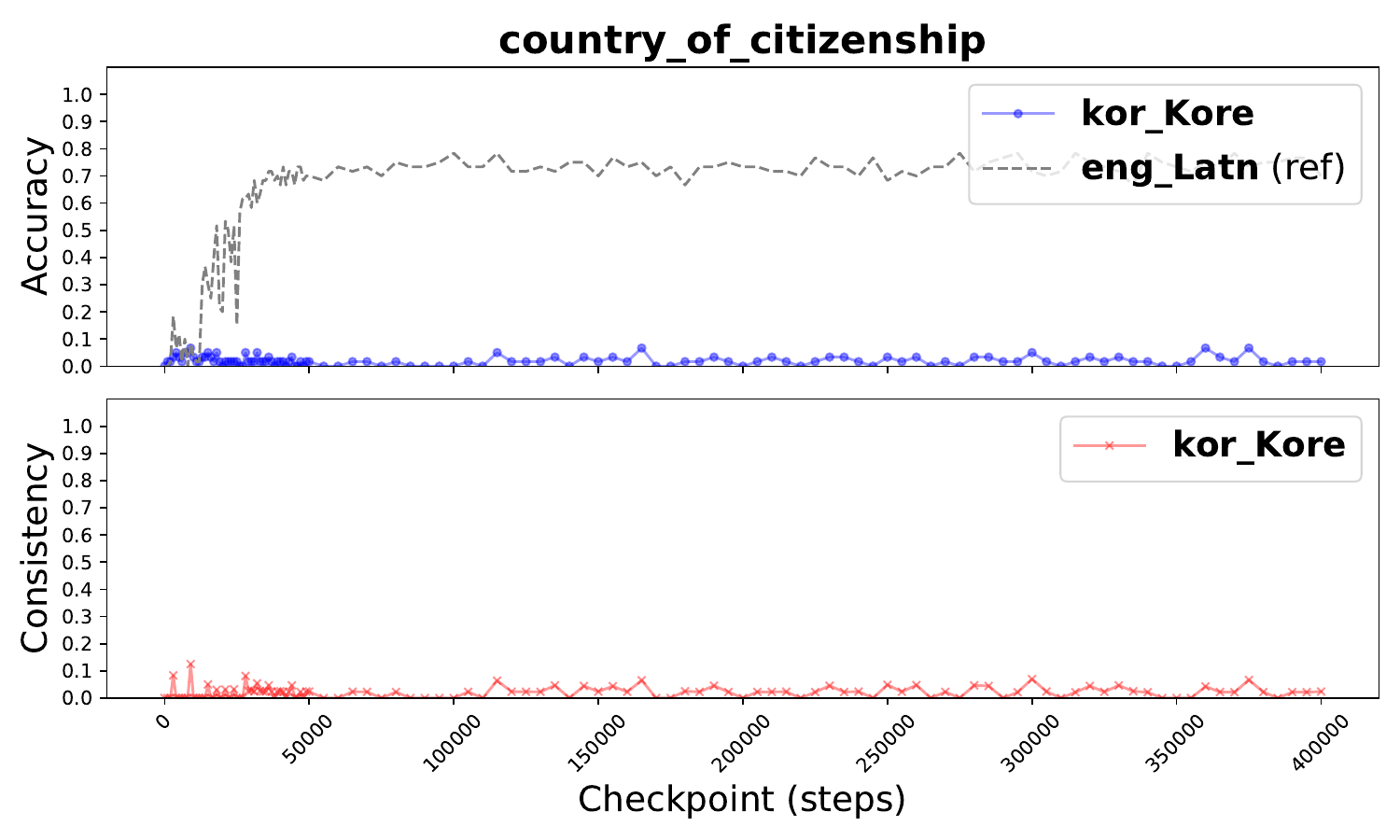}
    \includegraphics[width=0.24\textwidth]{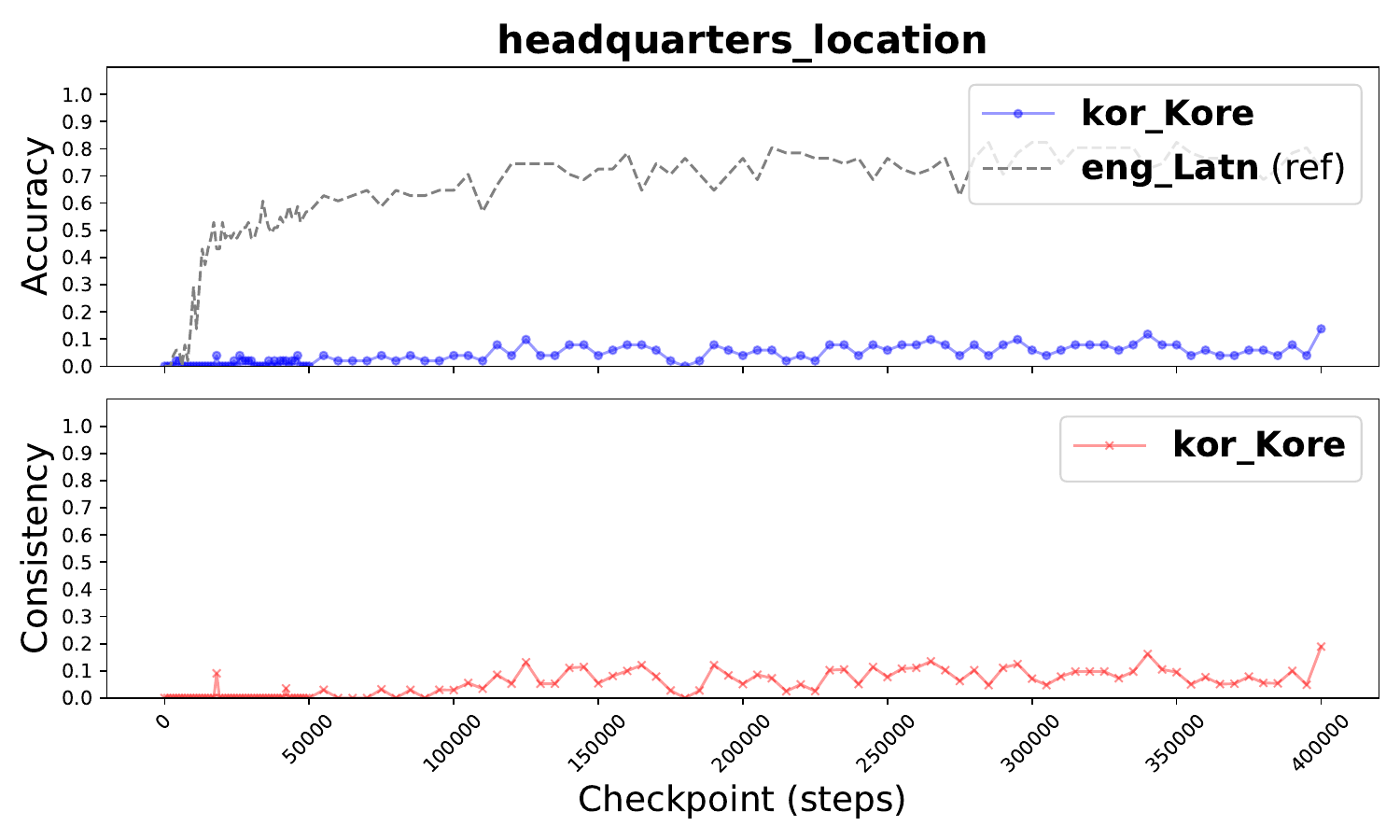}
    \includegraphics[width=0.24\textwidth]{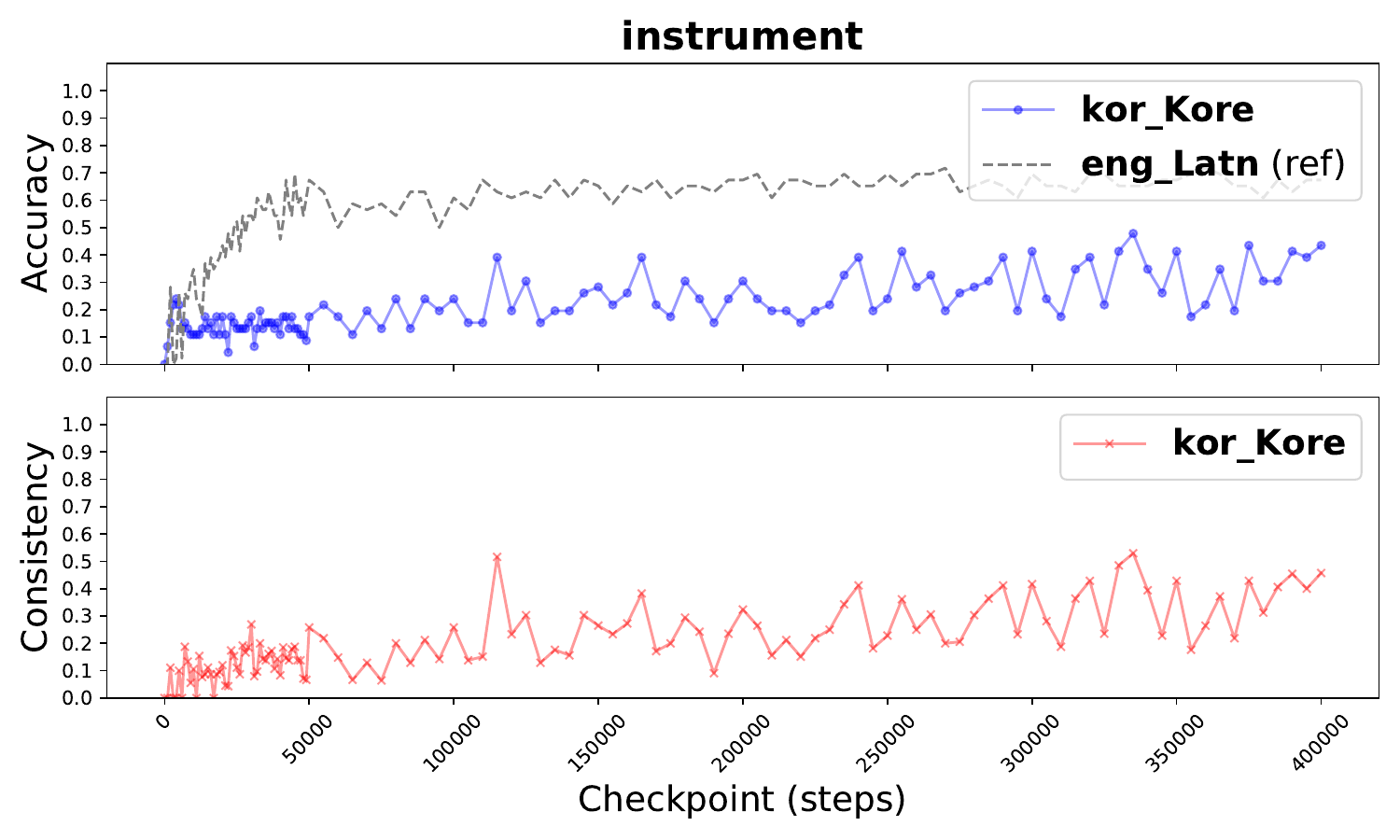}
    \includegraphics[width=0.24\textwidth]{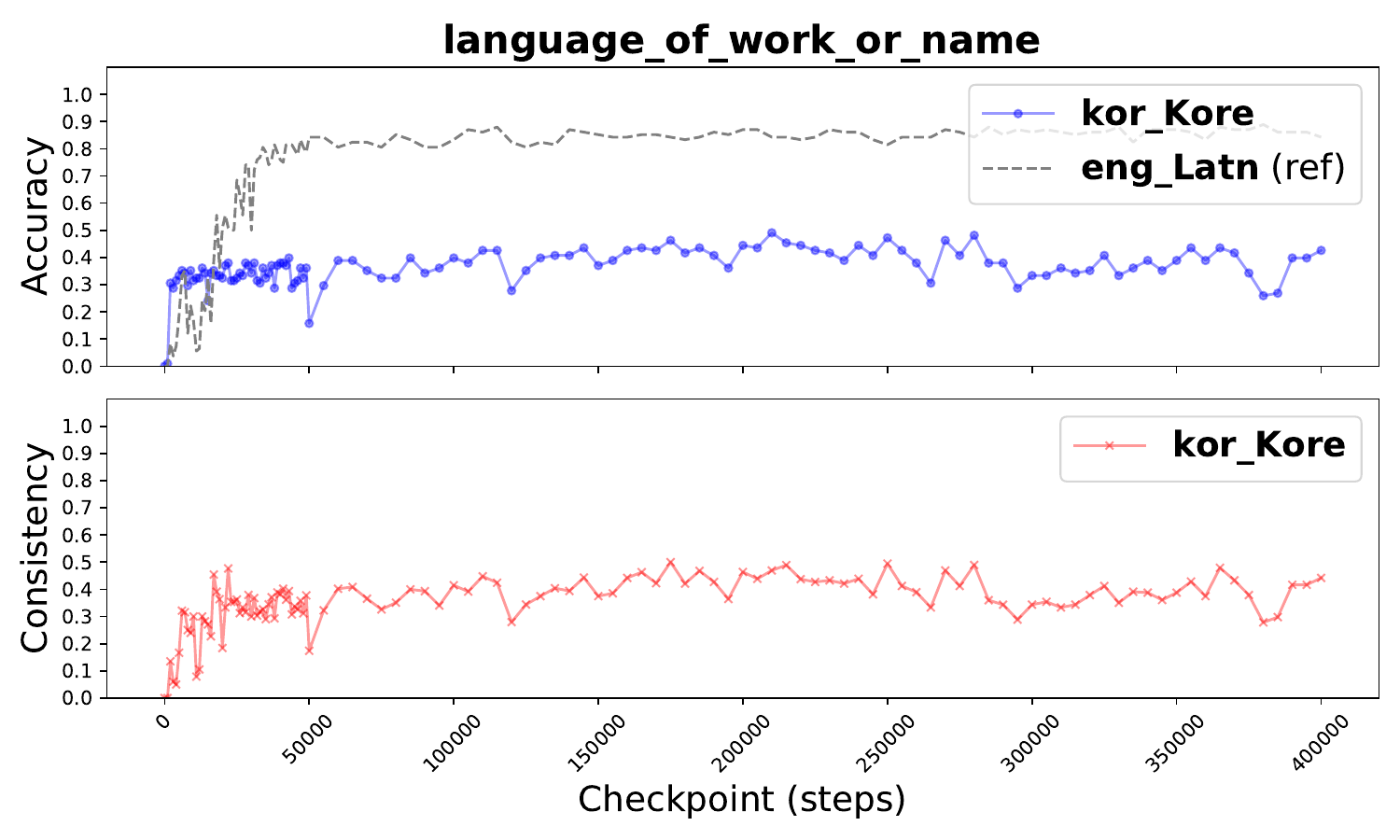}
    \includegraphics[width=0.24\textwidth]{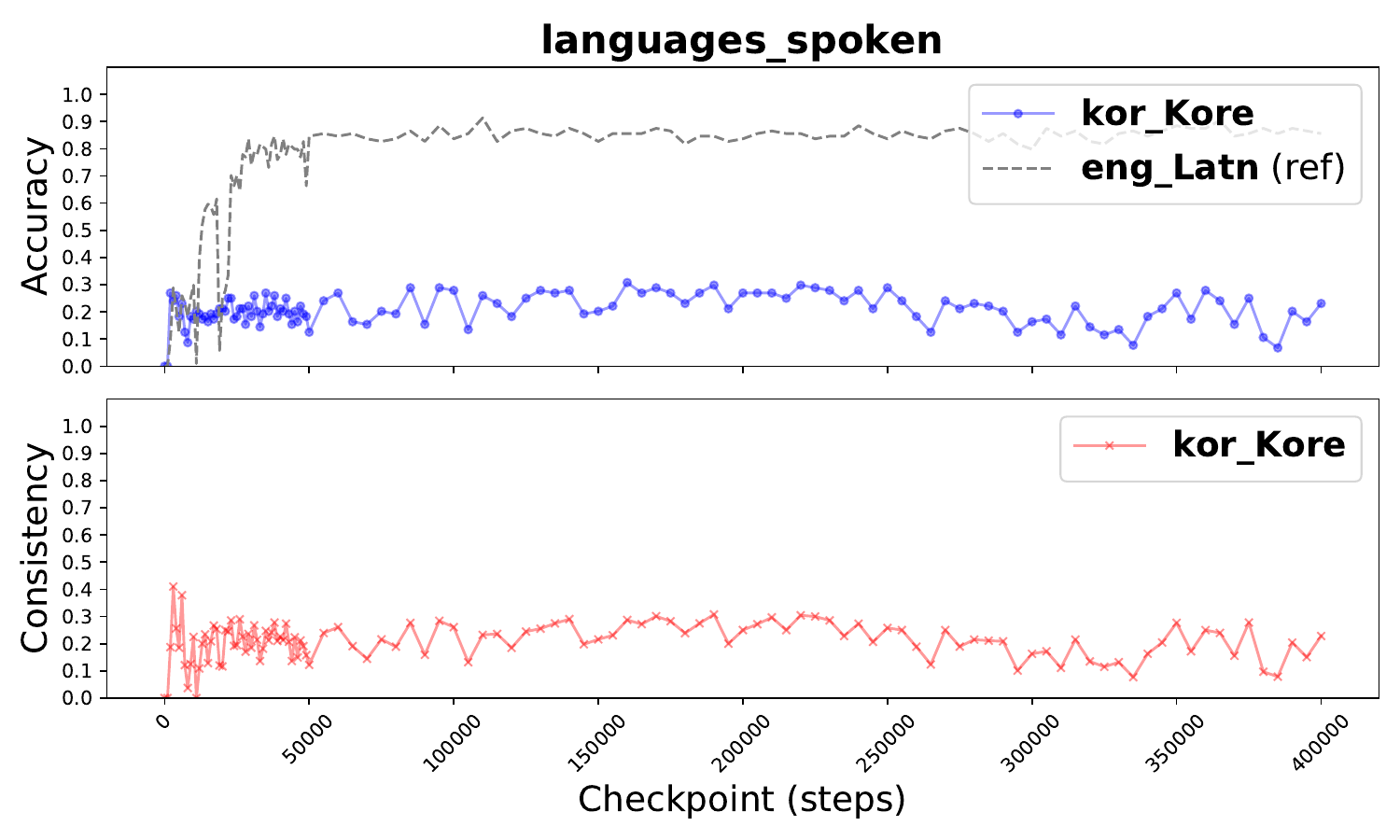}
    \includegraphics[width=0.24\textwidth]{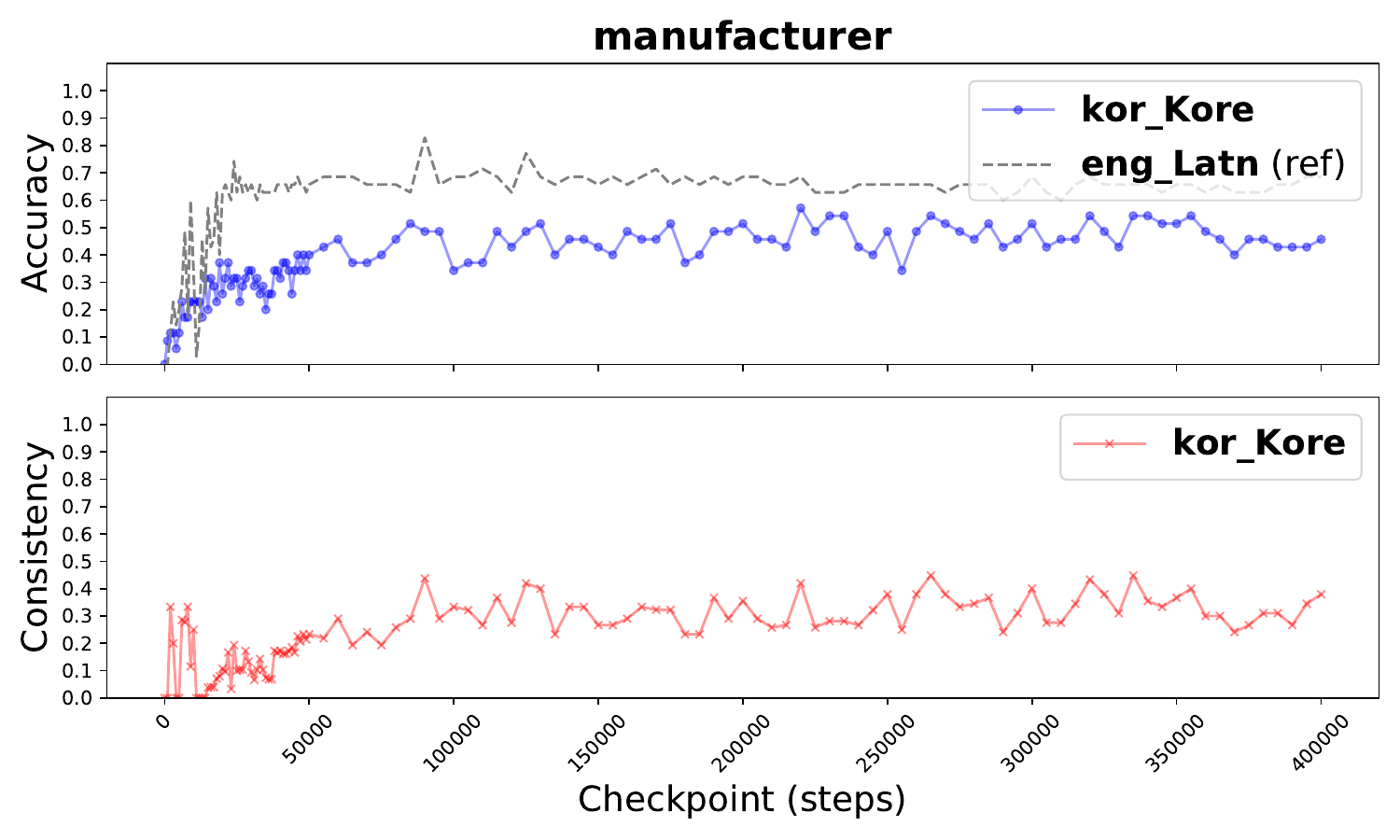}
    \includegraphics[width=0.24\textwidth]{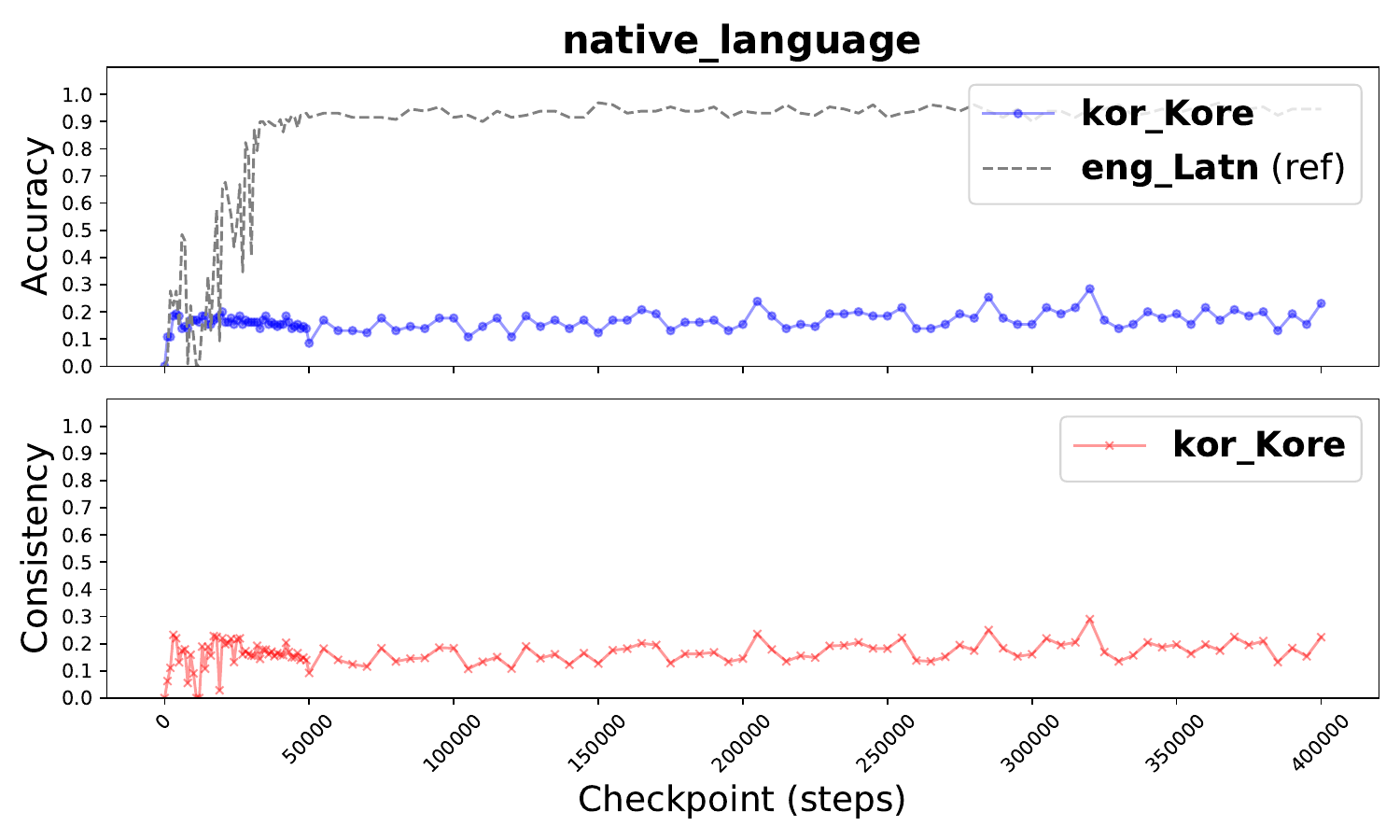}
    \includegraphics[width=0.24\textwidth]{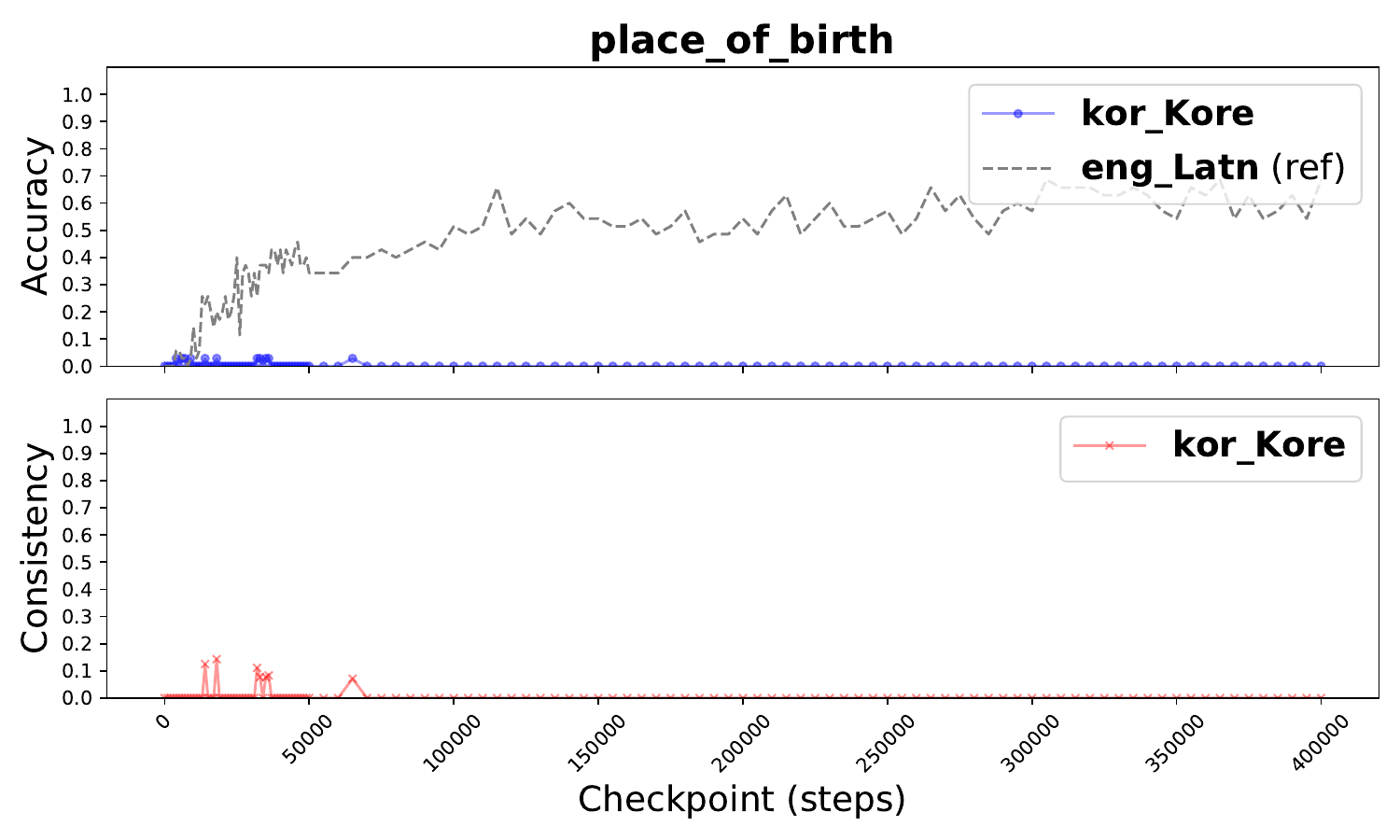}
    \includegraphics[width=0.24\textwidth]{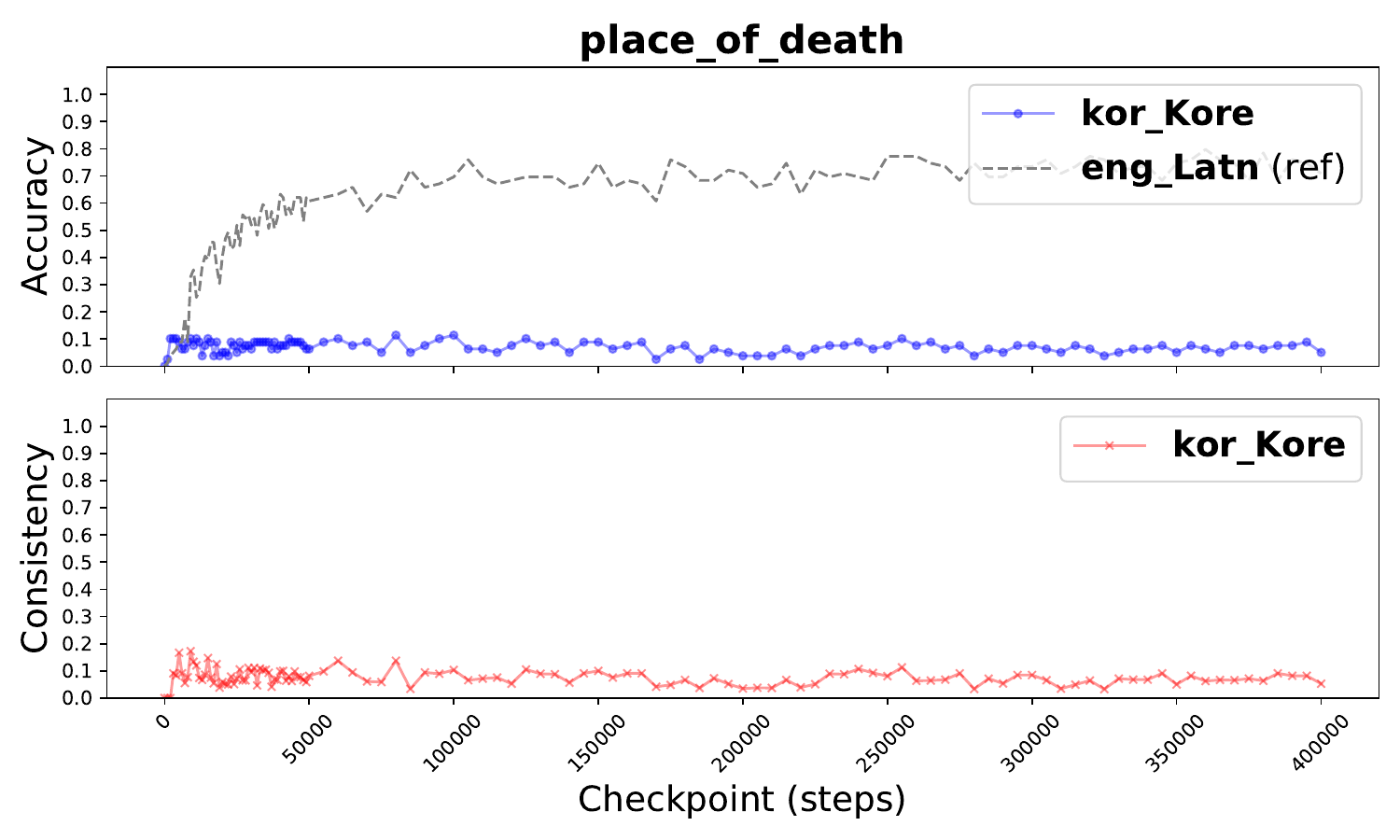}
    \includegraphics[width=0.24\textwidth]{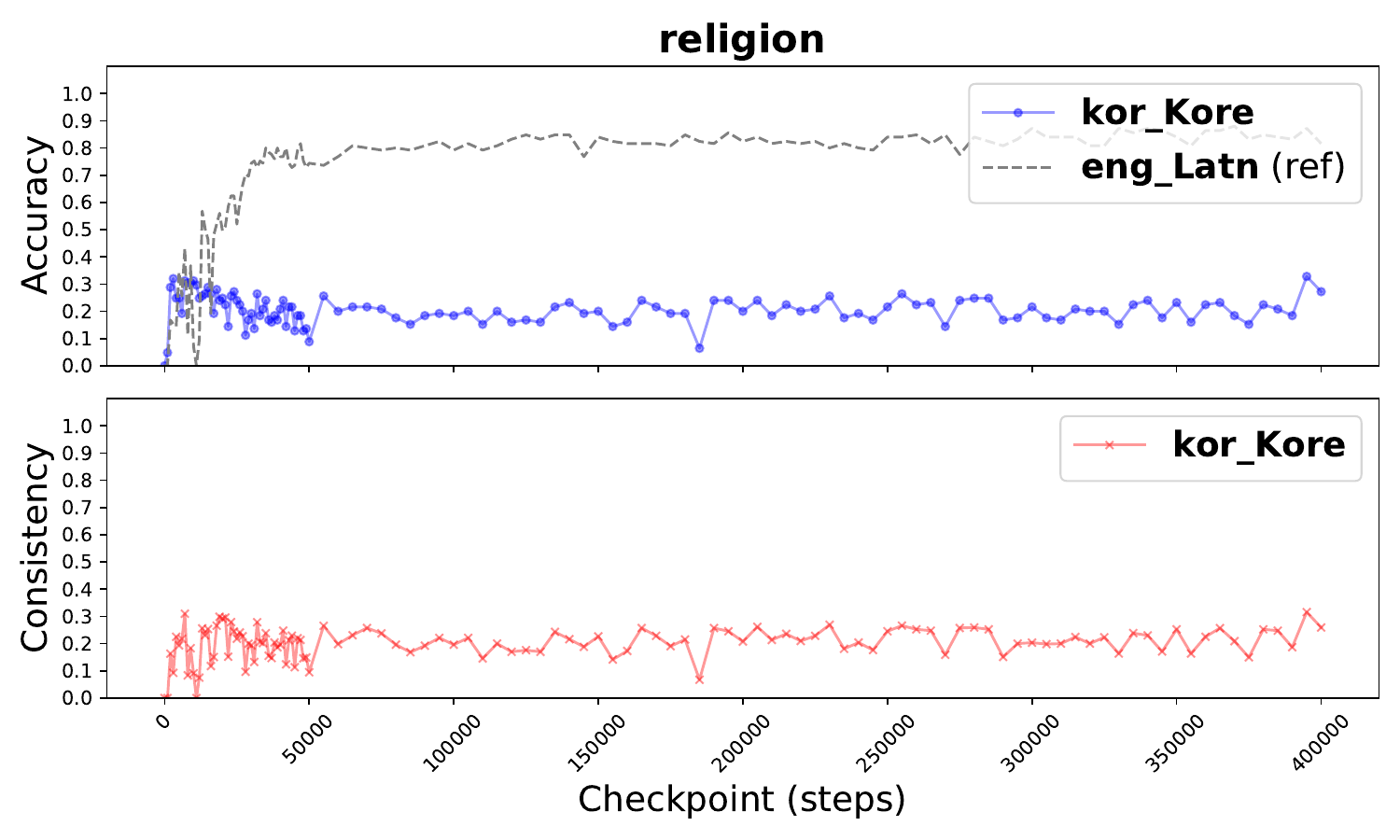}
    \caption{Factual accuracy (ACC) and crosslingual consistency (CO) for each relation type in \textbf{kor\_Kore}.}
    \label{fig:performance_over_checkpoints_ko}
\end{figure*}

\begin{figure*}
    \centering
    \includegraphics[width=0.24\textwidth]{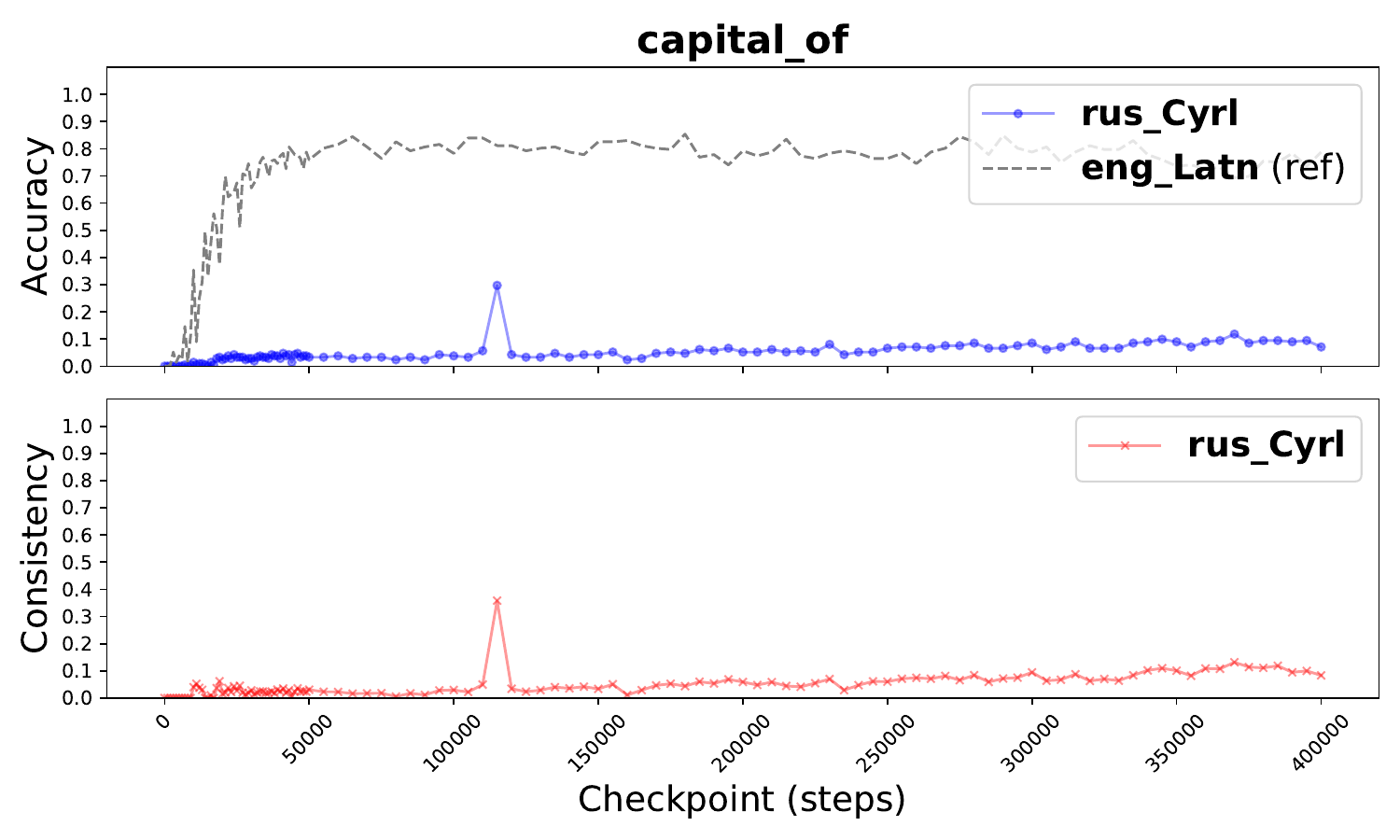}
    \includegraphics[width=0.24\textwidth]{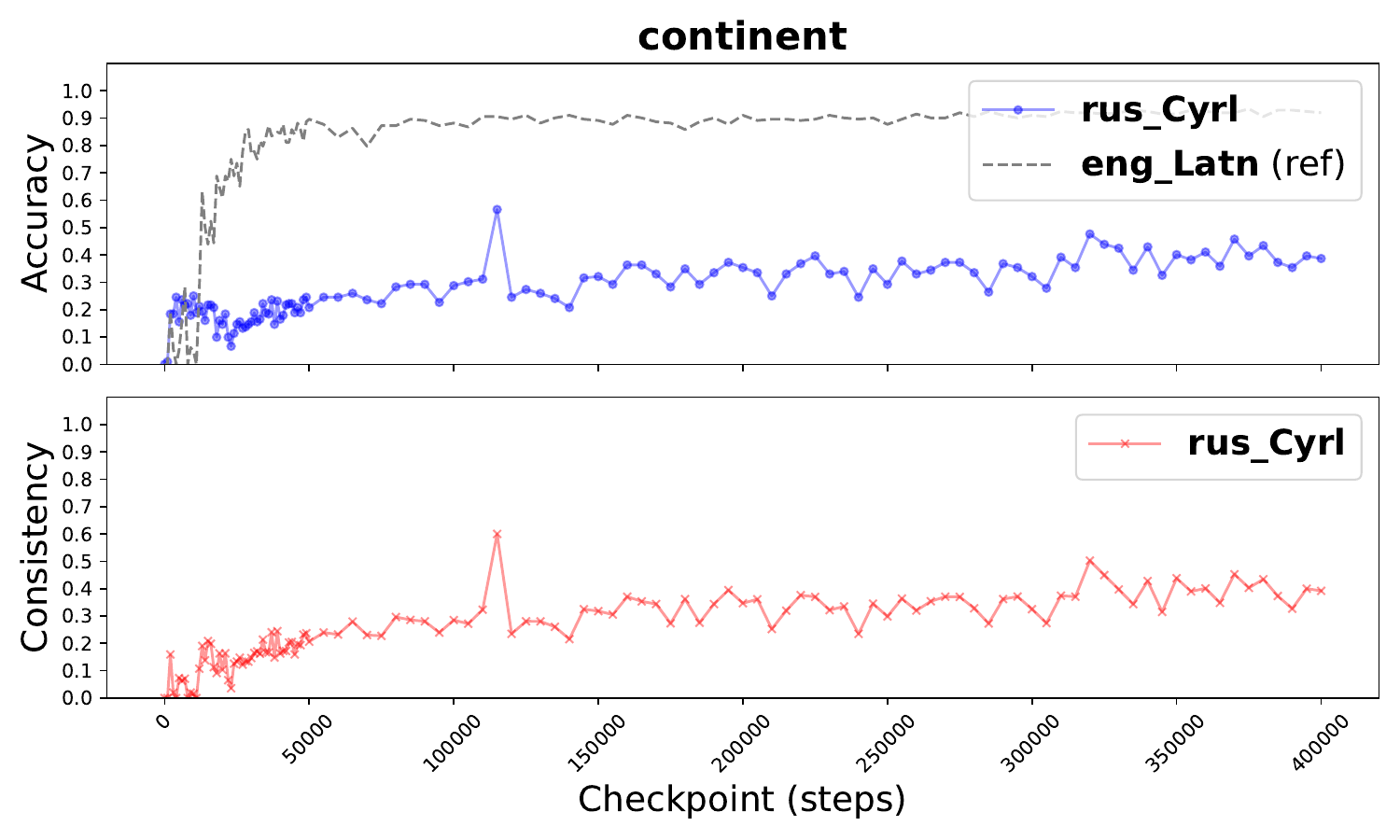}
    \includegraphics[width=0.24\textwidth]{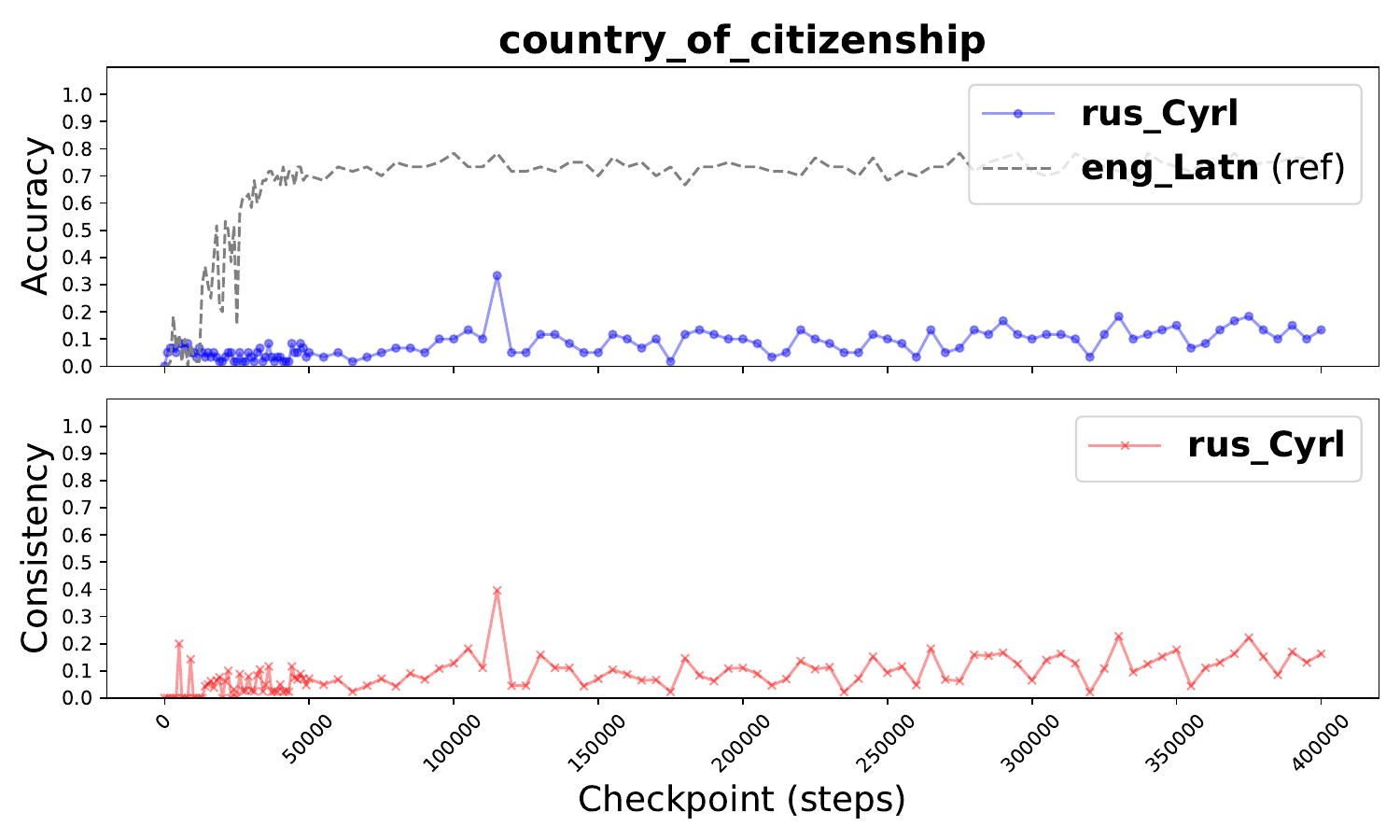}
    \includegraphics[width=0.24\textwidth]{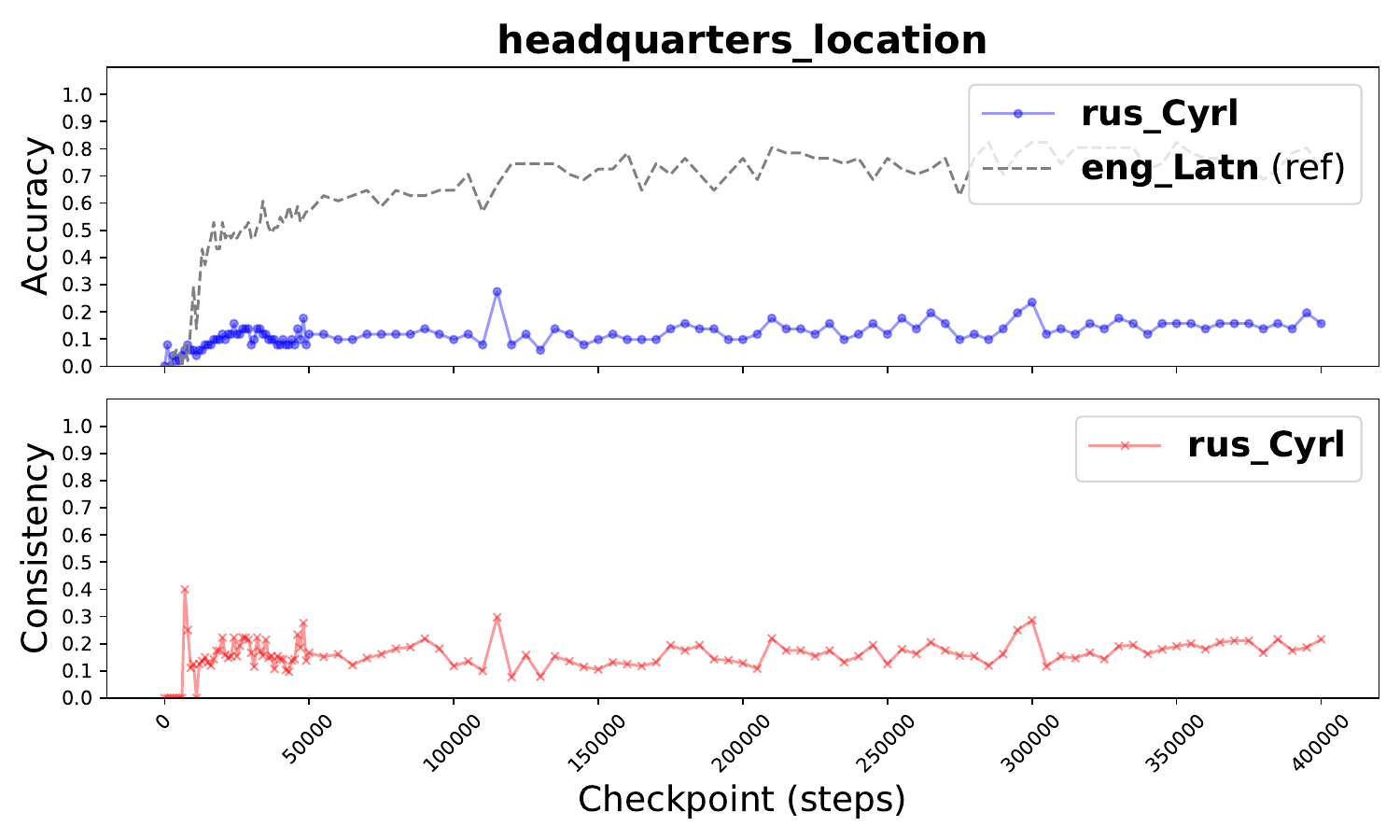}
    \includegraphics[width=0.24\textwidth]{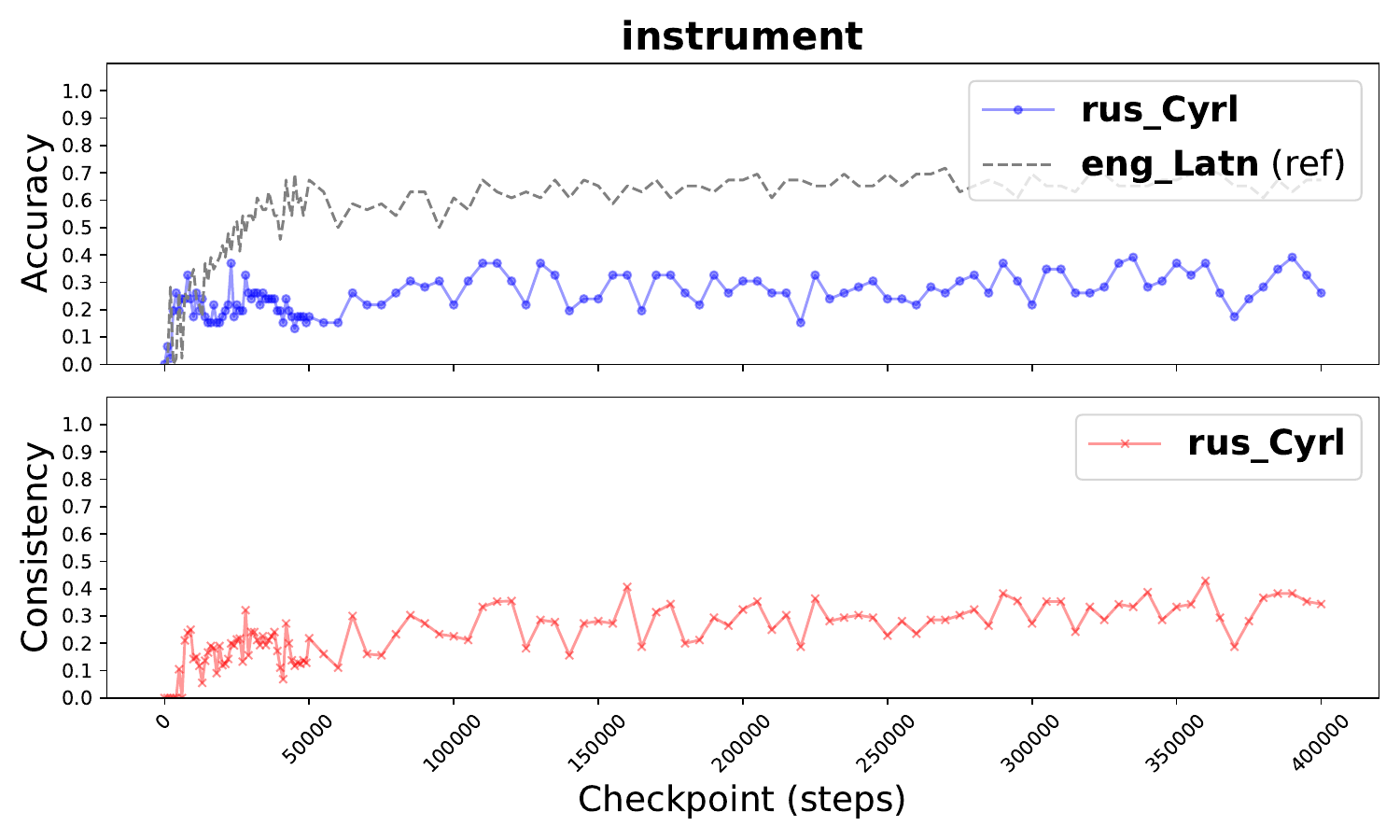}
    \includegraphics[width=0.24\textwidth]{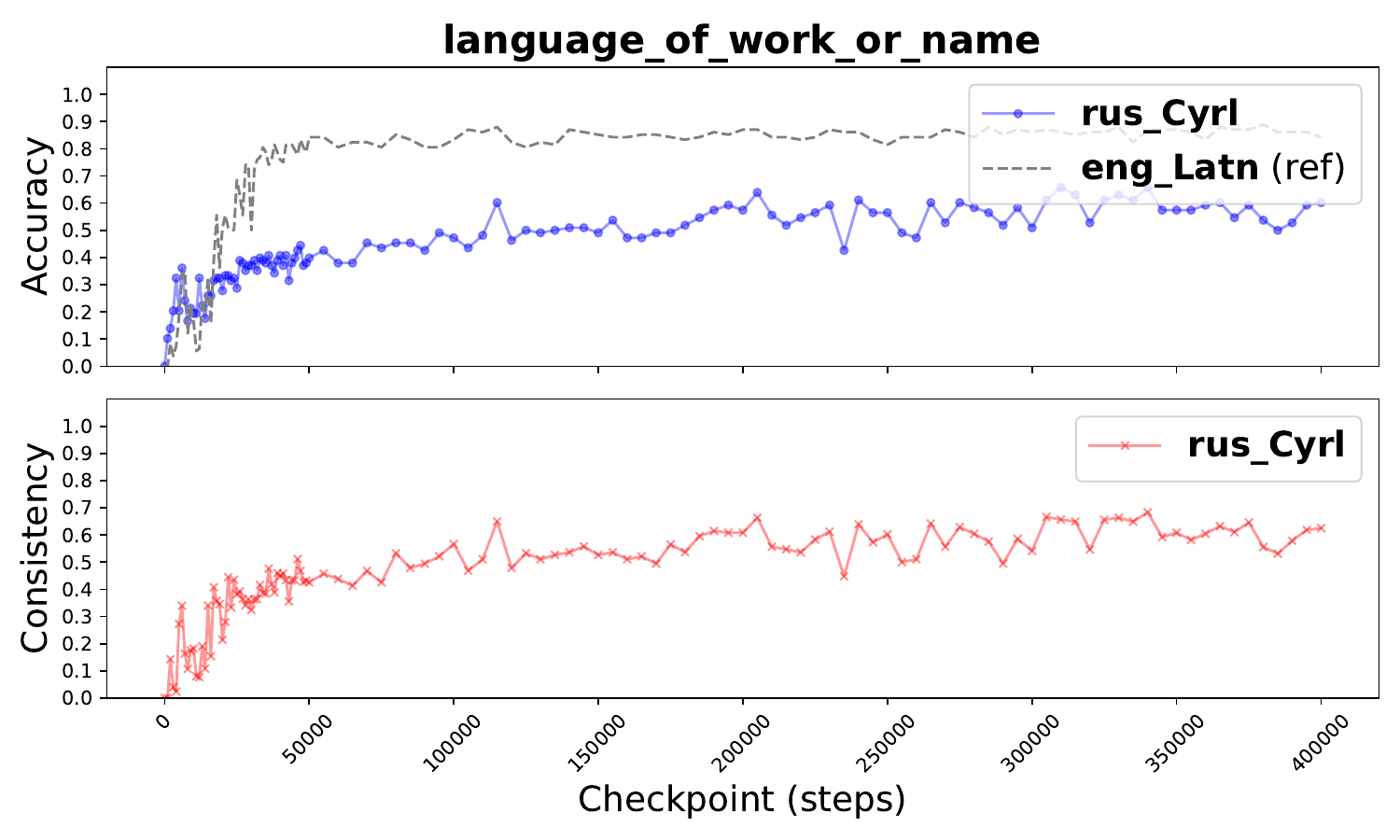}
    \includegraphics[width=0.24\textwidth]{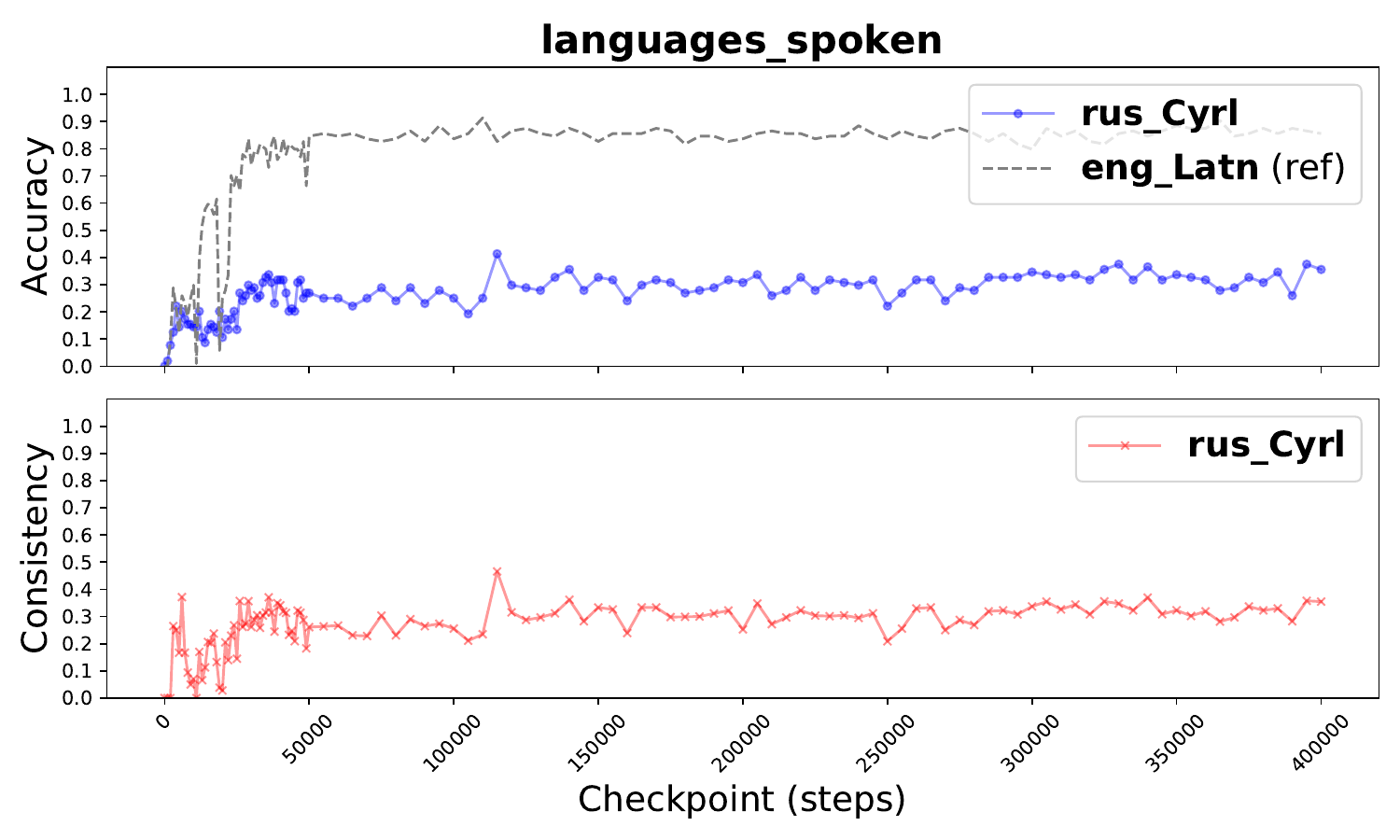}
    \includegraphics[width=0.24\textwidth]{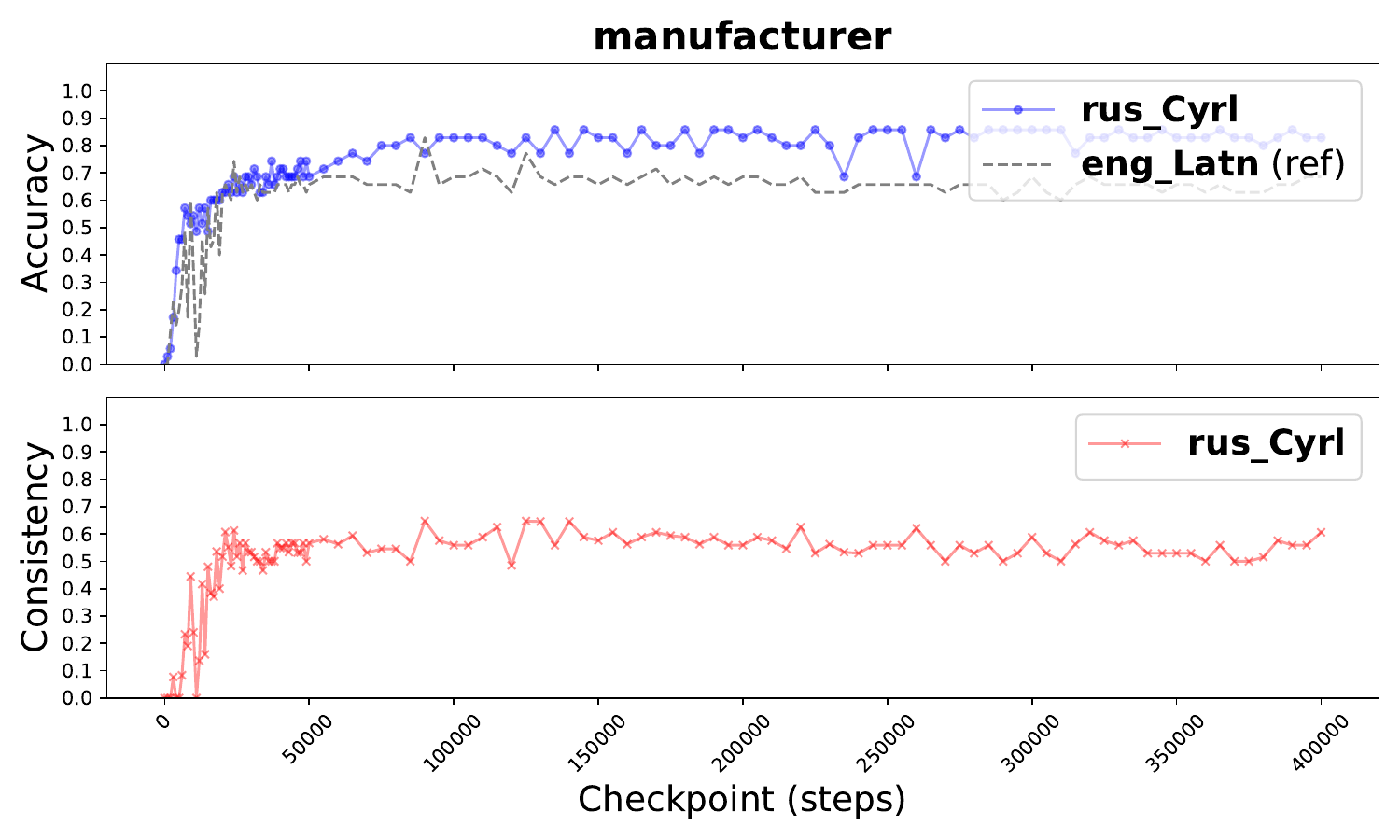}
    \includegraphics[width=0.24\textwidth]{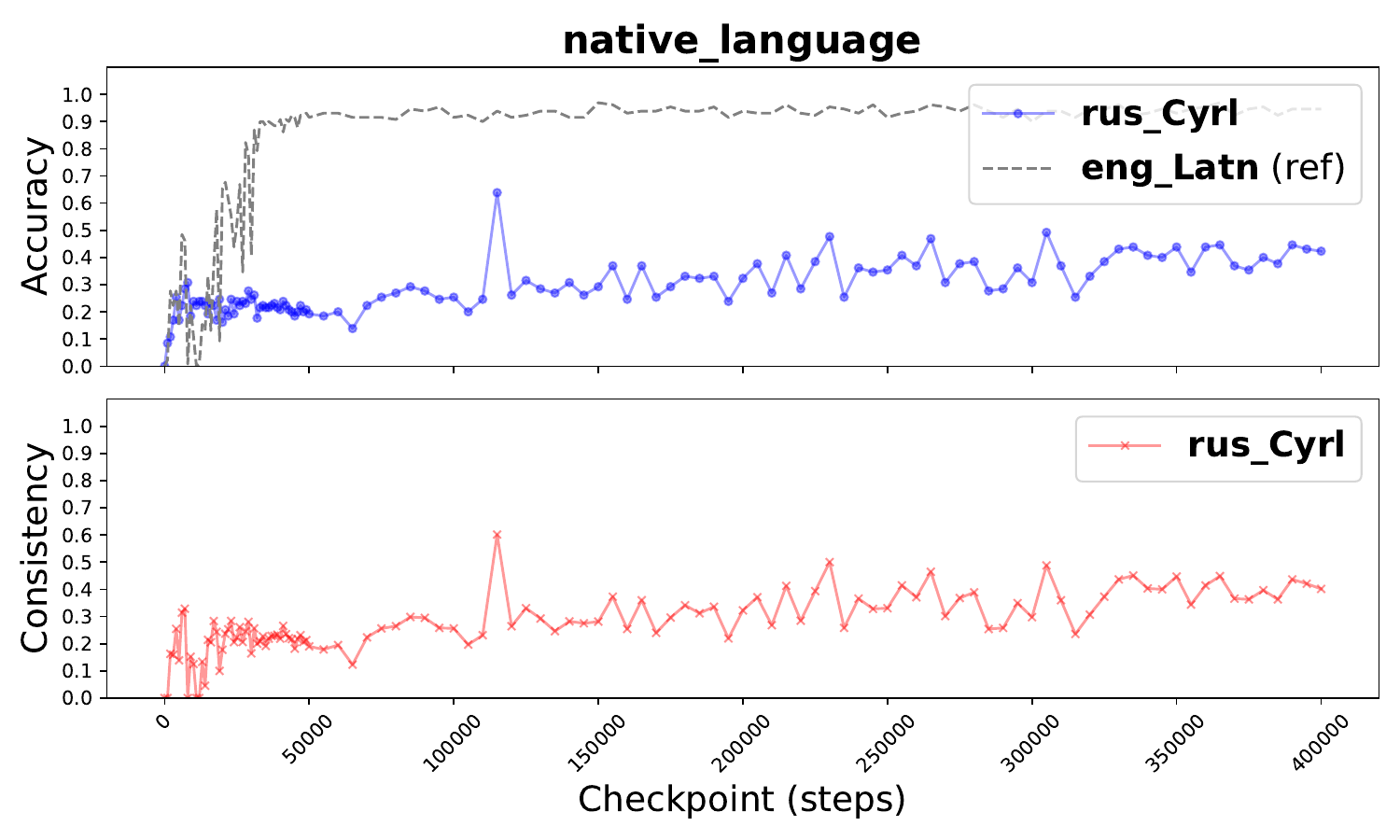}
    \includegraphics[width=0.24\textwidth]{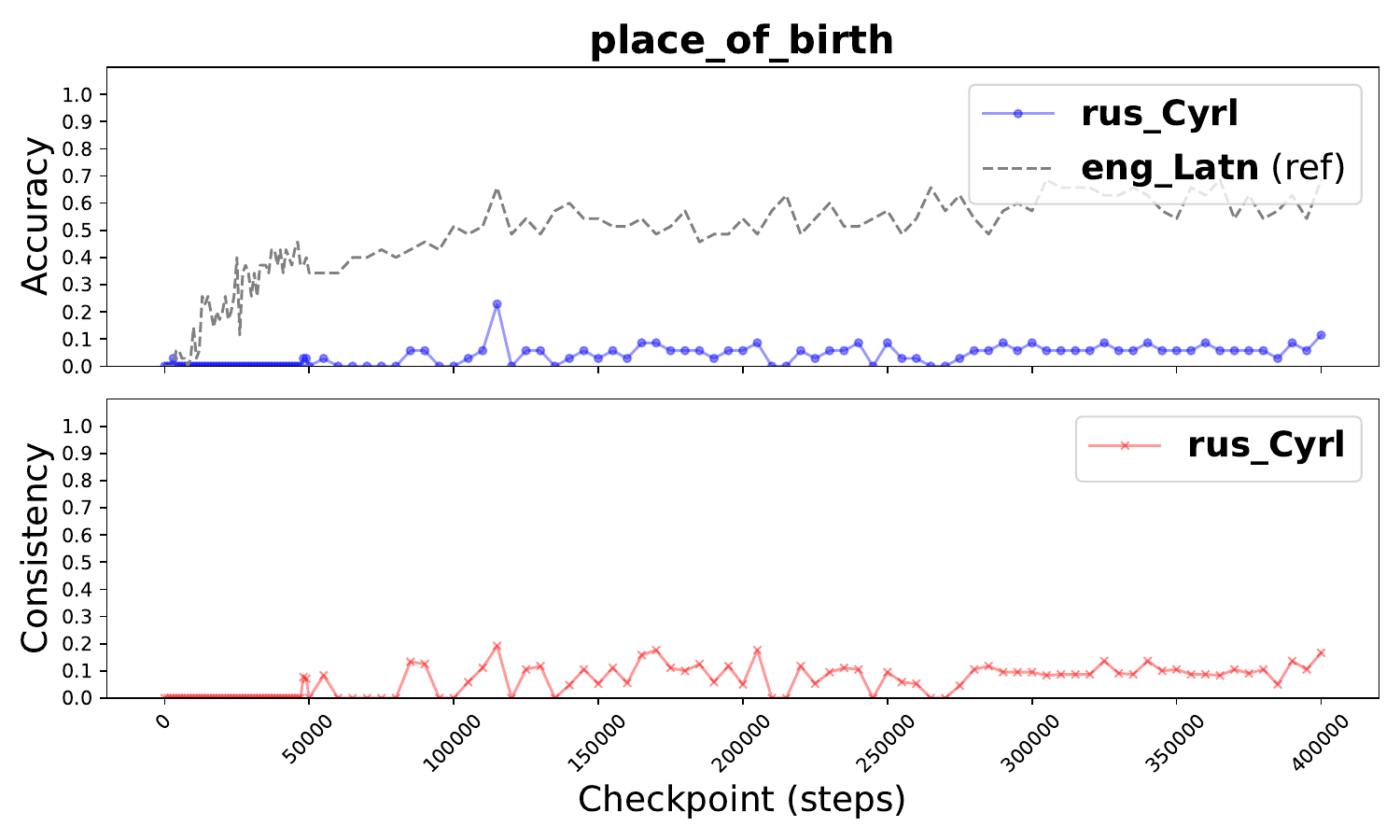}
    \includegraphics[width=0.24\textwidth]{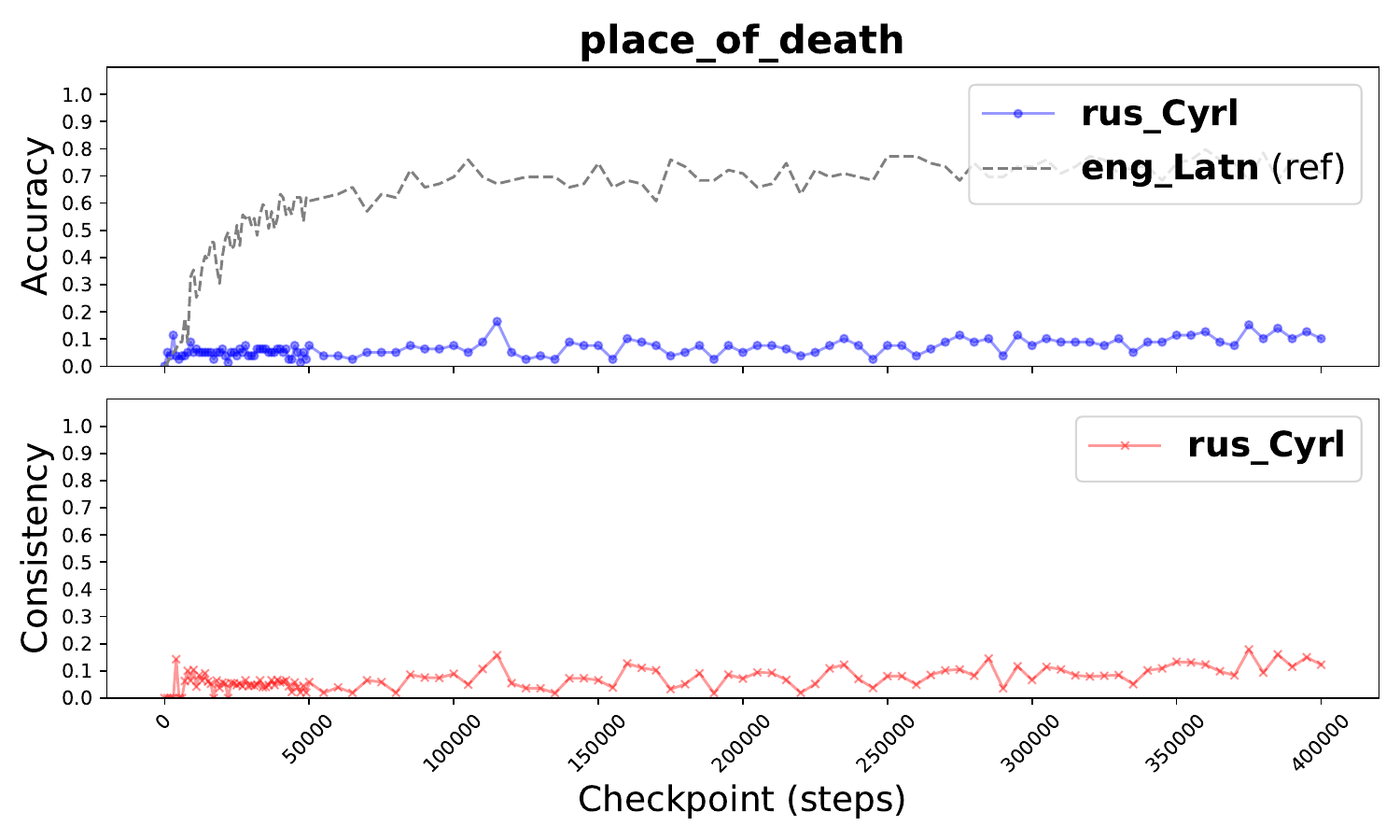}
    \includegraphics[width=0.24\textwidth]{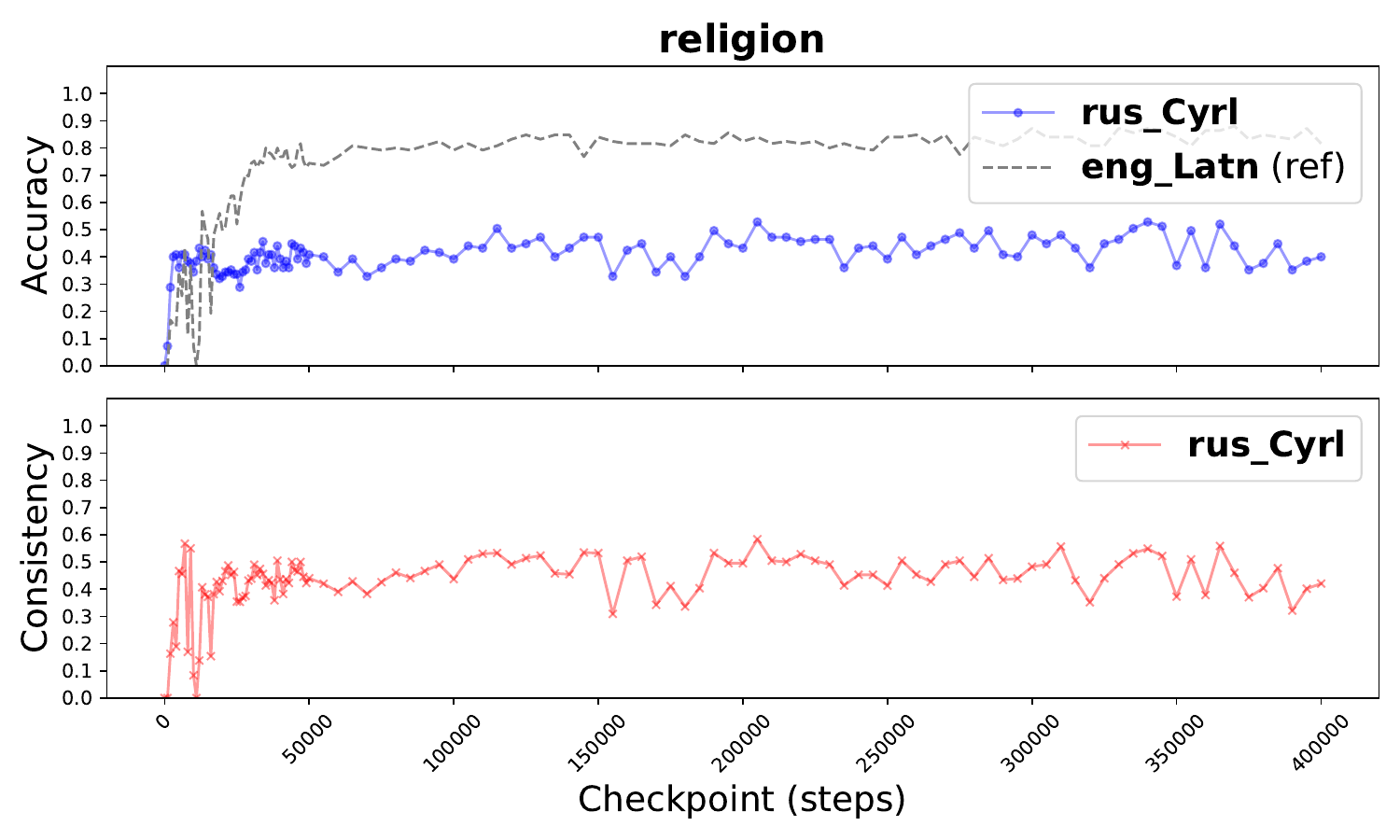}
    \caption{Factual accuracy (ACC) and crosslingual consistency (CO) for each relation type in \textbf{rus\_Cyrl}.}
    \label{fig:performance_over_checkpoints_ru}
\end{figure*}

\begin{figure*}
    \centering
    \includegraphics[width=0.24\textwidth]{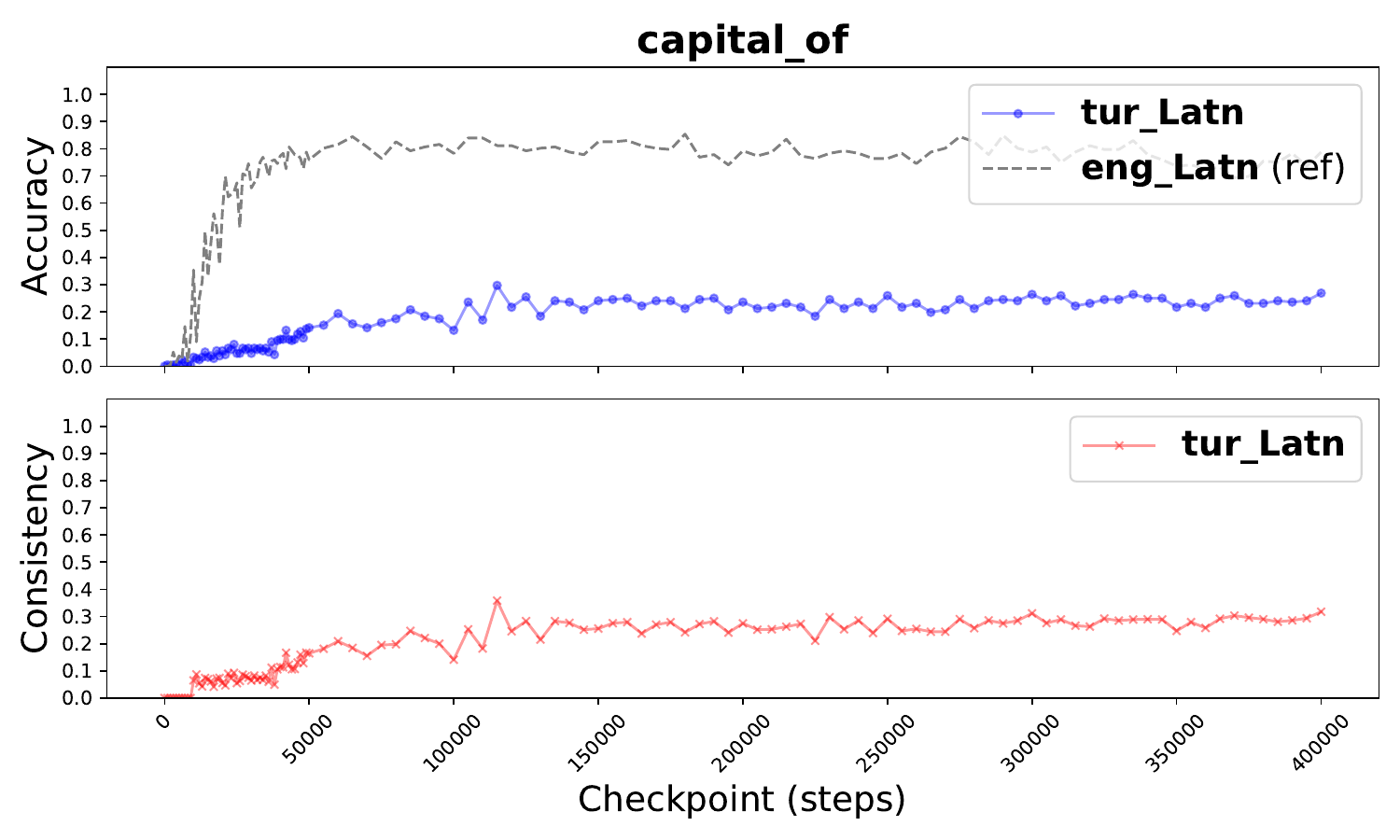}
    \includegraphics[width=0.24\textwidth]{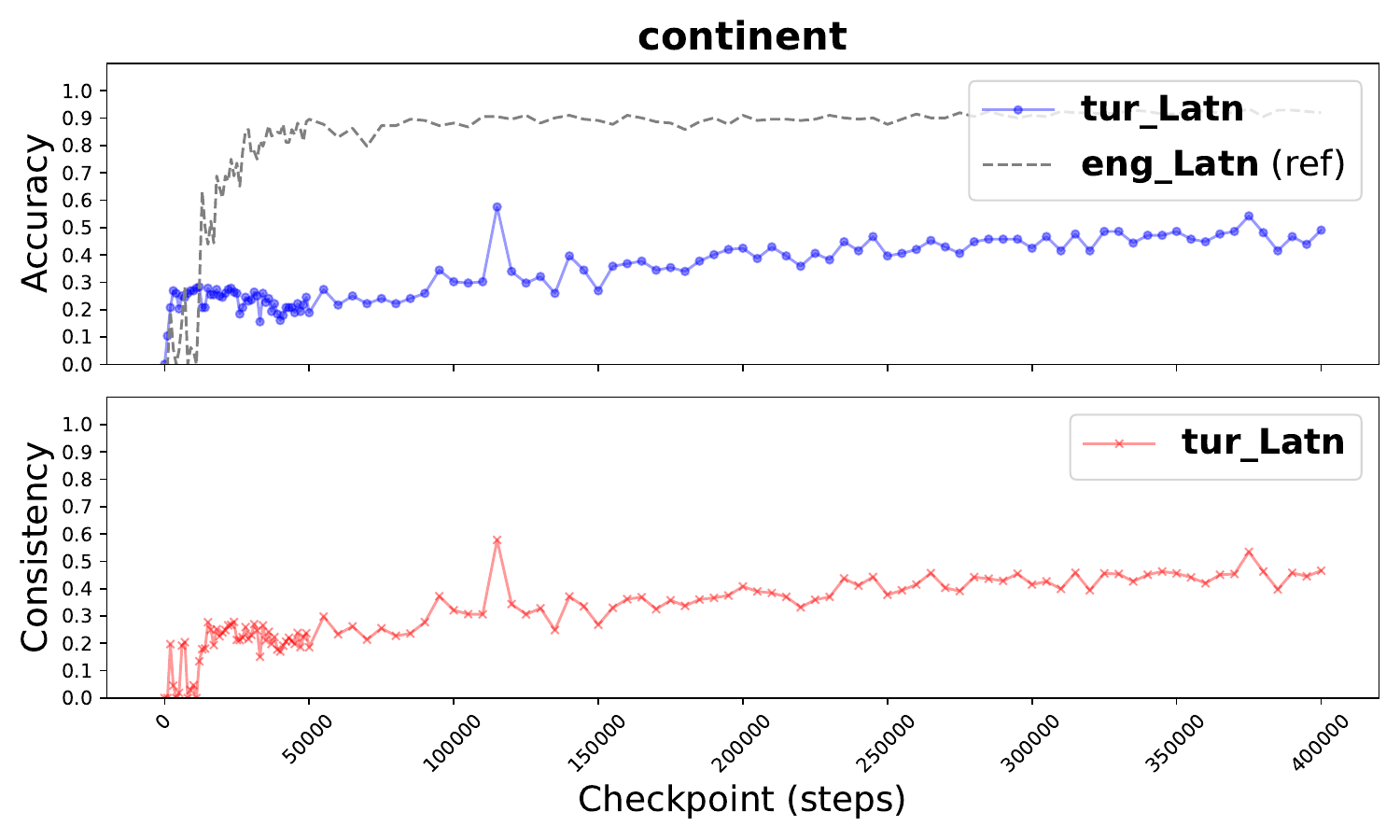}
    \includegraphics[width=0.24\textwidth]{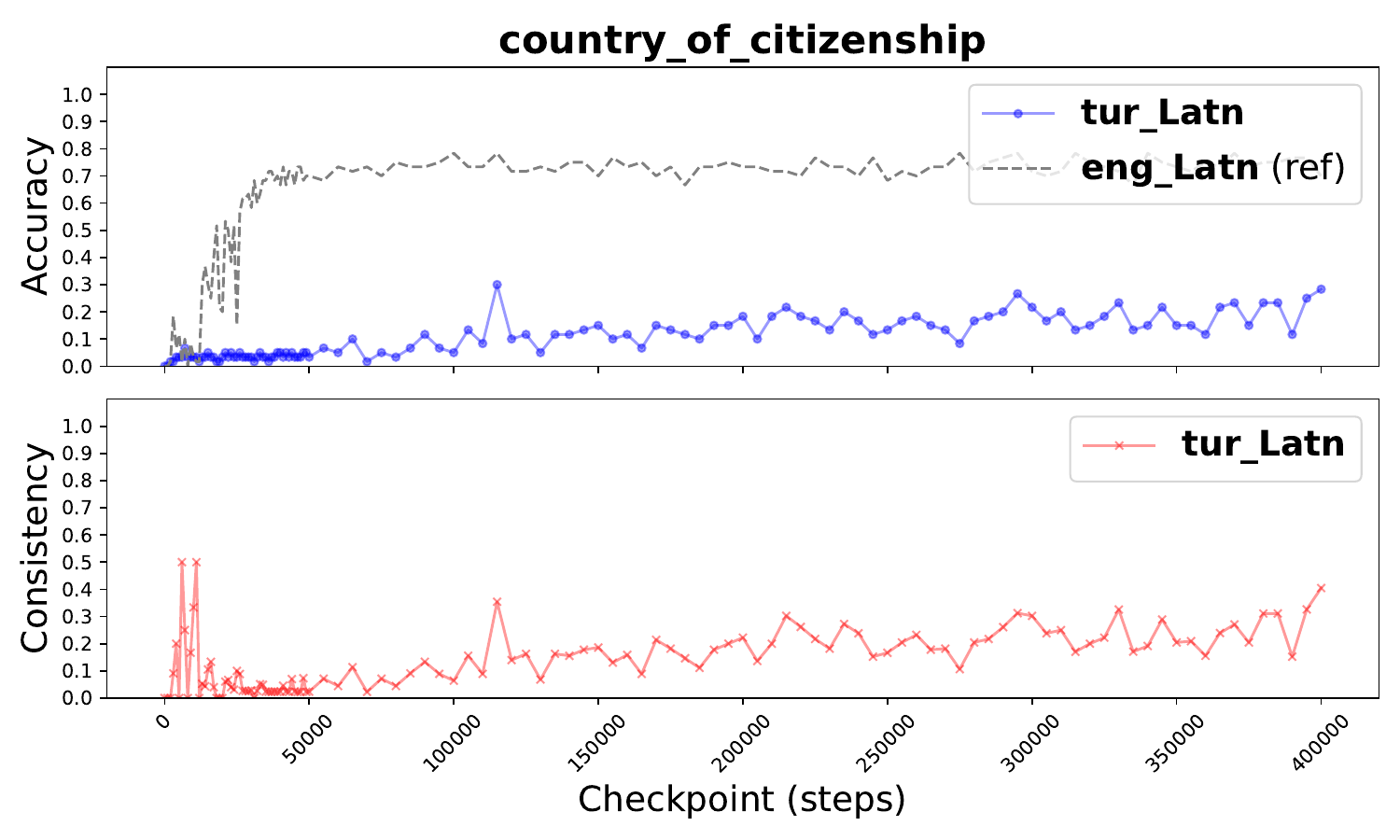}
    \includegraphics[width=0.24\textwidth]{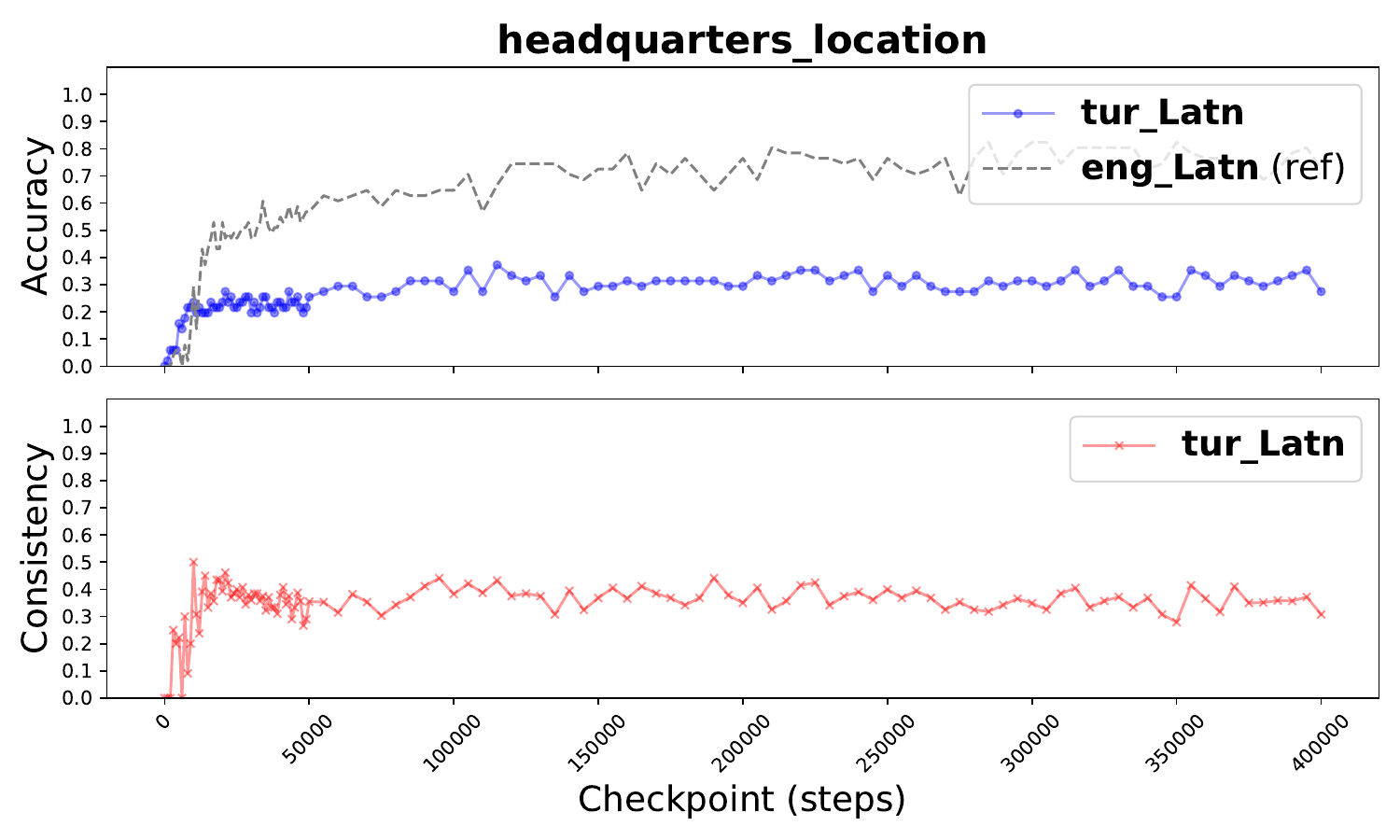}
    \includegraphics[width=0.24\textwidth]{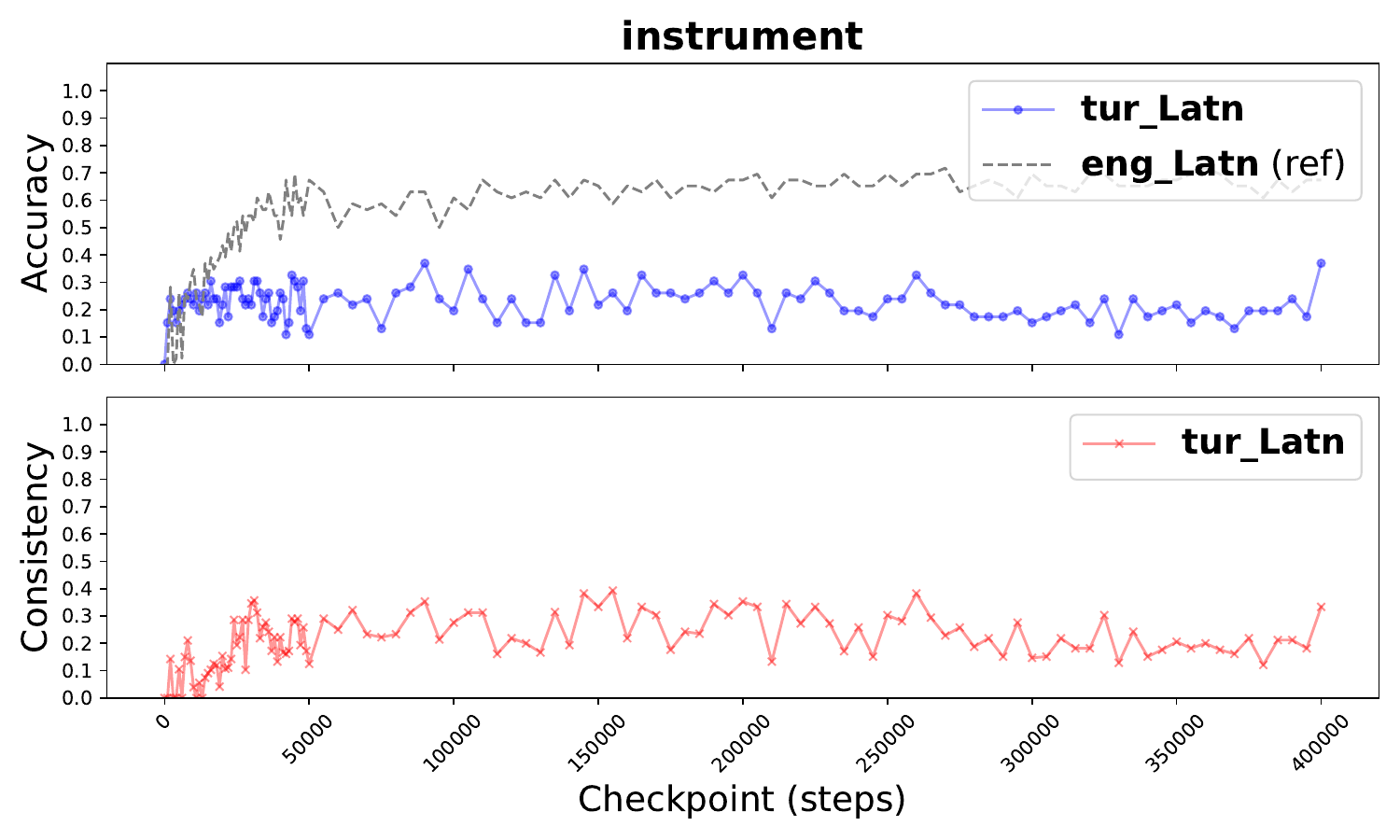}
    \includegraphics[width=0.24\textwidth]{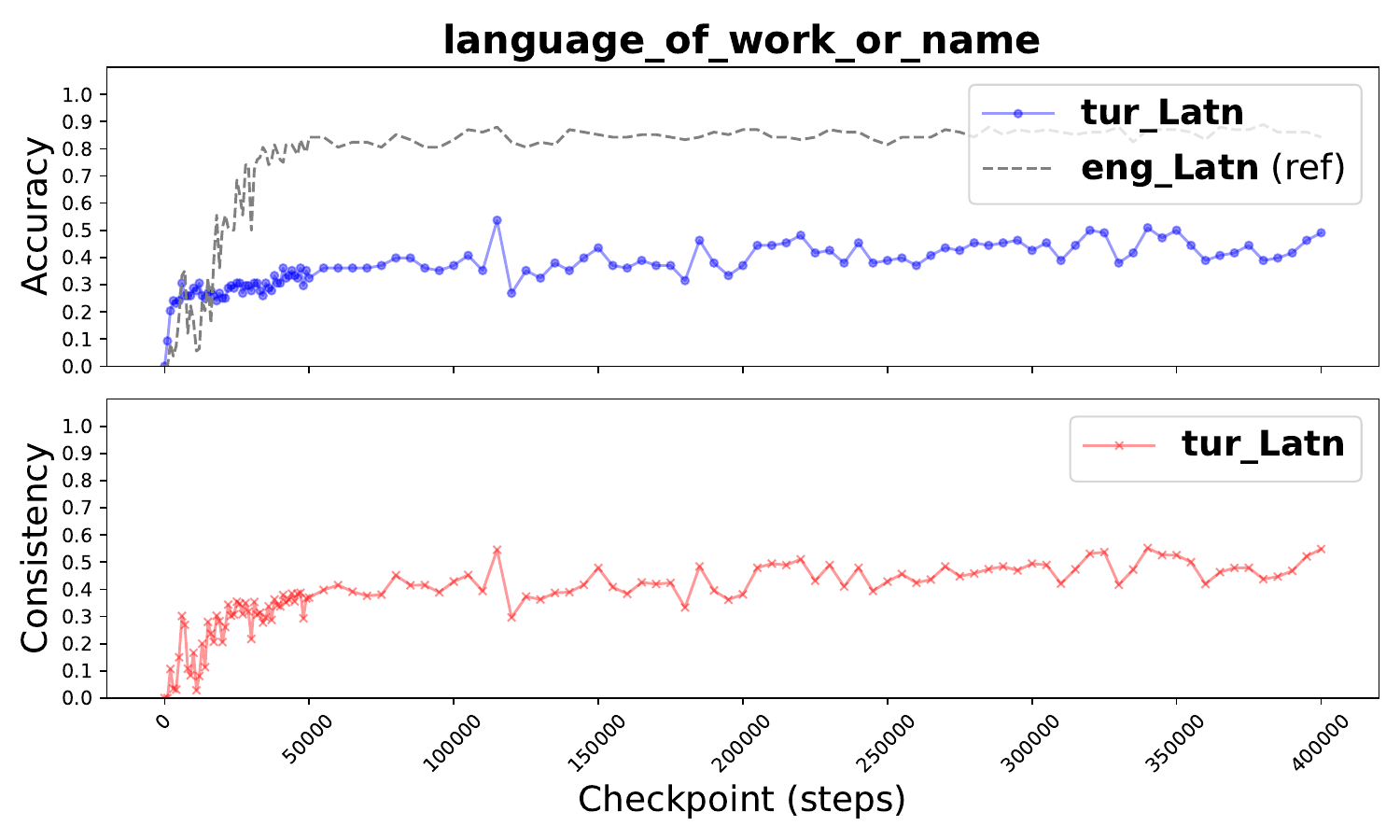}
    \includegraphics[width=0.24\textwidth]{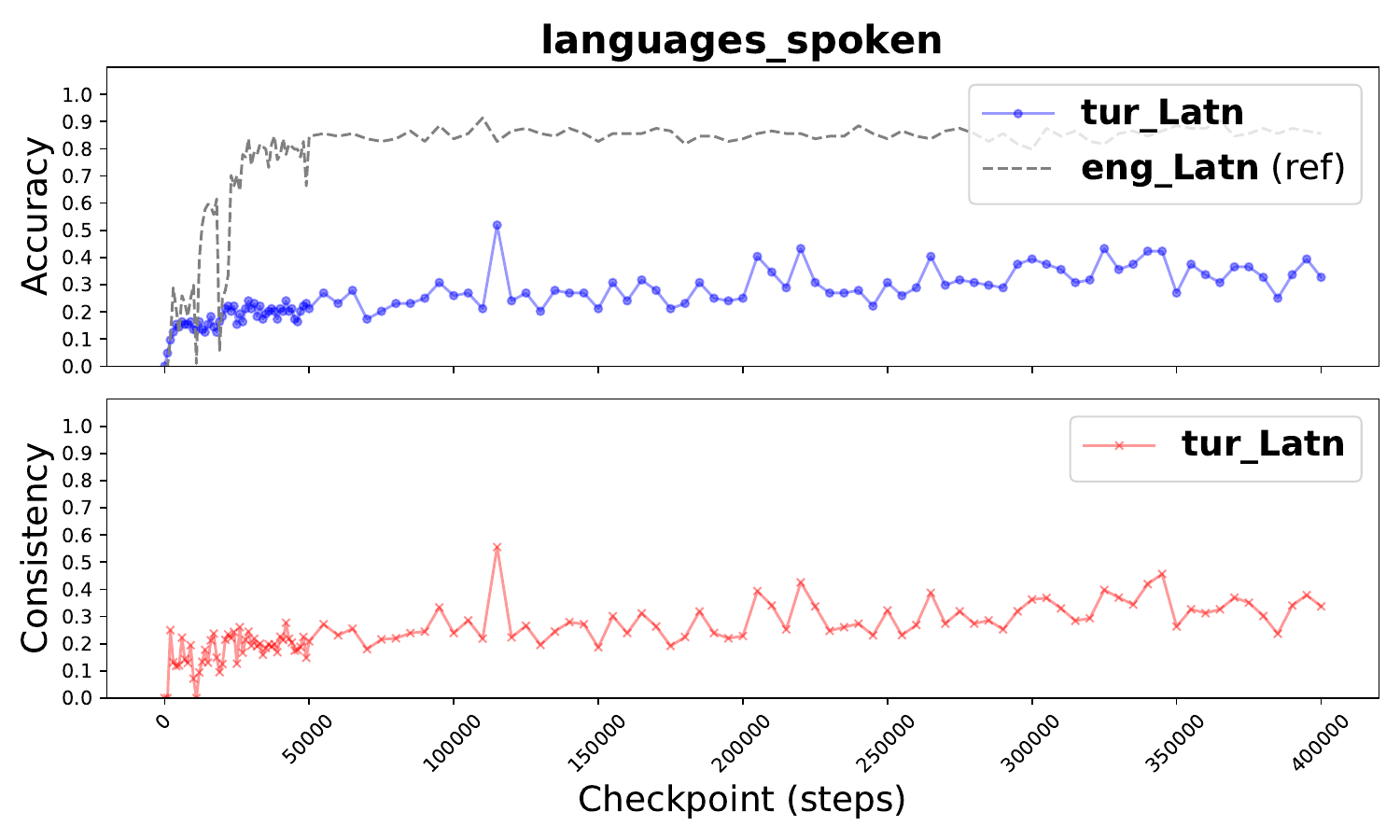}
    \includegraphics[width=0.24\textwidth]{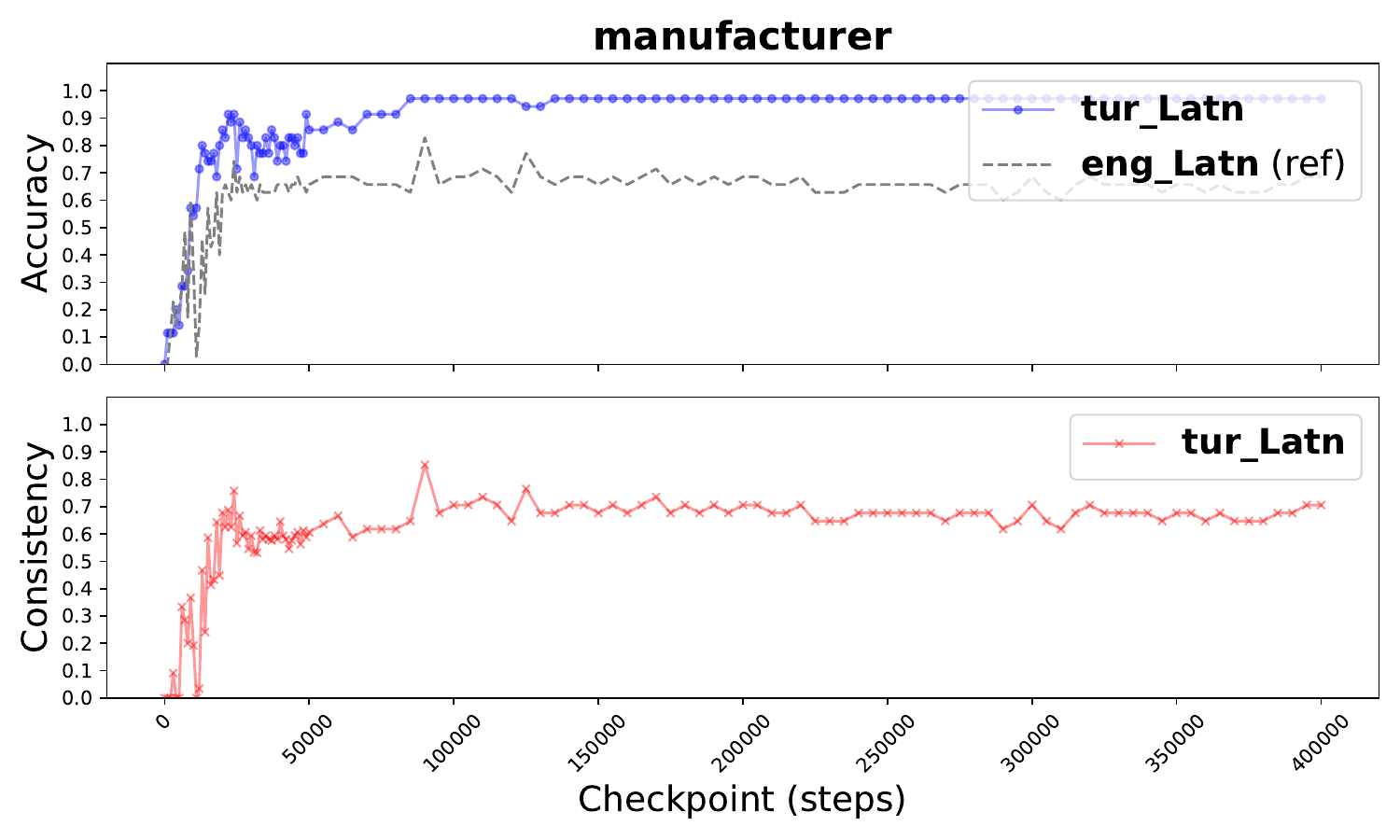}
    \includegraphics[width=0.24\textwidth]{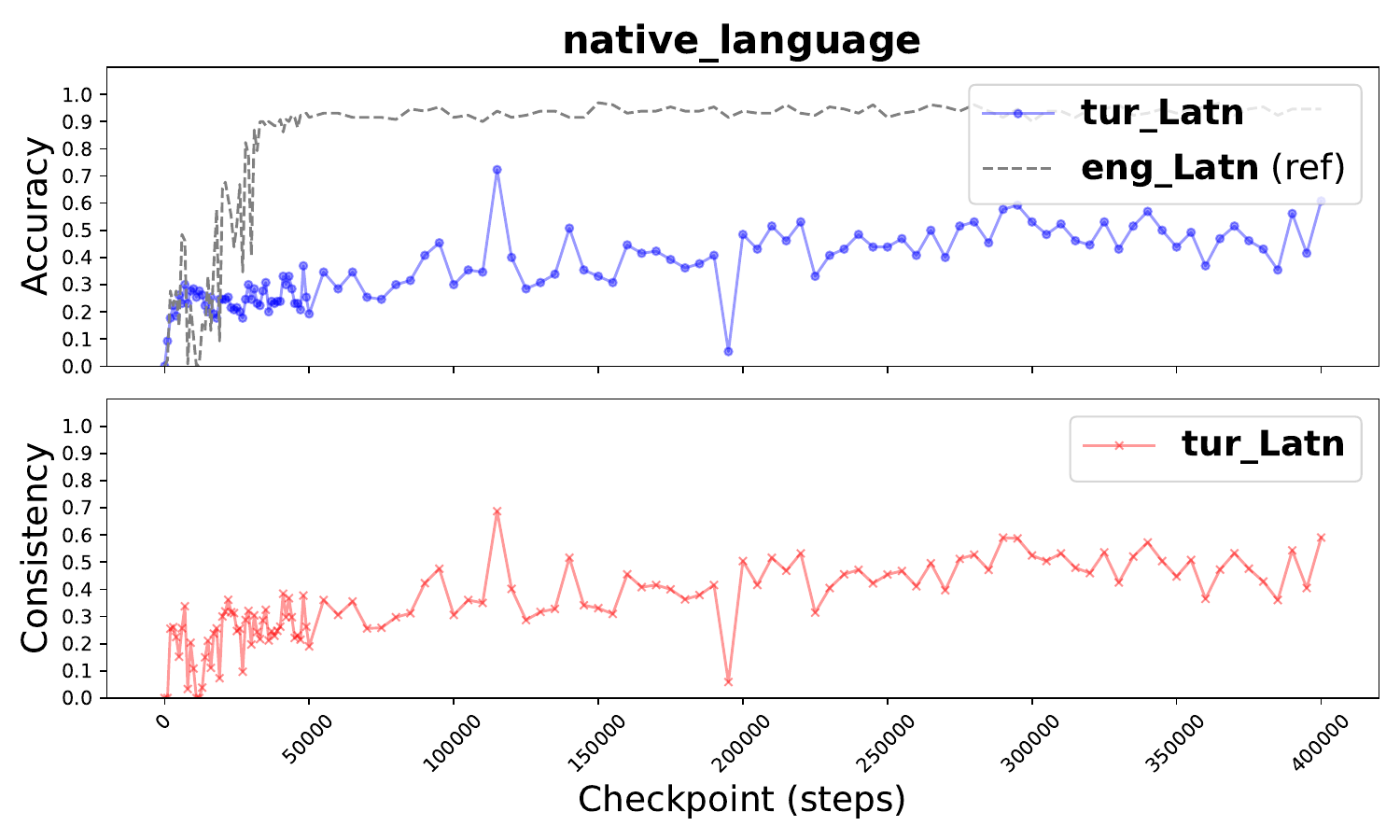}
    \includegraphics[width=0.24\textwidth]{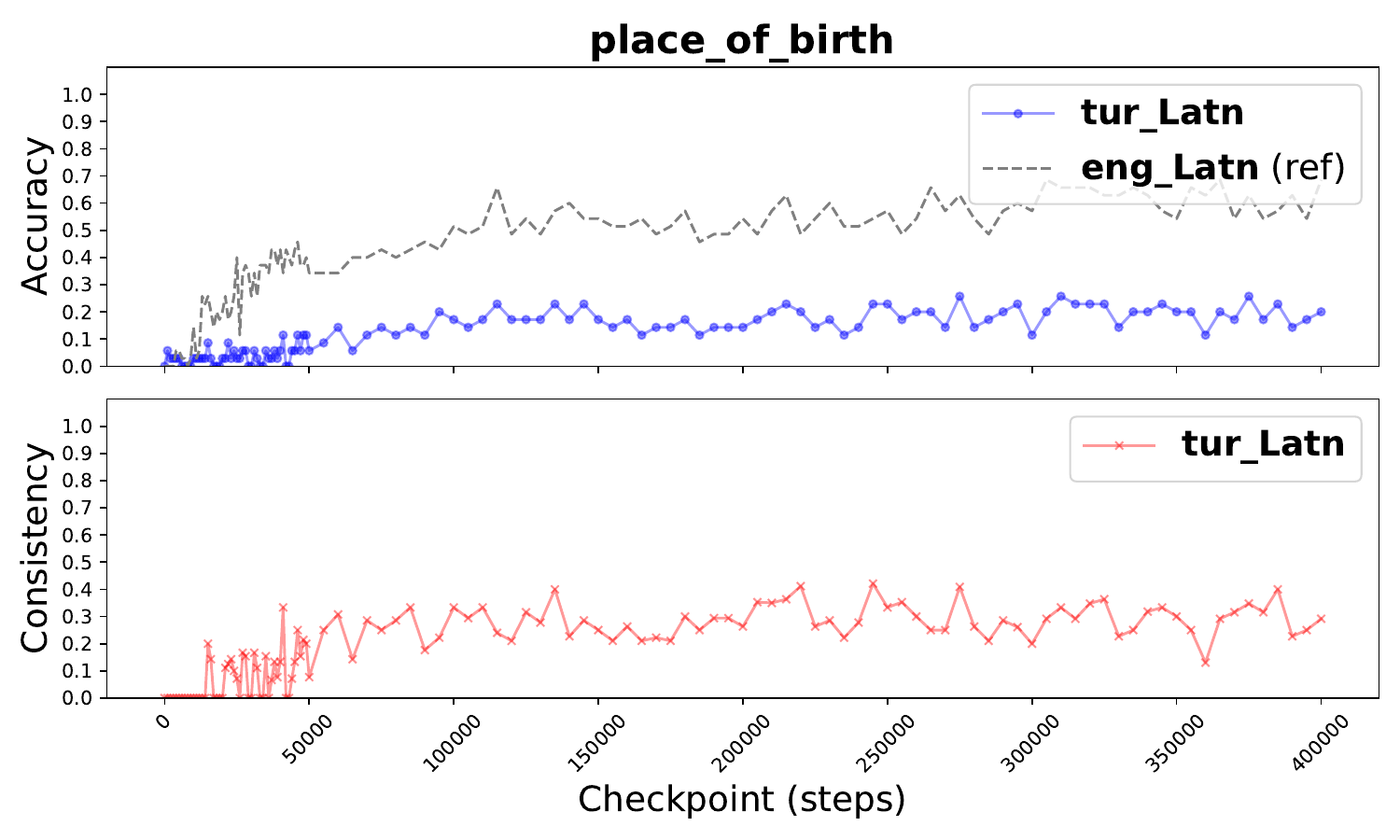}
    \includegraphics[width=0.24\textwidth]{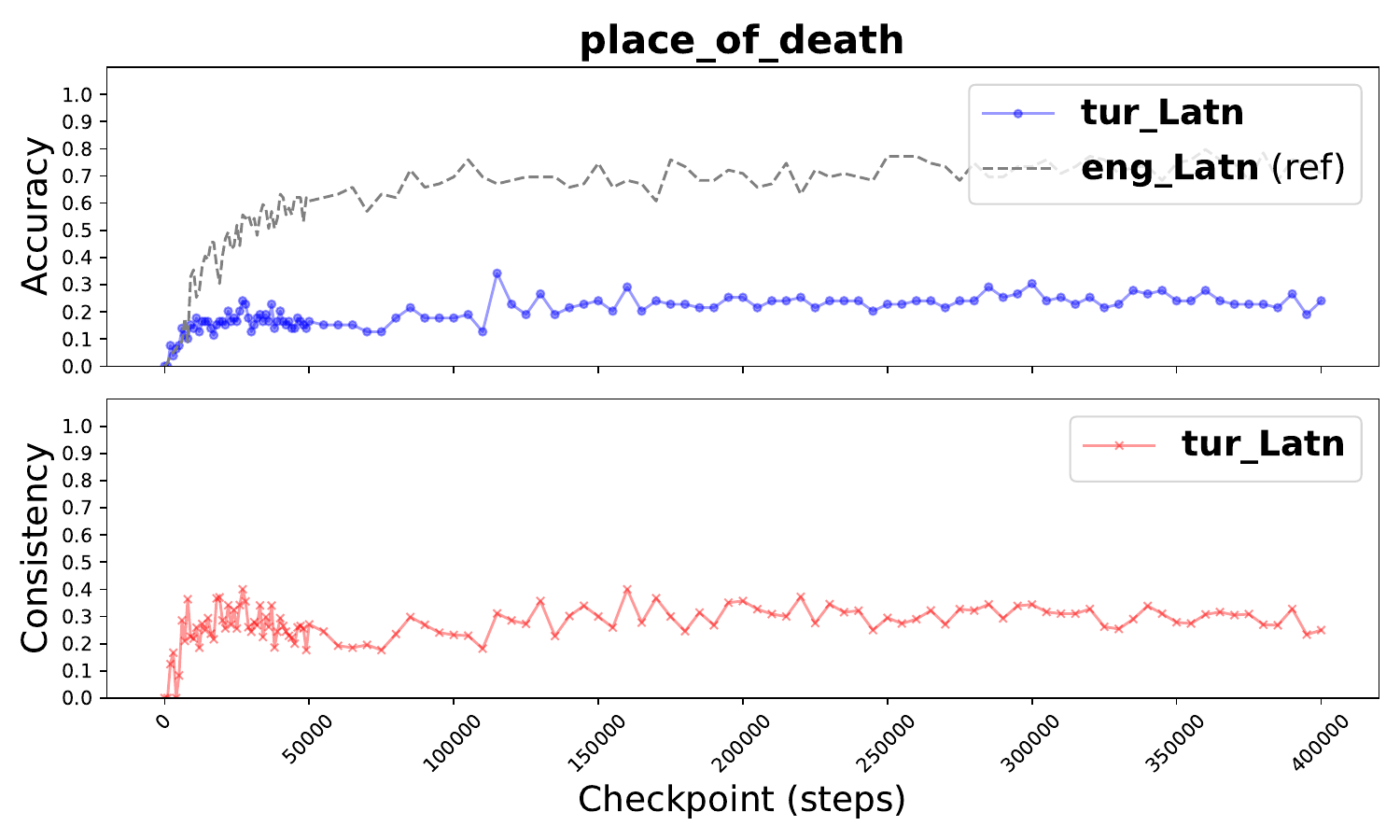}
    \includegraphics[width=0.24\textwidth]{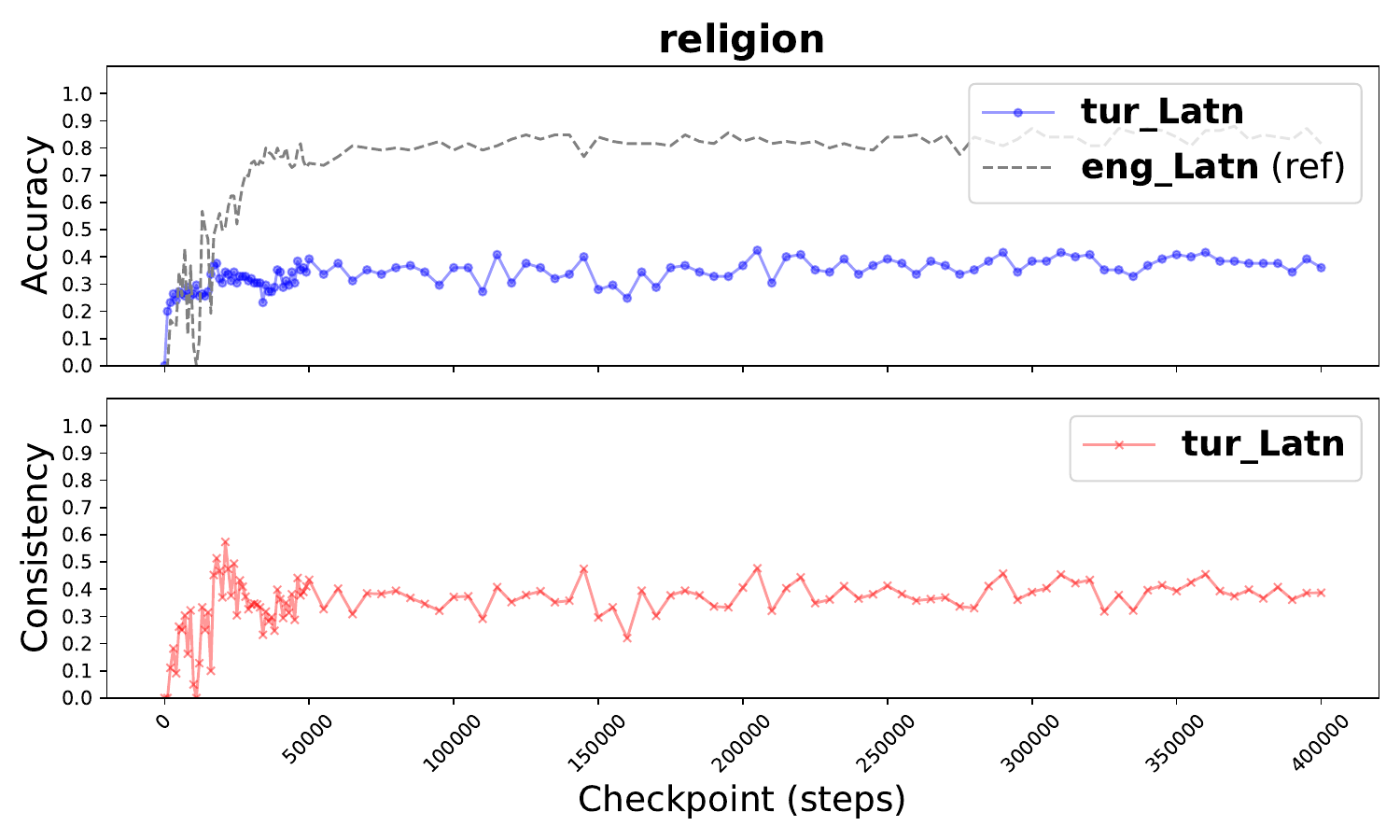}
    \caption{Factual accuracy (ACC) and crosslingual consistency (CO) for each relation type in \textbf{tur\_Latn}.}
    \label{fig:performance_over_checkpoints_tr}
\end{figure*}

\begin{figure*}
    \centering
    \includegraphics[width=0.24\textwidth]{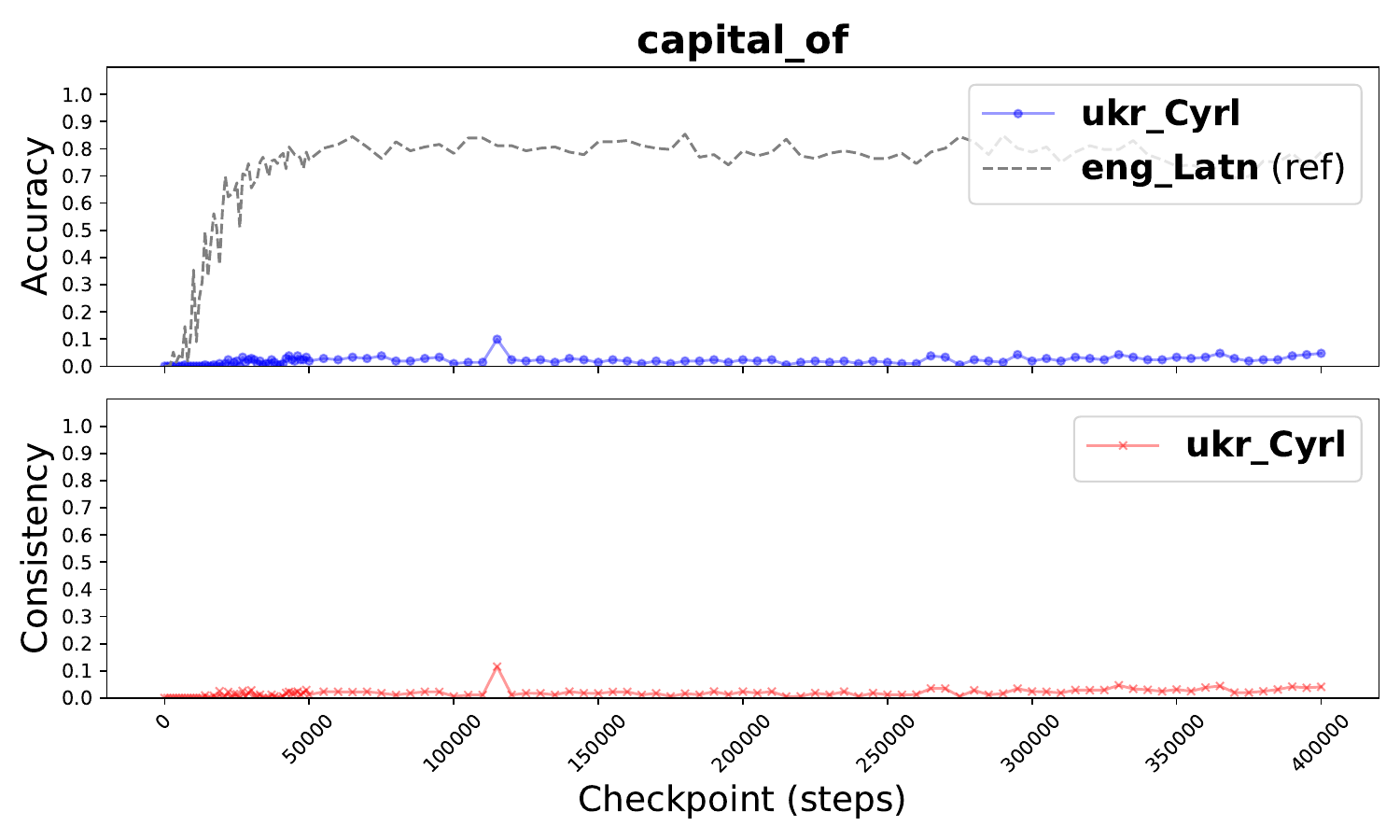}
    \includegraphics[width=0.24\textwidth]{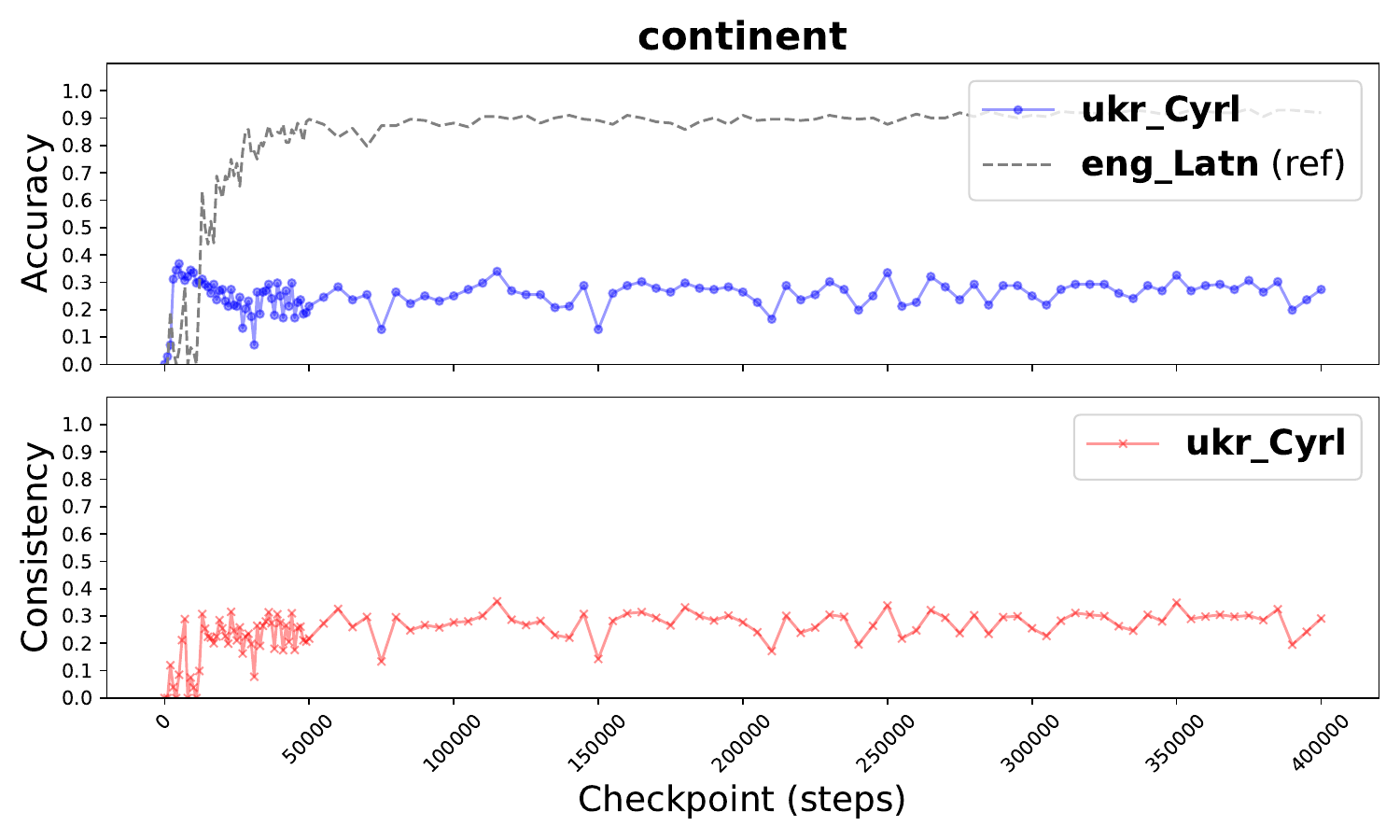}
    \includegraphics[width=0.24\textwidth]{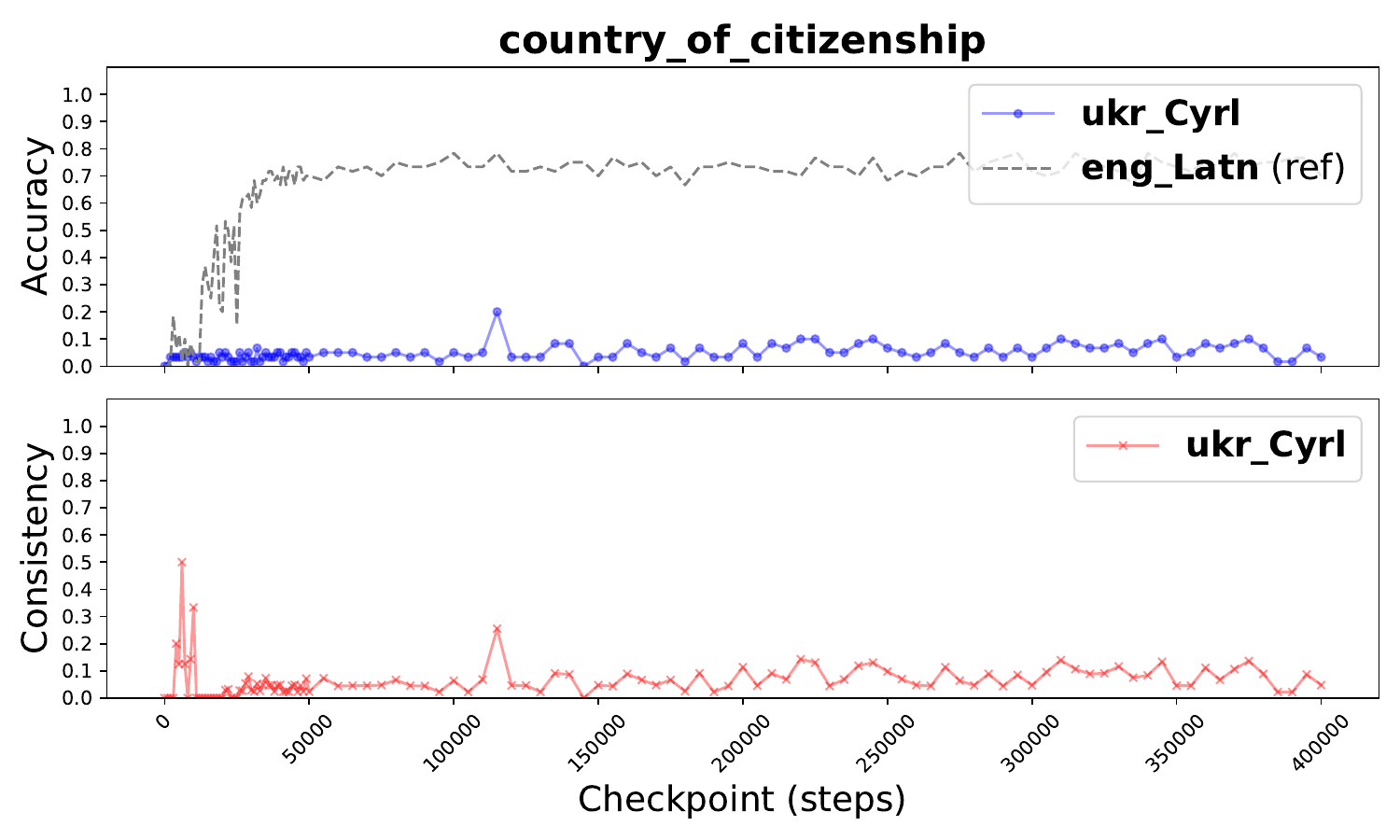}
    \includegraphics[width=0.24\textwidth]{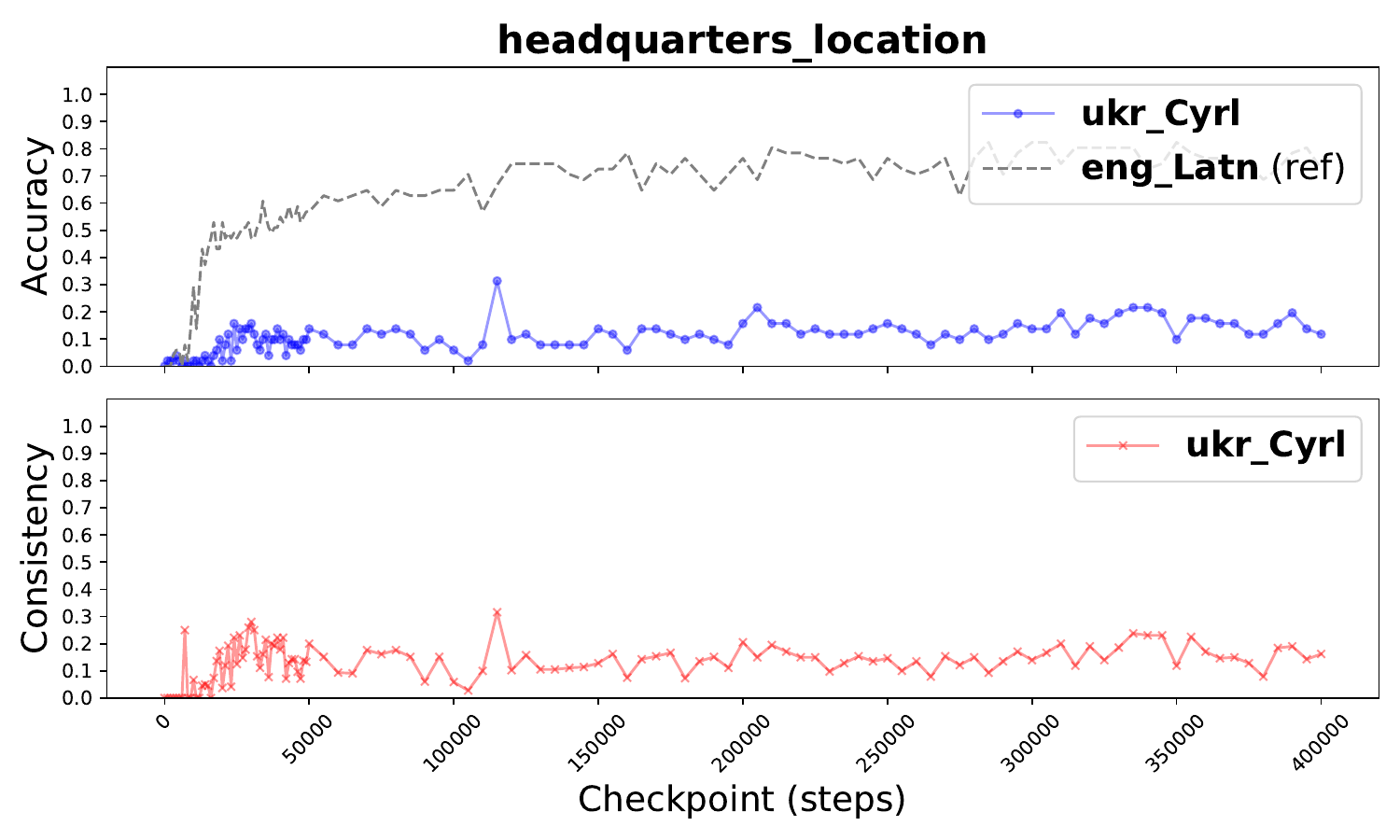}
    \includegraphics[width=0.24\textwidth]{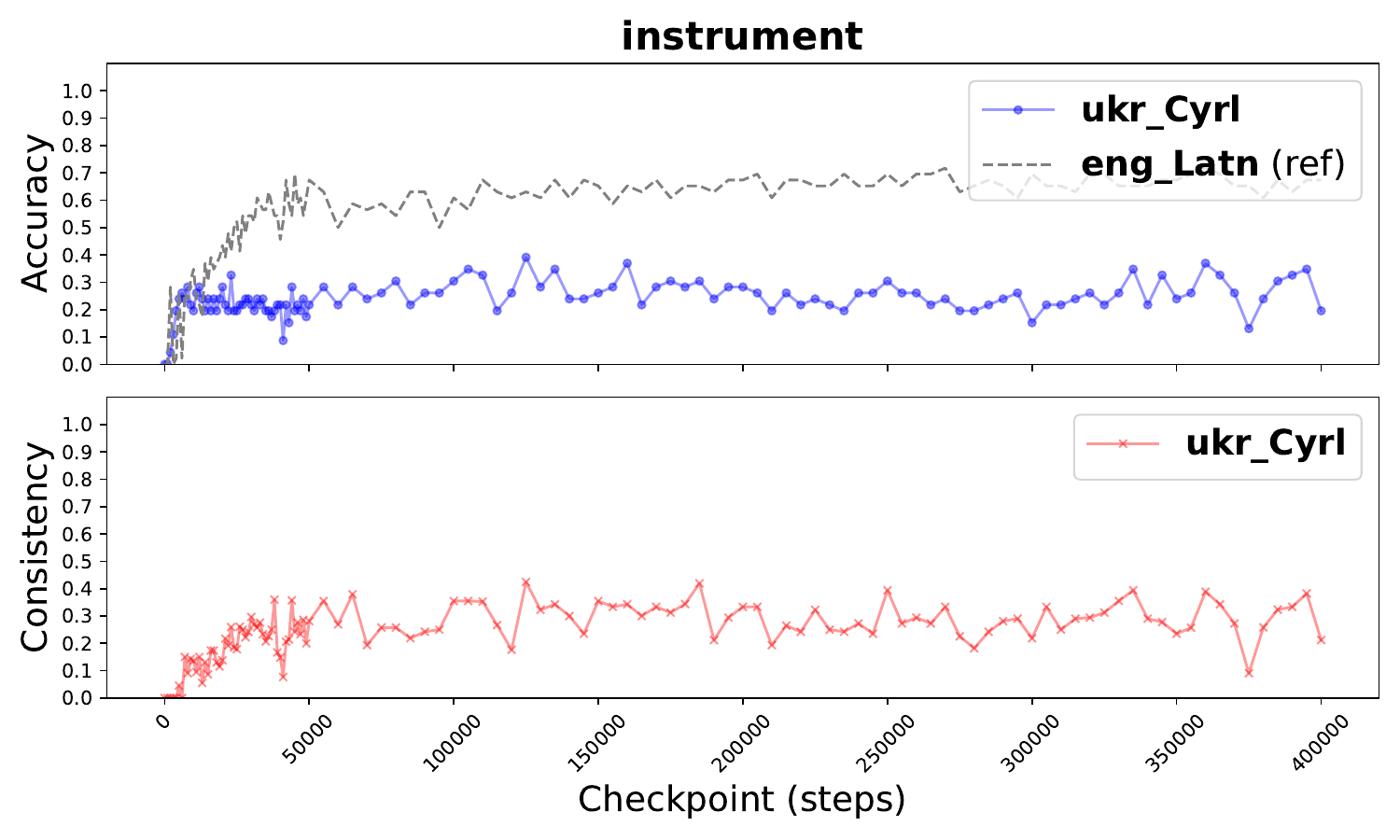}
    \includegraphics[width=0.24\textwidth]{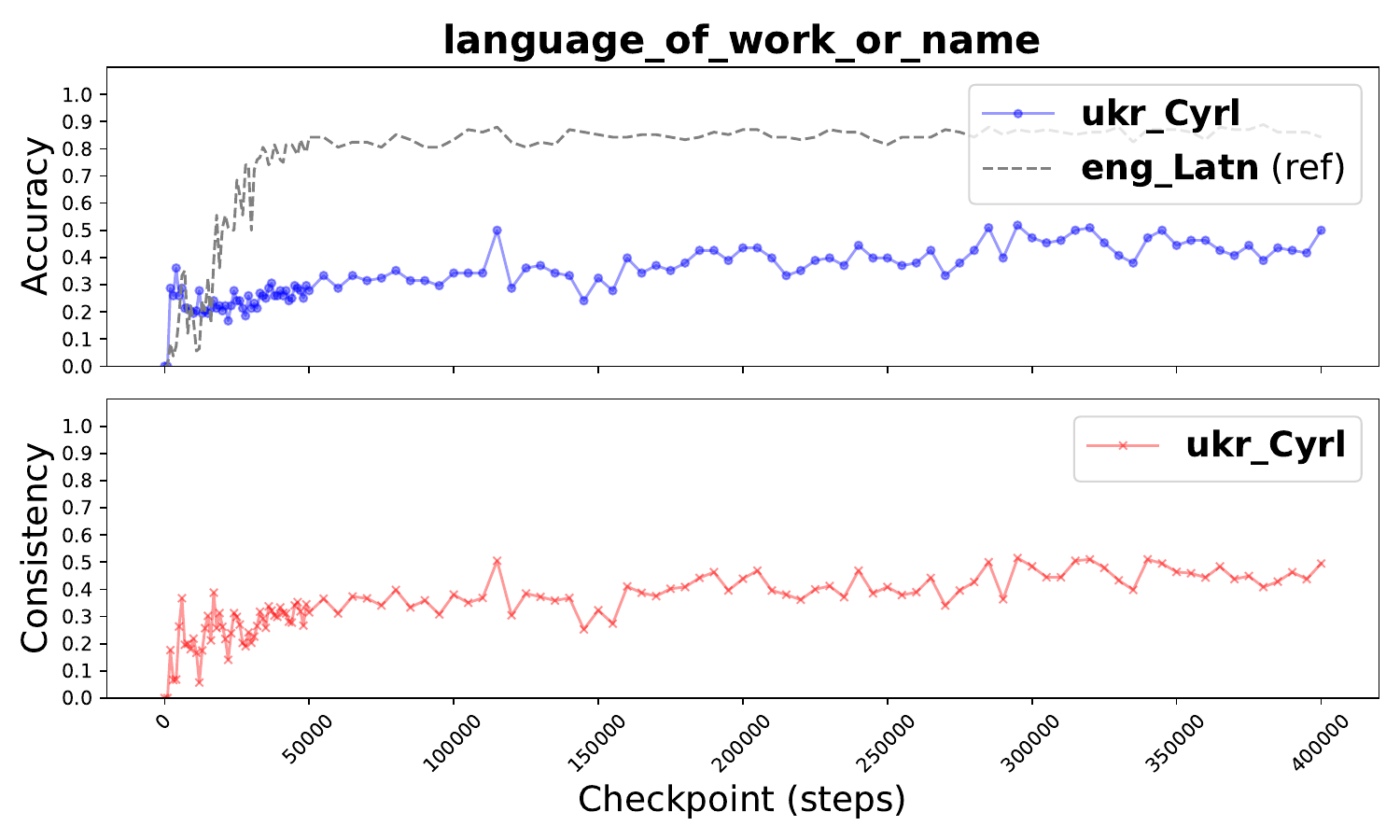}
    \includegraphics[width=0.24\textwidth]{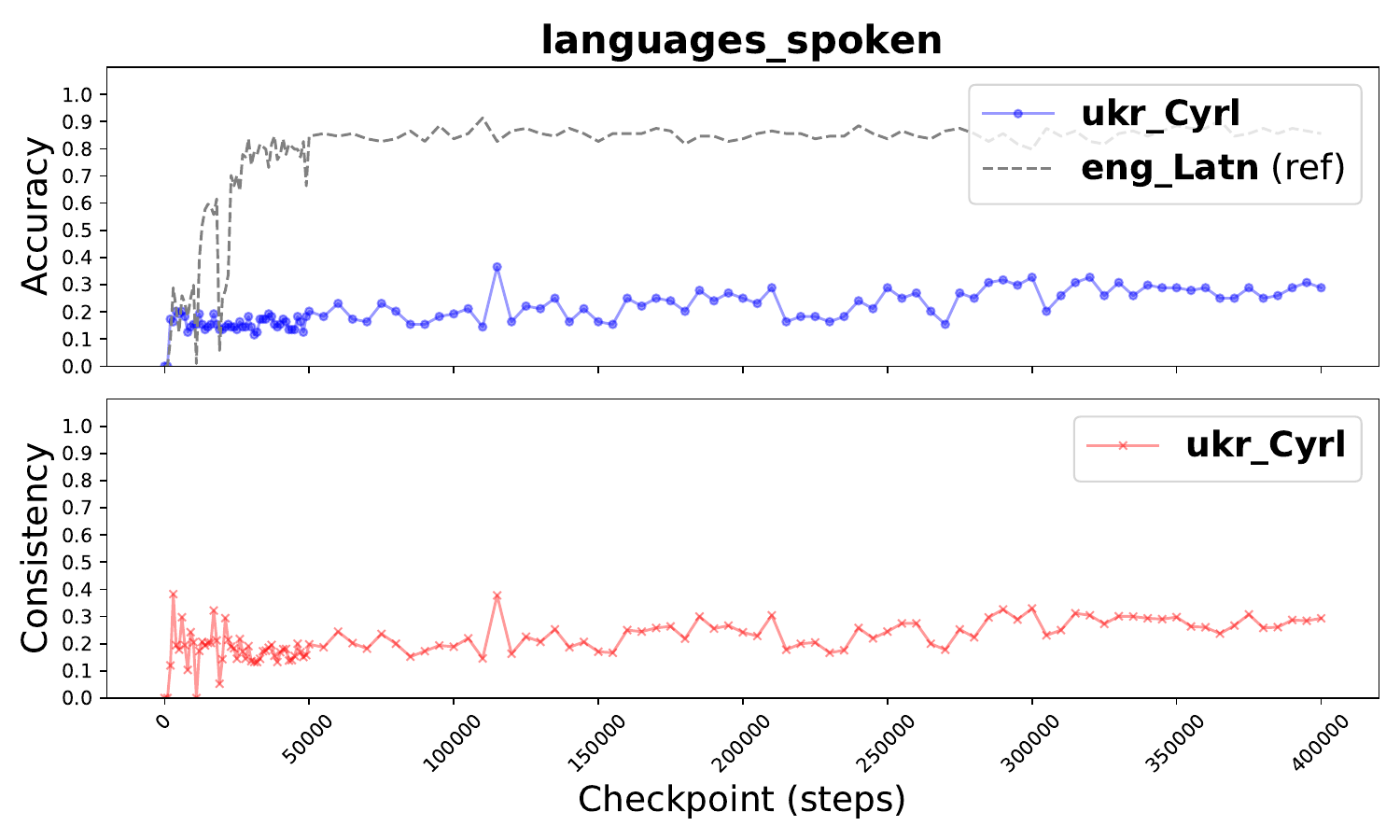}
    \includegraphics[width=0.24\textwidth]{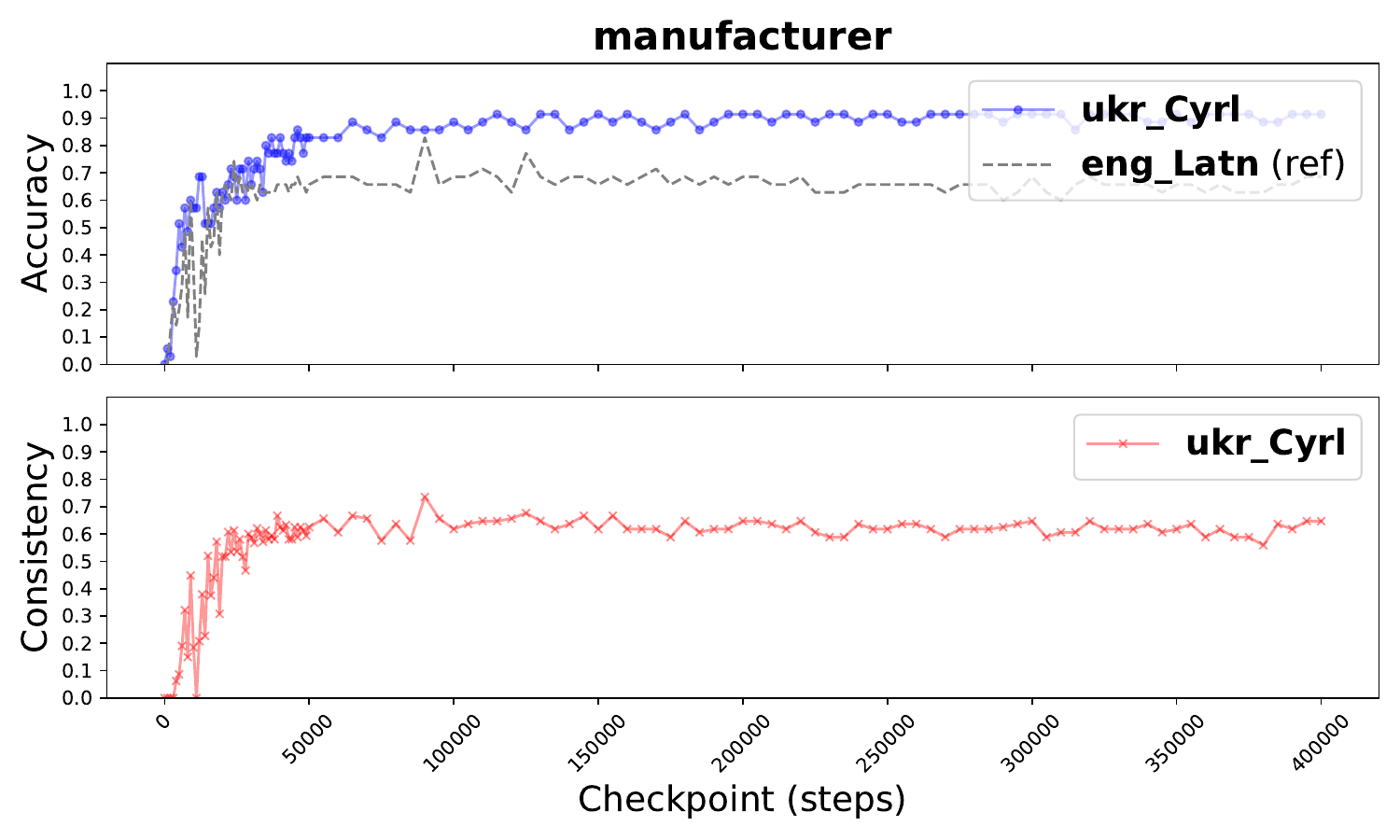}
    \includegraphics[width=0.24\textwidth]{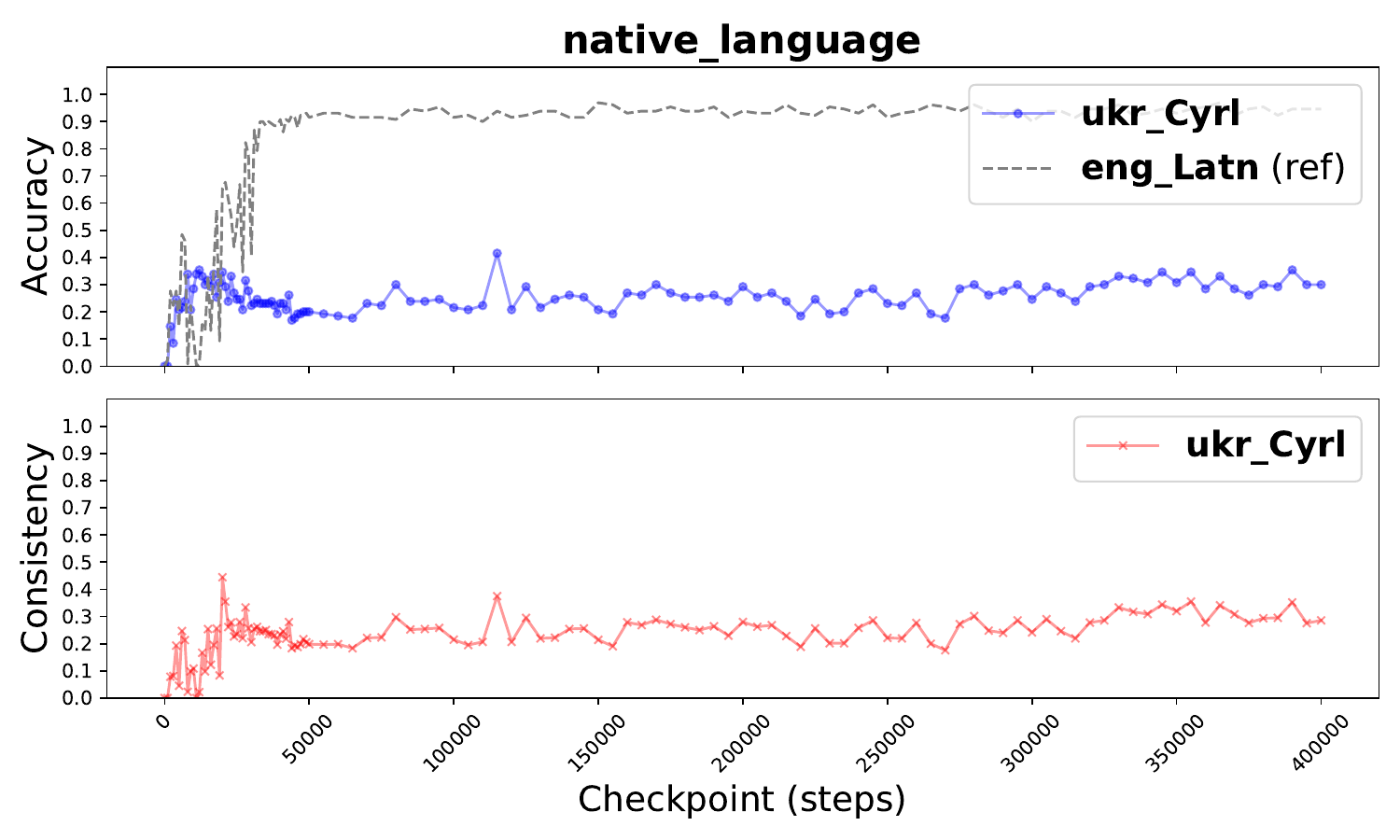}
    \includegraphics[width=0.24\textwidth]{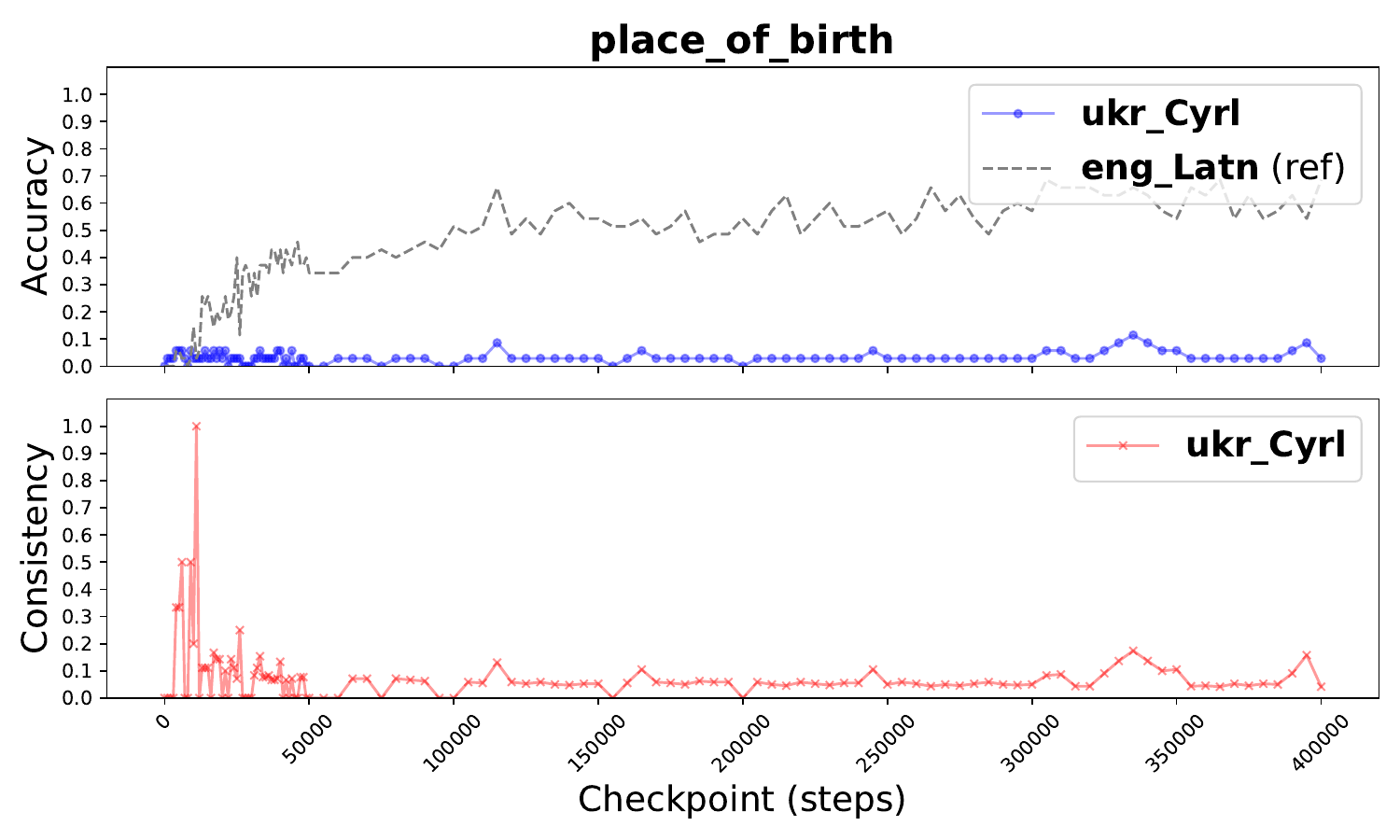}
    \includegraphics[width=0.24\textwidth]{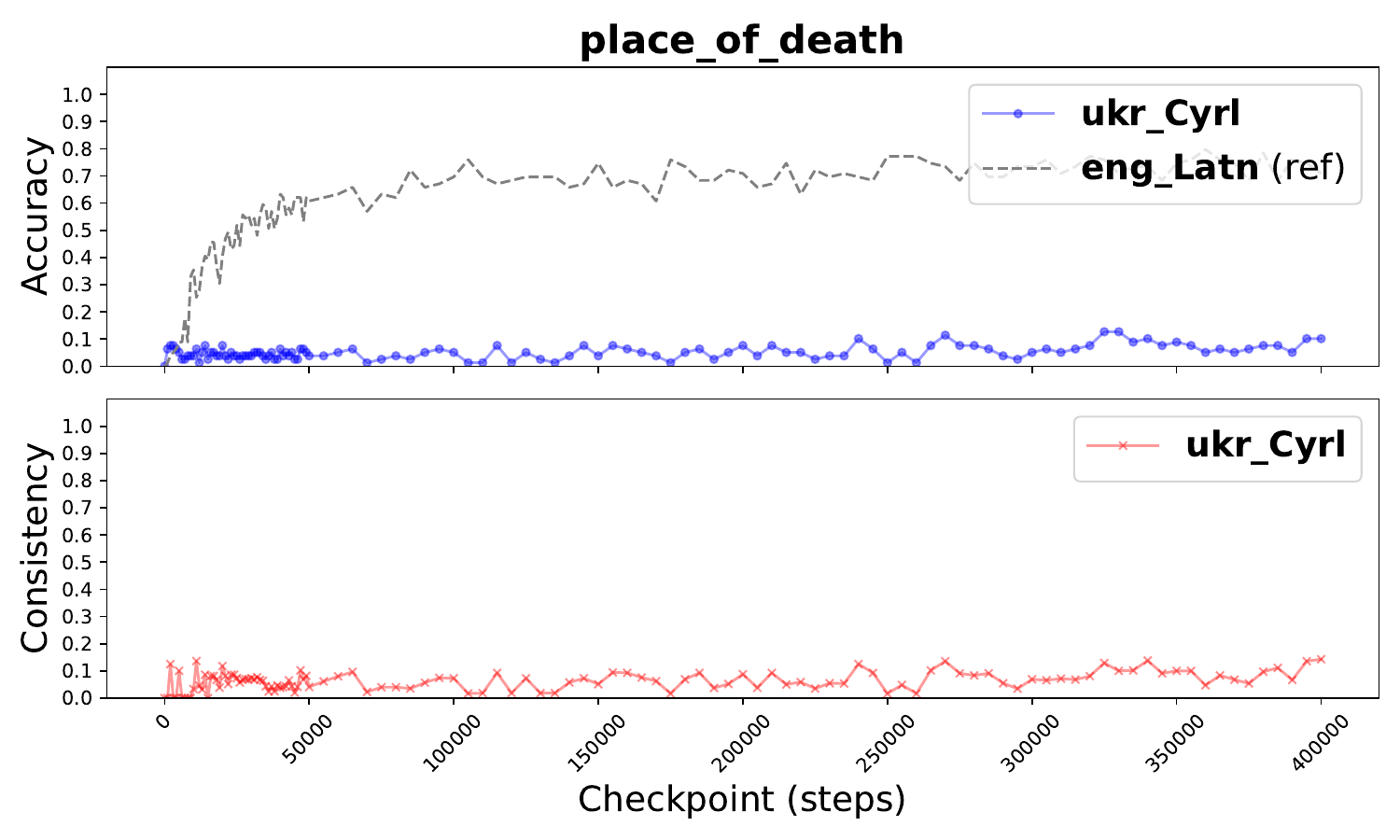}
    \includegraphics[width=0.24\textwidth]{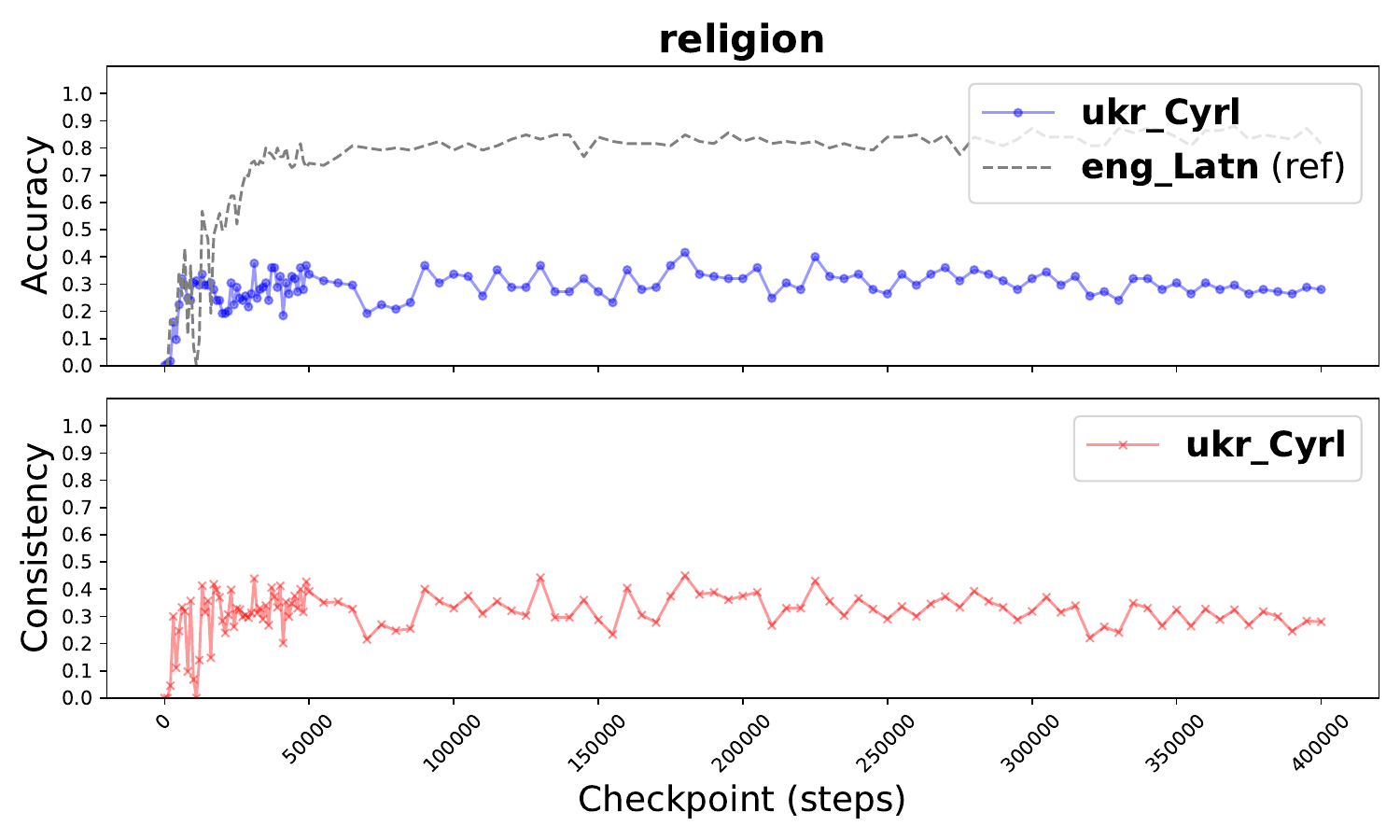}
    \caption{Factual accuracy (ACC) and crosslingual consistency (CO) for each relation type in \textbf{ukr\_Cyrl}.}
    \label{fig:performance_over_checkpoints_uk}
\end{figure*}

\begin{figure*}
    \centering
    \includegraphics[width=0.24\textwidth]{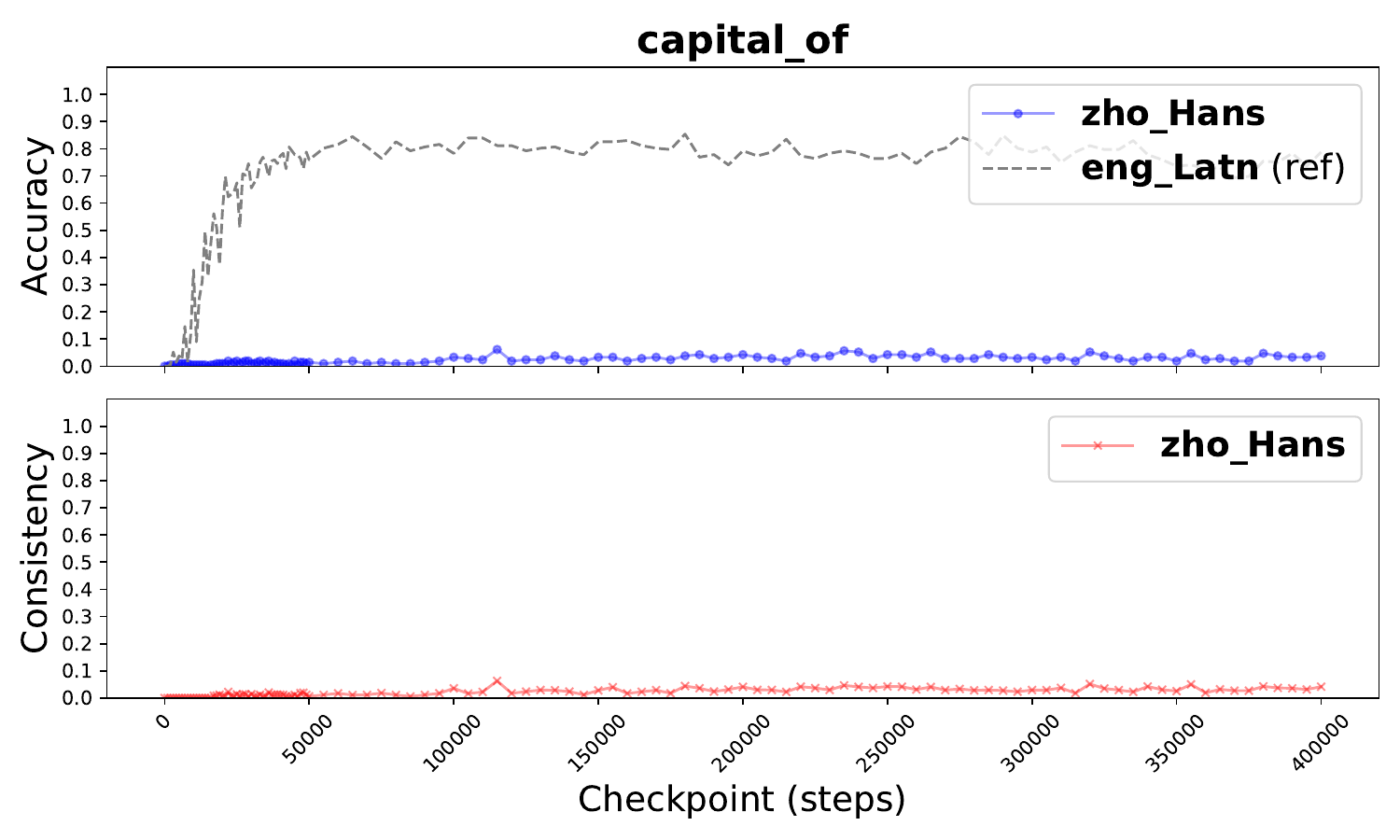}
    \includegraphics[width=0.24\textwidth]{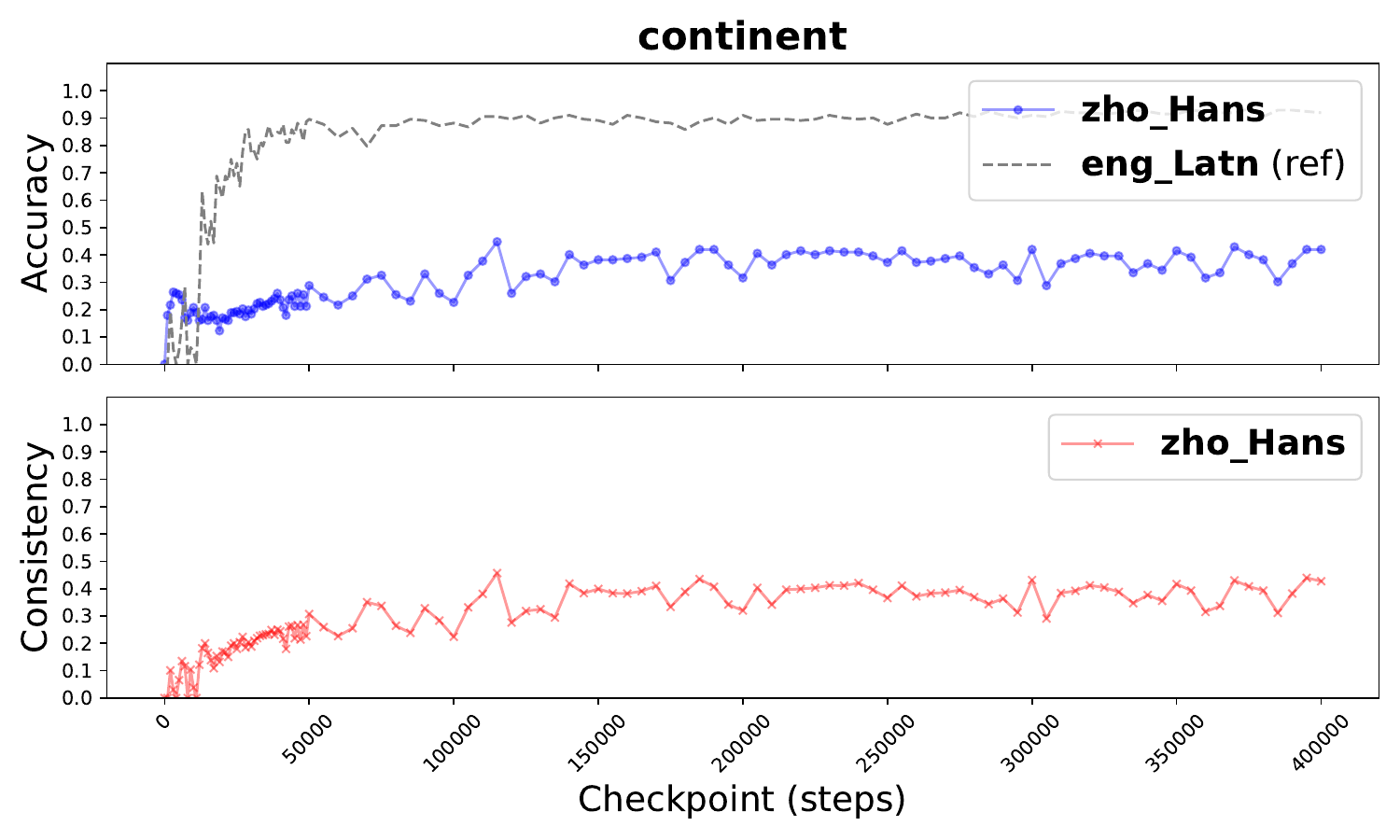}
    \includegraphics[width=0.24\textwidth]{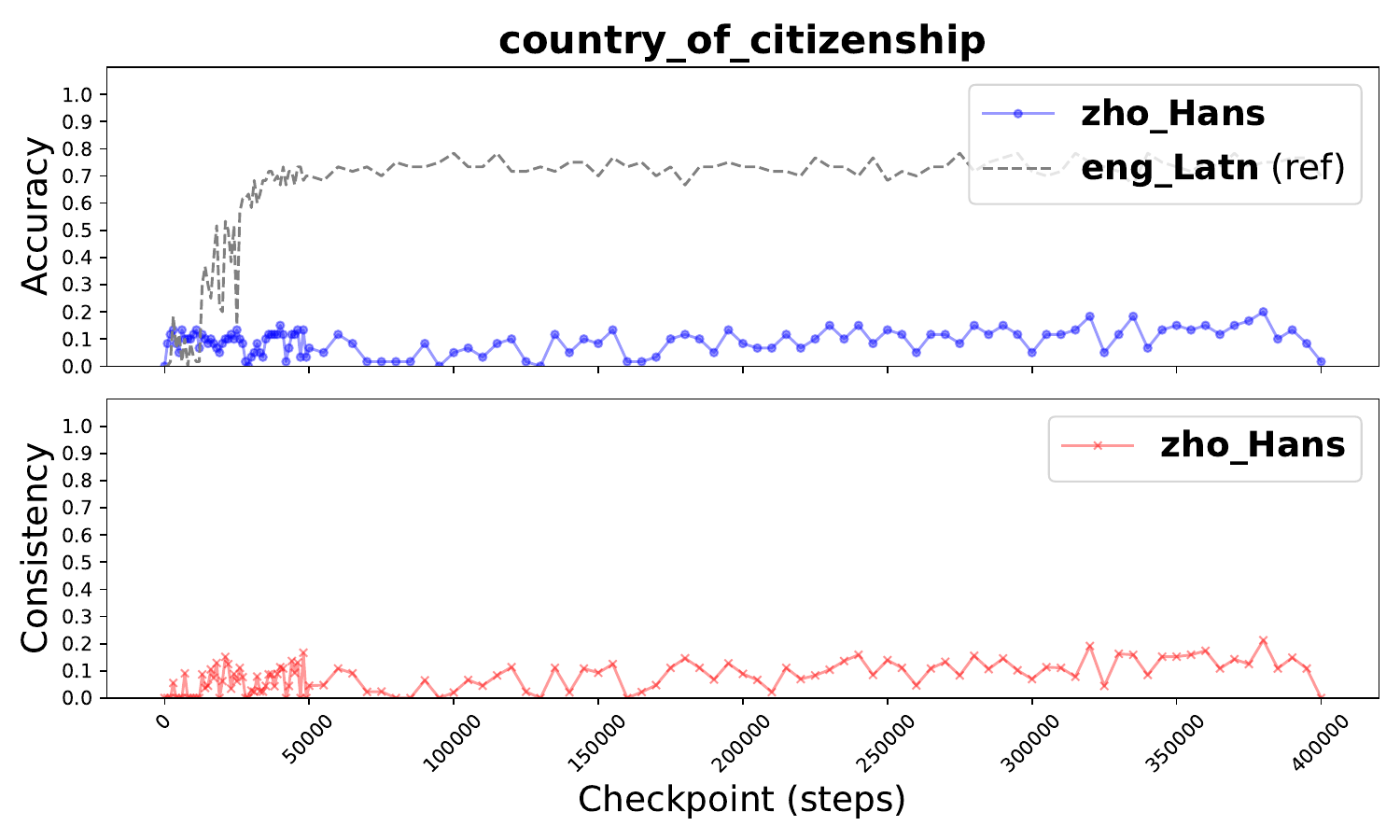}
    \includegraphics[width=0.24\textwidth]{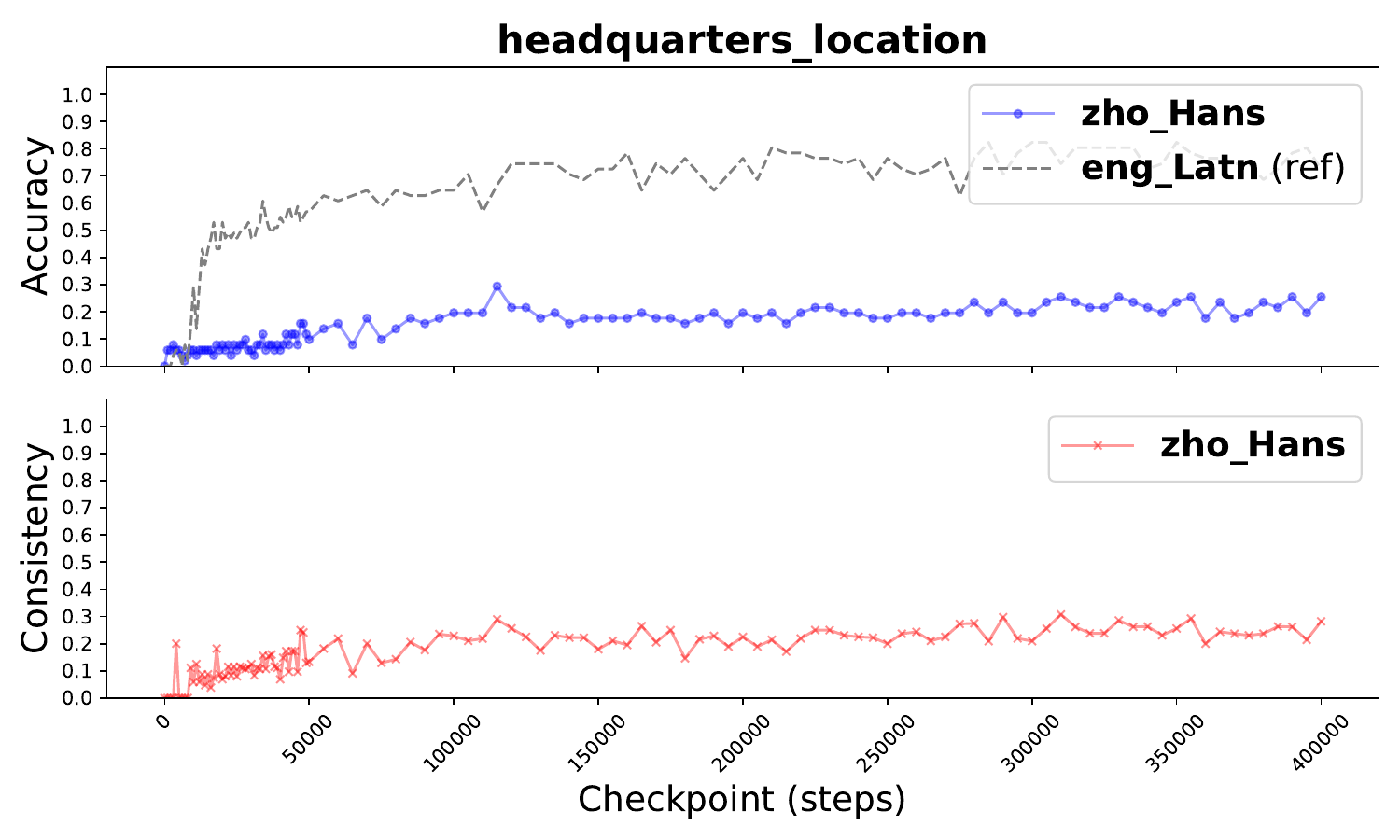}
    \includegraphics[width=0.24\textwidth]{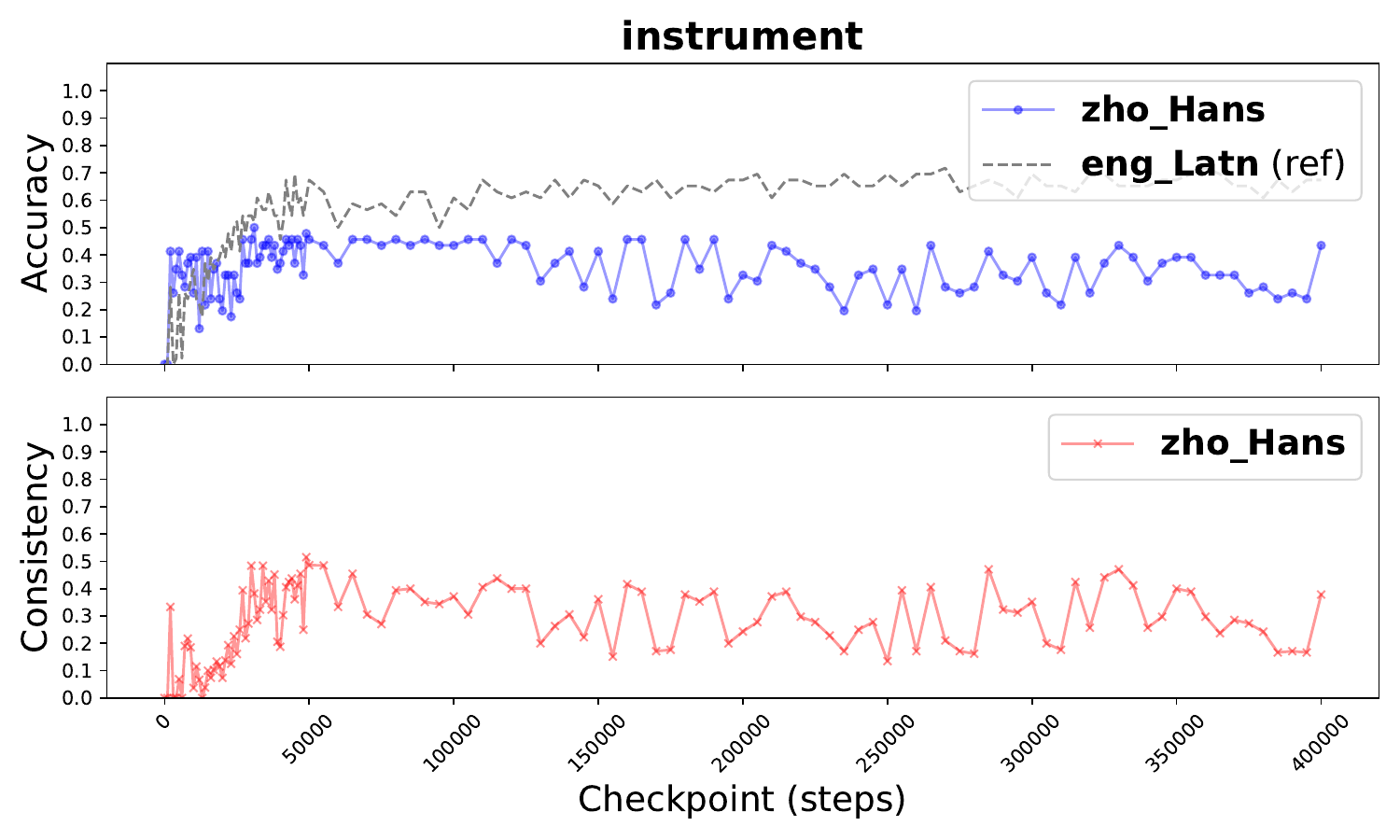}
    \includegraphics[width=0.24\textwidth]{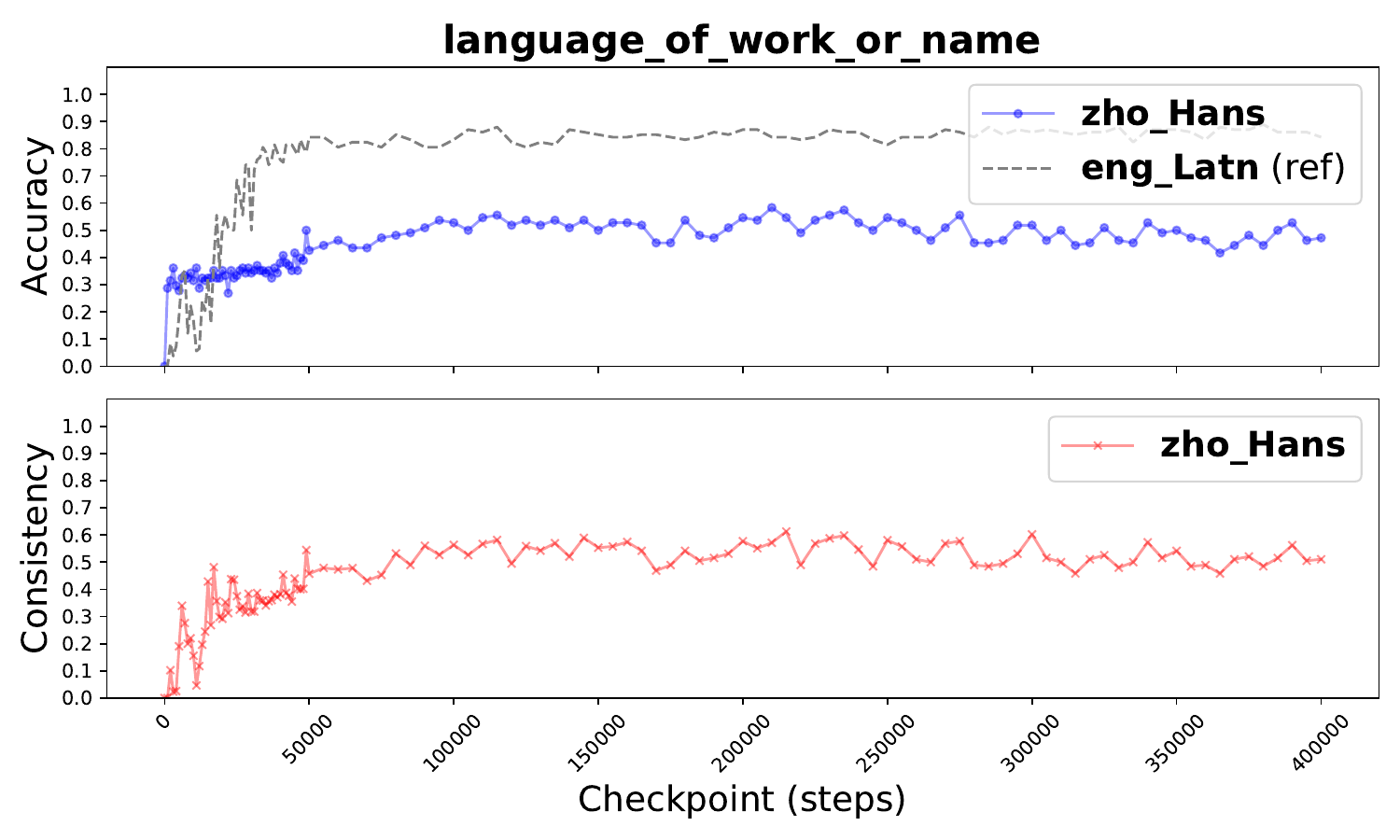}
    \includegraphics[width=0.24\textwidth]{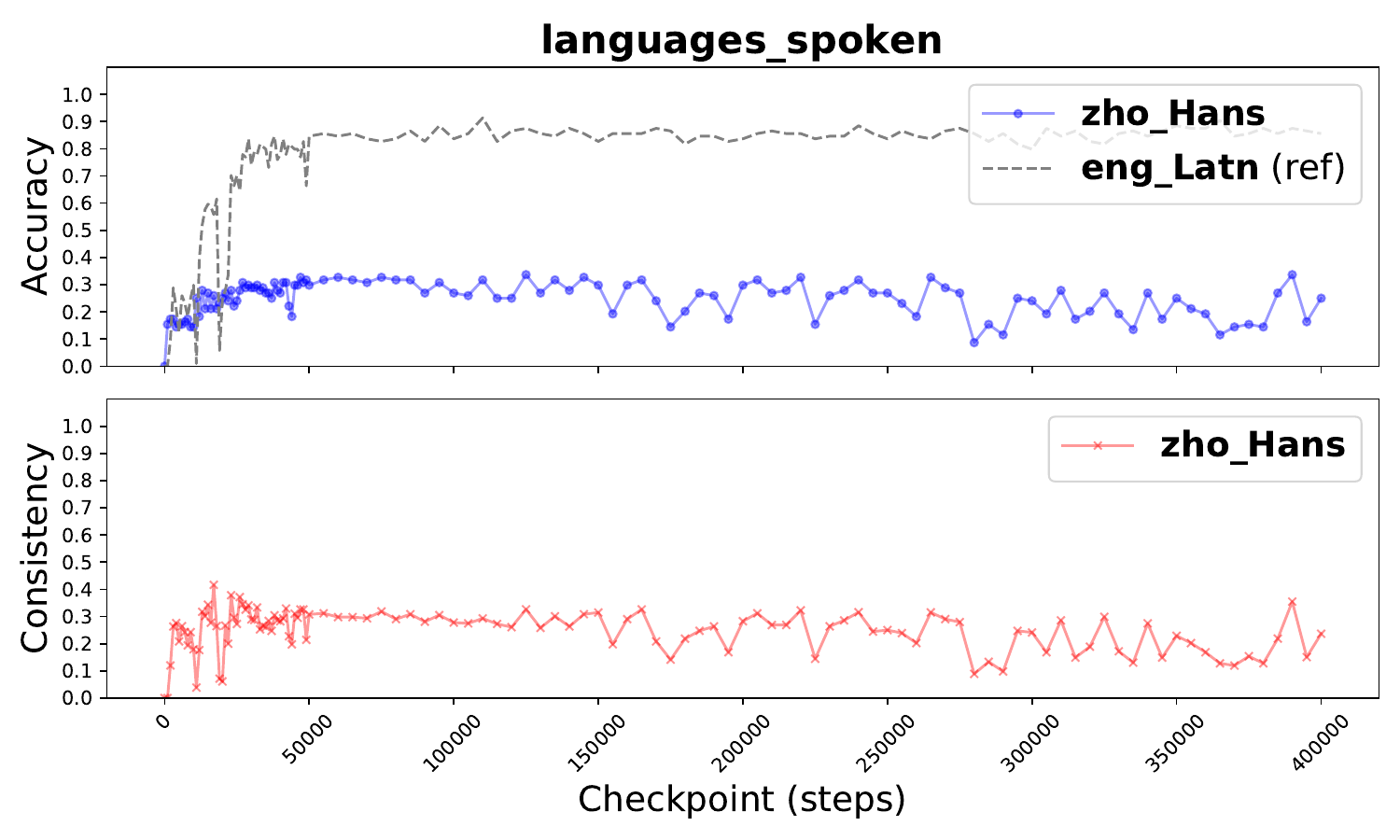}
    \includegraphics[width=0.24\textwidth]{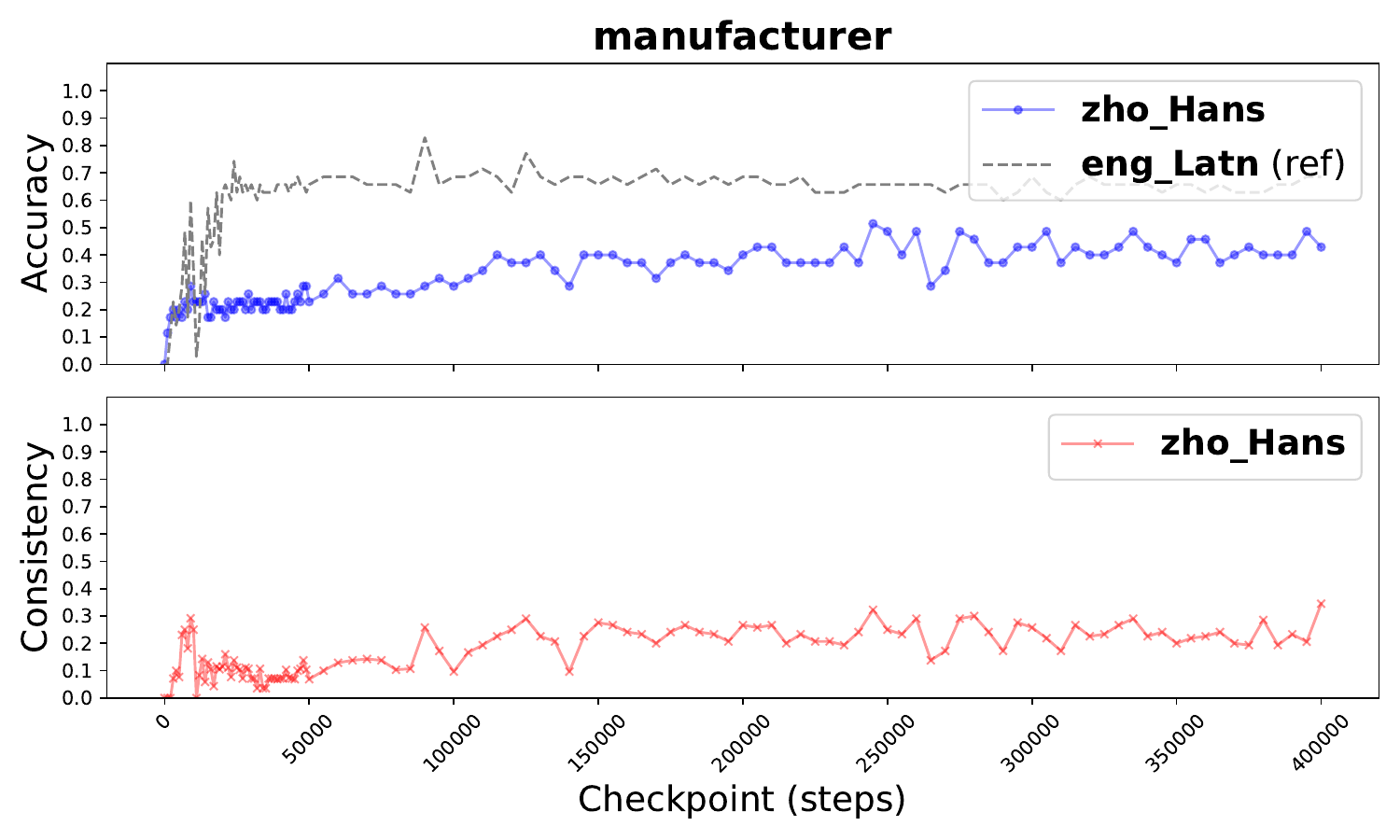}
    \includegraphics[width=0.24\textwidth]{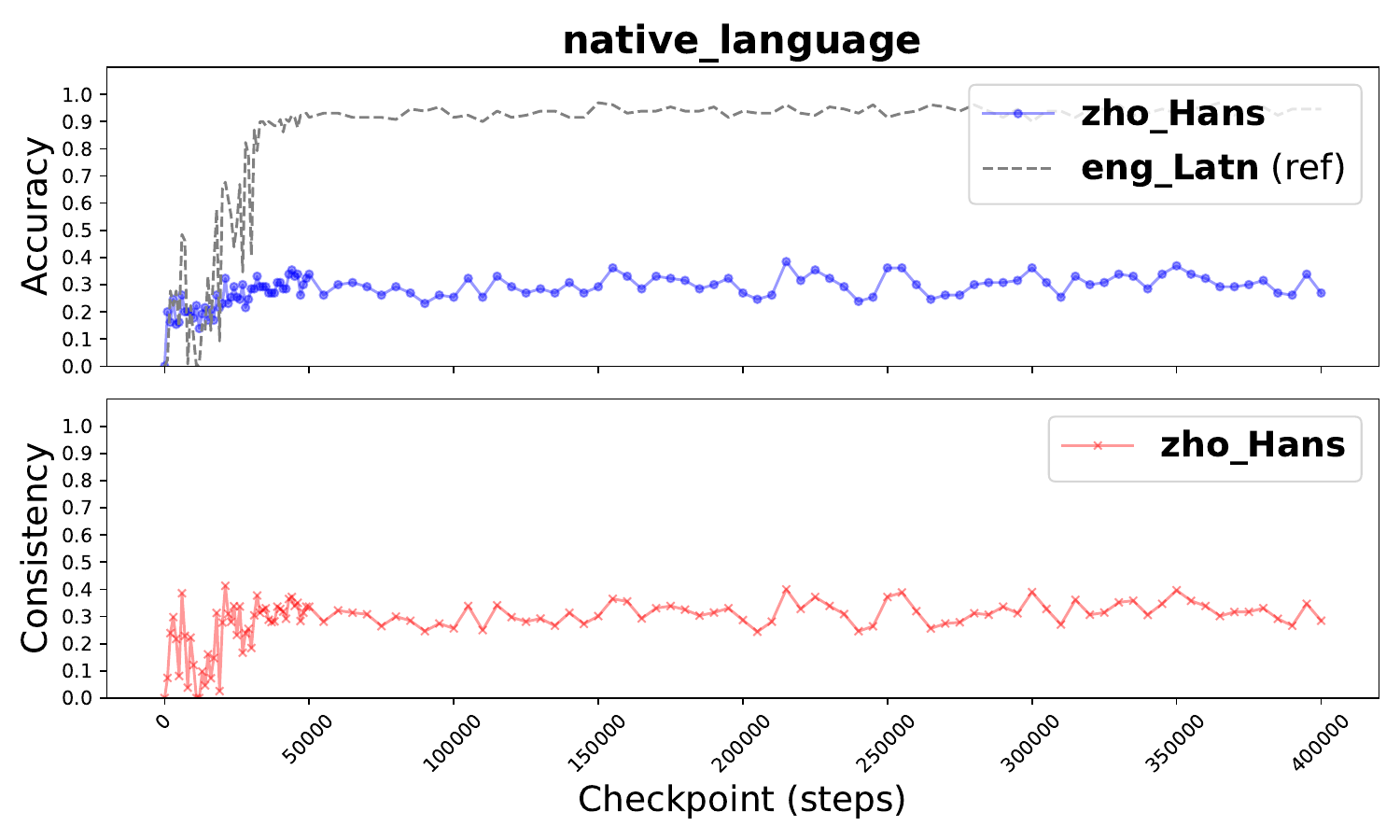}
    \includegraphics[width=0.24\textwidth]{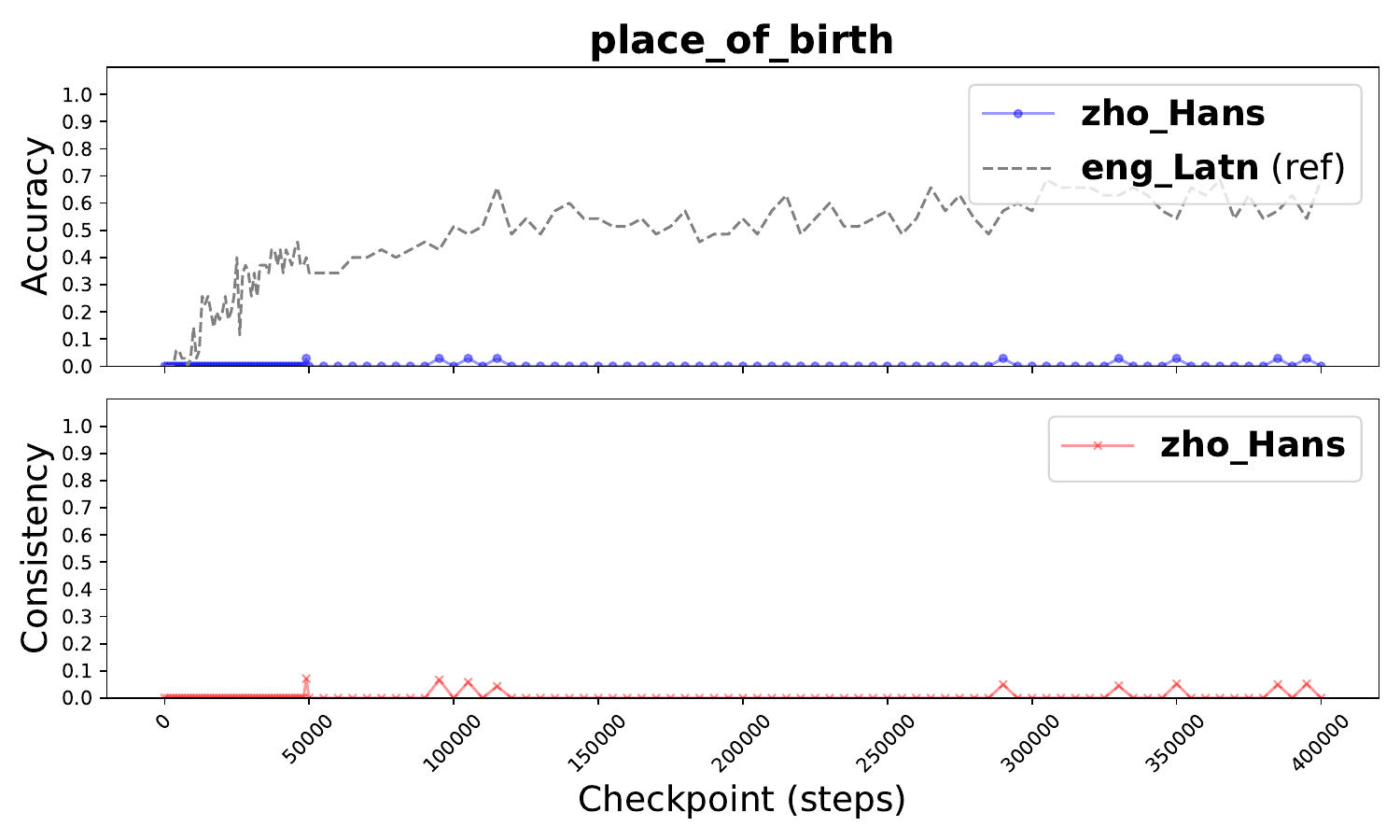}
    \includegraphics[width=0.24\textwidth]{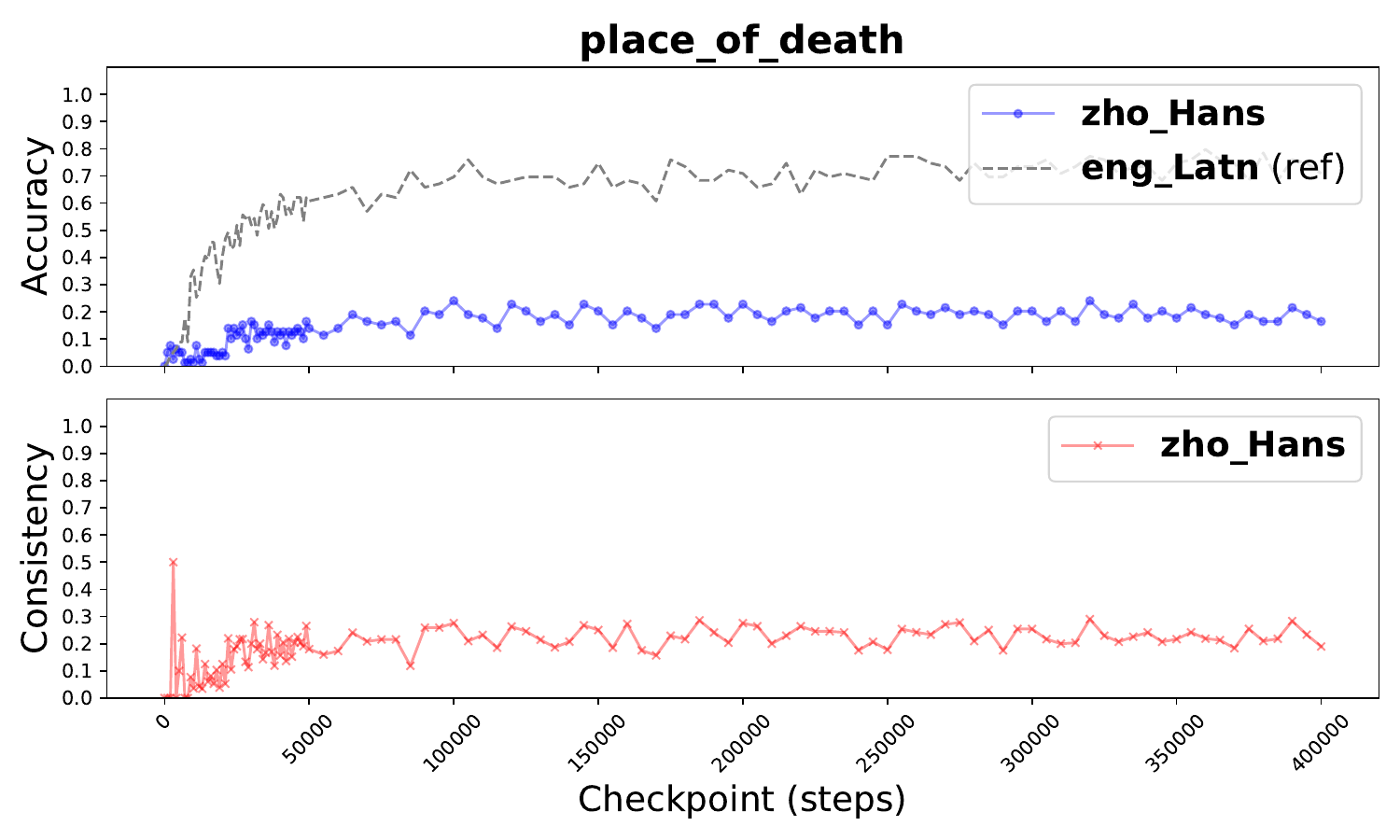}
    \includegraphics[width=0.24\textwidth]{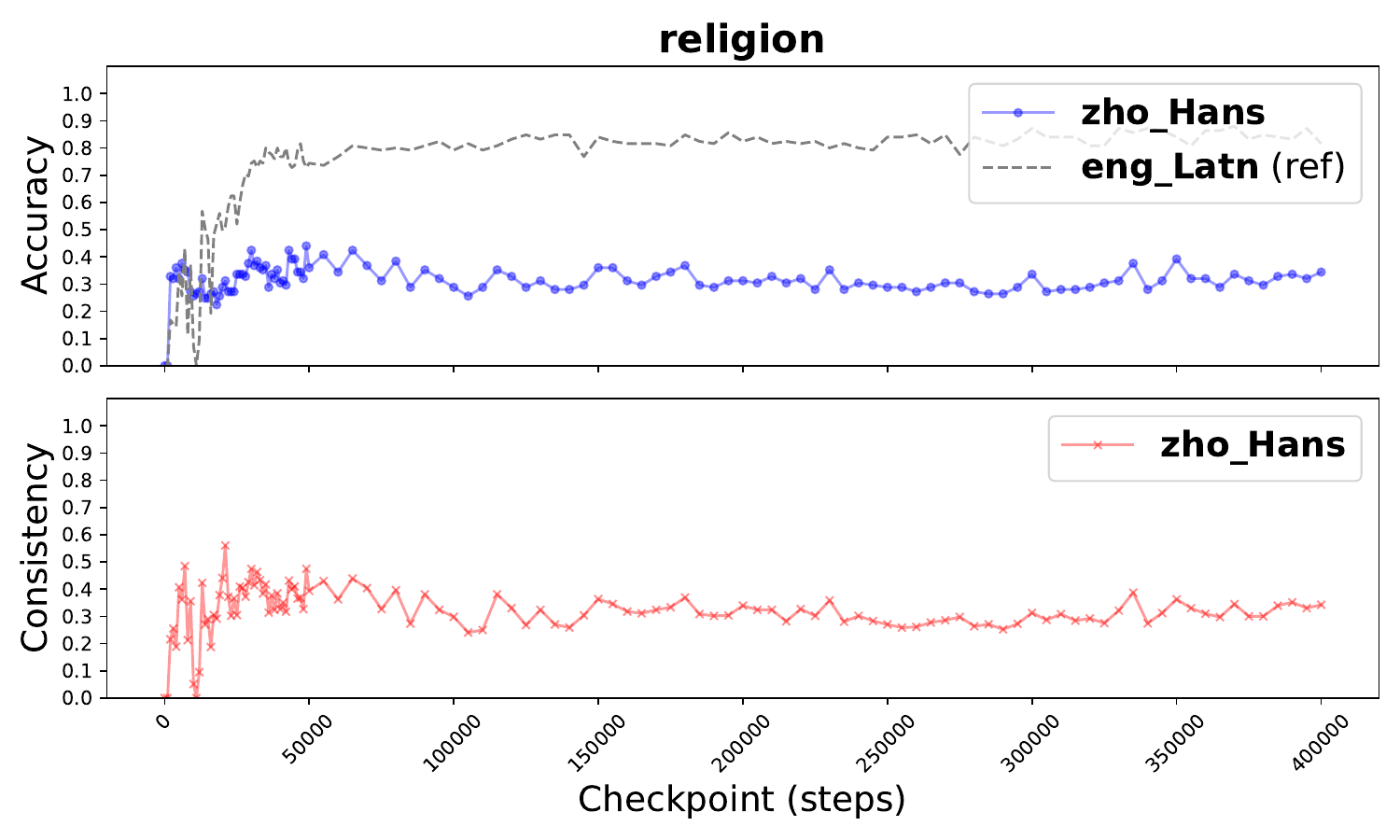}
    \caption{Factual accuracy (ACC) and crosslingual consistency (CO) for each relation type in \textbf{zho\_Hans}.}
    \label{fig:performance_over_checkpoints_zh}
\end{figure*}

In this section, we analyze factual recall accuracy and crosslingual consistency at the level of individual relations across languages, enabling us to examine how factual knowledge of different relation types evolves over the pretraining progression.
We report the results for ara\_Arab in Figure~\ref{fig:performance_over_checkpoints_ar}, cat\_Latn in Figure~\ref{fig:performance_over_checkpoints_ca}, ell\_Grek in Figure~\ref{fig:performance_over_checkpoints_el}, spa\_Latn in Figure~\ref{fig:performance_over_checkpoints_es}, fra\_Latn in Figure~\ref{fig:performance_over_checkpoints_fr}, jpn\_Jpan in Figure~\ref{fig:performance_over_checkpoints_ja}, kor\_Kore in Figure~\ref{fig:performance_over_checkpoints_ko}, rus\_Cryl in Figure~\ref{fig:performance_over_checkpoints_ru}, tur\_Latn in Figure~\ref{fig:performance_over_checkpoints_tr}, urk\_Cryl in Figure~\ref{fig:performance_over_checkpoints_uk}, and zho\_Hans in Figure~\ref{fig:performance_over_checkpoints_zh}.

We observe a similar trend as shown in \secref{dynamics}: the consistency in each relation is primarily driven by whether the fact is correctly recalled in each language $l \neq$ eng\_Latn, since the corresponding fact is almost always recalled in English.

The accuracy varies substantially across different relations within each language, with particularly large disparities in languages that use non-Latin scripts.
For example, ara\_Arab has nearly zero accuracy for \texttt{place\_of\_birth} relation.

%% file: prompt_variation.tex
\begin{table*}[h]
\centering
\footnotesize
\resizebox{\textwidth}{!}{%
\begin{tabular}{l|rrrrr|rrrrr}
\hline
& \multicolumn{5}{c|}{\textbf{Factual Recall (ACC, \%)}} & \multicolumn{5}{c}{\textbf{Consistency (CO, \%)}} \\
\cline{2-11}
\textbf{Language} & Template 1 & Template 2 & Template 3 & Template 4 & Template 5 & Template 1 & Template 2 & Template 3 & Template 4 & Template 5 \\
\hline
ara\_Arab & 17.5 & 18.2 & 18.1 & 18.3 & 18.1 & 18.9 & 19.3 & 19.5 & 19.5 & 18.9 \\
cat\_Latn & 44.2 & 45.3 & 43.5 & 45.4 & 44.7 & 48.5 & 49.4 & 47.6 & 49.3 & 48.5 \\
ell\_Grek & 18.8 & 17.1 & 18.6 & 17.2 & 18.8 & 20.0 & 18.0 & 19.7 & 18.2 & 19.7 \\
eng\_Latn & 82.0 & 81.9 & 81.4 & 81.9 & 82.2 & 100.0 & 100.0 & 100.0 & 100.0 & 100.0 \\
spa\_Latn & 56.4 & 55.6 & 55.4 & 55.7 & 56.8 & 62.0 & 60.2 & 61.8 & 61.2 & 61.7 \\
fra\_Latn & 62.0 & 62.7 & 62.7 & 62.0 & 61.3 & 68.1 & 68.1 & 69.3 & 67.8 & 67.5 \\
jpn\_Jpan & 18.5 & 18.1 & 17.8 & 19.5 & 20.3 & 19.6 & 19.1 & 18.3 & 20.6 & 21.2 \\
kor\_Kore & 20.0 & 18.9 & 18.8 & 19.8 & 19.1 & 20.7 & 19.3 & 19.4 & 20.8 & 20.5 \\
rus\_Cyrl & 31.2 & 31.1 & 32.1 & 30.6 & 32.2 & 33.3 & 32.6 & 33.9 & 31.7 & 33.5 \\
tur\_Latn & 40.1 & 40.8 & 40.4 & 41.3 & 39.9 & 42.3 & 43.4 & 42.8 & 43.8 & 42.7 \\
ukr\_Cyrl & 23.7 & 23.1 & 23.3 & 23.1 & 24.0 & 24.8 & 24.1 & 24.2 & 23.9 & 25.2 \\
zho\_Hans & 26.2 & 25.2 & 26.3 & 26.2 & 25.1 & 27.6 & 26.7 & 27.5 & 27.1 & 26.1 \\
\hline
\end{tabular}
}
\caption{Factual recall accuracy and crosslingual consistency (with respect to English) across five prompt templates.}
\label{tab:merged_acc_co_templates}
\end{table*}

\section{Effect of Prompt Template Variation}\seclabel{prompt_variation}

Prompt template variation, or prompt phrasing, can largely affect LLM behavior, particularly for open-ended or generative tasks. 
However, in the context of factual recall, where the expected output is typically a short, well-defined answer, the influence of prompt variation is more limited. 
To verify this, we conduct a case study using 5 different prompt templates provided by KLAR \citep{wang2025multilinguality} (Template 1 in them is used in the main text of this paper) on the step-400000 checkpoint.
The results of factual recall accuracy and consistency (with respect to English) are shown in Table~\ref{tab:merged_acc_co_templates}.
We can observe that the factual recall performance (accuracy and consistency) remains consistent across different prompts in all languages, confirming that the prompt variation does not have a substantial effect on the factual recall.
These complementary results indicate the robustness of our findings in the main text.